\pdfoutput=1
\documentclass[12pt,table,xcdraw]{cmuthesis}
\usepackage{helvet}
\usepackage{times}

\usepackage[T1]{fontenc}
\usepackage{fullpage}
\usepackage{graphicx}
\usepackage{amssymb}
\usepackage{amsmath}
\usepackage[numbers,sort]{natbib}
\usepackage[pageanchor=true,plainpages=false, pdfpagelabels, bookmarks,bookmarksnumbered,
pdfborder=0 0 0,  
]{hyperref}
\usepackage{wrapfig}
\usepackage{booktabs}
\usepackage{multirow}
\usepackage{lipsum}
\usepackage[skip=4pt,font=small,labelfont=bf]{caption}
\usepackage{appendix}
\usepackage{chngcntr}
\usepackage{etoolbox}
\usepackage{soul}
\usepackage{enumitem}
\usepackage{bm}
\usepackage{bbm}
\usepackage{amsthm}
\usepackage{hhline}
\usepackage{mathtools}
\usepackage{array}
\usepackage{color}
\usepackage{xcolor,colortbl}
\usepackage{MnSymbol}
\usepackage{makecell}
\usepackage{subcaption}
\usepackage{arydshln}
\usepackage{framed}
\usepackage[scientific-notation=true]{siunitx}
\usepackage{algorithm}
\usepackage{sidecap}
\usepackage{pifont}
\usepackage{fixltx2e}
\usepackage[flushleft]{threeparttable}
\usepackage{diagbox}
\usepackage{listings}
\usepackage{latexsym}
\usepackage[geometry]{ifsym}
\usepackage{amsfonts}
\usepackage{soul}
\usepackage{url}
\usepackage{algorithm}
\usepackage{algorithmic}

\newcommand{\cmark}{\ding{51}}%
\newcommand{\xmark}{\ding{55}}%
\definecolor{gg}{RGB}{15,150,15}
\definecolor{rr}{RGB}{250,15,15}
\definecolor{ggg}{RGB}{15,250,15}
\definecolor{rrr}{RGB}{250,15,15}
\definecolor{yyy}{RGB}{230,185,25}
\definecolor{yy}{RGB}{230,185,25}
\definecolor{darkpastelgreen}{rgb}{0.01, 0.75, 0.24}
\definecolor{amethyst}{rgb}{0.6, 0.4, 0.8}

\newcommand{\PreserveBackslash}[1]{\let\temp=\\#1\let\\=\temp}
\newcolumntype{C}[1]{>{\PreserveBackslash\centering}p{#1}}
\newcolumntype{R}[1]{>{\PreserveBackslash\raggedleft}p{#1}}
\newcolumntype{L}[1]{>{\PreserveBackslash\raggedright}p{#1}}

\makeatletter
\def\maketag@@@#1{\hbox{\m@th\normalfont\normalsize#1}}
\makeatother

\usepackage{amsmath,amsfonts,bm}

\def\eqref#1{eq~(\ref{#1})}

\def\1{\bm{1}}

\def\vx{{\bm{x}}}

\DeclareMathAlphabet{\mathsfit}{\encodingdefault}{\sfdefault}{m}{sl}
\SetMathAlphabet{\mathsfit}{bold}{\encodingdefault}{\sfdefault}{bx}{n}

\usepackage{hyperref}

\usepackage{booktabs}       %
\usepackage{amsfonts}       %
\usepackage{nicefrac}       %
\usepackage{microtype}      %
\usepackage{subcaption}

\usepackage{enumitem}
\setlist{nolistsep}
\setlist[itemize]{noitemsep, topsep=0pt}

\usepackage{times}

\usepackage{graphicx} %
\usepackage{color}

\usepackage{array}
\usepackage{amssymb}
\usepackage{amsmath}
\usepackage{xspace}
\usepackage{fancyhdr}
\usepackage{comment}

\newcolumntype{H}{>{\setbox0=\hbox\bgroup}c<{\egroup}@{}}

\newcommand{\noaistats}[1]{}  %

\definecolor{darkgreen}{rgb}{0,0.4,0.0}
\definecolor{darkblue}{rgb}{0,0.1,0.3}
\definecolor{darkred}{rgb}{0.7,0.0,0.0}

\newcommand\norm[1]{\left\lVert#1\right\rVert}

\newtheorem{theorem}{Theorem}

\newtheorem{definition}{Definition}

\DeclareMathOperator*{\argmax}{arg\,max}
\DeclareMathOperator*{\argmin}{arg\,min}

\newcommand\blfootnote[1]{%
  \begingroup
  \renewcommand\thefootnote{}\footnote{#1}%
  \addtocounter{footnote}{-1}%
  \endgroup
}

\newcommand{\pid}{\textsc{PID}}
\newcommand{\factorcl}{\textsc{FactorCL}}
\newcommand{\multiviz}{\textsc{MultiViz}}

\newcommand{\multibench}{\textsc{MultiBench}}
\newcommand{\multizoo}{\textsc{MultiZoo}}
\newcommand{\highmmt}{\textsc{HighMMT}}

\newcommand{\mult}{\textsc{MulT}}

\newcommand{\sentdebias}{\textsc{Sent-Debias}}
\newcommand{\ainlp}{\textsc{A-INLP}}

\AtBeginEnvironment{subappendices}{%
\chapter*{Appendix}
\addcontentsline{toc}{chapter}{Appendices}
\counterwithin{figure}{section}
\counterwithin{table}{section}
}

\newsavebox{\mybox}

\makeatletter

\newcommand*\wthelper[2]{%
        \hbox{\dimen@\accentfontxheight#1%
                \accentfontxheight#11.3\dimen@
                $\m@th#1\widetilde{#2}$%
                \accentfontxheight#1\dimen@
        }%
}

\newcommand*\accentfontxheight[1]{%
        \fontdimen5\ifx#1\displaystyle
                \textfont
        \else\ifx#1\textstyle
                \textfont
        \else\ifx#1\scriptstyle
                \scriptfont
        \else
                \scriptscriptfont
        \fi\fi\fi3
}
\makeatother

\makeatletter
\def\blfootnote{\gdef\@thefnmark{}\@footnotetext}
\makeatother

\usepackage[letterpaper,twoside,vscale=.8,hscale=.75,nomarginpar,hmarginratio=1:1]{geometry}

\begin {document} 
\sbox{\mybox}{\hbadness=10000 \parbox{2cm}{\lipsum[1]}}
\frontmatter

\pagestyle{empty}

\title{ 
{\bf Foundations of Multisensory Artificial Intelligence}}
\author{Paul Pu Liang}
\date{May 2024}
\Year{2024}
\trnumber{CMU-ML-24-103}
\committee{
Louis-Philippe Morency, Co-chair \\
Ruslan Salakhutdinov, Co-chair \\
Manuel Blum \\
Lenore Blum (UC Berkeley) \\
Trevor Darrell (UC Berkeley)
}

\support{This research was funded by: National Science Foundation awards IIS1722822 and IIS1750439; National Institutes of Health awards R01MH096951 and U01MH116923; graduate fellowships from Meta Platforms and Siebel Scholars; and grants from Meta Platforms, Nippon Telegraph and Telephone Corporation, Oculus VR, and Samsung Electronics.}



\maketitle

\let\cleardoublepage\clearpage

\begin{abstract}
\vspace{4mm}

Building multisensory artificial intelligence systems that learn from multiple sensory inputs such as text, speech, video, real-world sensors, wearable devices, and medical data holds great promise for impact in many scientific areas with practical benefits, such as in supporting human health and well-being, enabling multimedia content processing, and enhancing real-world autonomous agents.

However, the breadth of progress in multimodal research has made it difficult to identify the common themes and open questions in the field. By synthesizing a range of theoretical frameworks and application domains, this thesis aims to advance the foundations of multimodal machine learning. We start by defining three key principles of modality \textit{heterogeneity}, \textit{connections}, and \textit{interactions} often present in multimodal problems~\cite{liang2022foundations}. Using these principles as a foundation, we propose a taxonomy of six core challenges in multimodal research: \textit{representation}, \textit{alignment}, \textit{reasoning}, \textit{generation}, \textit{transference}, and \textit{quantification}. Recent technical achievements will be presented through this taxonomy, allowing researchers to understand the similarities and differences across approaches, and identifying open problems for future research.

The bulk of the thesis covers our recent progress towards tackling two key problems in multimodal learning: the machine learning foundations of multimodal interactions, as well as practical methods for building multisensory foundation models that generalize to many modalities and tasks in the real world.

In the first part, we study the foundations of multimodal interactions: the basic principle of how modalities combine to give rise to new information for a task. We present a theoretical framework formalizing how \textit{modalities interact} with each other to give rise to new information for a task, such as sarcasm identified from the incongruity between spoken words and vocal expressions~\cite{liang2023quantifying}.
Using this theoretical framework, we propose two practical estimators to quantify the interactions in real-world datasets. Quantifying the types of interactions a multimodal task requires enables researchers to decide which modality to collect~\cite{liang2023multimodal}, design suitable approaches to learn these interactions~\cite{liang2023factorized}, and analyze whether their model has succeeded in learning~\cite{liang2023multiviz}.

In the second part, we study the design of practical multimodal foundation models that generalize over many modalities and tasks, which presents a step toward grounding large language models to real-world sensory modalities.
We first introduce \multibench, a unified large-scale benchmark across a wide range of modalities, tasks, and research areas~\cite{liang2021multibench}.
We will also present the \textit{cross-modal attention}~\cite{liang2018multimodal,chen2017multimodal} and \textit{multimodal transformer}~\cite{tsai2019multimodal} architectures that now underpin many of today’s multimodal foundation models.
Scaling these architectures on \multibench\ enables the creation of general-purpose multimodal multitask models across a variety of tasks, and we have collaborated broadly with practitioners to apply these models for real-world impact on affective computing, mental health, and cancer prognosis.

We conclude this thesis by discussing how future work can leverage these ideas toward more general, interactive, and safe multimodal artificial intelligence.
\end{abstract}

\let\cleardoublepage\clearpage

\begin{acknowledgments}
\vspace{4mm}

I owe my greatest acknowledgments to my advisors, mentors, and thesis committee members for their invaluable guidance during my PhD. To Louis-Philippe Morency and Ruslan Salakutdinov, who have closely guided my research and personal development at every stage over the past 5 years. LP has mentored me closely in all aspects of research - brainstorming ideas, idea execution, and written and oral presentations. Some of the best memories I've had during my PhD have been whiteboard brainstorming sessions, coming up with good names for problems and models, and collaboratively figuring out the best ways to visually depict technical concepts. Russ's incredibly sharp insight and keen eye for impactful problems has shaped my thinking and forced me to work on problems that matter in practice, and I've thoroughly enjoyed our recent push towards interactive multimodal agents with other folks in the group and at CMU. Thank you LP and Russ for additionally giving me the opportunity to co-instruct and guest lecture CMU courses multimodal ML, deep learning, and socially intelligent AI. I also had the pleasure of working closely with Manuel Blum and Lenore Blum during the senior years of my PhD. I have learned a lot from our discussions at the intersection of artificial intelligence, consciousness, and neuroscience, which have changed how I look at long-term problems and approach them. Manuel and Lenore have also inspired me to think big and make broad impact across CS and beyond, giving me many opportunities to communicate my ideas and contributions to a wide audience in neuroscience, psychology, and more. Finally, Trevor Darrell has been a source of inspiration as a senior faculty and has given me great advice for my PhD research and broadly for my research career. Some of his early works in multimodal machine learning and multimodal interaction are still some of my favorite works in this space.

Beyond my committee members, I would like to acknowledge other CMU faculty and students with whom I've had fruitful collaborations, discussions, and received helpful feedback on ideas, paper drafts, and presentations. Fantastic students in LP and Russ's research groups: Hubert Tsai, Amir Zadeh, Chaitanya Ahuja, Volkan Cirik, Torsten Wortwein, Martin Ma, Alex Wilf, Leena Mathur, Victoria Lin, Yousouf Kebe, Devendra Chaplot, Bhuwan Dhingra, Lisa Lee, Shrimai Prabhumoye, Ben Eysenbach, Jing Yu Koh, Minji Yoon, Brandon Trabucco, Murtaza Dalal, Yue Wu, and Kelly He. In addition, Hai Pham, Shaojie Bai, and their advisors Barnabas Poczos, and Zico Kolter with whom I did some early work in multimodal representation learning. Yonatan Bisk, Daniel Fried, Albert Gu, Zack Lipton, Tom Mitchell, Graham Neubig, Mayank Goel, and Haiyi Zhu who have given me a lot of advice (both personal and professional) over the years. Roni Rosenfeld, Ryan Tibshirani, and Tai Sing Lee who mentored me on undergraduate research projects at CMU. Finally, CMU Machine Learning Department and Language Technologies Institute are some of the best places to do AI research, and this could not be possible without the fantastic support from staff like Diane Stidle, Dorothy Holland-Minkley, and John Friday.

\clearpage

Some folks outside CMU I would like to thank include: Faisal Mahmood's group at Harvard Medical School, especially students Richard Chen, Guillaume Jaume, and Anurag Vadiya for a series of fruitful collaborations regarding multimodal computational pathology; David Brent at UPMC, Nicholas Allen at University of Oregon, and Randy Auerbach at Columbia University for collaborations on daily mood assessment, markers of suicide ideation, and mobile health; Liangqiong Qu, Yuyin Zhou, Daniel Rubin, and James Zou at Stanford for investigations into multimodal and federated learning for biomedical applications; and most recently Jack Hessel, Yejin Choi, and Jae Sung Park at University of Washington/AI2 for many discussions regarding research and projects on vision-language commonsense reasoning.

I was also lucky to be mentored by several fantastic researchers in industry labs during my internships. To Manzil Zaheer at Google, you have made me a more mature researcher by reminding me to focus deeply on problems rather than jumping around during my junior researcher days. Our close collaborations have also strengthened my expertise in both fundamental and practical machine learning. To Yuke Zhu, Anima Anandkumar, and Sanja Fidler at Nvidia, you have done a great job setting up a vibrant and flexible research environment at Nvidia and I have learned a lot about the latest progress in multisensor robotics, AI for science, and vision-language models from our collaborations. To Makoto Yamada and Qibin Zhao at Riken AIP, where I learned more about tensors and kernels for multimodal learning. To Brandon Amos, Tim Rockt\"{a}schel, and Ed Grefenstette at Facebook AI, where I learned a lot about optimization, control, and reinforcement learning. And finally, to Dani Yogatama, Lisa Anne Hendricks and Aida Nematzadeh at DeepMind, where I gained practical experience training large-scale multimodal foundation models.

The most personally rewarding part of my PhD was definitely the many undergraduate, masters, and PhD students I have had the pleasure of advising - both at CMU and around the world: Adejuwon Fasanya, Akshay Goindani, Aviv Bick, Arav Agarwal, Chengfeng Mao, Chiyu Wu, Dong Won Lee, Edmund Tong, Gunjan Chhablani, Haofei Yu, Haoli Yin, Holmes Wu, Irene Li, Jiewen Hu, Jingyi Zhang, Jivat Neet, Katrina Jiao, Marian Qian, Peter Wu, Rana Shahroz, Richard Zhu, Rohan Pandey, Rulin Shao, Samuael Adnew, Samuel Yu, Seong Hyeon Park, Shentong Mo, Siyuan Wu, Talha Chafekar, Terrance Liu, Xiang Fan, Xiangru Tang, Yao Chong Lim, Ying Shen, Yiwei Lyu, Yudong Liu, Yun Cheng, Yuxin Xiao, Zhun Liu, Zihao Deng, Ziyin Liu. All of you have taught me so much and become experts in your own fields. I'm delighted to see all of you make great strides in PhD studies and industry, and look forward to hearing about your successes in the future.

Finally, I could not have done all this without the close support of my family and friends, especially from my mom, dad, sister, and grandparents, Jane, Truffle, and Tigger, Jane's family, close friends Chun Kai Ling, Yue Niu, Raahul Sriram, Dylan Sam, Rattana Pukdee, Jennifer Hsia, Clara Na, Cindy Wu, Pratyush Maini, Ananye Agarwal, Yiding Jiang, Sam Sokota, Alex Wilf, Leena Mathur, Yiwei Lyu, Chirag Gupta, Tom Yan, Helen Zhou, Manzil Zaheer, and many more.

\end{acknowledgments}

\let\cleardoublepage\clearpage

\tableofcontents
\listoffigures
\listoftables

\mainmatter

\chapter{Introduction}
\label{chap:intro}
Multimodal artificial intelligence is a vibrant multi-disciplinary research field that aims to design computer agents that can perceive, reason, and interact through multiple communicative modalities, including linguistic, acoustic, visual, tactile, sensory, and physiological messages~\cite{baltruvsaitis2018multimodal,liang2022foundations}. Multimodal AI systems can bring great impact in many scientific areas with practical benefits, such as in supporting human health and well-being~\cite{liang2018computational,morency2011towards,zadeh2019social}, enabling multimedia content processing~\cite{agrawal2017vqa,plummer2015flickr30k,rui1999image}, and enhancing real-world autonomous agents~\cite{bisk2020piqa,chen2020soundspaces,lee2020multimodal,savva2019habitat,shridhar2020alfred}.

However, the breadth of progress in multimodal research has made it difficult to identify the common themes and open questions in the field. By synthesizing a broad range of theoretical frameworks and application domains from both historical and recent perspectives, this thesis is designed to advance the theoretical and computational foundations of multimodal machine learning. We start by defining three key principles of modality \textit{heterogeneity}, \textit{connections}, and \textit{interactions} often present in multimodal problems which brings unique challenges to machine learning. The heterogeneity of multimodal data makes learning challenging, for example, language is often seen as symbolic while audio and video are represented as continuous signals. At the same time, these modalities contain overlapping connected information, and interact to give rise to new information relevant for a task. It is crucial to learn these connections and interactions for systems to perform well. Using these principles as a foundation, we propose a taxonomy of six core challenges in multimodal research: \textit{representation}, \textit{alignment}, \textit{reasoning}, \textit{generation}, \textit{transference}, and \textit{quantification}. Recent technical achievements will be presented through the lens of this taxonomy, allowing researchers to understand the similarities and differences across new approaches, and enabling us to identify key open problems for future research.

\begin{figure}[tbp]
\centering
\includegraphics[width=\linewidth]{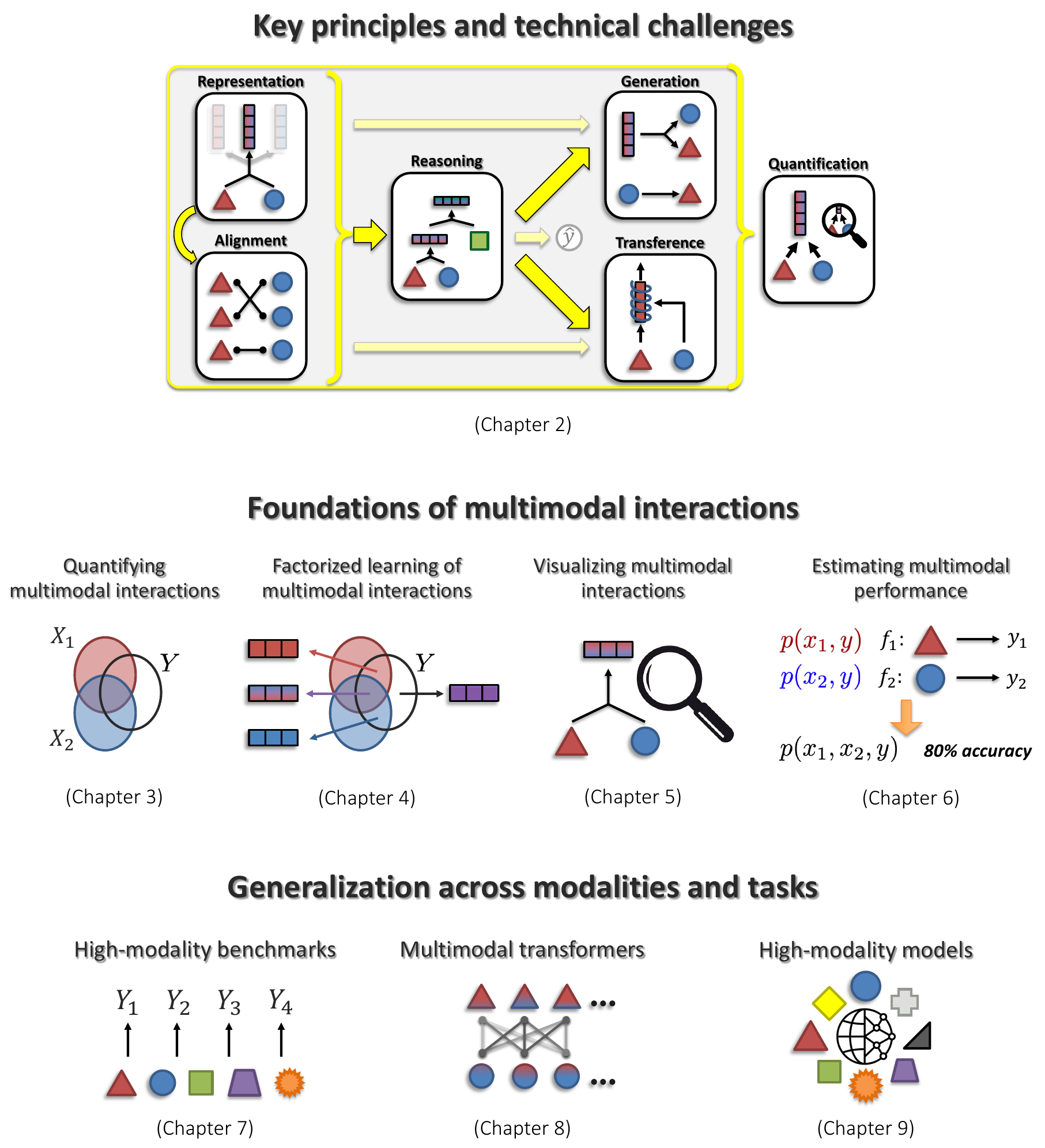}
\vspace{0mm}
\caption{This thesis is designed to advance the theoretical and computational foundations of multimodal machine learning, and enable the creation of next-generation multimodal technologies. It starts by identifying the common themes and open questions in the field, through a taxonomy of six \textbf{core challenges} in multimodal research: representation, alignment, reasoning, generation, transference, and quantification. The bulk of the thesis studies two core challenges in multimodal learning: (1) building a \textbf{foundation for multimodal interactions} that enables the quantification of multimodal interactions in data and their principled modeling using machine learning methods, and (2) the data requirements and model building blocks enabling \textbf{generalization} of knowledge between modalities, tasks, and their representations.}
\label{fig:intro_overview}
\vspace{-0mm}
\end{figure}

Using our taxonomy for multimodal machine learning, we highlight two key challenges that are important for progress in multimodal learning: (1) building the \textbf{foundations} of multimodal interactions so we can quantify the interactions present in datasets and model these interactions correctly using machine learning methods, and (2) constructing multimodal models and datasets that enable \textbf{generalization} across a large number of modalities and tasks for real-world societal impact (Figure~\ref{fig:intro_overview}).

\section{Foundations of Multimodal Interactions}

Multimodal interactions can be categorized into redundancy, uniqueness, and synergy: \textit{redundancy} quantifies information shared between modalities, such as smiling while telling an overtly humorous joke; \textit{uniqueness} quantifies the information present in only one, such as each medical sensor designed to provide new information; and \textit{synergy} quantifies the emergence of new information using both, such as conveying sarcasm through disagreeing verbal and nonverbal cues~\cite{liang2022foundations}.
These interactions are the basic principles of how modalities combine to give rise to new information for a task, which is present in all multimodal problems.
While there have been intuitive definitions of these multimodal interactions, we still lack a formal foundation and systematic understanding of how to learn these interactions from data. As a result, there remain basic open questions like:
\begin{center}
    \textit{What interactions are in my data?}\\
    \textit{What interactions are learned by different models?}\\
    \textit{What models are suitable for my data?}\\
\end{center}

To answer these questions, the first part of the thesis presents a theoretical framework formalizing the \textit{useful information} in each modality and how \textit{modalities interact} with each other to give rise to new information for a task~\cite{liang2023quantifying}. Based on this theoretical framework, we propose two practical estimators to quantify the interactions in high-dimensional datasets, which can also be used more broadly for estimating information-theoretic quantities in real-world distributions. These estimators allow us to understand the information and interactions in multimodal datasets, and design the right models that provably learn the desired interactions in data.

We further show several broader implications that quantifying multimodal interactions can have on practitioners. Firstly, we operationalize the learning of multimodal interactions through a new approach called Factorized Contrastive Learning to capture both shared and unique information across modalities~\cite{liang2023factorized}. Secondly, a formal definition of multimodal interactions also enables us to analyze through qualitative visualizations whether a trained model has succeeded in learning the desired interactions from data~\cite{liang2023multiviz}. Finally, we show how to use this information-theoretic framework to estimate the performance of optimal multimodal models given only unimodal data, which can inform practitioners which modalities to collect, and whether multimodal modeling is worth it for maximum increase in performance~\cite{liang2023multimodal}. We release all code for quantifying multimodal interaction (both exact and approximate), and their implications on understanding datasets and models at \url{https://github.com/pliang279/PID}, code for Factorized Contrastive Learning at \url{https://github.com/pliang279/FactorCL}, and code for visualizing and debugging multimodal models at \url{https://github.com/pliang279/MultiViz}, which can help practitioners navigate the multimodal modeling pipeline.

\section{Multisensory Foundation Models}

There has been substantial impact of foundation models (e.g., large language models) trained on vast amounts of unlabeled data to obtain general-purpose capabilities over many prediction tasks. The future will lie in multisensory foundation models that are \textit{grounded in the world}: being able to simultaneously process a large number of modalities beyond language, to vision, audio~\cite{agrawal2017vqa,liang2018computational,lin2014microsoft,ramesh2021zero}, and leveraging advances in sensing technologies such as cellphones~\cite{liang2021learning}, wearable devices~\cite{hamisu2011accessible}, autonomous vehicles~\cite{yeong2021sensor}, healthcare technologies~\citep{MIMIC}, and robots~\cite{belpaeme2018social,kim2013social}. The large number of heterogeneous modalities creates challenges in building multisensory foundation models. For example, the healthcare domain typically collects tabular data and high-frequency sensors~\citep{MIMIC}, and it remains an open question how to best combine large language models with tabular data and sensors~\cite{shwartz2022tabular}. In the second part of this thesis, we take steps towards both data and modeling requirements to build the next generation of multisensory foundation models: 
\begin{center}
    \textit{What data sources do we need to train foundation models over many heterogeneous modalities?}\\
    \textit{What modeling architectures are suitable for scaling to many heterogeneous modalities?}
\end{center}
To answer the first question, we introduce \multibench, the largest and most comprehensive multimodal benchmark enabling the training of multisensory foundation models. \multibench\ collects and standardizes $15$ realistic datasets across $10$ diverse modalities, $20$ prediction tasks, and $6$ research areas from multimedia, affective computing, robotics, HCI, finance, and healthcare. \multibench\ is publicly available at \url{https://github.com/pliang279/MultiBench}, and has been broadly used in the community to train and evaluate multimodal architectures.

On the modeling side, prior work on multimodal learning has focused on a fixed set of modalities (e.g., image and text), without tackling generalization to many heterogeneous modalities and tasks necessary for truly multisensory models. To tackle the heterogeneity across many different modalities, we treat modalities in their most general form as sequences of elements, and present the \textit{cross-modal attention}~\cite{liang2018multimodal,chen2017multimodal} and \textit{multimodal transformer}~\cite{tsai2019multimodal} architectures to learn interactions between all sequences of elements. These multimodal transformers are scalable and achieve strong results over a wide range of modalities, and we show their applications to image, text, video, sensors, and medical data. Finally, using \multibench, we scale multimodal transformers to the high-modality setting, resulting in a single model architecture with the same set of parameters that can function across a large number of modalities partially observed for different tasks~\cite{liang2022highmmt} (e.g., image and text on the internet, video and audio in human communication, video and sensors in robotics, and so on). This represents the most realistic setting of how humans process the multisensory world, and we believe that general-purpose AI systems will also need to be trained in the high-modality setting. Our collection of high-modality models, available at \url{https://github.com/pliang279/HighMMT}, has already been extended for learning over many modalities in the medical, internet-of-things, and affective computing domains.

We end the thesis by discussing our collaborative efforts in applying these multisensory models for real-world impact on affective computing, mental health, and cancer prognosis.

\section{Summary of Contributions}

In this section, we provide a highlight of our main thesis contributions.
\begin{enumerate}
    \item \textbf{Literature survey and taxonomy of multimodal challenges (Chapter~\ref{chap:background})}
    \begin{enumerate}
        \item \textbf{Three key principles}: We begin by defining three key principles that have driven technical challenges and innovations: (1) modalities are \textit{heterogeneous} because the information present often shows diverse qualities, structures, and representations, (2) modalities are \textit{connected} since they are often related and share commonalities, and (3) modalities \textit{interact} to give rise to new information when used for task inference. 
        \item \textbf{Six technical challenges}: Building upon these principles, we propose a new taxonomy of six core challenges in multimodal learning: (1) \textit{Representation} studies how to summarize multimodal data to reflect the heterogeneity and interconnections between individual modality elements, before (2) \textit{alignment} captures the connections and interactions between multiple local elements according to their structure. After representation and alignment comes (3) \textit{reasoning}, which aims to combine the information from multimodal evidence in a principled way that respects the structure of the problem to give more robust and interpretable predictions. While most systems aim to predict the label $y$, there are also cases where the goal is (4) \textit{generation}, to learn a generative process to produce raw modalities that reflect cross-modal interactions, structure, and coherence, or (5) \textit{transference}, to transfer information from high-resource modalities to low-resource ones and their representations. Finally, (6) \textit{quantification} revisits the previous challenges to give deeper empirical and theoretical understanding of modality heterogeneity, interconnections, and the learning process.
        \item \textbf{Current work and open directions}: For each challenge, we create a taxonomy of subchallenges and categorize recent advances in the field. This new taxonomy will enable researchers to better understand the state of research, and we identify several key directions for future work.
    \end{enumerate}
    
    \item \textbf{Foundations of multimodal interactions (Chapter~\ref{chap:foundations1})}
    \begin{enumerate}
        \item \textbf{Multimodal interactions} can be categorized into redundancy, uniqueness, and synergy: \textit{redundancy} quantifies information shared between modalities, such as smiling while telling an overtly humorous joke; \textit{uniqueness} quantifies the information present in only one, such as each medical sensor designed to provide new information; and \textit{synergy} quantifies the emergence of new information using both, such as conveying sarcasm through disagreeing verbal and nonverbal cues~\cite{liang2022foundations}.
        \item \textbf{Formal framework and estimation}: By introducing a new connection between information theory and multimodal interactions~\cite{liang2023quantifying}, I designed \textit{scalable estimators to quantify the interactions in large-scale multimodal datasets and those learned by multimodal models}. These estimators are based on max-entropy convex optimization and a scalable end-to-end estimator suitable for high-dimensional continuous data.
        \item \textbf{Model selection}: We show that quantifying the interactions enables practitioners to analyze their datasets and select the most appropriate model that captures the right interactions in the data. We implemented these methods in two real-world case studies in mental health assessment~\cite{liang2021learning} and cancer prognosis~\cite{liang2023quantifying} from multimodal data. Domain experts appreciated the transparency that these methods convey as opposed to black-box neural networks, resulting in trust and adoption in real-world practice.
    \end{enumerate}
    
    \item \textbf{Learning multimodal interactions using self-supervised learning (Chapter~\ref{chap:foundations2})}
    \begin{enumerate}
        \item \textbf{From estimation to learning}: Naturally, a formal definition of multimodal interactions also translates to new training objectives to learn these interactions using neural networks. We show how to better learn task-relevant \textit{unique information}~\cite{liang2023factorized,tsai2019learning} using self-supervised learning, going beyond shared information between modalities.
        \item \textbf{Factorized learning of each interaction}: \factorcl\ is built from three new contributions: (1) factorizing task-relevant information into shared and unique representations, (2) capturing task-relevant information via maximizing MI lower bounds and removing task-irrelevant information via minimizing MI upper bounds, and (3) multimodal data augmentations to approximate task relevance without labels.
        \item \textbf{Real-world settings with unique information}: On large-scale real-world datasets, \factorcl\ captures both shared and unique information and achieves state-of-the-art results on six benchmarks, including tasks involving medical sensors or robotics with force sensors that provide unique information, or cartoon images and figurative captions (i.e., not literal but metaphoric or idiomatic descriptions of the images).
    \end{enumerate}
    
    \item \textbf{Visualizing multimodal interactions in trained models (Chapter~\ref{chap:foundations3})}
    \begin{enumerate}
        \item \textbf{Interpreting multimodal models}: \multiviz\ is a framework for visualizing and understanding multimodal models across multiple stages: (1) modality importance, (2) multimodal interactions, and (3) multimodal reasoning. It includes tools to visualize what the model has learned about each stage of the prediction process.
        \item \textbf{Model simulation}: To evaluate the fidelity of \multiviz\ visualizations, we worked with real-world stakeholders to judge the accuracy of explanations at each fine-grained stage to determine if it helps users gain a deeper understanding of model behavior.
        \item \textbf{Model debugging}: Furthermore, we ran user studies to show \multiviz\ as a tool to highlight errors made by models and help users debug multimodal models for real-world deployment.
    \end{enumerate}
    
    \item \textbf{Estimating multimodal performance for modality selection (Chapter~\ref{chap:foundations4})}
    \begin{enumerate}
        \item \textbf{Modality selection}: We extended our analysis to quantify interactions in a semi-supervised setting with only labeled unimodal data $(x_1,y),(x_2,y)$ and naturally co-occurring multimodal data $(x_1,x_2)$ (e.g., unlabeled images and captions, video and corresponding audio) but when labeling them is time-consuming~\cite{liang2023multimodal}. We show how to approximately estimate the multimodal interactions in the unseen full distribution $(x_1,x_2,y)$, which enables practitioners to prioritize collecting data for modalities that has the most synergy with existing ones. 
        \item \textbf{Estimating performance}: Our approximation is based on lower and upper bounds for synergy: a lower bound based on the \textit{disagreement} between modality predictors, and an upper bound based on a connection to \textit{min-entropy couplings}. Lower and upper bounds on synergistic information translate to bounds on multimodal performance.
        \item \textbf{On disagreement}: Finally, we show that disagreement is a critical quality that can result in synergy between modalities, and propose a learning algorithm that captures disagreement between modalities beyond agreement that is typically done.
    \end{enumerate}

    \item \textbf{\multibench: A benchmark for real-world generalization (Chapter~\ref{chap:models1})}
    \begin{enumerate}
        \item \textbf{Real-world benchmarks}: We describe \multibench, the largest unified benchmark for multimodal representation learning~\cite{liang2021multibench}. \multibench\ provides an end-to-end machine learning pipeline that simplifies and standardizes data loading, experimental setup, and model evaluation, while ensuring reproducibility and ease of use.
        \item \textbf{Standardized building blocks}: To accompany this benchmark, we also provide a standardized implementation of $20$ core approaches in multimodal learning spanning innovations in fusion paradigms, optimization objectives, and training approaches.
        \item \textbf{Benefits of standardization}: We find that standardizing and sharing methods proposed in different research areas can improve performance on several datasets. \multibench\ also provides a better understanding of the capabilities and limitations of multimodal models.
    \end{enumerate}
       
    \item \textbf{Learning multimodal interactions across time (Chapter~\ref{chap:models2})}
    \begin{enumerate}
        \item \textbf{Temporal interactions}: To tackle heterogeneity across many different modalities, we treat modalities in their most general form as a sequence of elements, such as words in a sentence, patches in an image, frames in a video, and time steps in time-series data. This introduces a critical challenge of learning multimodal interactions across sequences, such as relating a word with a facial expression within a long video.
        \item \textbf{Recurrent cross-modal attention}: While prior work summarized temporal modalities into a single static feature before fusion, I developed a new method for \textit{fine-grained temporal fusion} to learn interactions between all elements across the sequence, such as between individual words, gestures, and vocal expressions~\cite{liang2018multimodal,chen2017multimodal}. We call this module recurrent cross-modal attention, by using attention weights to recursively learn interactions based on the current input and previous signals.
        \item \textbf{Multimodal transformers}: We extended recurrent attention into multimodal transformers that learn all interactions across sequences in parallel~\cite{tsai2019multimodal}. The multimodal transformer learns a cross-modal attention matrix to highlight related signals across time (e.g., rolling eyes and sighing). This matrix is used to learn a new representation for each modality fused with other modalities in parallel over the entire sequence, which provides huge efficiency gains when trained on modern GPUs.
    \end{enumerate}
    
    \item \textbf{Multimodal and multitask foundation models (Chapter~\ref{chap:models3})}
    \begin{enumerate}
        \item \textbf{High-modality learning}: Chapter~\ref{chap:models3} builds upon the diverse modalities and tasks provided by \multibench\ by designing methods for high-modality learning: where there are a large number of modalities partially observed for different tasks~\cite{liang2022highmmt}. This represents the most realistic setting of how humans process the multisensory world, and we believe that general-purpose AI systems will also need to be multisensory.
        
        \item \textbf{A single model for many modalities and tasks}: We propose \highmmt, a single shared high-modality model that achieves generalization over more than $10$ modalities and $15$ tasks, and transfers to new modalities and tasks.
        
        \item \textbf{Tackling extreme heterogeneity}: We've seen two extremes - full parameter sharing across everything, and no sharing at all across modality and task-specific models. A key idea in \highmmt\ is to find the optimal amount of parameter sharing balancing performance and efficiency. We do this by defining a new measure of which modalities are similar, and which modality pairs interact similarly, to inform parameter sharing.
    \end{enumerate}
    
\end{enumerate}

\section{Other Contributions}

\noindent I have also pursued the following selected research directions during my Ph.D. studies, which are excluded from this thesis. The first major direction lies in datasets and methods for learning representations from a fixed set of input modalities (i.e., without modeling generalization). To that end, I have contributed core resources and models for multimodal representation learning, especially in the application domain of modeling human communication. I have also engaged in collaborations with real-world stakeholders particularly in the healthcare and affective computing space where multimodal learning paradigms offer opportunities to learn from high-dimensional multimodal data. Finally, I have also worked on addressing the real-world societal concerns these models, such as improving their robustness, fairness, and privacy.

\begin{figure}[tbp]
\centering
\vspace{-4mm}
\includegraphics[width=0.6\linewidth]{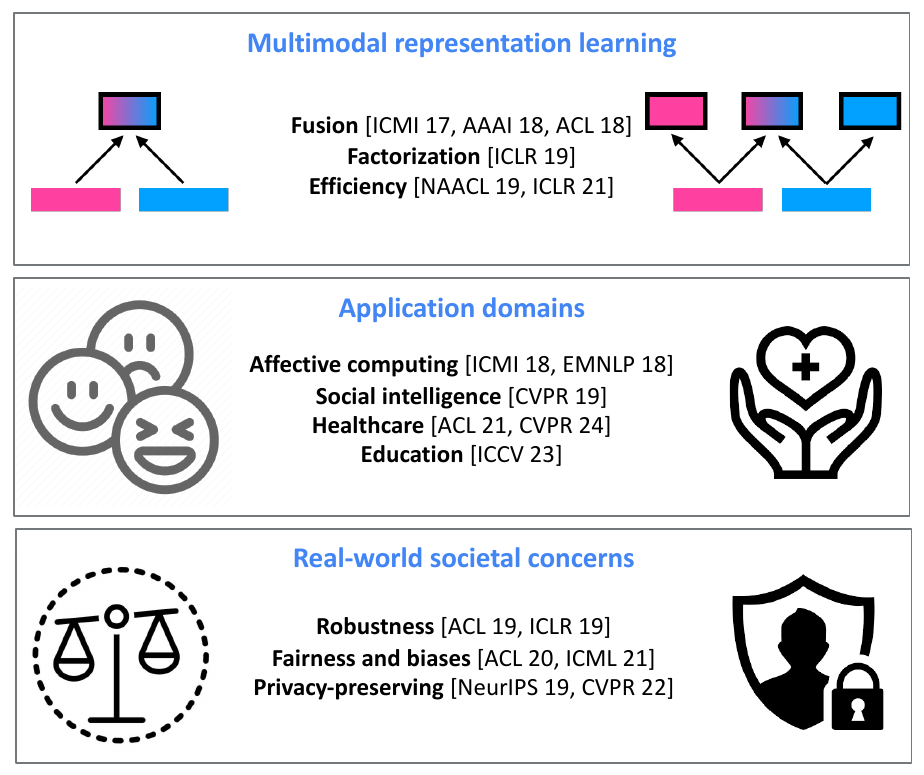}
\vspace{1mm}
\caption{I have also pursued the following directions during my Ph.D. studies: (1) new machine learning and deep learning models to learn multimodal representations (without modeling generalization), (2) collaborating with real-world stakeholders to apply these methods in affective computing, socially intelligent AI, healthcare, and education, and (3) mitigating real-world issues of deploying multimodal models in the face of real-world noise topologies, dataset biases, and privacy concerns.}
\label{fig:broader}
\vspace{-0mm}
\end{figure}

\subsection{Multimodal representation learning}

\textbf{Computational modeling of human multimodal language}: From a computational perspective, the modeling of human communication across both verbal and nonverbal behaviors enables real-world tasks such as multimodal sentiment analysis~\citep{morency2011towards}, emotion recognition~\citep{Busso2008IEMOCAP:Interactiveemotionaldyadic}, and personality traits recognition~\citep{Park:2014:CAP:2663204.2663260}. To comprehend human communication, there is a need for 1) large multimodal resources with diversity in training samples, topics, speakers, and annotations, as well as 2) powerful models for multimodal communication.

As a first step, we have worked towards addressing the lack of multimodal resources by collecting and releasing the largest dataset of multimodal sentiment and emotion recognition enabling generalizable studies of human communication. CMU-MOSEI contains $23,500$ annotated video segments from $1,000$ distinct speakers and $250$ topics. The diversity in topics, speakers, annotations, and modalities allows for generalizable studies of speaker and topic-independent features. The multimodal dataset and a general multimodal data loading framework are provided to the scientific community to encourage valuable research in human communication analysis~\citep{liang2018computational,zadeh2018multimodal}. Since then, the dataset has also been the subject of two workshop challenges in modeling human multimodal language at ACL 2018 and ACL 2020, and has been a standard benchmark dataset for the multimodal machine learning community.

\textbf{Multimodal gated fusion}: With the increasing popularity of video sharing websites such as YouTube and Facebook, multimodal sentiment analysis has received increasing attention from the scientific community~\citep{morency2011towards,perez2013utterance,wollmer2013youtube}. We develop a novel deep architecture for multimodal sentiment analysis that performs modality fusion at the word level~\cite{chen2017multimodal}. We proposed the GME-LSTM model that is composed of 2 modules. The Gated Multimodal Embedding alleviates the difficulties of fusion when there are noisy modalities. The LSTM with Temporal Attention performs word level fusion at a finer fusion resolution between input modalities and attends to the most important time steps. As a result, the GME-LSTM is able to better model the multimodal structure of speech through time and perform better sentiment comprehension. We demonstrate the effectiveness of this approach by achieving state-of-the-art sentiment classification and regression results. Qualitative analysis on our model emphasizes the importance of the Temporal Attention Layer in sentiment prediction because the additional acoustic and visual modalities are noisy. We also demonstrate the effectiveness of the Gated Multimodal Embedding layer in selectively filtering these noisy modalities out. Our results and analysis open new areas in the study of sentiment analysis in human communication and provide new models for multimodal fusion.

\textbf{Factorized multimodal representations}: Using \multibench\ and other related multimodal benchmarks enables us a deeper study of the desiderata for multimodal representations beyond discriminative performance~\citep{tsai2019learning}. While the two main pillars of research in multimodal representation learning have considered discriminative~\citep{chen2017multimodal,chaplot2017gated,frome2013devise,socher2013zero,zadeh2017tensor} and generative~\citep{ngiam2011multimodal,srivastava2012multimodal,sohn2014improved,suzuki2016joint,pham2019found} objectives individually, we demonstrate that factorizing multimodal representations into multimodal discriminative and modality-specific generative factors marries the strengths of discriminative learning of joint features across modalities that achieves state-of-the-art performance for affect analysis with controllable generation of human language based on individual factors, robustness to partially missing modalities, and interpretable local contributions from each modality during prediction. Our resulting Multimodal Factorization Model (\textsc{MFM}) defines a flexible latent variable framework balancing prediction with robustness and understandability for real-world human multimodal language.

\textbf{Efficient statistical baselines}: The constraints of real-world edge devices have created a demand for data and compute-efficient multimodal learning via simple yet strong models~\cite{strubell2019energy}. We proposed an approach based on stronger statistical baselines rather than black-box neural networks. By assuming a fully-factorized probabilistic generative model of multimodal data from a latent representation, careful model design allows us to capture expressive unimodal, bimodal, and trimodal interactions while at the same time retaining simplicity and efficiency during learning and inference~\citep{liang2019strong}. These models show strong performance on both supervised and semi-supervised multimodal prediction, as well as significant ($10$ times) speedups over neural models during inference.

\subsection{\mbox{Applications in affective computing, social intelligence, and healthcare}}

Improving the generalization and quantification of multimodal models enables a step towards real-world models capturing the benefits of multimodal data while mitigating its risks. However, tangible real-world impact requires direct collaboration with real-world stakeholders to determine their precise computational needs. During my PhD, I have had the pleasure of collaborating on the following real-world applications:

\textbf{Multimodal affective computing}: As an application-specific instantiation of multimodal learning, we studied the problem of continuous-time human affect analysis and proposed a new perspective by modeling both person-independent and person-dependent signals through insights from human psychology~\citep{liang2018ranking}. Some emotional expressions are almost universal person-independent behaviors and can be recognized directly from a video~\cite{kim2002development,dhall2011emotion}. For example, an open mouth with raised eyebrows and a loud voice is likely to be associated with surprise. However, emotions are also expressed in a person-dependent fashion with idiosyncratic behaviors where it may not be possible to directly estimate absolute emotion intensities. Instead, it would be easier to compare two video segments of the same person and judge whether there was a relative change in emotion intensities~\citep{cogprints730,Stewart2005AbsoluteIB,doi:10.1111/j.1460-9568.2008.06202.x}. For example, a person could have naturally furrowed eyebrows and we should not always interpret this as a display of anger, but rather compare two video segments to determine relative changes in anger. By designing a model combining both signals, we are able to achieve state-of-the-art audio-visual emotion recognition performance and allow for fine-grained investigation of person-independent and person-dependent behaviors.

\textbf{Social intelligence question-answering}: As intelligent systems increasingly blend into our everyday life, artificial social intelligence becomes a prominent area of research. Intelligent systems must be socially intelligent in order to comprehend human intents and maintain a rich level of interaction with humans~\cite{hunt1928measurement,kihlstrom2000social,thorndike1920intelligence,vernon1933some}. Human language offers a unique unconstrained approach to probe \textit{through questions} and reason \textit{through answers} about social situations~\cite{agrawal2017vqa,LEVELT198278}. This unconstrained approach extends previous attempts to model social intelligence through numeric supervision (e.g. sentiment and emotions labels). We introduced the Social-IQ dataset~\cite{zadeh2019social}, an unconstrained benchmark specifically designed to train and evaluate socially intelligent technologies. By providing a rich source of open-ended questions and answers, Social-IQ opens the door to explainable social intelligence. The dataset contains rigorously annotated and validated videos, questions and answers, as well as annotations for the complexity level of each question and answer. Social-IQ contains $1,250$ natural in-the-wild social situations, $7,500$ questions and $52,500$ correct and incorrect answers. Although humans can reason about social situations with very high accuracy ($95.08\%$), existing state-of-the-art computational models struggle on this task. As a result, Social-IQ brings novel challenges that will spark future research in social intelligence modeling, visual reasoning, and multimodal question answering (QA).

\textbf{Privacy-preserving mood prediction from mobile data}: Mental health conditions remain underdiagnosed even in countries with common access to advanced medical care~\citep{franklin2017risk,large2017patient}. The ability to accurately and efficiently predict mood from easily collectible data has several important implications for the early detection, intervention, and treatment of mental health disorders~\citep{glenn2014improving,nahum2018just}. One promising data source to help monitor human behavior is daily smartphone usage~\citep{prendes2018personalized}. However, care must be taken to summarize behaviors without identifying the user through personal (e.g., personally identifiable information) or protected (e.g., race, gender) attributes~\citep{kosinski2013private,liang2020fair,sharma2018toward,vsuster2017short}. Through data collected via a collaboration with psychiatrists and psychologists at the University of Oregon, Columbia University, and the University of Pittsburgh, we study behavioral markers of daily mood using a recent dataset of mobile behaviors from adolescent populations at high risk of suicidal behaviors~\cite{liang2021learning}. Using computational models, we find that language and multimodal representations of mobile \textit{typed text} (spanning typed characters, words, keystroke timings, and app usage) are predictive of daily mood. However, we find that models trained to predict mood often also capture private user identities in their intermediate representations. To tackle this problem, we evaluate approaches that obfuscate user identity while remaining predictive. By combining multimodal representations with privacy-preserving learning, we are able to push forward the performance-privacy frontier.

\subsection{Real-world robustness, fairness, and privacy}

Finally, the third major direction studies the real-world concerns of deploying multimodal models in the face of real-world noise topologies, dataset biases, and privacy concerns.

\textbf{Robustness to noisy modalities}: Different modalities often display different noise topologies, and real-world multimodal signals possibly suffer from missing or noisy data in at least one of the modalities~\cite{baltruvsaitis2018multimodal,ding2014latent,lee2020detect}. Human-centric data is also often imperfect due to personal idiosyncrasies which affect the contribution of certain modalities during social interactions~\cite{golder2007rhythms,schadenberg2021predictable}. For example, multimodal dialogue systems trained on acted TV shows are susceptible to poor performance when deployed in the real world where users might be less expressive in using facial gestures. This calls for robust models that can still make accurate predictions despite only having access to a (possibly noisy) subset of signals.

As a step towards robustness, we propose a tensor representation learning method to deal with noisy modalities in time-series data (e.g., text, videos, audio)~\citep{liang2019tensor}. This method is based on the observation that multimodal time series data often exhibits correlations across time and modalities which lead to low-rank multimodal representations~\citep{Hidaka:2010:AMT:1891903.1891968,Lakshmanan2015ExtractingLL,Yang_2017_CVPR}. However, the presence of noise or incomplete values breaks these correlations and results in tensor representations of higher rank. Regularizing the rank of tensor representations therefore provides a denoising effect which achieves strong results across various levels of imperfection. We show how to integrate an upper-bound of tensor rank minimization as a simple regularizer for training in the presence of imperfect data, thereby combining the strength of temporal non-linear transformations of multimodal data with principled regularization on tensor structures. Through experiments on multimodal video data, our results back up our intuitions that imperfect data increases tensor rank and demonstrates strong results across various levels of imperfection.

\textbf{Learning fair sentence representations}: To safely deploy human-centric multimodal models in real-world scenarios such as healthcare, legal systems, and social science, it is also necessary to recognize the role they play in shaping social biases and stereotypes. Previous work has revealed the presence of \textit{representational biases} in widely used word embeddings - harmful biases resulting from stereotyping that propagate negative generalizations involving gender, race, religion, and other social constructs~\cite{blodgett2020language,gehman2020realtoxicityprompts,hendrycks2020aligning,nadeem2020stereoset,sap2020social,sheng2019woman}. While some methods were proposed to debias these word-level embeddings~\cite{bolukbasi2016man,manzini2019black}, there is a need to perform debiasing at the sentence-level given the recent shift towards new contextualized sentence representations such as ELMo~\cite{peters-etal-2018-deep} and BERT~\cite{devlin2019bert} which have become core components in both real-world language~\cite{Alsentzer2019PubliclyAC,Huang2019ClinicalBERTMC,wang2019multi} and multimodal prediction systems~\cite{li2019visualbert,lu2019vilbert}. We investigated the presence of social biases in sentence-level representations and proposed a new method, \sentdebias, to reduce these biases~\cite{liang2020fair}. We show that \sentdebias\ is effective in reducing biases from the geometry of contextual representation spaces, and at the same time, preserves performance on sentence-level downstream NLP tasks such as sentiment analysis, linguistic acceptability, and natural language understanding.

\textbf{Mitigating social biases in language models}: In addition to sentence representations deployed primarily for discriminative tasks, large-scale pretrained language models (LMs) have also become widely-deployed for generative applications such as text generation~\cite{radford2019language}, dialog systems~\cite{zhang2019dialogpt}, recommendation systems~\cite{shakespeare2020exploring}, and search engines~\cite{baeza2016data,otterbacher2018investigating}. Recent work has found that these language models can potentially generate text propagating negative generalizations about particular social groups~\cite{nadeem2020stereoset}, language that is denigrating to particular social groups~\cite{sheng2019woman}, and toxic speech~\cite{gehman2020realtoxicityprompts}, while at the same time also being unable to reason about human-aligned values such as ethics~\cite{hendrycks2020aligning}, social bias implications~\cite{sap2020social}, and allocational harms across social groups~\cite{liu2021mitigating}. As a step towards improving the fairness of LMs, we carefully defined several sources of representational biases before proposing new benchmarks and metrics to measure them~\cite{liang2021towards}. With these tools, we propose \ainlp, an approach towards post-hoc debiasing of large pretrained LMs. The key to our approach lies in dynamically finding bias-sensitive tokens rather than relying on a predefined set of bias-sensitive words that are common in existing literature~\cite{bolukbasi2016man}. Our empirical results and human evaluation on large language models such as GPT-2 demonstrate effectiveness in mitigating bias while retaining crucial context information for high-fidelity text generation, thereby pushing forward the performance-fairness Pareto frontier. These steps are critical towards improving the safety of language and multimodal models.

\textbf{Privacy-preserving federated learning}: More broadly, federated learning is a method of training models on private data distributed over multiple devices~\cite{li2020federated,DBLP:journals/corr/McMahanMRA16,DBLP:journals/corr/abs-1902-01046,DBLP:journals/corr/SmithCST17}. To keep device data private, a single global model is trained by only communicating parameters and updates which poses scalability challenges for large models~\cite{Nilsson:2018:PEF:3286490.3286559}. Furthermore, current approaches use the same model architecture across all local models and the global aggregated model, which causes federated learning to struggle with data heterogeneity across devices~\cite{li2020federated,hsu2019measuring,zhao2018federated}. This is made worse when each device contains multimodal data sources that are used unequally across users~\cite{DBLP:journals/corr/abs-1902-00146}. To this end, we propose a new federated learning algorithm, Local Global Federated Averaging (\textsc{LG-FedAvg}), that jointly learns compact \textit{local representations} on each device and a global model across all devices~\cite{liang2019think}. As a result, the global model can be smaller since it only operates on local representations, reducing the number of communicated parameters. Furthermore, well-designed local models enable learning of personalized representations for user-specific behavior modeling while enjoying the benefit of global model learning across many users' data. Theoretically, we provide a generalization analysis which shows that a combination of local and global models reduces both variance in the data as well as variance across device distributions. Empirically, we demonstrate that local models enable communication-efficient training while retaining performance. We also evaluate on the task of personalized mood prediction from real-world mobile data where privacy is key. Finally, we show that local models handle heterogeneous data from new devices, and learn fair representations that obfuscate protected attributes such as race, age, and gender~\cite{bolukbasi2016man}.

\chapter{Literature Survey and Taxonomy of Multimodal Challenges}
\label{chap:background}
\vspace{-2mm}
\section{Introduction}
\vspace{-1mm}

It has always been a grand goal of artificial intelligence to develop computer agents with intelligent capabilities such as understanding, reasoning, and learning through multimodal experiences and data, similar to how humans perceive and interact with our world using multiple sensory modalities. With recent advances in embodied autonomous agents~\citep{brodeur2017home,savva2019habitat}, self-driving cars~\citep{xiao2020multimodal}, image and video understanding~\citep{alayrac2022flamingo,sun2019videobert}, image and video generation~\citep{ramesh2021zero,singer2022make}, and multisensor fusion in application domain such as robotics~\citep{lee2019making,marge2022spoken} and healthcare~\citep{MIMIC,liang2021multibench}, we are now closer than ever to intelligent agents that can integrate and learn from many sensory modalities. This vibrant multi-disciplinary research field of multimodal machine learning brings unique challenges given the heterogeneity of the data and the interconnections often found between modalities, and has widespread applications in multimedia~\citep{1667983}, affective computing~\citep{PORIA201798}, robotics~\citep{kirchner2019embedded,lee2019making}, human-computer interaction~\citep{obrenovic2004modeling,sharma2002toward}, and healthcare~\citep{cai2019survey,muhammad2021comprehensive}.

\begin{figure}[t]
\centering
\vspace{-0mm}
\includegraphics[width=0.8\linewidth]{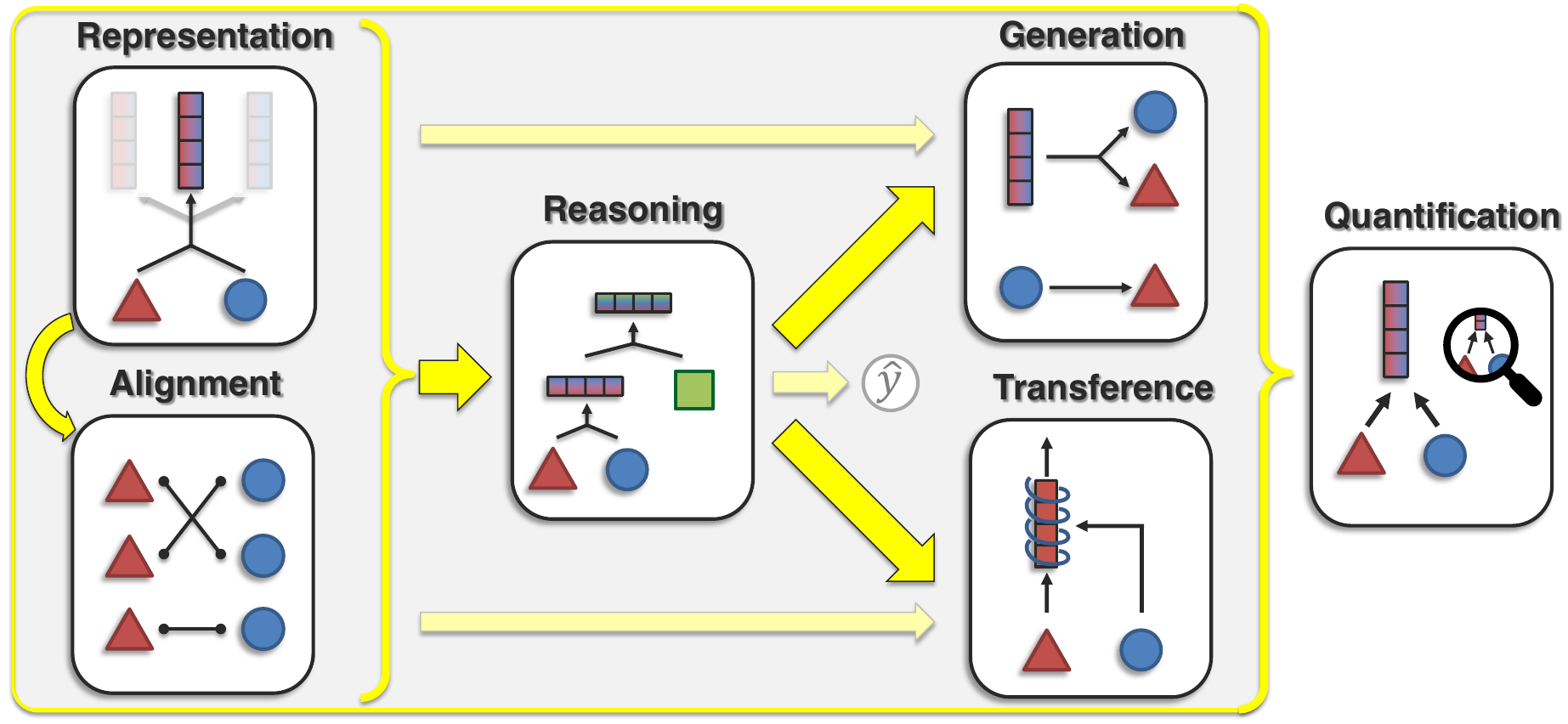}
\caption{Core research challenges in multimodal learning: Every multimodal problem typically requires tackling representation and alignment: (1) \textit{Representation} studies how to summarize multimodal data to reflect the heterogeneity and interconnections between individual modality elements, before (2) \textit{alignment} captures the connections and interactions between multiple local elements according to their structure. After representation and alignment comes (3) \textit{reasoning}, which aims to combine the information from multimodal evidence in a principled way that respects the structure of the problem to give more robust and interpretable predictions.  While most systems aim to predict the label $y$, there are also cases where the goal is (4) \textit{generation}, to learn a generative process to produce raw modalities that reflect cross-modal interactions, structure, and coherence, or (5) \textit{transference}, to transfer information from high-resource modalities to low-resource ones and their representations. Finally, (6) \textit{quantification} revisits the previous challenges to give deeper empirical and theoretical understanding of modality heterogeneity, interconnections, and the learning process.}
\label{fig:intro_challenges}
\vspace{-2mm}
\end{figure}

However, the rate of progress in multimodal research has made it difficult to identify the common themes underlying historical and recent work, as well as the key open questions in the field. By synthesizing a broad range of research, this paper is designed to provide an overview of the methodological, computational, and theoretical foundations of multimodal machine learning. We begin by defining (in \S\ref{sec:definitions}) three key principles that have driven technical challenges and innovations: (1) modalities are \textit{heterogeneous} because the information present often shows diverse qualities, structures, and representations, (2) modalities are \textit{connected} since they are often related and share commonalities, and (3) modalities \textit{interact} to give rise to new information when used for task inference. Building upon these definitions, we propose a new taxonomy of six core challenges in multimodal learning: \textit{representation}, \textit{alignment}, \textit{reasoning}, \textit{generation}, \textit{transference}, and \textit{quantification} (see Figure~\ref{fig:intro_challenges}). These core multimodal challenges are understudied in conventional unimodal machine learning and need to be tackled in order to progress the field forward:
\begin{enumerate}[noitemsep,topsep=0pt,nosep,leftmargin=*,parsep=0pt,partopsep=0pt]
    \item \textbf{Representation (\S\ref{sec:representation}):} Can we learn representations that reflect heterogeneity and interconnections between modality elements? We will cover approaches for (1) \textit{representation fusion}: integrating information from two or more modalities to capture cross-modal interactions, (2) \textit{representation coordination}: interchanging cross-modal information to keep the same number of representations but improve multimodal contextualization, and (3) \textit{representation fission}: creating a larger set of disjoint representations that reflects knowledge about internal structure such as data clustering or factorization.

    \item \textbf{Alignment (\S\ref{sec:alignment}):} How can we identify the connections and interactions between modality elements? Alignment is challenging since it may depend on long-range dependencies, involves ambiguous segmentation (e.g., words or utterances), and could be either one-to-one, many-to-many, or not exist at all. We cover (1) \textit{discrete alignment}: identifying connections between discrete elements across modalities, (2) \textit{continuous alignment}: modeling alignment between continuous modality signals with ambiguous segmentation, and (3) \textit{contextualized representations}: learning better representations by capturing cross-modal interactions between elements.

    \item \textbf{Reasoning (\S\ref{sec:reasoning})} is defined as composing knowledge, usually through multiple inferential steps, that exploits the problem structure for a specific task. Reasoning involves (1) \textit{modeling the structure} over which composition occurs, (2) the \textit{intermediate concepts} in the composition process, (3) understanding the \textit{inference paradigm} of more abstract concepts, and (4) leveraging large-scale \textit{external knowledge} in the study of structure, concepts, and inference.

    \item \textbf{Generation (\S\ref{sec:generation})} involves learning a generative process to produce raw modalities. We categorize its subchallenges into (1) \textit{summarization}: summarizing multimodal data to reduce information content while highlighting the most salient parts of the input, (2) \textit{translation}: translating from one modality to another and keeping information content while being consistent with cross-modal connections, and (3) \textit{creation}: simultaneously generating multiple modalities to increase information content while maintaining coherence within and across modalities.
    
    \item \textbf{Transference (\S\ref{sec:transference})} aims to transfer knowledge between modalities, usually to help the target modality, which may be noisy or with limited resources. Transference is exemplified by (1) \textit{cross-modal transfer}: adapting models to tasks involving the primary modality, (2) \textit{co-learning}: transferring information from secondary to primary modalities by sharing representation spaces between both modalities, and (3) \textit{model induction}: keeping individual unimodal models separate but transferring information across these models.
    
    \item \textbf{Quantification (\S\ref{sec:quantification}):} The sixth and final challenge involves empirical and theoretical studies to better understand (1) the dimensions of \textit{heterogeneity} in multimodal datasets and how they subsequently influence modeling and learning, (2) the presence and type of modality \textit{connections and interactions} in multimodal datasets and captured by trained models, and (3) the \textit{learning} and optimization challenges involved with heterogeneous data.
\end{enumerate}

Finally, we conclude this paper with a long-term perspective on multimodal learning by motivating open research questions identified by this taxonomy. This survey was also presented by the authors in a visual medium through tutorials at \href{https://cmu-multicomp-lab.github.io/mmml-tutorial/cvpr2022/}{CVPR 2022} and \href{https://cmu-multicomp-lab.github.io/mmml-tutorial/naacl2022/}{NAACL 2022}, as well as courses \href{https://cmu-multicomp-lab.github.io/mmml-course/fall2022/}{11-777 Multimodal Machine Learning} and \href{https://cmu-multicomp-lab.github.io/adv-mmml-course/spring2023/}{11-877 Advanced Topics in Multimodal Machine Learning} at CMU. The reader is encouraged to refer to these public video recordings, additional readings, and discussion probes for more mathematical depth on certain topics, visual intuitions and explanations, and more open research questions in multimodal learning.

This paper is designed to complement other surveys that belong broadly to the study of multiple modalities or views: multi-view learning~\citep{nguyen2020multiview,sun2013survey,yan2021deep} is concerned with settings where different views (e.g., camera views) typically provide overlapping (redundant) information but not the other core challenges we cover, surveys on multimodal foundation models~\citep{du2022survey,gan2022vision} go into detail on tackling representation, fusion, and alignment using large-scale pretraining but do not cover other core challenges, and several application-oriented surveys in vision-language models~\citep{uppal2022multimodal}, language and reinforcement learning~\citep{luketina2019survey}, multimedia analysis~\citep{atrey2010multimodal}, and multimodal human-computer interaction~\citep{jaimes2007multimodal} discuss specific multimodal challenges faced in these applications. This survey presents a telescoping overview suitable as a starting point for researchers who can then diver deeper into methodology or application-specific research areas.

\vspace{-2mm}
\subsection{Key modalities and application domains}
\vspace{-1mm}

In this subsection, we first contextualize our subsequent discussion of multimodal machine learning by listing some key modalities of interest, standard multimodal datasets and toolkits, and major applications of multimodal learning in the real world.

\textbf{Affective computing} studies the perception of human affective states such as emotions, sentiment, and personalities from multimodal human communication: spoken language, facial expressions and gestures, body language, vocal expressions, and prosody~\citep{picard2000affective}. Some commonly studied tasks involve predicting sentiment~\citep{soleymani2017survey,zadeh2016mosi}, emotions~\citep{zadeh2018multimodal}, humor~\cite{hasan2019ur}, and sarcasm~\citep{castro2019towards} from multimodal videos of social interactions.

\textbf{Healthcare:} Machine learning can help integrate complementary medical signals from lab tests, imaging reports, patient-doctor conversations, and multi-omics data to assist doctors in the clinical process~\cite{acosta2022multimodal,amisha2019overview,lipkova2022artificial}. Multimodal physiological signals recorded regularly from smartphones and wearable devices can also provide non-invasive health monitoring~\cite{de2015multimodal,garcia2018mental,liang2021learning}. Public datasets include MIMIC~\citep{MIMIC} with patient tabular data, medical reports, and medical sensor readings, question answering on pathology~\cite{he2020pathvqa} and radiology~\cite{lau2018dataset} images, and multi-omics data integration~\cite{tran2021openomics}.

\textbf{Robotics} systems are often equipped with multiple sensors to aid in robust decision-making for real-world physical tasks such as grasping, cleaning, and delivery. These sensors can include vision (RGB and depth), force, and proprioception~\citep{lee2019making}. These multi-sensor robots have been successfully applied in haptic~\citep{pai2005multisensory,seminara2019active} and surgical robots~\citep{abiri2019multi,bethea2004application}. More generally, language~\citep{luketina2019survey} and audio~\citep{dean2020see} have also emerged as useful signals for robot learning.

\textbf{Interactive agents} in the virtual world can assist humans in multimedia web tasks and computer tasks~\citep{furuta2023multimodal} as well as in the social world through virtual agents~\citep{pelachaud2009modelling,pelachaud2021multimodal}. These agents need to understand human commands and behaviors, process various forms of visual, tabular, and multimedia content, use external web tools and APIs, and interact in multi-step decision-making tasks. Webshop~\citep{yao2022webshop} and WebAreana~\citep{zhou2023webarena} are recent environments testing the capabilities of AI agents in navigating image and text content to solve web tasks.

\textbf{Multimedia} data spanning text, images, videos, audio, and music is abundant on the internet and has fueled a significant body of multimodal research~\cite{atrey2010multimodal}, such as classification~\citep{zhang2001audio}, retrieval~\citep{rasiwasia2010new}, and recommendation~\citep{mitrovic2010features,rui1999image,zhen2019deep} of multimedia content, image and video question answering~\citep{agrawal2017vqa,krishna2017visual,lei2018tvqa} and captioning~\citep{drossos2020clotho,vinyals2016show}), multimedia and entertainment content description~\citep{shah2017multimodal} (including movies~\citep{arevalo2017gated}, memes~\cite{chhavi2020memotion,kiela2020hateful}, and cartoons~\citep{hessel2022androids}), and more recently in automatic creation of text~\citep{zhang2023survey}, images~\citep{rombach2022high}, videos~\citep{wu2023tune}, music~\citep{agostinelli2023musiclm}, and more.

\textbf{Human-computer interaction} has sought to endow computers with multimodal capabilities to provide more natural, powerful, and compelling interactive user experiences~\citep{turk2014multimodal}. These systems have leveraged speech, touch, vision, gestures, affective states~\citep{pantic2003toward} and affordable wearable and mobile sensors~\citep{jaimes2007multimodal,oviatt1999ten,turk2014multimodal}. Public datasets have enabled the study of multimodal user interfaces~\cite{leiva2020enrico,wang2021screen2words}, speech and gesture interactions~\citep{escalera2013multi}, and human sensing~\citep{chatzitofis2020human4d,delpreto2022actionsense,schmidt2018introducing}.

\textbf{Science and environment}: Deepening our knowledge of the natural sciences and physical environments can bring about impactful changes in scientific discovery, sustainability, and conservation. This requires processing modalities such as chemical molecules~\citep{su2022molecular}, protein structures~\citep{zhang2019multimodal}, satellite images~\cite{cheng2017remote,yang2010bag}, remote sensing~\citep{hong2020more,li2022deep}, wildlife movement~\citep{lopes2008development}, scientific diagrams and texts~\cite{lu2022learn}, and various physical sensors~\citep{mo2023multiiot}.

\textbf{Education}: AI can broaden access to educational content by digitizing lecture slides and videos, creating personalized tutors, and designing interactive learning curricula. It introduces challenges in processing recorded lecture slides and videos~\cite{lee2022multimodal}, and modeling student learning via asked questions, spoken feedback and non-verbal gestures~\cite{chango2022review,sumer2021multimodal,xu2019mutla}.

\vspace{-2mm}
\section{Foundational Principles in Multimodal Research}
\label{sec:definitions}
\vspace{-0mm}

A \textit{modality} refers to a way in which a natural phenomenon is perceived or expressed. For example, modalities include speech and audio recorded through microphones, images and videos captured via cameras, and force and vibrations captured via haptic sensors.
Modalities can be placed along a spectrum from \textit{raw} to \textit{abstract}: raw modalities are those more closely detected from a sensor, such as speech recordings from a microphone or images captured by a camera. Abstract modalities are those farther away from sensors, such as language extracted from speech recordings, objects detected from images, or even abstract concepts like sentiment intensity and object categories.

\textit{Multimodal} refers to situations where multiple modalities are involved. From a research perspective, multimodal entails the computational study of \textit{heterogeneous} and \textit{interconnected} modalities. Firstly, modalities are \textit{heterogeneous} because the information present in different modalities will often show diverse qualities, structures, and representations. Secondly, these modalities are not independent entities but rather share \textit{connections} due to complementary information. Thirdly, modalities \textit{interact} in different ways when they are integrated for a task. We expand on these three foundational principles of multimodal research in the following subsections.

\begin{figure}[t]
\centering
\includegraphics[width=0.8\linewidth]{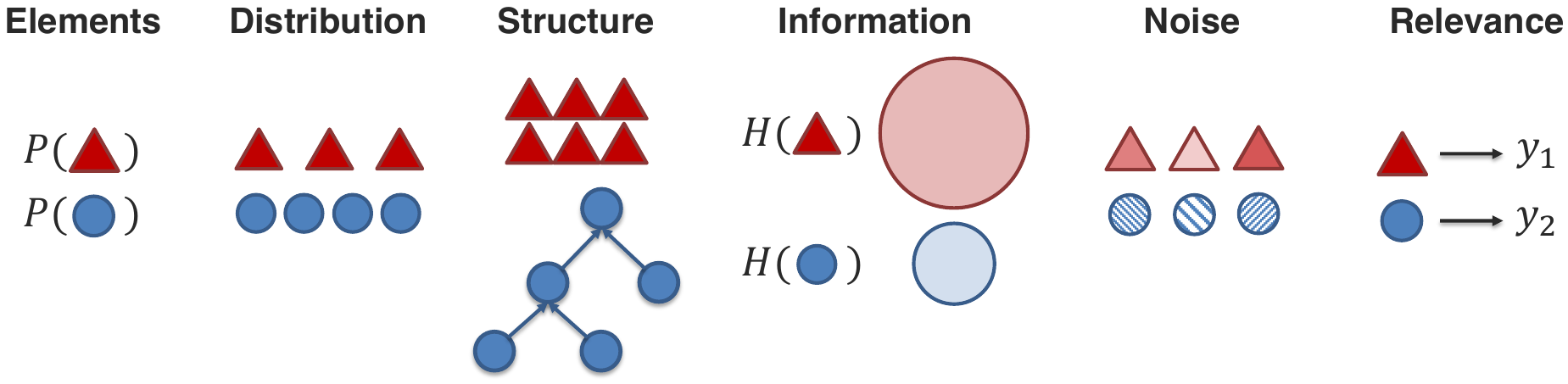}
\caption{The information present in different modalities will often show diverse qualities, structures, and representations. \textbf{Dimensions of heterogeneity} can be measured via differences in individual elements and their distribution, the structure of elements, as well as modality information, noise, and task relevance.}
\label{fig:hetero}
\vspace{-2mm}
\end{figure}

\vspace{-2mm}
\subsection{Principle 1: Modalities are heterogeneous}
\vspace{-1mm}

The principle of heterogeneity reflects the observation that the information present in different modalities will often show diverse qualities, structures, and representations. Heterogeneity should be seen as a spectrum: two images from the same camera that capture the same view modulo camera wear and tear are closer to homogeneous, two different languages that capture the same meaning but from different language families are slightly heterogeneous, language and vision are even more heterogeneous, and so on.
In this section, we present a non-exhaustive list of dimensions of heterogeneity (see Figure~\ref{fig:hetero} for an illustration). These dimensions are complementary and may overlap; each multimodal problem likely involves heterogeneity in multiple dimensions.

\begin{enumerate}[noitemsep,topsep=0pt,nosep,leftmargin=*,parsep=0pt,partopsep=0pt]
    \item \textbf{Element representation}: Each modality is typically comprised of a set of elements - the most basic unit of data which cannot (or rather, the user chooses to not) be broken down into further units~\citep{barthes1977image,liang2022brainish}. For example, typed text is recorded via a set of characters, videos are recorded via a set of frames, and graphs are recorded via a set of nodes and edges. What are the basic elements present in each modality, and how can we represent them? Formally, this dimension measures heterogeneity in the sample space or representation space of modality elements.

    \item \textbf{Distribution} refers to the frequency and likelihood of modality elements. Elements typically follow a unique distribution, with words in a linguistic corpus following Zipf's Law~\citep{zipf2016human} as an example. Distribution heterogeneity refers to the differences in frequencies and likelihoods of elements, such as different frequencies in recorded signals and the density of elements.

    \item \textbf{Structure}: Natural data exhibits structure in the way individual elements are composed to form entire modalities~\citep{bronstein2021geometric}. For example, images exhibit spatial structure across objects, language is hierarchically composed of words, and signals exhibit temporal structure across time. Structure heterogeneity refers to differences in this underlying structure.
    
    \item \textbf{Information} measures the total information content present in each modality. Subsequently, information heterogeneity measures the differences in information content across modalities, which could be formally measured by information theoretic metrics~\citep{shannon1948mathematical}.

    \item \textbf{Noise}: Noise can be introduced at several levels across naturally occurring data and also during the data recording process. Natural data noise includes occlusions, imperfections in human-generated data (e.g., imperfect keyboard typing or unclear speech), or data ambiguity due to sensor failures~\citep{liang2021multibench}. Noise heterogeneity measures differences in noise distributions across modalities, as well as differences in signal-to-noise ratio.

    \item \textbf{Relevance}: Finally, each modality shows different relevance toward specific tasks and contexts - certain modalities may be more useful for certain tasks than others~\citep{gat2021perceptual}. Task relevance describes how modalities can be used for inference, while context relevance describes how modalities are contextualized with other modalities.
\end{enumerate}
It is useful to take these dimensions of heterogeneity into account when studying both unimodal and multimodal data. In the unimodal case, specialized encoders are typically designed to capture these unique characteristics in each modality~\citep{bronstein2021geometric}. In the multimodal case, modeling heterogeneity is useful when learning representations and capturing alignment~\citep{zamir2018taskonomy}, and is a key subchallenge in quantifying multimodal models~\citep{liang2022highmmt}.

\vspace{-2mm}
\subsection{Principle 2: Modalities are connected}
\vspace{-1mm}

\begin{figure}[t]
\centering
\includegraphics[width=0.65\linewidth]{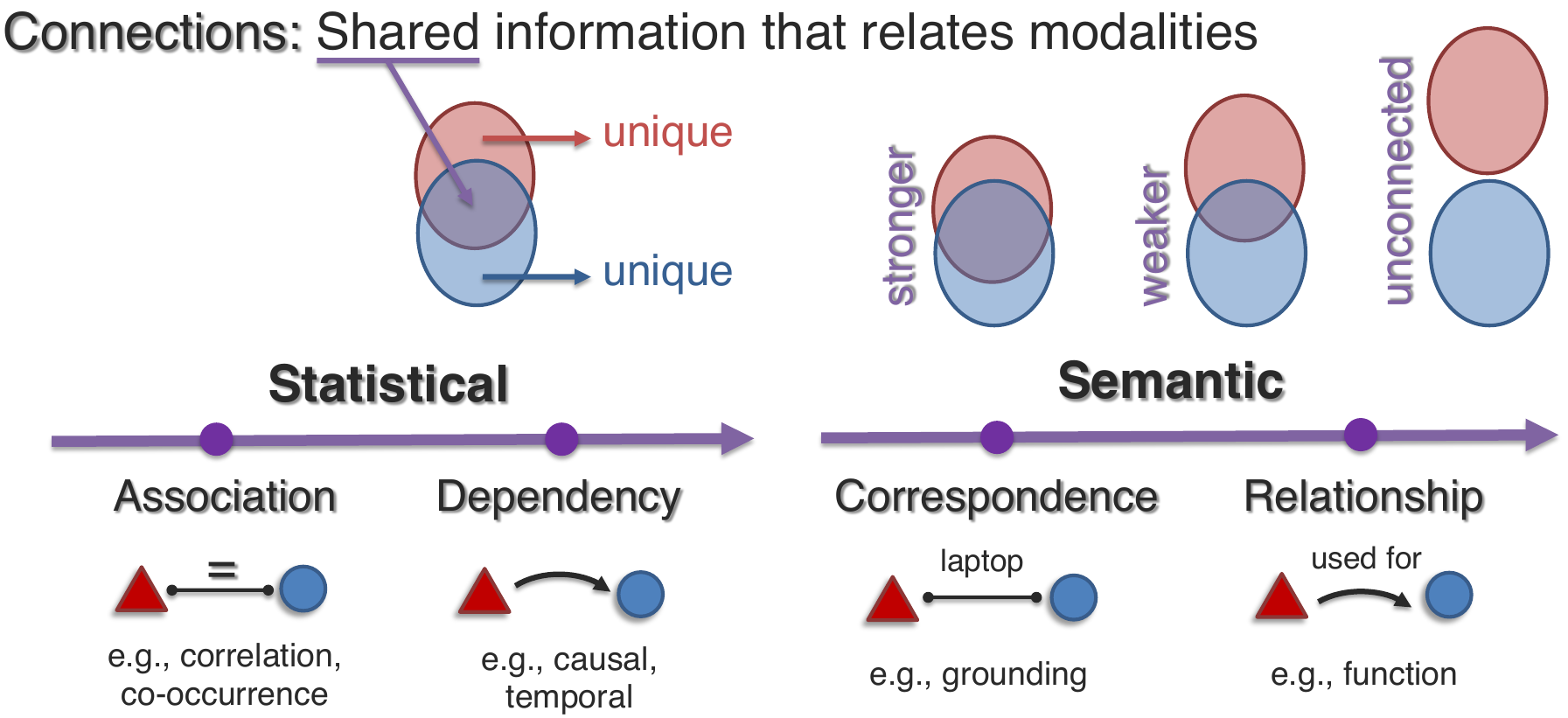}
\caption{\textbf{Modality connections} describe how modalities are related and share commonalities, such as correspondences between the same concept in language and images or dependencies across spatial and temporal dimensions. Connections can be studied through both statistical and semantic perspectives.}
\label{fig:connections}
\vspace{-2mm}
\end{figure}

Although modalities are heterogeneous, they are often connected due to shared complementary information. The presence of \textit{shared} information is often in contrast to \textit{unique} information that exists solely in a single modality~\citep{williams2010nonnegative}. Modality connections describe the extent and dimensions to which information can be shared across modalities. When reasoning about the connections in multimodal data, it is helpful to think about both bottom-up (statistical) and top-down (semantic) approaches (see Figure~\ref{fig:connections}). From a statistical data-driven perspective, connections are identified from distributional patterns in multimodal data, while semantic approaches define connections based on our domain knowledge about how modalities share and contain unique information.

\begin{enumerate}[noitemsep,topsep=0pt,nosep,leftmargin=*,parsep=0pt,partopsep=0pt]
    \item \textbf{Statistical association} exists when the values of one variable relate to the values of another. For example, two elements may co-occur with each other, resulting in a higher frequency of both occurring at the same time. Statistically, this could lead to correlation - the degree to which elements are linearly related, or other non-linear associations. From a data-driven perspective, discovering which elements are associated with each other is important for modeling the joint distributions across modalities during multimodal representation and alignment~\citep{tian2020makes}.

    \item \textbf{Statistical dependence} goes deeper than association and requires an understanding of the exact type of statistical dependency between two elements. For example, is there a causal dependency from one element to another, or an underlying confounder causing both elements to be present at the same time? Other forms of dependencies could be spatial or temporal: one element occurring above the other, or after the other. Typically, while statistical association can be estimated purely from data, understanding the nature of statistical dependence requires some knowledge of the elements and their underlying relationships~\cite{nickel2015review,turney2005corpus}.
    
    \item \textbf{Semantic correspondence} can be seen as the problem of ascertaining which elements in one modality share the same semantic meaning as elements in another modality~\citep{otto2020characterization}. Identifying correspondences is fundamental in many problems related to language grounding~\cite{chandu2021grounding}, translation and retrieval~\citep{plummer2015flickr30k}, and cross-modal alignment~\citep{tan2019lxmert}.
    
    \item \textbf{Semantic relations}: Finally, semantic relations generalize semantic correspondences: instead of modality elements sharing the same exact meaning, semantic relations include an attribute describing the exact nature of the relationship between two modality elements, such as semantic, logical, causal, or functional relations. Identifying these semantically related connections is important for higher-order reasoning~\citep{marsh2003taxonomy,barthes1977image}.
\end{enumerate}

\vspace{-2mm}
\subsection{Principle 3: Modalities interact}
\vspace{-1mm}

\begin{figure}[t]
\centering
\vspace{-0mm}
\includegraphics[width=0.7\linewidth]{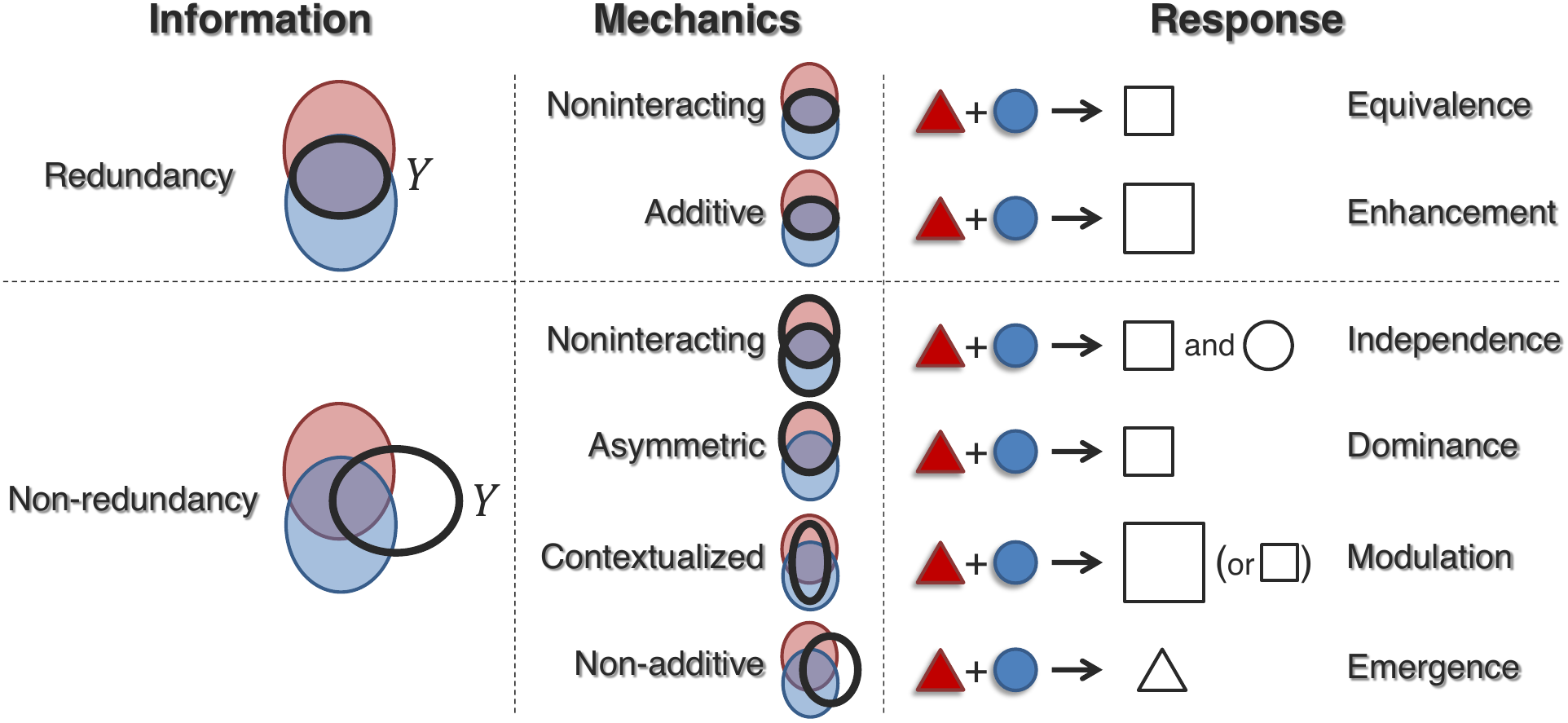}
\vspace{-2mm}
\caption{\textbf{Several dimensions of modality interactions}: (1) Interaction information studies whether common redundant information or unique non-redundant information is involved in interactions; (2) interaction mechanics study the manner in which interaction occurs, and (3) interaction response studies how the inferred task changes in the presence of multiple modalities.}
\label{fig:interactions}
\vspace{-4mm}
\end{figure}

Modality interactions study how modality elements interact to give rise to new information when integrated together for task \textit{inference}.
We note an important difference between modality connections and interactions: connections exist within multimodal data itself, whereas interactions only arise when modalities are integrated and processed together to bring a new response. In Figure~\ref{fig:interactions}, we provide a high-level illustration of some dimensions of interactions that can exist.

\begin{enumerate}[noitemsep,topsep=0pt,nosep,leftmargin=*,parsep=0pt,partopsep=0pt]
    \item \textbf{Interaction information} investigates the type of connected information that is involved in an interaction. When an interaction involves shared information common to both modalities, the interaction is \textit{redundant}, while a \textit{non-redundant} interaction is one that does not solely rely on shared information, and instead relies on different ratios of shared, unique, or possibly even synergistic information~\citep{liang2023quantifying,williams2010nonnegative}.

    \item \textbf{Interaction mechanics} are the functional operators involved when integrating modality elements for task inference. For example, interactions can be expressed as statistically additive, non-additive, and non-linear forms~\citep{jayakumar2020multiplicative}, as well as from a semantic perspective where two elements interact through a logical, causal, or temporal operation~\citep{unsworth2014multimodality}.

    \item \textbf{Interaction response} studies how the inferred response changes in the presence of multiple modalities. For example, through sub-dividing redundant interactions, we can say that two modalities create an equivalence response if the multimodal response is the same as responses from either modality, or enhancement if the multimodal response displays higher confidence. On the other hand, non-redundant interactions such as modulation or emergence happen when there exist different multimodal versus unimodal responses~\cite{partan1999communication}.
\end{enumerate}

\vspace{-2mm}
\subsection{Core technical challenges}
\vspace{-1mm}

\begin{table*}[t]
\fontsize{9}{11}\selectfont
\setlength\tabcolsep{0.0pt}
\vspace{-0mm}
\caption{This table summarizes our taxonomy of $6$ core challenges in multimodal machine learning, their subchallenges, categories of corresponding approaches, and representative examples. We believe that this taxonomy can help to catalog rapid progress in this field and better identify the open research questions.}
\centering
\footnotesize
\vspace{0mm}
\begin{tabular}{c|c|ccccccc}
\hline \hline
Challenge & Subchallenge & Approaches \& key examples \\
\hline

\multirow{3}{*}{Representation (\ref{sec:representation})} & \multirow{1}{*}{Fusion (\ref{sec:representation1})} & Abstract~\cite{jayakumar2020multiplicative,zadeh2017tensor} \& raw~\cite{barnum2020benefits,rajagopalan2016extending} fusion \\
\Xcline{2-5}{0.5\arrayrulewidth}
& \multirow{1}{*}{Coordination (\ref{sec:representation2})} & Strong~\cite{frome2013devise,radford2021learning} \& partial~\cite{vendrov2015order,zhang2016learning} coordination \\
\Xcline{2-5}{0.5\arrayrulewidth}
& \multirow{1}{*}{Fission (\ref{sec:representation3})} & Modality-level~\cite{hessel2020does,tsai2019learning} \& fine-grained~\cite{abavisani2018deep,chen2021multimodal} fission \\
\hline \hline

\multirow{3}{*}{Alignment (\ref{sec:alignment})} & \multirow{1}{*}{Discrete connections (\ref{sec:alignment1})} & Local~\cite{cirik2018visual,hsu2018unsupervised} \& global~\cite{li2022clip} alignment \\
\Xcline{2-5}{0.5\arrayrulewidth}
& \multirow{1}{*}{Continuous alignment (\ref{sec:alignment2})} & Warping~\cite{hu2019multimodal,haresh2021learning} \& segmentation~\cite{sun2019videobert} \\
\Xcline{2-5}{0.5\arrayrulewidth}
& \multirow{1}{*}{Contextualization (\ref{sec:alignment3})} & Joint~\cite{li2019visualbert}, cross-modal~\cite{hendricks2021decoupling,lu2019vilbert} \& graphical~\cite{yang2021mtag} &  \\
\hline \hline

\multirow{4}{*}{Reasoning (\ref{sec:reasoning})} & \multirow{1}{*}{Structure modeling (\ref{sec:reasoning1})} & Hierarchical~\cite{andreas2016neural}, temporal~\cite{xiong2016dynamic}, interactive~\cite{luketina2019survey}, discovery~\cite{perez2019mfas} \\
\Xcline{2-5}{0.5\arrayrulewidth}
& \multirow{1}{*}{Intermediate concepts (\ref{sec:reasoning2})} & Attention~\cite{xu2015show}, discrete symbols~\cite{amizadeh2020neuro,vedantam2019probabilistic}, language~\cite{hudson2019learning,zeng2022socratic} \\
\Xcline{2-5}{0.5\arrayrulewidth}
& \multirow{1}{*}{Inference paradigm (\ref{sec:reasoning3})} & Logical~\cite{gokhale2020vqa,suzuki2019multimodal} \& causal~\cite{agarwal2020towards,niu2021counterfactual,yi2019clevrer} \\
\Xcline{2-5}{0.5\arrayrulewidth}
& \multirow{1}{*}{External knowledge (\ref{sec:reasoning4})} & Knowledge graphs~\cite{gui2021kat,zhu2015building} \& commonsense~\cite{park2020visualcomet,zellers2019vcr} \\
\hline \hline

\multirow{3}{*}{Generation (\ref{sec:generation})} & \multirow{1}{*}{Summarization (\ref{sec:generation1})} & Extractive~\cite{chen2018extractive,uzzaman2011multimodal} \& abstractive~\cite{li2017multi,palaskar2019multimodal} \\
\Xcline{2-5}{0.5\arrayrulewidth}
& \multirow{1}{*}{Translation (\ref{sec:generation2})} & Exemplar-based~\cite{karpathy2014deep,lebret2015phrase} \& generative~\cite{ahuja2020style,jamaludin2019you,ramesh2021zero} \\
\Xcline{2-5}{0.5\arrayrulewidth}
& \multirow{1}{*}{Creation (\ref{sec:generation3})} & Conditional decoding~\cite{denton2018stochastic,oord2018parallel,zhu2021arbitrary} \\
\Xcline{2-5}{0.5\arrayrulewidth}
\hline \hline

\multirow{3}{*}{Transference (\ref{sec:transference})} & \multirow{1}{*}{Cross-modal transfer (\ref{sec:transference1})} & Tuning~\cite{rahman2020integrating,tsimpoukelli2021multimodal}, multitask~\cite{singh2022flava,liang2022highmmt} \& transfer~\cite{lu2021pretrained} \\
\Xcline{2-5}{0.5\arrayrulewidth}
& \multirow{1}{*}{Co-learning (\ref{sec:transference2})} & Representation~\cite{jia2021scaling,zadeh2020foundations} \& generation~\cite{pham2019found,tan2020vokenization} \\
\Xcline{2-5}{0.5\arrayrulewidth}
& \multirow{1}{*}{Model Induction (\ref{sec:transference3})} & Co-training~\cite{blum1998combining,dunnmon2020cross} \& co-regularization~\cite{sridharan2008information,yang2019comprehensive} &  \\
\hline \hline

\multirow{3}{*}{Quantification (\ref{sec:quantification})} & \multirow{1}{*}{Heterogenity (\ref{sec:quantification1})} & Importance~\cite{gat2021perceptual,park2018multimodal}, bias~\cite{hendricks2018women,pena2020faircvtest} \& noise~\cite{ma2021smil} \\
\Xcline{2-5}{0.5\arrayrulewidth}
& \multirow{1}{*}{Interconnections (\ref{sec:quantification2})} & Connections~\cite{aflalo2022vl,cao2020behind,thrush2022winoground} \& interactions~\cite{hessel2020does,liang2023multiviz,wang2021m2lens} \\
\Xcline{2-5}{0.5\arrayrulewidth}
& \multirow{1}{*}{Learning (\ref{sec:quantification3})} & Generalization~\cite{liang2022highmmt,reed2022generalist}, optimization~\cite{wang2020makes,wu2022characterizing}, tradeoffs~\cite{liang2021multibench} \\
\hline \hline
\end{tabular}
\vspace{-4mm}
\label{table:all}
\end{table*}

Building on these three core principles and our detailed review of recent work, we propose a new taxonomy to characterize the core technical challenges in multimodal research: representation, alignment, reasoning, generation, transference, and quantification. In Table~\ref{table:all} we summarize our full taxonomy of these six core challenges, their subchallenges, categories of corresponding approaches, and recent examples in each category. In the following sections, we describe our new taxonomy in detail and also revisit the principles of heterogeneity, connections, and interactions to see how they pose research questions and inspire research in each of these six challenges.  

\vspace{-2mm}
\section{Challenge 1: Representation}
\label{sec:representation}
\vspace{-1mm}

\begin{figure}[t]
\centering
\includegraphics[width=0.7\linewidth]{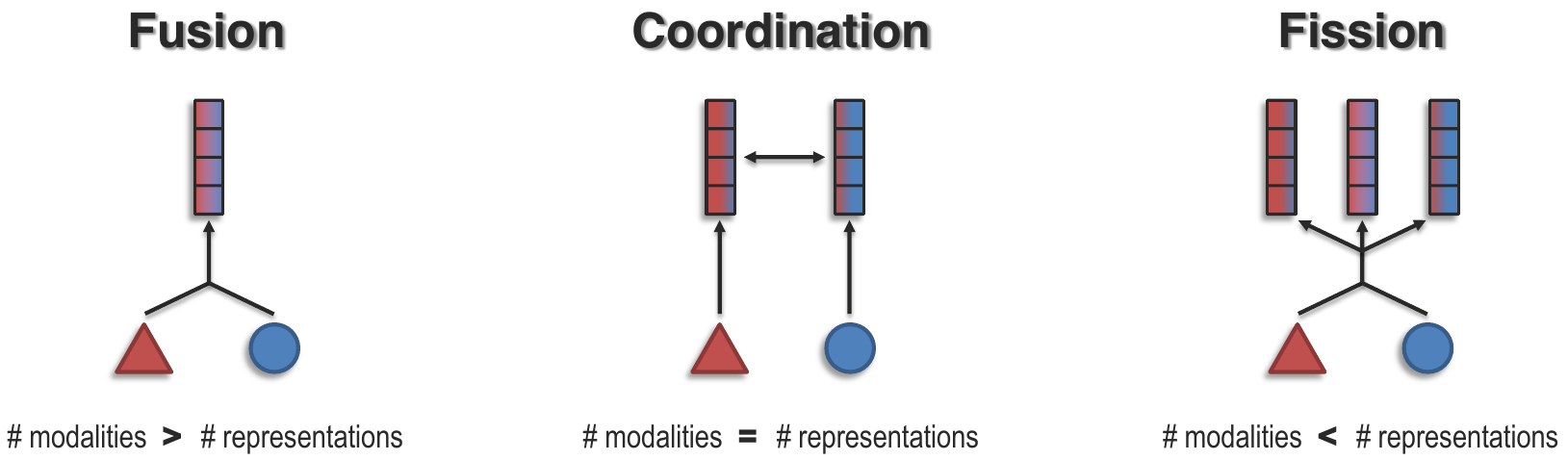}
\caption{Challenge 1 aims to learn \textbf{representations} that reflect cross-modal interactions between individual modality elements, through (1) \textit{fusion}: integrating information to reduce the number of separate representations, (2) \textit{coordination}: interchanging cross-modal information by keeping the same number of representations but improving multimodal contextualization, and (3) \textit{fission}: creating a larger set of decoupled representations that reflects knowledge about internal structure.}
\label{fig:rep}
\vspace{-2mm}
\end{figure}

The first fundamental challenge is to learn representations that reflect cross-modal interactions between individual elements across different modalities. This challenge can be seen as learning a `local' representation between elements, or a representation using holistic features.
This section covers (1) \textit{representation fusion}: integrating information from 2 or more modalities, effectively reducing the number of separate representations, (2) \textit{representation coordination}: interchanging cross-modal information by keeping the same number of representations but improving multimodal contextualization, and (3) \textit{representation fission}: creating a new decoupled set of representations, usually larger number than the input set, that reflects knowledge about internal structure such as data clustering or factorization (Figure~\ref{fig:rep}).

\vspace{-2mm}
\subsection{Subchallenge 1a: Representation fusion}
\label{sec:representation1}
\vspace{-1mm}

\begin{figure}[t]
\centering
\vspace{-0mm}
\includegraphics[width=0.6\linewidth]{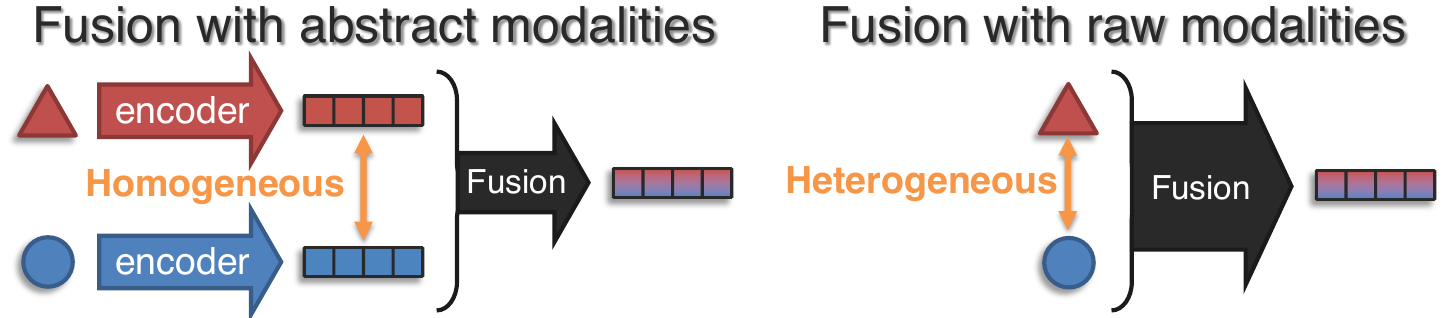}
\caption{We categorize \textbf{representation fusion} approaches into (1) \textit{fusion with abstract modalities}, where unimodal encoders first capture a holistic representation of each element before fusion at relatively homogeneous representations, and (2) \textit{fusion with raw modalities} which entails representation fusion at very early stages, perhaps directly involving heterogeneous raw modalities.}
\label{fig:rep1}
\vspace{-2mm}
\end{figure}

Representation fusion aims to learn a joint representation that models cross-modal interactions between individual elements of different modalities, effectively \textit{reducing} the number of separate representations. We categorize these approaches into \textit{fusion with abstract modalities} and \textit{fusion with raw modalities} (Figure~\ref{fig:rep1}). In fusion with abstract modalities, suitable unimodal encoders are first applied to capture a holistic representation of each element (or modality entirely), after which several building blocks for representation fusion are used to learn a joint representation. As a result, fusion happens at the abstract representation level. On the other hand, fusion with raw modalities entails representation fusion at very early stages with minimal preprocessing, perhaps even involving raw modalities themselves.

\textbf{Fusion with abstract modalities}: We begin our treatment of representation fusion of abstract representations with \textit{additive and multiplicative interactions}. These operators can be seen as differentiable building blocks combining information from two streams of data that can be flexibly inserted into almost any unimodal machine learning pipeline. Given unimodal data or features $\mathbf{x}_1$ and $\mathbf{x}_2$, additive fusion can be seen as learning a new joint representation $\mathbf{z}_\textrm{mm} = w_0 + w_1 \mathbf{x}_1 + w_2 \mathbf{x}_2 + \epsilon$, where $w_1$ and $w_2$ are the weights learned for additive fusion of $\mathbf{x}_1$ and $\mathbf{x}_2 $, $w_0$ the bias term, and $\epsilon$ the error term. If the joint representation $\mathbf{z}_\textrm{mm}$ is directly taken as a prediction $\hat{y}$, then additive fusion resembles late or ensemble fusion $\hat{y} = f_1(\mathbf{x}_1) + f_2(\mathbf{x}_2)$ with unimodal predictors $f_1$ and $f_2$~\citep{friedman2008predictive}. Otherwise, the additive representation $\mathbf{z}_\textrm{mm}$ can also undergo subsequent unimodal or multimodal processing~\citep{baltruvsaitis2018multimodal}.
Multiplicative interactions extend additive interactions to include a cross term $w_3 (\mathbf{x}_1 \times \mathbf{x}_2)$. These models have been used extensively in statistics, where it can be interpreted as a \textit{moderation} effect of $\mathbf{x}_1$ affecting the linear relationship between $\mathbf{x}_2$ and $y$~\cite{baron1986moderator}. Overall, purely additive interactions $\mathbf{z}_\textrm{mm} = w_0 + w_1 \mathbf{x}_1 + w_2 \mathbf{x}_2$ can be seen as a first-order polynomial between input modalities $\mathbf{x}_1$ and $\mathbf{x}_2$, combining additive and multiplicative $\mathbf{z}_\textrm{mm} = w_0 + w_1 \mathbf{x}_1 + w_2 \mathbf{x}_2 + w_3 (\mathbf{x}_1 \times \mathbf{x}_2)$ captures a second-order polynomial.

To further go beyond first and second-order interactions, \textit{tensors} are specifically designed to explicitly capture higher-order interactions across modalities~\cite{zadeh2017tensor}. Given unimodal data $\mathbf{x}_1, \mathbf{x}_2$, tensors are defined as $\mathbf{z}_\textrm{mm} = \mathbf{x}_{1}\otimes \mathbf{x}_{2}$ where $\otimes$ denotes an outer product~\citep{ben2017mutan,fukui2016multimodal}. Tensor products of higher order represent polynomial interactions of higher order between elements~\citep{hou2019deep}.
However, computing tensor products is expensive since their dimension scales exponentially with the number of modalities, so several efficient approximations based on low-rank decomposition have been proposed~\cite{hou2019deep,liu2018efficient}. Finally, \textit{Multiplicative Interactions (MI)} generalize additive and multiplicative operators to include learnable parameters that capture second-order interactions~\cite{jayakumar2020multiplicative}. In its most general form, MI defines a bilinear product $\mathbf{z}_\textrm{mm} = \mathbf{x}_1 \mathbb{W} \mathbf{x}_2 + \mathbf{x}_1^\top \mathbf{U} + \mathbf{V} \mathbf{x}_2 + \mathbf{b}$ where $\mathbb{W}, \mathbf{U}, \mathbf{Z}$, and $\mathbf{b}$ are trainable parameters.

\textit{Multimodal gated units/attention units} learn representations that dynamically change for every input~\cite{chaplot2017gated,wang2020makes}. Its general form can be written as $\mathbf{z}_\textrm{mm} = \mathbf{x}_1 \odot h(\mathbf{x}_2)$, where $h$ represents a function with sigmoid activation and $\odot$ denotes element-wise product. $h(\mathbf{x}_2)$ is commonly referred to as `attention weights' learned from $\mathbf{x}_2$ to attend on $\mathbf{x}_1$. Recent work has explored more expressive forms of learning attention weights such as using Query-Key-Value mechanisms~\cite{tsai2019multimodal}, fully-connected neural network layers~\citep{arevalo2017gated,chaplot2017gated}, or even hard gated units for sharper attention~\citep{chen2017multimodal}.

\textbf{Fusion with raw modalities} entails representation fusion at very early stages, perhaps even involving raw modalities themselves. These approaches typically bear resemblance to early fusion~\citep{baltruvsaitis2018multimodal}, which performs concatenation of input data before applying a prediction model (i.e., $\mathbf{z}_\textrm{mm} = \left[ \mathbf{x}_1, \mathbf{x}_2\right]$). Fusing at the raw modality level is more challenging since raw modalities are likely to exhibit more dimensions of heterogeneity. Nevertheless,~\citet{barnum2020benefits} demonstrated robustness benefits of fusion at early stages, while~\citet{gadzicki2020early} also found that complex early fusion can outperform abstract fusion. To account for the greater heterogeneity during complex early fusion, many approaches rely on generic encoders that are applicable to both modalities, such as convolutional layers~\citep{barnum2020benefits,gadzicki2020early} and Transformers~\citep{liang2022highmmt,likhosherstov2022polyvit}. However, do these complex non-additive fusion models actually learn non-additive interactions between modality elements? Not necessarily, according to~\citet{hessel2020does}. We cover these fundamental analysis questions and more in the quantification challenge (\S\ref{sec:quantification}).

\vspace{-2mm}
\subsection{Subchallenge 1b: Representation coordination}
\label{sec:representation2}
\vspace{-1mm}

\begin{figure}[t]
\centering
\vspace{-0mm}
\includegraphics[width=0.6\linewidth]{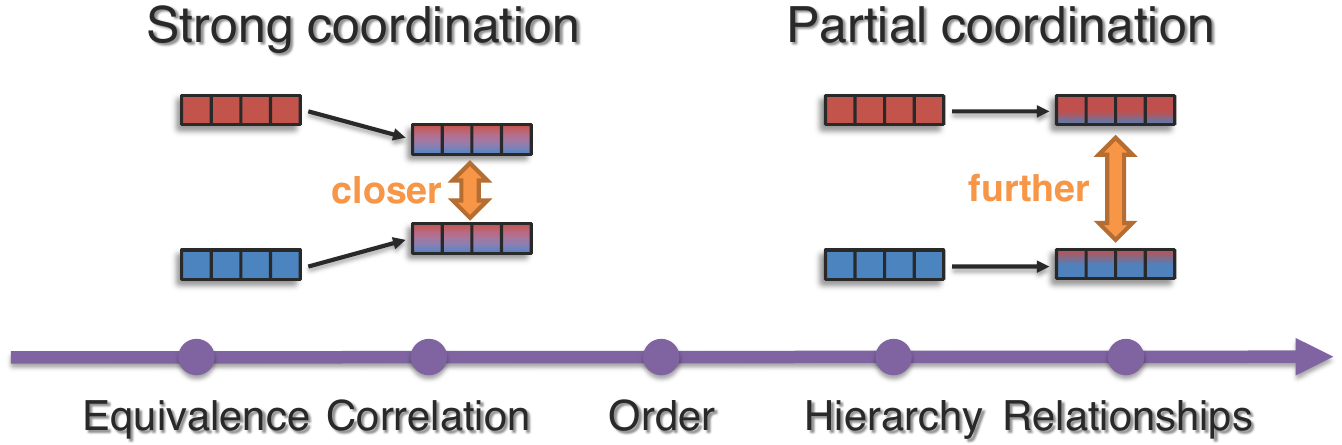}
\caption{There is a spectrum of \textbf{representation coordination} functions: \textit{strong coordination} aims to enforce strong equivalence in all dimensions, whereas in \textit{partial coordination} only certain dimensions may be coordinated to capture more general connections such as correlation, order, hierarchies, or relationships.}
\label{fig:rep2}
\vspace{-2mm}
\end{figure}

Representation coordination aims to learn multimodal contextualized representations that are coordinated through their interconnections (Figure~\ref{fig:rep2}). In contrast to representation fusion, coordination keeps the same number of representations but improves multimodal contextualization. We start our discussion with \textit{strong coordination} that enforces strong equivalence between modality elements, before moving on to \textit{partial coordination} that captures more general connections such as correlation, order, hierarchies, or relationships beyond similarity.

\textbf{Strong coordination} aims to bring semantically corresponding modalities close together in a coordinated space, thereby enforcing strong \textit{equivalence} between modality elements. For example, these models would encourage the representation of the word `dog' and an image of a dog to be close (i.e., semantically positive pairs), while the distance between the word `dog' and an image of a car to be far apart (i.e., semantically negative pairs)~\cite{frome2013devise}. The coordination distance is typically cosine distance~\cite{mekhaldi2007multimodal} or max-margin losses~\cite{hu2019deep}. Recent work has explored large-scale representation coordination by scaling up contrastive learning of image and text pairs~\cite{radford2021learning}, and also found that contrastive learning provably captures redundant information across the two views~\citep{tian2020contrastive,tosh2021contrastive} (but not non-redundant information). In addition to contrastive learning, several approaches instead learn a coordinated space by mapping corresponding data from one modality to another~\cite{dyer2014notes}. For example,~\citet{socher2013zero} maps image embeddings into word embedding spaces for zero-shot image classification. Similar ideas were used to learn coordinated representations between text, video, and audio~\cite{pham2019found}, as well as between pretrained language models and image features~\cite{tan2020vokenization}.

\textbf{Partial coordination}: Instead of capturing strong equivalences, partial coordination captures more general modality connections such as correlation, order, hierarchies, or relationships. Partially coordinated models enforce different types of constraints on the representation space beyond semantic similarity, and perhaps only on certain dimensions of the representation.

\textit{Canonical correlation analysis} (CCA) computes a linear projection that maximizes the correlation between two random variables while enforcing each dimension in a new representation to be orthogonal to each other~\citep{thompson2000canonical}.
CCA models have been used extensively for cross-modal retrieval~\cite{rasiwasia2010new} audio-visual signal analysis~\cite{Sargin2007}, and emotion recognition~\cite{nemati2019hybrid}. To increase the expressiveness of CCA, several nonlinear extensions have been proposed including Kernel CCA~\cite{lai2000kernel}, Deep CCA~\cite{andrew2013deep}, and CCA Autoencoders~\cite{wang2015deep}.

\textit{Ordered and hierarchical spaces}: Another example of representation coordination comes from order-embeddings of images and language~\cite{vendrov2015order}, which aims to capture a partial order on the language and image embeddings to enforce a hierarchy in the coordinated space. A similar model using denotation graphs was also proposed by~\citet{Young2014} where denotation graphs are used to induce such a partial ordering hierarchy.

\textit{Relationship coordination}: In order to learn a coordinated space that captures semantic relationships between elements beyond correspondences,~\citet{zhang2016learning} use structured representations of text and images to create multimodal concept taxonomies.~\citet{delaherche2010multimodal} learn coordinated representations capturing hierarchical relationships, while~\citet{alviar2020multimodal} apply multiscale coordination of speech and music using partial correlation measures. Finally,~\citet{xu2015multi} learn coordinated representations using a Cauchy loss to strengthen robustness to outliers.

\vspace{-2mm}
\subsection{Subchallenge 1c: Representation fission}
\label{sec:representation3}
\vspace{-1mm}

\begin{figure}[t]
\centering
\includegraphics[width=0.5\linewidth]{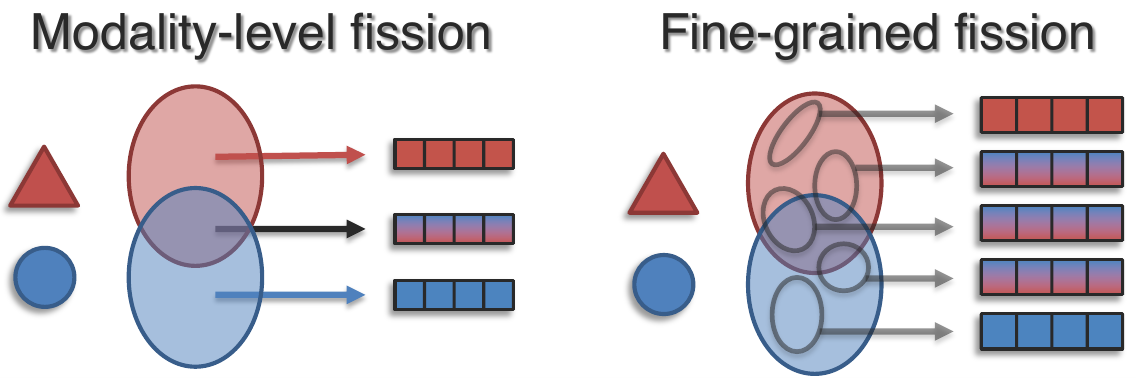}
\caption{\textbf{Representation fission} creates a larger set of decoupled representations that reflects knowledge about internal structure. (1) \textit{Modality-level fission} factorizes into modality-specific information primarily in each modality, and multimodal information redundant in both modalities, while (2) \textit{fine-grained fission} attempts to further break multimodal data down into individual subspaces.}
\label{fig:rep3}
\vspace{-2mm}
\end{figure}

Finally, representation fission aims to create a new decoupled set of representations (usually a larger number than the input representation set) that reflects knowledge about internal multimodal structure such as data clustering, independent factors of variation, or modality-specific information. In comparison with joint and coordinated representations, representation fission enables careful interpretation and fine-grained controllability. Depending on the granularity of decoupled factors, methods can be categorized into \textit{modality-level} and \textit{fine-grained} fission (Figure~\ref{fig:rep3}).

\textbf{Modality-level fission} aims to factorize into modality-specific information primarily in each modality and multimodal information redundant in both modalities~\cite{hsu2018disentangling,liang2023factorized,tsai2019learning}. \textit{Disentangled representation learning} aims to learn mutually independent latent variables that each explain a particular variation of the data~\cite{Bengio:2013:RLR:2498740.2498889,higgins2016beta}, and has been useful for modality-level fission by enforcing independence constraints on modality-specific and multimodal latent variables~\cite{hsu2018disentangling,tsai2019learning}.~\citet{tsai2019learning} and~\citet{hsu2018disentangling} study factorized multimodal representations and demonstrate the importance of modality-specific and multimodal factors towards generation and prediction.~\citet{shi2019variational} study modality-level fission in multimodal variational autoencoders using a mixture-of-experts layer, while~\citet{wu2018multimodal} instead use a product-of-experts layer.

\textit{Post-hoc representation disentanglement} is suitable when it is difficult to retrain a disentangled model, especially for large pretrained multimodal models. Empirical multimodally-additive function projection (EMAP)~\cite{hessel2020does} is an approach for post-hoc disentanglement of the effects of unimodal (additive) contributions from cross-modal interactions in multimodal tasks, which works for arbitrary multimodal models and tasks. EMAP is also closely related to the use of Shapley values for feature disentanglement and interpretation~\cite{merrick2020explanation}, which can also be used for post-hoc representation disentanglement in general models.

\textbf{Fine-grained fission}: Beyond factorizing only into individual modality representations, fine-grained fission attempts to further break multimodal data down into the individual subspaces covered by the modalities~\cite{vidal2011subspace}. \textit{Clustering} approaches that group data based on semantic similarity~\cite{madhulatha2012overview} have been integrated with multimodal networks for end-to-end representation fission and prediction. For example,~\citet{hu2019deep} combine $k$-means clustering in representations with unsupervised audiovisual learning.~\citet{chen2021multimodal} combine $k$-means clustering with self-supervised contrastive learning on videos. Subspace clustering~\cite{abavisani2018deep,khan2021multi}, manifold learning~\cite{li2021multi} approximate graph Laplacians~\cite{khan2019approximate}, conjugate mixture models~\cite{khalidov2011conjugate}, and dictionary learning~\cite{kim2016joint} have also been integrated with multimodal models. \textit{Matrix factorization} techniques have also seen several applications in multimodal fission for  prediction~\cite{aktukmak2019probabilistic} and cross-modal retrieval~\cite{caicedo2012online}.
\vspace{-2mm}
\section{Challenge 2: Alignment}
\label{sec:alignment}
\vspace{-1mm}

\begin{figure}[t]
\centering
\vspace{-0mm}
\includegraphics[width=0.7\linewidth]{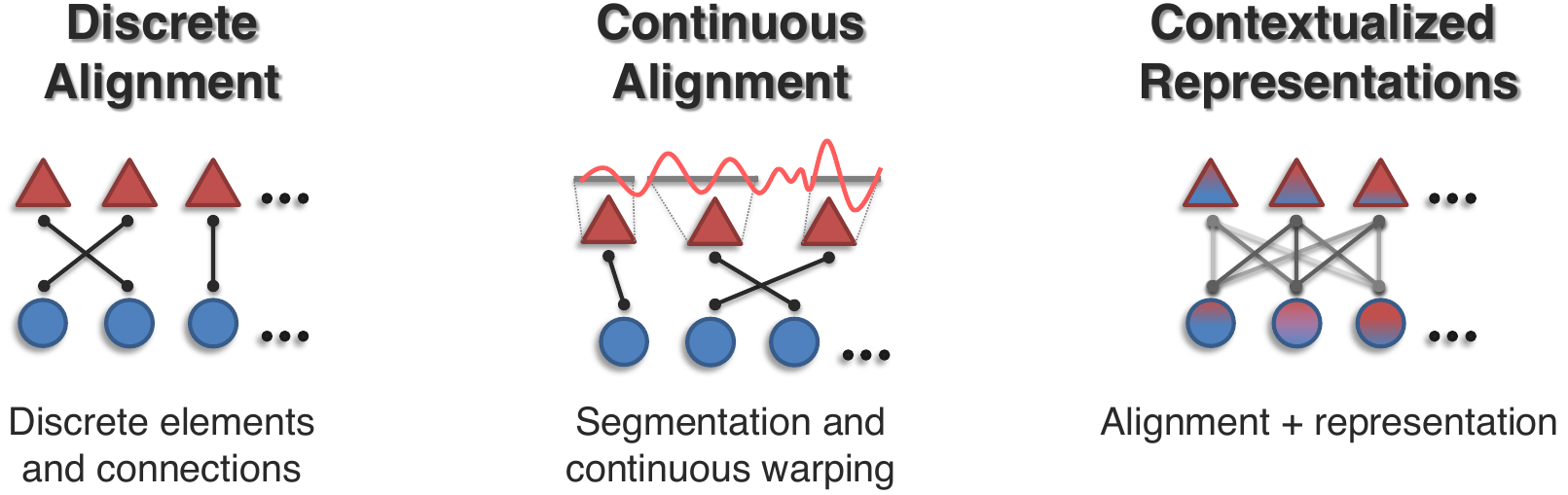}
\caption{\textbf{Alignment} aims to identify cross-modal connections and interactions between modality elements. Recent work has involved (1) \textit{discrete alignment} to identify connections among discrete elements, (2) \textit{continuous alignment} of continuous signals with ambiguous segmentation, and (3) \textit{contextualized representation} learning to capture these cross-modal interactions between connected elements.}
\label{fig:align}
\vspace{-2mm}
\end{figure}

A second challenge is to identify cross-modal connections and interactions between elements of multiple modalities. For example, when analyzing the speech and gestures of a human subject, how can we align specific gestures with spoken words or utterances?
Alignment between modalities is challenging since it may depend on long-range dependencies, involves ambiguous segmentation (e.g., words or utterances), and could be either one-to-one, many-to-many, or not exist at all.
This section covers recent work in multimodal alignment involving (1) \textit{discrete alignment}: identifying connections between discrete elements across modalities, (2) \textit{continuous alignment}: modeling alignment between continuous modality signals with ambiguous segmentation, and (3) \textit{contextualized representations}: learning better multimodal representations by capturing cross-modal interactions between elements (Figure~\ref{fig:align}).

\vspace{-2mm}
\subsection{Subchallenge 2a: Discrete alignment}
\label{sec:alignment1}
\vspace{-1mm}

\begin{figure}[t]
\centering
\vspace{-0mm}
\includegraphics[width=0.4\linewidth]{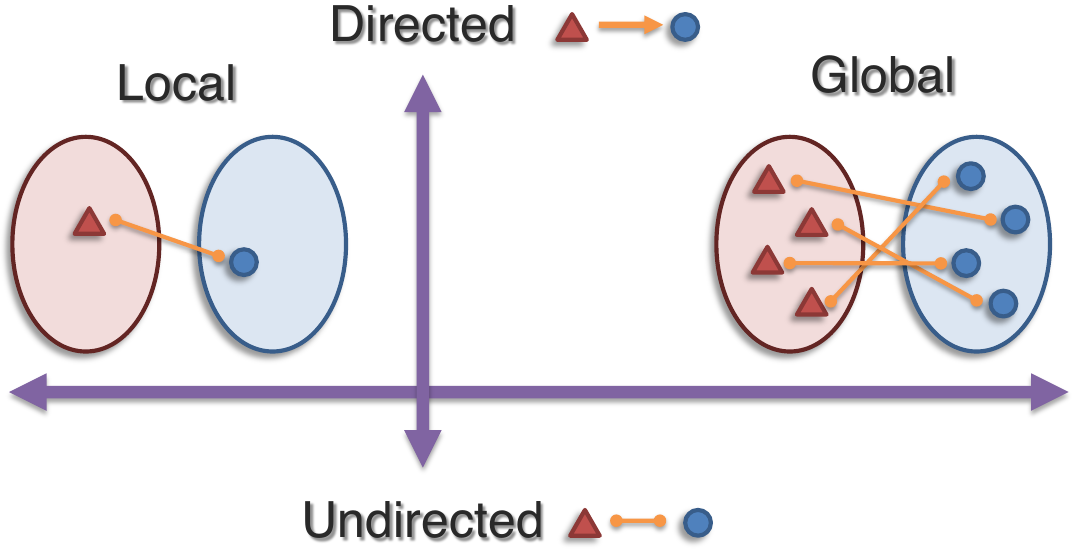}
\caption{\textbf{Discrete alignment} identifies connections between discrete elements, spanning (1) \textit{local alignment} to discover connections given matching pairs, and (2) \textit{global alignment} where alignment must be performed globally to learn both the connections and matchings between modality elements.}
\label{fig:align1}
\vspace{-2mm}
\end{figure}

The first subchallenge aims to identify connections between discrete elements of multiple modalities. We describe recent work in (1) \textit{local alignment} to discover connections between a given matching pair of modality elements, and (2) \textit{global alignment} where alignment must be performed globally to learn both the connections and matchings (Figure~\ref{fig:align1}).

\textbf{Local alignment} between connected elements is particularly suitable for multimodal tasks where there is clear segmentation into discrete elements such as words in text or object bounding boxes in images or videos (e.g., tasks such as visual coreference resolution~\cite{kottur2018visual}, visual referring expression recognition~\cite{cirik2018using,cirik2020refer360}, and cross-modal retrieval~\cite{frome2013devise,pandey2023cross,plummer2015flickr30k}). When we have supervised data in the form of connected modality pairs, \textit{contrastive learning} is a popular approach where the goal is to match representations of the same concept expressed in different modalities~\cite{baltruvsaitis2018multimodal}. Several objective functions for learning aligned spaces from varying quantities of paired~\cite{cao2017transitive,huang2017cross} and unpaired~\cite{grave2019unsupervised} data have been proposed. Many of the ideas that enforce strong~\cite{frome2013devise,liang2021cross} or partial~\citep{andrew2013deep,vendrov2015order,zhang2016learning} representation coordination (\S\ref{sec:representation2}) are also applicable for local alignment. Several examples include aligning books with their corresponding movies/scripts~\cite{zhu2015aligning}, matching referring expressions to visual objects~\cite{mao2016generation}, and finding similarities between image regions and their descriptions~\cite{hu2016natural}. Methods for local alignment have also enabled the learning of shared semantic concepts not purely based on language but also on additional modalities such as vision~\cite{huang2017cross}, sound~\cite{cirik2018visual,socher2013zero}, and multimedia~\cite{zhu2015aligning} that are useful for downstream tasks.

\textbf{Global alignment}: When the ground-truth modality pairings are not available, alignment must be performed globally between all elements across both modalities. Optimal transport (OT)-based approaches~\citep{villani2009optimal} (which belong to a broader set of matching algorithms) are a potential solution since they jointly optimize the coordination function and optimal coupling between modality elements by posing alignment as a divergence minimization problem. These approaches are useful for aligning multimodal representation spaces~\cite{pramanick2022multimodal,li2022clip}. To alleviate computational issues, several recent advances have integrated them with neural networks~\cite{chen2020graph}, approximated optimal transport with entropy regularization~\cite{wei2018unsupervised}, and formulated convex relaxations for efficient learning~\cite{grave2019unsupervised}.

\vspace{-2mm}
\subsection{Subchallenge 2b: Continuous alignment}
\label{sec:alignment2}
\vspace{-1mm}

So far, one important assumption we have made is that modality elements are already segmented and discretized. While certain modalities display clear segmentation (e.g., words/phrases in a sentence or object regions in an image), there are many cases where the segmentation is not readily provided, such as in continuous signals (e.g, financial or medical time-series), spatiotemporal data (e.g., satellite or weather images), or data without clear semantic boundaries (e.g., MRI images). In these settings, methods based on warping and segmentation have been recently proposed:

\textbf{Continuous warping} aims to align two sets of modality elements by representing them as continuous representation spaces and forming a bridge between these representation spaces, such as aligning continuous audio and video data~\cite{gao20192,tian2018audio,tian2020unified}. \textit{Adversarial training} is a popular approach to warp one representation space into another. Initially used in domain adaptation~\cite{ben2006analysis}, adversarial training learns a domain-invariant representation across domains where a domain classifier is unable to identify which domain a feature came from~\cite{ajakan2014domain}. These ideas have been extended to align multimodal spaces~\cite{hsu2018unsupervised,hu2019multimodal,munro2020multi}.~\citet{hsu2018unsupervised} use adversarial training to align images and medical reports,~\citet{hu2019multimodal} design an adversarial network for cross-modal retrieval, and~\citet{munro2020multi} design both self-supervised alignment and adversarial alignment objectives for multimodal action recognition. \textit{Dynamic time warping (DTW)}~\cite{kruskal1983overview} segments and aligns multi-view time-series data by maximizing their similarity via time warping (inserting frames) such that they are aligned across time. For multimodal tasks, it is necessary to design similarity metrics between modalities~\cite{anguera2014audio,tapaswi2015book2movie}, such as combining DTW with CCA or other coordination functions~\cite{trigeorgis2017deep}.

\begin{figure}[t]
\centering
\includegraphics[width=0.4\linewidth]{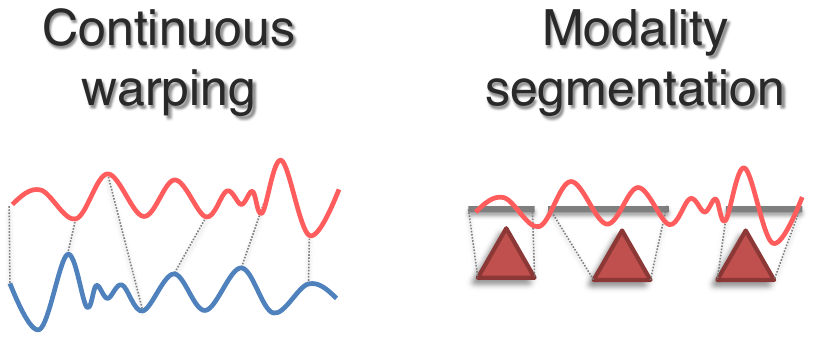}
\caption{\textbf{Continuous alignment} tackles the difficulty of aligning continuous signals where element segmentation is not readily available. We cover related work in (1) \textit{continuous warping} of representation spaces and (2) \textit{modality segmentation} of continuous signals into discrete elements at an appropriate granularity.}
\label{fig:align2}
\vspace{-2mm}
\end{figure}

\textbf{Modality segmentation} involves dividing high-dimensional data into elements with semantically meaningful boundaries. A common problem involves \textit{temporal segmentation}, where the goal is to discover the temporal boundaries across sequential data. Several approaches for temporal segmentation include forced alignment, a popular approach to align discrete speech units with individual words in a transcript~\cite{P2FA}.~\citet{malmaud2015s} explore multimodal alignment using a factored hidden Markov model to align ASR transcripts to the ground truth. \textit{Clustering} approaches have also been used to group continuous data based on semantic similarity~\cite{madhulatha2012overview}. Clustering-based discretization has recently emerged as an important preprocessing step for generalizing language-based pretraining (with clear word/byte pair segmentation boundaries and discrete elements) to video or audio-based pretraining (without clear segmentation boundaries and continuous elements). By clustering raw video or audio features into a discrete set, approaches such as VideoBERT~\citep{sun2019videobert} perform masked pretraining on raw video and audio data. Similarly, approaches such as DALL.E~\citep{ramesh2021zero}, VQ-VAE~\citep{van2017neural}, and CMCM~\citep{liu2022cross} also utilize discretized intermediate layers obtained via vector quantization and showed benefits in modality alignment.

\vspace{-2mm}
\subsection{Subchallenge 2c: Contextualized representations}
\label{sec:alignment3}
\vspace{-1mm}

\begin{figure}[t]
\centering
\vspace{-0mm}
\includegraphics[width=0.6\linewidth]{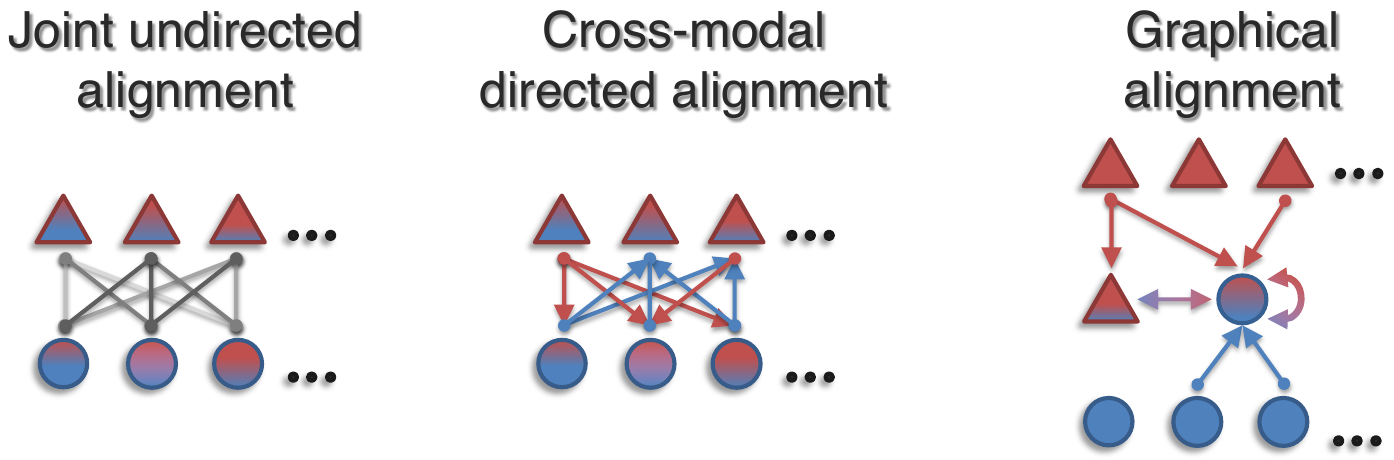}
\caption{\textbf{Contextualized representation} learning aims to model modality connections to learn better representations. Recent directions include (1) \textit{joint undirected alignment} that captures undirected symmetric connections, (2) \textit{cross-modal directed alignment} that models asymmetric connections in a directed manner, and (3) \textit{graphical alignment} that generalizes the sequential pattern into arbitrary graph structures.}
\label{fig:align3}
\vspace{-2mm}
\end{figure}

Finally, contextualized representation learning aims to model all modality connections and interactions to learn better representations. Contextualized representations have been used as an intermediate (often latent) step enabling better performance on a number of downstream tasks including speech recognition, machine translation, media description, and visual question-answering. We categorize work in contextualized representations into (1) \textit{joint undirected alignment}, (2) \textit{cross-modal directed alignment}, and (3) \textit{alignment with graph networks} (Figure~\ref{fig:align3}).

\textbf{Joint undirected alignment} aims to capture undirected connections across pairs of modalities, where the connections are symmetric in either direction. This is commonly referred to in the literature as unimodal, bimodal, trimodal interactions, and so on~\cite{macaluso2005multisensory}. Joint undirected alignment is typically captured by parameterizing models with alignment layers and training end-to-end for a multimodal task. These alignment layers can include attention weights~\cite{chaplot2017gated}, tensor products~\cite{liu2018efficient,zadeh2017tensor}, and multiplicative interactions~\cite{jayakumar2020multiplicative}. More recently, transformer models~\cite{vaswani2017attention} have emerged as powerful encoders for sequential data by automatically aligning and capturing complementary features at different time steps. Building upon the initial text-based transformer model, multimodal transformers have been proposed that perform joint alignment using a full self-attention over modality elements concatenated across the sequence dimension (i.e., early fusion)~\cite{li2019visualbert,sun2019videobert}. As a result, all modality elements become jointly connected to all other modality elements similarly (i.e., modeling all connections using dot-product similarity kernels).

\textbf{Cross-modal directed alignment} relates elements from a source modality in a directed manner to a target modality, which can model asymmetric connections. For example, \textit{temporal attention models} use alignment as a latent step to improve many sequence-based tasks~\cite{xiong2016dynamic,zeng2017leveraging}. These attention mechanisms are typically directed from the output to the input so that the resulting weights reflect a soft alignment distribution over the input. \textit{Multimodal transformers} perform directed alignment using query-key-value attention mechanisms to attend from one modality's sequence to another, before repeating in a bidirectional manner. This results in two sets of asymmetric contextualized representations to account for the possibly asymmetric connections between modalities~\cite{lu2019vilbert,tan2019lxmert,tsai2019multimodal}. 
These methods are useful for sequential data by automatically aligning and capturing complementary features at different time-steps~\cite{tsai2019multimodal}. 

\textbf{Large vision-language foundation models} have emerged as powerful models capable of learning contextualized representations for multiple tasks involving natural language, images, video, and audio~\citep{gan2022vision,liang2022highmmt,openai2023gpt4,radford2021learning,reed2022generalist}. These models typically build on top of pretrained language models~\citep{radford2019language}, pretrained visual encoders~\citep{dosovitskiy2020image} combined with an alignment layer. Alignment can be done via end-to-end training with multimodal transformers~\citep{xu2023multimodal} (e.g., Flamingo~\citep{alayrac2022flamingo}, OpenFlamingo~\citep{awadalla2023openflamingo}, Kosmos~\citep{peng2023kosmos}), or keeping the language and vision parts frozen and only training a post-hoc alignment layer (e.g., MiniGPT-4~\cite{zhu2023minigpt}, BLIP-2~\cite{li2023blip}, InstructBLIP~\cite{dai_instructblip:_2023}, LLaMA-Adapter V2~\cite{gao2023llama}). Self-supervised pretraining has emerged as an effective way to train these architectures to learn general-purpose representations from larger-scale unlabeled multimodal data before transferring to specific downstream tasks via supervised fine-tuning~\cite{driess2023palm,li2019visualbert,zhu2023minigpt}. Pretraining objectives typically consist of unimodal language modeling~\citep{radford2019language,raffel2020exploring}, image-to-text or text-to-image alignment~\cite{hendricks2021decoupling,zhu2023minigpt}, and multimodal instruction tuning~\cite{dai_instructblip:_2023,liu2023visual,lu2023empirical}. We refer the reader to recent survey papers discussing these large vision-language models in more detail~\cite{du2022survey,gan2022vision}.

\textbf{Graphical alignment} generalizes the sequential pattern seen in undirected or directed alignment into arbitrary graph structures between elements. This has several benefits since it does not require all elements to be connected, and allows the user to choose different edge functions for different connections. Graph neural networks~\citep{velivckovic2018graph} can be used to recursively learn element representations contextualized with the elements in locally connected neighborhoods~\citep{scarselli2008graph,velivckovic2018graph}, such as in MTAG~\citep{yang2021mtag} and F2F-CL~\citep{wilf2022face} for multimodal and multi-speaker videos.
\begin{figure}[t]
\centering
\vspace{-0mm}
\includegraphics[width=0.6\linewidth]{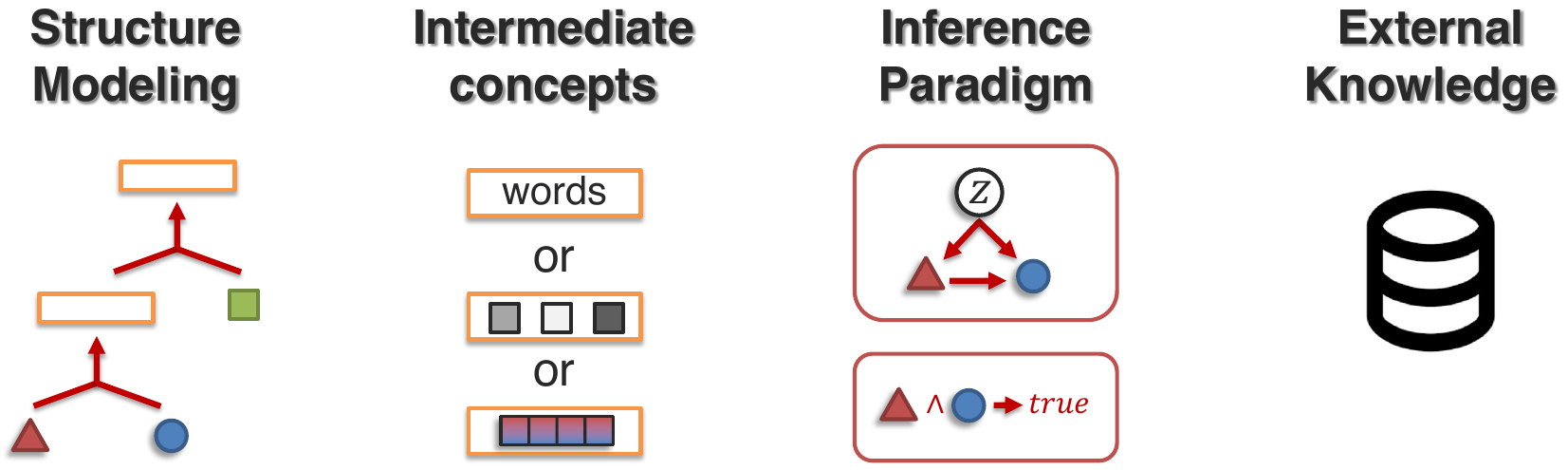}
\caption{\textbf{Reasoning} aims to combine knowledge, usually through multiple inferential steps, exploiting the problem structure. Reasoning involves (1) \textit{structure modeling}: defining or learning the relationships over which reasoning occurs, (2) the \textit{intermediate concepts} used in reasoning, (3) \textit{inference} of increasingly abstract concepts from evidence, and (4) leveraging \textit{external knowledge} in the study of structure, concepts, and inference.}
\label{fig:reason}
\vspace{-2mm}
\end{figure}

\vspace{-2mm}
\section{Challenge 3: Reasoning}
\label{sec:reasoning}
\vspace{-1mm}

Reasoning is defined as combining knowledge, usually through multiple inferential steps, exploiting multimodal alignment and the problem structure. We categorize work towards multimodal reasoning into $4$ subchallenges of structure modeling, intermediate concepts, inference paradigm, and external knowledge (Figure~\ref{fig:reason}). (1) \textit{Structure modeling} involves defining or learning the relationships over which reasoning occurs, (2) \textit{intermediate concepts} studies the parameterization of individual multimodal concepts in the reasoning process, (3) \textit{inference paradigm} learns how increasingly abstract concepts are inferred from individual multimodal evidence, and (4) \textit{external knowledge} aims to leverage large-scale databases in the study of structure, concepts, and inference.

\vspace{-2mm}
\subsection{Subchallenge 3a: Structure modeling}
\label{sec:reasoning1}
\vspace{-1mm}

\begin{figure}[t]
\centering
\vspace{-0mm}
\includegraphics[width=0.6\linewidth]{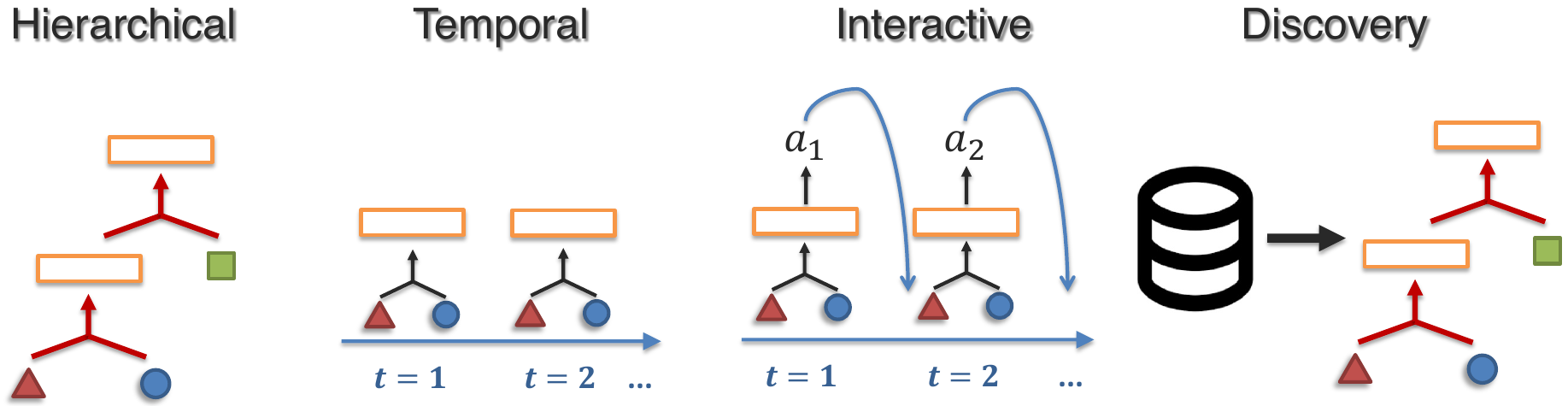}
\caption{\textbf{Structure modeling} aims to define the relationship over which composition occurs, which can be (1) \textit{hierarchical} (i.e., more abstract concepts are defined as a function of less abstract ones), (2) \textit{temporal} (i.e., organized across time), (3) \textit{interactive} (i.e., where the state changes depending on each step's decision), and (4) \textit{discovered} when the latent structure is unknown and instead directly inferred from data and optimization.}
\label{fig:reason1}
\vspace{-2mm}
\end{figure}

Structure modeling aims to capture the hierarchical relationship over which composition occurs, usually via a data structure parameterizing atoms, relations, and the reasoning process. Commonly used data structures include trees~\cite{hong2019learning}, graphs~\cite{Yu2019HeterogeneousGL}, or neural modules~\cite{andreas2016neural}. We cover recent work in modeling latent \textit{hierarchical}, \textit{temporal}, and \textit{interactive} structure, as well as \textit{structure discovery} when the latent structure is unknown (Figure~\ref{fig:reason1}).

\textbf{Hierarchical structure} defines a system of organization where abstract concepts are defined as a function of less abstract ones. Hierarchical structure is present in many tasks involving language syntax, visual syntax, or higher-order reasoning. These approaches typically construct a graph based on predefined node and edge categories before using (heterogeneous variants of) graph neural networks to capture a representation of structure~\cite{shi2016survey}, such as using language syntactic structure to guide visual modules that discover specific information in images~\cite{andreas2016neural,cirik2018using}. Graph-based reasoning approaches have been applied for visual commonsense reasoning~\cite{lin2019kagnet}, visual question answering~\cite{saqur2020multimodal}, machine translation~\cite{yin2020novel}, recommendation systems~\cite{tao2020mgat}, web image search~\cite{wang2012multimodal}, and social media analysis~\cite{schinas2015multimodal}.

\textbf{Temporal structure} extends the notion of compositionality to elements across time, which is necessary when modalities contain temporal information, such as in video, audio, or time-series data. Explicit memory mechanisms have emerged as a popular choice to accumulate multimodal information across time so that long-range cross-modal interactions can be captured through storage and retrieval from memory.~\citet{rajagopalan2016extending} explore various memory representations including multimodal fusion, coordination, and factorization. Insights from key-value memory~\cite{xiong2016dynamic} and attention-based memory~\cite{zadeh2018memory} have also been successfully applied to applications including question answering, video captioning, emotion recognition, and sentiment analysis.

\textbf{Interactive structure} extends the challenge of reasoning to interactive settings, where the state of the reasoning agent changes depending on the local decisions made at every step. Typically formalized by the sequential decision-making framework, the challenge lies in maximizing long-term cumulative reward despite only interacting with the environment through short-term actions~\cite{sutton2018reinforcement}. To tackle the challenges of interactive reasoning, the growing research field of multimodal reinforcement learning (RL) has emerged from the intersection of language understanding, embodiment in the visual world, deep reinforcement learning, and robotics. We refer the reader to the extensive survey paper by~\citet{luketina2019survey} and the position paper by~\citet{bisk2020experience} for a full review of this field.~\citet{luketina2019survey} separate the literature into multimodal-conditional RL (in which multimodal interaction is necessitated by the problem formulation itself, such as instruction following~\cite{chaplot2017gated,wang2019reinforced}) and language-assisted RL (in which multimodal data is optionally used to facilitate learning, such as reading instruction manuals~\cite{narasimhan2018grounding}).

\textbf{Structure discovery}: It may be challenging to define the structure of multimodal composition without some domain knowledge of the given task. As an alternative approach, recent work has also explored using differentiable strategies to automatically search for the structure in a fully data-driven manner. To do so, one first needs to define a candidate set of reasoning atoms and relationships, before using a `meta' approach such as architecture search to automatically search for the ideal sequence of compositions for a given task~\cite{perez2019mfas,xu2021mufasa}. These approaches can benefit from optimization tricks often used in the neural architecture search literature. Memory, Attention, and Composition (MAC) similarly search for a series of attention-based reasoning steps from data in an end-to-end approach~\cite{hudson2018compositional}. Finally,~\citet{hu2017learning} extend the predefined reasoning structure obtained through language parsing in~\citet{andreas2016neural} by instead using policy gradients to automatically optimize a compositional structure over a discrete set of neural modules.

\vspace{-2mm}
\subsection{Subchallenge 3b: Intermediate concepts}
\label{sec:reasoning2}
\vspace{-1mm}

The second subchallenge studies how we can parameterize individual multimodal concepts within the reasoning process. While intermediate concepts are usually dense vector representations in standard neural architectures, there has also been substantial work towards interpretable attention maps, discrete symbols, and language as an intermediate medium for reasoning.

\textbf{Attention maps} are a popular choice for intermediate concepts since they are, to a certain extent, human-interpretable, while retaining differentiability. For example,~\citet{andreas2016neural} design individual modules such as `attend', `combine', `count', and `measure' that are each parametrized by attention operations on the input image for visual question answering.~\citet{xu2015show} explore both soft and hard attention mechanisms for reasoning in image captioning generation. Related work has also used attention maps through dual attention architectures~\citep{nam2017dual} or stacked latent attention architectures~\citep{fan2018stacked} for multimodal reasoning. These are typically applied for problems involving complex reasoning steps such as CLEVR~\citep{johnson2017clevr} or VQA~\citep{zhang2020multimodal}.

\textbf{Discrete symbols}: A further level of discretization beyond attention maps involves using discrete symbols to represent intermediate concepts. Recent work in neuro-symbolic learning aims to integrate these discrete symbols as intermediate steps in multimodal reasoning in tasks such as visual question answering~\cite{andreas2016neural,mao2018neuro,vedantam2019probabilistic} or referring expression recognition~\cite{cirik2018using}. A core challenge in this approach lies in maintaining the differentiability of discrete symbols, which has been tackled via logic-based differentiable reasoning~\cite{amizadeh2020neuro,serafini2016logic}.

\textbf{Language as a medium}: Finally, perhaps the most human-understandable form of intermediate concepts is natural language (through discrete words or phrases) as a medium. Recently,~\citet{zeng2022socratic} explored using language as an intermediate medium to coordinate multiple separate pretrained models in a zero-shot manner. Several approaches also used language phrases obtained from external knowledge graphs to facilitate interpretable reasoning~\citep{gui2021kat,zhu2015building}.~\citet{hudson2019learning} designed a neural state machine to simulate the execution of a question being asked about an image, while using discrete words as intermediate concepts.

\vspace{-2mm}
\subsection{Subchallenge 3c: Inference paradigms}
\label{sec:reasoning3}
\vspace{-1mm}

The third subchallenge in multimodal reasoning defines how increasingly abstract concepts are inferred from individual multimodal evidence. While advances in local representation fusion (such as additive, multiplicative, tensor-based, attention-based, and sequential fusion, see \S\ref{sec:representation1} for a full review) are also generally applicable here, the goal of reasoning is to be more interpretable in the inference process through domain knowledge about the multimodal problem. To that end, we cover recent directions in explicitly modeling the inference process via logical and causal operators as examples of recent trends in this direction.

\textbf{Logical inference}: Logic-based differentiable reasoning has been widely used to represent knowledge in neural networks~\cite{amizadeh2020neuro,serafini2016logic}. Many of these approaches use differentiable fuzzy logic~\cite{van2022analyzing} which provides a probabilistic interpretation of logical predicates, functions, and constants to ensure differentiability. These logical operators have been applied for visual question answering~\cite{gokhale2020vqa} and visual reasoning~\cite{amizadeh2020neuro}. Among the greatest benefits of logical reasoning lies in its ability to perform interpretable and compositional multi-step reasoning~\cite{hudson2019gqa}. Logical frameworks have also been useful for visual-textual entailment~\cite{suzuki2019multimodal} and geometric numerical reasoning~\cite{chen2021geoqa}, fields where logical inductive biases are crucial toward strong performance.

\textbf{Causal inference} extends the associational level of reasoning to interventional and counterfactual levels~\cite{pearl2009causality}, which requires extensive knowledge of the world to imagine counterfactual effects. For example,~\citet{yi2019clevrer} propose the CLEVRER benchmark focusing on four specific elements of reasoning on videos: descriptive (e.g., ‘what color’), explanatory (‘what’s responsible for’), predictive (‘what will happen next’), and counterfactual (‘what if’). Beyond CLEVRER, recent work has also proposed Causal VQA~\cite{agarwal2020towards} and Counterfactual VQA~\cite{niu2021counterfactual} to measure the robustness of VQA models under controlled interventions to the question as a step towards mitigating language bias in VQA models. Methods inspired by integrating causal reasoning capabilities into neural network models have also been shown to improve robustness and reduce biases~\cite{wang2020visual}.

\vspace{-2mm}
\subsection{Subchallenge 3d: External knowledge}
\label{sec:reasoning4}
\vspace{-1mm}

The final subchallenge studies the derivation of knowledge in the study of defining composition and structure. Knowledge can refer to any data source that is complementary to the limited supervised training data that models typically see, which encapsulates larger banks of unlabeled internet data (e.g., textbooks, Wikipedia, videos), curated knowledge graphs and knowledge bases, and expert domain knowledge for specific tasks such as healthcare and robotics.

\textbf{Multimodal knowledge graphs} extend classic work in language and symbolic knowledge graphs (e.g., Freebase~\cite{bollacker2008freebase}, DBpedia~\cite{auer2007dbpedia}, YAGO~\cite{suchanek2007yago}, WordNet~\cite{miller1995wordnet}) to semantic networks containing multimodal concepts as nodes and multimodal relationships as edges~\cite{zhu2022multi}. Multimodal knowledge graphs are important because they enable the grounding of structured information in the visual and physical world~\cite{bisk2020piqa,yu2022pacs}. For example,~\citet{liu2019mmkg} constructs multimodal knowledge graphs containing both numerical features and images for entities. Visual Genome is another example containing dense annotations of objects, attributes, and relationships in images and text~\cite{krishna2017visual}. These multimodal knowledge bases have been shown to benefit visual question answering~\cite{wu2016ask,zhu2015building}, knowledge base completion~\cite{pezeshkpour2018embedding}, and image captioning~\cite{melas2018training}.~\citet{gui2021kat} integrates knowledge into vision-and-language transformers for automatic reasoning over both knowledge sources. Another promising approach is multimodal knowledge expansion~\citep{radevski2023multimodal,wu2023multimodal,xue2021multimodal} using knowledge distillation to expand knowledge from unimodal data to multimodal settings. We refer the reader to a comprehensive survey by~\citet{zhu2022multi} for additional references.

\textbf{Multimodal commonsense reasoning} requires deeper real-world knowledge potentially spanning logical, causal, and temporal relationships between concepts. For example, elements of causal reasoning are required to answer the questions regarding images in VCR~\cite{zellers2019vcr} and VisualCOMET~\cite{park2020visualcomet}, while other works have also introduced datasets with video and text inputs to test for temporal reasoning (e.g., MovieQA~\cite{tapaswi2016movieqa}, MovieFIB~\cite{maharaj2017dataset}, TVQA~\cite{lei2018tvqa}). Benchmarks for multimodal commonsense typically require leveraging external knowledge from knowledge bases~\cite{song2021kvl} or pretraining paradigms on large-scale datasets~\cite{lu2019vilbert,zellers2021merlot}.
\vspace{-2mm}
\section{Challenge 4: Generation}
\label{sec:generation}
\vspace{-1mm}

The fourth challenge involves learning a generative process to produce raw modalities that reflect cross-modal interactions, structure, and coherence, through \textit{summarization}, \textit{translation}, and \textit{creation} (Figure~\ref{fig:gen}). These three categories are distinguished based on the information change from input to output modalities, following categorizations in text generation~\citep{deng2021compression}. We will cover recent advances as well as the evaluation of generated content.

\begin{figure}[t]
\centering
\vspace{-0mm}
\includegraphics[width=0.6\linewidth]{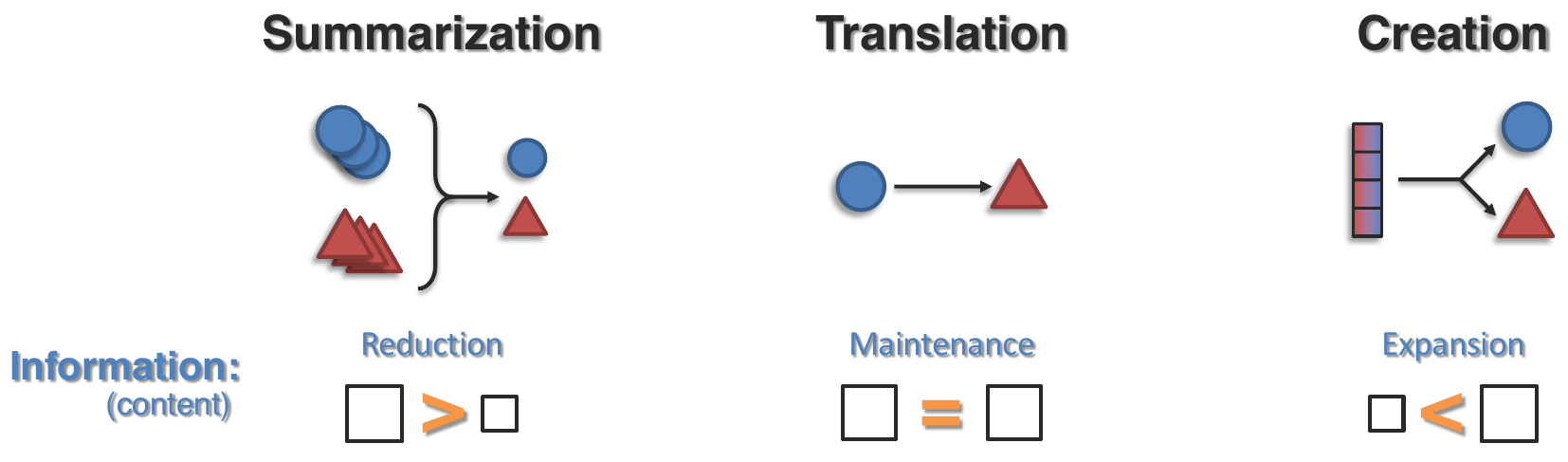}
\caption{How can we learn a generative process to produce raw modalities that reflect cross-modal interactions, structure, and coherence? \textbf{Generation} involves (1) \textit{summarizing} multimodal data to highlight the most salient parts, (2) \textit{translating} from one modality to another while being consistent with modality connections, and (3) \textit{creating} multiple modalities simultaneously while maintaining coherence.}
\label{fig:gen}
\vspace{-2mm}
\end{figure}

\vspace{-2mm}
\subsection{Subchallenge 4a: Summarization}
\label{sec:generation1}
\vspace{-1mm}

Summarization aims to compress data to create an abstract that represents the most important or relevant information within the original content. Recent work has explored various input modalities to guide text summarization, such as images~\cite{chen2018abstractive}, video~\cite{li2020vmsmo}, and audio~\cite{evangelopoulos2013multimodal,jangra2020text,li2017multi}. Recent trends in multimodal summarization include \textit{extractive} and \textit{abstractive} approaches. Extractive approaches aim to filter words, phrases, and other unimodal elements from the input to create a summary~\cite{chen2018extractive,jangra2020text,li2017multi}. Beyond text as output, video summarization is the task of producing a compact version of the video (visual summary) by encapsulating the most informative parts~\cite{sah2017semantic}.~\citet{li2017multi} collected a dataset of news videos and articles paired with manually annotated summaries as a benchmark towards multimodal summarization. Finally,~\citet{uzzaman2011multimodal} aim to simplify complex sentences by extracting multimodal summaries for accessibility. On the other hand, abstractive approaches define a generative model to generate the summary at multiple levels of granularity~\cite{chen2018abstractive,li2019keep}. Although most approaches only focus on generating a textual summary from multimodal data~\cite{palaskar2019multimodal}, several directions have also explored generating summarized images to supplement the generated textual summary~\cite{chen2018abstractive,li2020vmsmo}.

\vspace{-2mm}
\subsection{Subchallenge 4b: Translation}
\label{sec:generation2}
\vspace{-1mm}

Translation aims to map one modality to another while respecting semantic connections and information content~\citep{vinyals2016show}. For example, generating a descriptive caption of an image can help improve the accessibility of visual content for blind people~\citep{gurari2018vizwiz}. Multimodal translation brings about new difficulties involving the generation of high-dimensional structured data as well as their evaluation. Recent approaches can be classified as \textit{exemplar-based}, which are limited to retrieving from training instances to translate between modalities but guarantee fidelity~\cite{farhadi2010every}, and \textit{generative} models which can translate into arbitrary instances interpolating beyond the data but face challenges in quality, diversity, and evaluation~\cite{koh2021text,ramesh2021zero,tsimpoukelli2021multimodal}. Despite these challenges, recent progress in large-scale generative models has yielded impressive results in text-to-image~\citep{rombach2022high,ramesh2021zero}, text-to-video~\citep{singer2022make}, audio-to-image~\cite{jamaludin2019you}, text-to-speech~\cite{ren2019fastspeech}, speech-to-gesture~\citep{ahuja2020style}, speaker-to-listener~\citep{ng2022learning}, language to pose~\citep{ahuja2019language2pose}, and speech and music generation~\cite{agostinelli2023musiclm,copet2023simple,oord2018parallel}.

\vspace{-2mm}
\subsection{Subchallenge 4c: Creation}
\label{sec:generation3}
\vspace{-1mm}

Creation aims to generate novel high-dimensional data (which could span text, images, audio, video, and other modalities) from small initial examples or latent conditional variables. This \textit{conditional decoding} process is extremely challenging since it needs to be (1) conditional: preserve semantically meaningful mappings from the initial seed to a series of long-range parallel modalities, (2) synchronized: semantically coherent across modalities, (3) stochastic: capture many possible future generations given a particular state, and (4) auto-regressive across possibly long ranges.
Many modalities have been considered as targets for creation.
Language generation has been explored for a long time~\cite{radford2019language}, and recent work has explored high-resolution speech and sound generation using neural networks~\cite{oord2018parallel}. Photorealistic image generation has also recently become possible due to advances in large-scale generative modeling~\cite{karras2020analyzing}. Furthermore, there have been a number of attempts at generating abstract scenes~\cite{tan2019text2scene}, computer graphics~\cite{mildenhall2020nerf}, and talking heads~\cite{zhu2021arbitrary}. While there has been some progress toward video generation~\citep{singer2022make}, complete synchronized generation of realistic video, text, and audio remains a challenge.

Finally, one of the biggest challenges facing multimodal generation is difficulty in evaluating generated content, especially when there exist serious ethical issues when fake news~\citep{bender2021dangers}, hate speech~\cite{gehman2020realtoxicityprompts,abid2021persistent}, deepfakes~\cite{hancock2021social}, and lip-syncing videos~\citep{suwajanakorn2017synthesizing} can be easily generated. 
While the ideal way to evaluate generated content is through user studies, it is time-consuming, costly, and can potentially introduce subjectivity bias~\citep{geva2019we}. Several automatic proxy metrics have been proposed~\cite{anderson2016spice,chen2020comprises} by none are universally robust across many generation tasks.
\vspace{-2mm}
\section{Challenge 5: Transference}
\label{sec:transference}
\vspace{-1mm}

\begin{figure}[t]
\centering
\vspace{-0mm}
\includegraphics[width=0.6\linewidth]{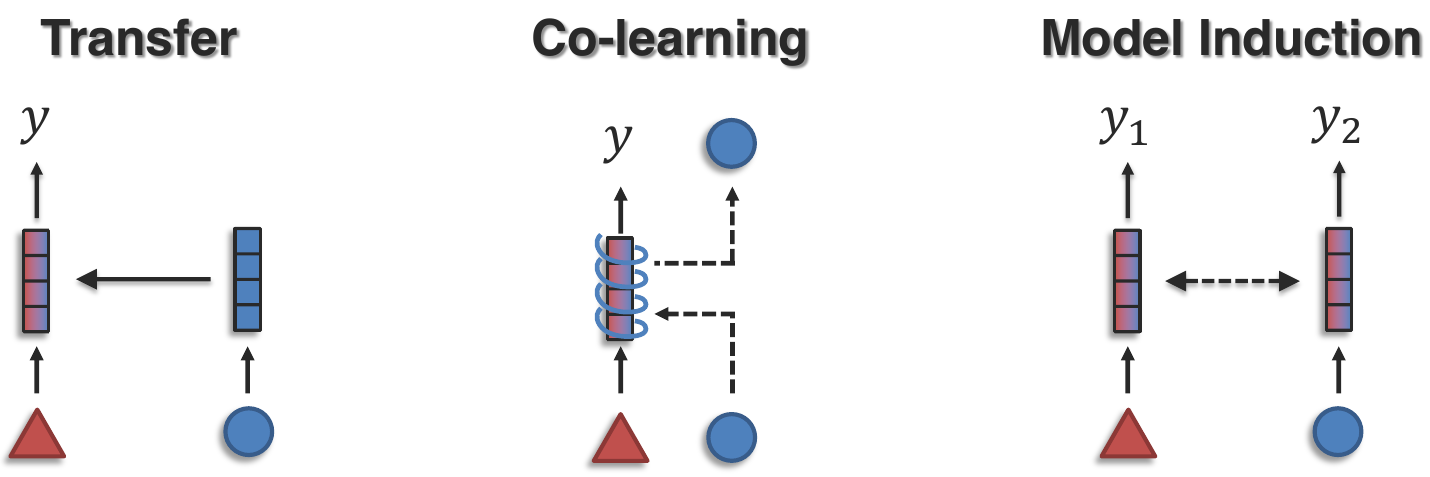}
\caption{\textbf{Transference} studies the transfer of knowledge between modalities, usually to help a noisy or limited primary modality, via (1) \textit{cross-modal transfer} from models trained with abundant data in the secondary modality, (2) \textit{multimodal co-learning} to share information across modalities by sharing representations, and (3) \textit{model induction} that keeps individual unimodal models separate but induces behavior in separate models.}
\label{fig:transference}
\vspace{-2mm}
\end{figure}

Transference aims to transfer knowledge between modalities and their representations, and is often used when there is a \textit{primary modality} that we care about making predictions on but suffers from limited resources - a lack of annotated data, noisy inputs, or unreliable labels, and a \textit{secondary modality} with more abundant or reliable data. How can knowledge learned from a secondary modality (e.g., predicted labels or representation) help a model trained on a primary modality? We call this challenge transference, since the transfer of information from the secondary modality gives rise to new behaviors previously unseen in the primary modality. We identify three types of approaches: (1) \textit{cross-modal transfer}, (2) \textit{multimodal co-learning}, and (3) \textit{model induction} (Figure~\ref{fig:transference}).

\vspace{-2mm}
\subsection{Subchallenge 5a: Cross-modal transfer}
\label{sec:transference1}
\vspace{-1mm}

In most settings, it may be easier to collect either labeled or unlabeled data in the secondary modality and train strong supervised or pretrained models. These models can then be conditioned or fine-tuned for a downstream task involving the primary modality. In other words, this line of research extends unimodal transfer and fine-tuning to cross-modal settings.

\textbf{Tuning}: Inspired by prior work in NLP involving prefix tuning~\cite{li2021prefix} and prompt tuning~\cite{lester2021power}, recent work has also studied the tuning of pretrained language models to condition on visual and other modalities. For example,~\citet{tsimpoukelli2021multimodal} quickly conditions a pretrained, frozen language model on images for image captioning. Related work has also adapted prefix tuning for image captioning~\cite{chen2021visualgpt}, multimodal fusion~\cite{hasan2021humor}, and summarization~\cite{yu2021vision}. While prefix tuning is simple and efficient, it provides the user with only limited control over how information is transferred. Representation tuning goes a level deeper by modifying the inner representations of the language model via contextualization with other modalities. For example,~\citet{ziegler2019encoder} includes additional self-attention layers between language model layers and external modalities.~\citet{rahman2020integrating} design a shifting gate to adapt language model layers with audio and visual information.

\textbf{Multitask learning} aims to use multiple large-scale tasks to improve performance as compared to learning on individual tasks. Several models such as Perceiver~\cite{jaegle2021perceiver}, MultiModel~\cite{kaiser2017one}, ViT-BERT~\cite{li2021towards}, and PolyViT~\cite{likhosherstov2022polyvit} have explored the possibility of using the same unimodal encoder architecture for different inputs across unimodal tasks (i.e., language, image, video, or audio-only). The Transformer architecture has emerged as a popular choice due to its suitability for serialized inputs such as text (sequence of tokens)~\cite{devlin2019bert}, images (sequence of patches)~\cite{dosovitskiy2020image}, video (sequence of images)~\cite{sun2019videobert}, and other time-series data (sequence of timesteps)~\cite{lim2021temporal}.
There have also been several attempts to build a single model that works well on a suite of multimodal tasks, including both not limited to HighMMT~\cite{liang2022highmmt}, VATT~\cite{akbari2021vatt}, FLAVA~\cite{singh2022flava}, and Gato~\citep{reed2022generalist}.

\textbf{Transfer learning}: While more research has focused on transfer within the same modality with external information~\cite{socher2013zero,NIPS2019_8731,zadeh2020foundations},~\citet{liang2021cross} studies transfer to new modalities using small amounts of paired but unlabeled data.~\citet{lu2021pretrained} found that Transformers pretrained on language transfer to other sequential modalities as well.~\citet{liang2022highmmt} builds a single multimodal model capable of transferring to completely new modalities and tasks. Recently, there has also been a line of work investigating the transfer of pretrained language models for planning~\cite{huang2022language}, interactive decision-making~\cite{li2022pre}, and robotics~\cite{brohan2023rt}.

\vspace{-2mm}
\subsection{Subchallenge 5b: Multimodal co-learning}
\label{sec:transference2}
\vspace{-1mm}

Multimodal co-learning aims to transfer information learned through secondary modalities to target tasks involving the primary modality by sharing intermediate representation spaces between both modalities. These approaches essentially result in a single joint model across all modalities.

\textbf{Co-learning via representation} aims to learn a joint or coordinated representation space using both modalities as input. Typically, this involves adding secondary modalities during the training process, designing a suitable representation space, and investigating how the multimodal model transfers to the primary modality during testing. For example, DeViSE learns a coordinated space between image and text to improve image classification~\cite{frome2013devise}.~\citet{marino2016more} use knowledge graphs for image classification via a graph-based joint representation.~\citet{jia2021scaling} improve image classifiers with contrastive learning between images and noisy captions.
Finally,~\citet{zadeh2020foundations} showed that implicit co-learning is also possible without explicit co-learning objectives.

\textbf{Co-learning via generation} instead learns a translation model from  the primary to secondary modality, resulting in enriched representations of the primary modality that can predict both the label and `hallucinate' secondary modalities containing shared information. Classic examples in this category includes language modeling by mapping contextualized text embeddings into images~\citep{tan2020vokenization}, image classification by projecting image embeddings into word embeddings~\citep{socher2013zero}, and language sentiment analysis by translating language into video and audio~\citep{pham2019found}.

\vspace{-2mm}
\subsection{Subchallenge 5c: Model induction}
\label{sec:transference3}
\vspace{-1mm}

In contrast to co-learning, model induction approaches keep individual unimodal models across primary and secondary modalities separate but transfer information across them. There are two general ways of doing so. The first is co-training, where each unimodal model's predictions on their own modality are used to pseudo-label new unlabeled examples in the other modality, thereby enlarging the training set of the other modality~\cite{blum1998combining}. The second is co-regularization~\citep{sindhwani2005co,sridharan2008information}, in which the predictions from separate unimodal classifiers are regularized to be similar, thereby encouraging both classifiers to share information (i.e., redundancy). Therefore, information is transferred across modalities through model predictions instead of shared representation spaces.

\textbf{Multimodal co-training} extends co-training by jointly learning classifiers for multiple modalities~\cite{hinami2018multimodal}.~\citet{guillaumin2010multimodal} use a classifier on both image and text to pseudo-label unlabeled images before training a final classifier on both labeled and unlabeled images.~\citet{cheng2016semi} performs semi-supervised multimodal learning using a diversity-preserving co-training algorithm. Finally,~\citet{dunnmon2020cross} applies ideas from data programming to the problem of cross-modal weak supervision, where weak labels derived from a secondary modality (e.g., text) are used to train models over the primary modality (e.g., images).

\textbf{Co-regularization} methods employs a regularizer that penalizes functions from either modality that disagree with each other. These methods are useful in controlling model complexity by preferring hypothesis classes with redundancy across the two modalities~\citep{sindhwani2005co}.~\citet{sridharan2008information} provide guarantees for these approaches using an information-theoretic framework. More recently, similar co-regularization approaches have also been applied for multimodal feature selection~\citep{hsieh2019adaptive}, semi-supervised multimodal learning~\citep{yang2019comprehensive}, and video summarization~\citep{morere2015co}.
\begin{figure}[t]
\centering
\vspace{-0mm}
\includegraphics[width=0.7\linewidth]{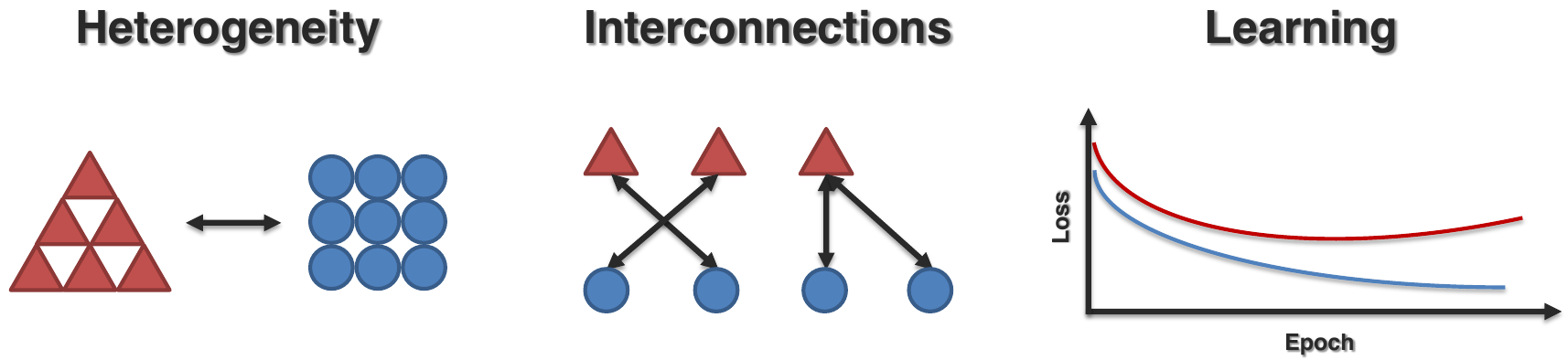}
\caption{\textbf{Quantification}: what are the empirical and theoretical studies we can design to better understand (1) the dimensions of \textit{heterogeneity}, (2) the presence and type of \textit{interconnections}, and (3) the \textit{learning} and optimization challenges?}
\label{fig:quant}
\vspace{-2mm}
\end{figure}

\vspace{-2mm}
\section{Challenge 6: Quantification}
\label{sec:quantification}
\vspace{-1mm}

Quantification aims to provide a deeper empirical and theoretical study of multimodal models to gain insights and improve their robustness, interpretability, and reliability in real-world applications. We break down quantification into 3 sub-challenges: (1) quantifying the \textit{dimensions of heterogeneity} and how they subsequently influence modeling and learning, (2) quantifying the presence and type of \textit{connections and interactions} in multimodal datasets and trained models, and (3) characterizing the \textit{learning and optimization} challenges involved when learning from heterogeneous data (Figure~\ref{fig:quant}).

\vspace{-2mm}
\subsection{Subchallenge 6a: Dimensions of heterogeneity}
\label{sec:quantification1}
\vspace{-1mm}

\begin{figure}[t]
\centering
\vspace{-0mm}
\includegraphics[width=0.5\linewidth]{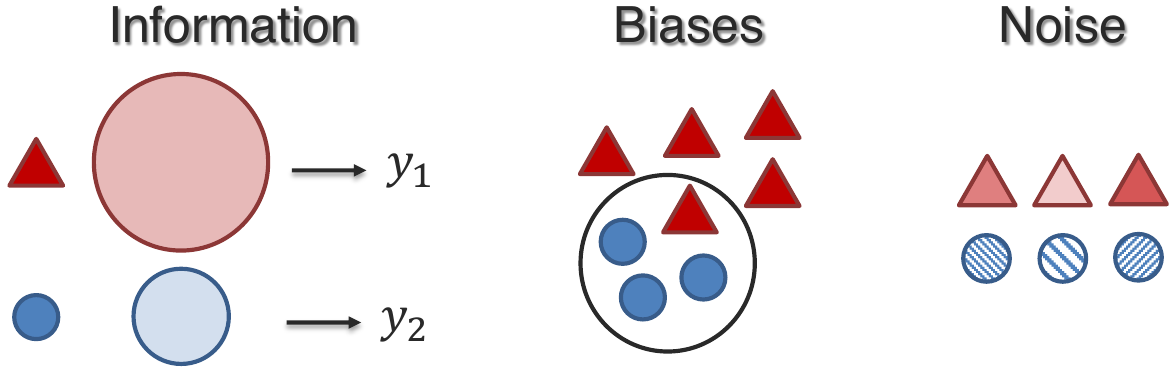}
\caption{The subchallenge of \textbf{heterogeneity} quantification aims to understand the dimensions of heterogeneity commonly encountered in multimodal research, such as (1) different quantities and usages of \textit{modality information}, (2) the presence of \textit{modality biases}, and (3) quantifying and mitigating \textit{modality noise}.}
\label{fig:quant1}
\vspace{-2mm}
\end{figure}

This subchallenge aims to understand the dimensions of heterogeneity commonly encountered in multimodal research, and how they subsequently influence modeling and learning (Figure~\ref{fig:quant1}).

\textbf{Modality information}: Understanding the information of modalities and their constituents is important for determining which parts contributed to subsequent modeling. Recent work can be categorized into (1) interpretable methods that explicitly model how each modality is used~\cite{park2018multimodal,tsai2020multimodal,zadeh2018multimodal} or (2) post-hoc explanations of black-box models~\cite{chandrasekaran2018explanations,goyal2016towards}. In the former, methods such as Concept Bottleneck Models~\cite{koh2020concept} and fitting sparse linear layers~\cite{wong2021leveraging} or decision trees~\cite{wan2020nbdt} on top of deep feature representations have emerged as promising choices.
In the latter, gradient-based visualizations~\cite{simonyan2013deep,goyal2016towards,selvaraju2017grad}) and feature attributions (e.g., modality contribution~\citep{gat2021perceptual}, LIME~\cite{ribeiro2016should}, and Shapley values~\cite{merrick2020explanation}) have been used to highlight regions of modality importance.

\textbf{Modality biases} are unintended correlations between input and outputs that could be introduced during data collection~\cite{birhane2021multimodal,bolukbasi2016man}, modeling~\cite{geirhos2020shortcut}, or during human annotation~\cite{devillers2005challenges}. Modality biases can lead to unexpectedly poor performance in the real world~\cite{sakaguchi2020winogrande}, or even more dangerously, potential for harm towards underrepresented groups~\citep{pena2020faircvtest,hendricks2018women}. For example,~\citet{goyal2017making} found \textit{unimodal biases} in the language modality of VQA tasks, resulting in mistakes due to ignoring visual information~\cite{agrawal2016analyzing}. Subsequent work has developed carefully curated diagnostic benchmarks to mitigate data collection biases, like VQA 2.0~\cite{goyal2017making}, GQA~\cite{hudson2019gqa}, and NLVR2~\cite{suhr2019nlvr2}. Recent work has also found compounding \textit{social biases} in multimodal systems~\cite{ross2020measuring,srinivasan2021worst,cho2022dall} stemming from gender bias in both language and visual modalities~\citep{buolamwini2018gender,sheng2019woman}, which may cause danger when deployed~\citep{pena2020faircvtest}.

\textbf{Modality noise topologies and robustness}: The study of modality noise topologies aims to benchmark and improve how multimodal models perform in the presence of real-world data imperfections. Each modality has a unique noise topology, which determines the distribution of noise and imperfections that it commonly encounters. For example, images are susceptible to blurs and shifts, typed text is susceptible to typos following keyboard positions, and multimodal time-series data is susceptible to correlated imperfections across synchronized time steps.~\citet{liang2021multibench} collect a comprehensive set of targeted noisy distributions unique to each modality. In addition to natural noise topologies~\cite{ma2022multimodal,lee2023multimodal}, related work has also explored adversarial attacks~\cite{ding2021multimodal} and distribution shifts~\cite{foltyn2021towards} in multimodal systems. Finally, there have been recent efforts on incomplete multimodal learning~\citep{liang2022highmmt,ma2021smil,wang2020multimodal,yang2018semi} to account for noisy or missing modalities, such as modality imputation using probabilistic models~\cite{ma2021smil}, autoencoders~\citep{DBLP:conf/cvpr/Tran0ZJ17}, translation models~\cite{pham2019found}, low-rank approximations~\cite{liang2019tensor}, or knowledge distillation~\citep{wang2020multimodal}, or training general models with a wide range of modalities so they can still operate on partial subsets~\citep{liang2022highmmt,reed2022generalist}. However, they may run the risk of possible error compounding and require knowing which modalities are imperfect beforehand.

\vspace{-2mm}
\subsection{Subchallenge 6b: Modality interconnections}
\label{sec:quantification2}
\vspace{-1mm}

\begin{figure}[t]
\centering
\vspace{-0mm}
\includegraphics[width=0.5\linewidth]{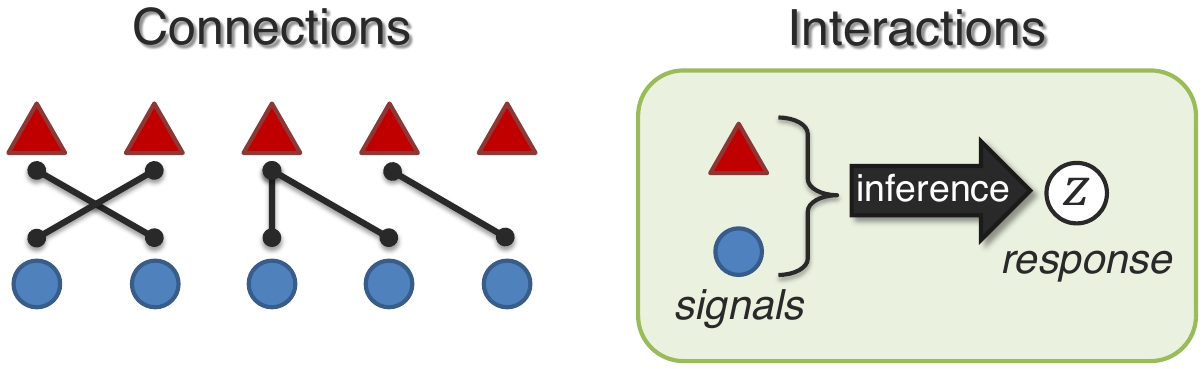}
\caption{Quantifying \textbf{modality interconnections} studies (1) \textit{connections}: can we discover what modality elements are related to each other and why, and (2) \textit{interactions}: can we understand how modality elements interact during inference?}
\label{fig:quant2}
\vspace{-2mm}
\end{figure}

Modality connections and interactions are an essential component of multimodal models, which has inspired an important line of work in visualizing and understanding the nature of modality interconnections in datasets and trained models. We divide recent work into quantifying (1) \textit{connections}: how modalities are related and share commonality, and (2) \textit{interactions}: how modality elements interact during inference (Figure~\ref{fig:quant2}).

\textbf{Connections}: Recent work has explored the quantification of modality connections through visualization tools on joint representation spaces~\citep{Itkina2020EvidentialSO} or attention maps~\citep{aflalo2022vl}. Perturbation-based analysis perturbs the input and observes changes in the output to understand internal connections~\citep{liang2023multiviz,niu2021counterfactual}. Finally, specifically curated diagnostic datasets are also useful in understanding semantic connections: Winoground~\citep{thrush2022winoground} probes vision and language models for visio-linguistic compositionality, and PaintSkills~\citep{cho2022dall} measures the connections necessary for visual reasoning.

\textbf{Interactions}: One common categorization of interactions involves redundancy, uniqueness, and synergy~\citep{williams2010nonnegative}. Redundancy describes task-relevant information shared among features, uniqueness studies the task-relevant information present in only one of the features, and synergy investigates the emergence of new information when both features are present. From a statistical perspective, measures of redundancy include mutual information~\cite{blum1998combining,balcan2004co} and contrastive learning estimators~\cite{tosh2021contrastive,tsai2020self}. Other approaches have studied these measures in isolation, such as redundancy via distance between prediction logits using either feature~\cite{mazzetto21a}, statistical distribution tests on input features~\cite{auffarth2010comparison}, or via human annotations~\cite{ruiz2006examining}.
From the semantic view, recent work in Causal VQA~\cite{agarwal2020towards} and Counterfactual VQA~\cite{niu2021counterfactual} seek to understand the interactions captured by trained models by measuring their robustness under controlled semantic edits to the question or image. Finally, recent work has formalized definitions of non-additive interactions to quantify their presence in trained models~\citep{sorokina2008detecting,tsang2018detecting,xue2022modality}. Parallel research such as EMAP~\cite{hessel2020does}, DIME~\cite{lyu2022dime}, M2Lens~\cite{wang2021m2lens}, and MultiViz~\cite{liang2023multiviz} take a more visual approach to visualize the interactions in real-world multimodal datasets and models through higher-order gradient activations of learned representations. Despite this, accurately visualizing multimodal information and interactions remains a challenge due to the brittleness of interpretation methods~\citep{ghorbani2019interpretation}, difficulty in evaluation~\citep{krishna2022disagreement}, and challenges in extending visualization methods to applications such as biomedical data integration, imaging, intelligent systems and user interfaces.

\vspace{-2mm}
\subsection{Subchallenge 6c: Multimodal learning process}
\label{sec:quantification3}
\vspace{-1mm}

\begin{figure}[t]
\centering
\vspace{-0mm}
\includegraphics[width=0.7\linewidth]{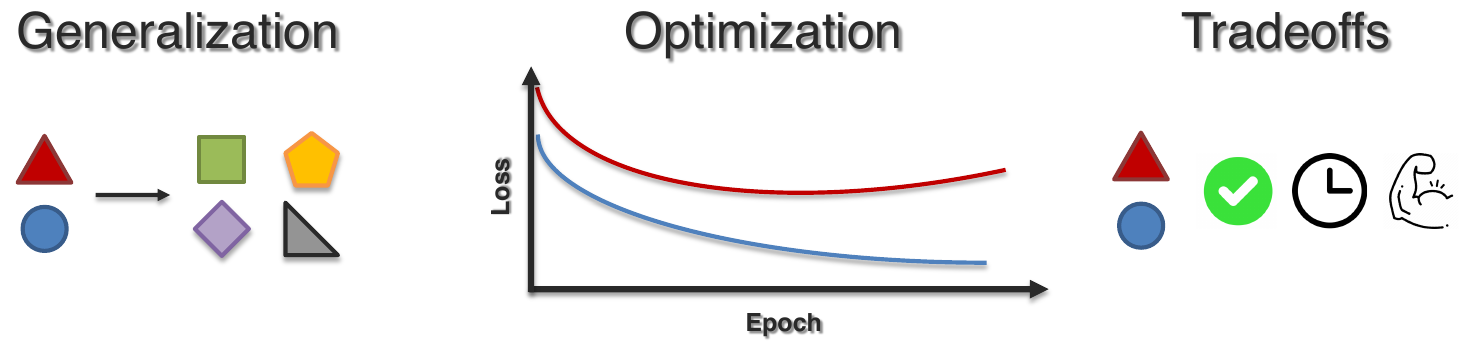}
\caption{Studying the multimodal \textbf{learning process} involves understanding (1) \textit{generalization} across modalities and tasks, (2) \textit{optimization} for balanced and efficient training, and (3) \textit{tradeoffs} between performance, robustness, and complexity in the real-world deployment of multimodal models.}
\label{fig:quant3}
\vspace{-2mm}
\end{figure}

Finally, there is a need to characterize the learning and optimization challenges involved when learning from heterogeneous data. This section covers recent work in (1) \textit{generalization} across modalities and tasks, (2) better \textit{optimization} for balanced and efficient training, and (3) balancing the \textit{tradeoffs} between performance, robustness, and complexity in real-world deployment (Figure~\ref{fig:quant3}).

\textbf{Generalization}: With advances in sensing technologies, many real-world platforms such as cellphones, smart devices, self-driving cars, healthcare technologies, and robots now integrate a much larger number of sensors beyond the prototypical text, video, and audio modalities~\cite{huang2019multimodal}. Recent work has studied generalization across paired modality inputs~\citep{liang2021cross,radford2021learning} and in unpaired scenarios where each task is defined over only a small subset of all modalities~\cite{liang2022highmmt,lu2021pretrained,reed2022generalist}.

\textbf{Optimization challenges}: Related work has also explored the optimization challenges of multimodal learning, where multimodal networks are often prone to overfitting due to increased capacity, and different modalities overfit and generalize at different rates so training them jointly with a single optimization strategy is sub-optimal~\cite{wang2020makes}. Subsequent work has studied why joint training of multimodal networks may be difficult and proposed methods to improve the optimization process via weighting approaches~\citep{,wu2022characterizing}, adaptive learning~\citep{huang2021makes,huang2022modality}, or contrastive learning~\cite{liang2022mind}.

\textbf{Modality Tradeoffs}: In real-world deployment, a balance between performance, robustness, and complexity is often required. Therefore, one often needs to balance the utility of additional modalities with the additional complexity in data collection and modeling~\cite{liang2021multibench} as well as increased susceptibility to noise and imperfection in the additional modality~\cite{pham2019found}.
How can we formally quantify the utility and risks of each input modality, while balancing these tradeoffs for reliable real-world usage? There have been several attempts toward formalizing the semantics of a multimodal representation and how these benefits can transfer to downstream tasks~\cite{liang2022brainish,tsai2020self,thomason2016learning}, while information-theoretic arguments have also provided useful insights~\cite{blum1998combining,sridharan2008information}.

\chapter{Machine Learning Foundations of Multimodal Interactions}
\label{chap:foundations1}
\newcommand{\name}{\textsc{PID}}

\newcommand{\gmmcartesian}{\textsf{Cartesian}}
\newcommand{\gmmpolar}{\textsf{Polar}}
\newcommand\fplus{\ding{58}}
\newcommand\fcross{\ding{54}}
\newcommand{\gmmcolR}[1]{\textcolor{blue}{#1}}
\newcommand{\gmmcolUone}[1]{\textcolor{darkpastelgreen}{#1}}
\newcommand{\gmmcolUtwo}[1]{\textcolor{red}{#1}}
\newcommand{\gmmcolS}[1]{\textcolor{orange}{#1}}
\newcommand{\gmmcolTot}[1]{\textcolor{amethyst}{#1}}

\newcommand{\esta}{\textsc{CVX}}
\newcommand{\estb}{\textsc{Batch}}

\newcommand{\red}{$R$}
\newcommand{\uone}{$U_1$}
\newcommand{\utwo}{$U_2$}
\newcommand{\syn}{$S$}

\newcommand{\acc}{Acc}

A core challenge in machine learning lies in capturing the interactions between multiple input modalities. Learning different types of multimodal interactions is often quoted as motivation for many successful multimodal modeling paradigms, such as contrastive learning to capture redundancy~\cite{kim2021vilt,radford2021learning}, modality-specific representations to retain unique information~\cite{tsai2019learning}, as well as tensors and multiplicative interactions to learn higher-order interactions~\cite{jayakumar2020multiplicative,liang2019tensor,zadeh2017tensor}. However, several fundamental research questions remain: \textit{How can we quantify the interactions that are necessary to solve a multimodal task?} \textit{Subsequently, what are the most suitable multimodal models to capture these interactions?}
This paper aims to formalize these questions by proposing an approach to quantify the \textit{nature} (i.e., which type) and \textit{degree} (i.e., the amount) of modality interactions, a fundamental principle underpinning our understanding of multimodal datasets and models~\cite{liang2022foundations}.

\begin{wrapfigure}{R}{0.52\textwidth}
    \begin{minipage}{0.52\textwidth}
    \vspace{-6mm}
    \includegraphics[width=\linewidth]{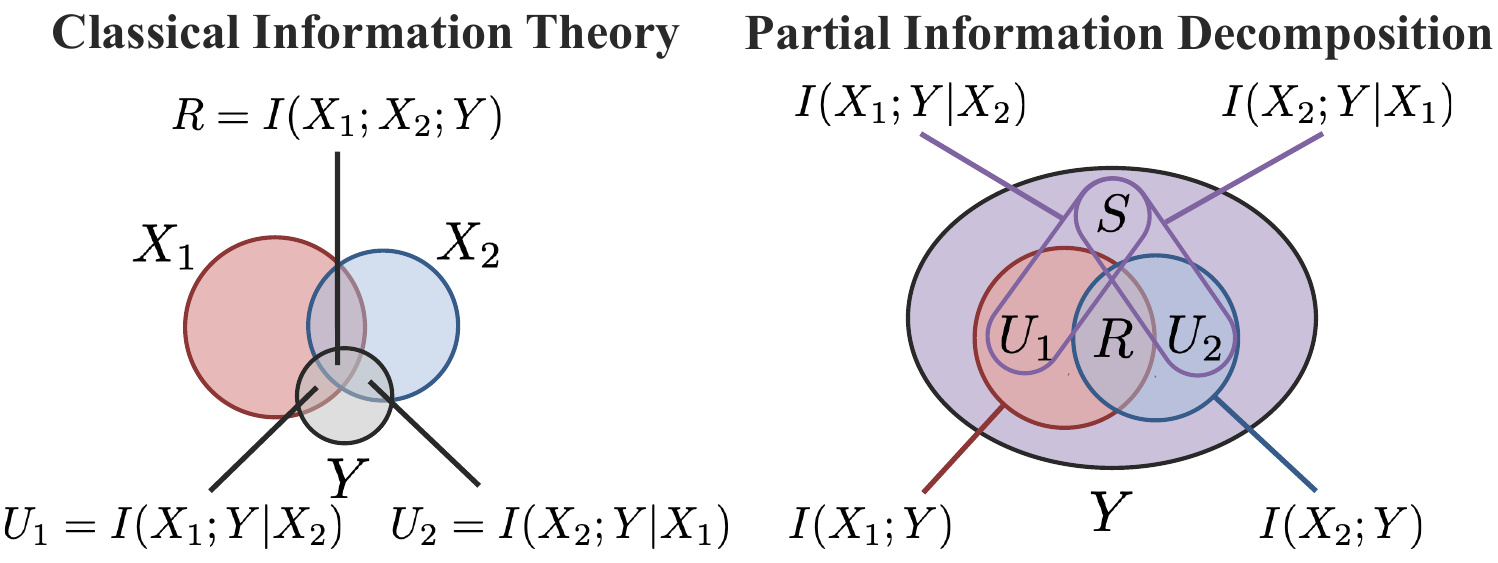}
    \vspace{-4mm}
    \caption{PID decomposes $I(X_1,X_2;Y)$ into redundancy $R$ between $X_1$ and $X_2$, uniqueness $U_1$ in $X_1$ and $U_2$ in $X_2$, and synergy $S$ in both $X_1$ and $X_2$.}
    \label{fig:info}
    \end{minipage}
\vspace{1mm}
\end{wrapfigure}

By bringing together two previously disjoint research fields of Partial Information Decomposition (PID) in information theory~\cite{bertschinger2014quantifying,griffith2014quantifying,williams2010nonnegative} and multimodal machine learning~\cite{baltruvsaitis2018multimodal,liang2022foundations}, we provide precise definitions categorizing interactions into \textit{redundancy}, \textit{uniqueness}, and \textit{synergy}. Redundancy quantifies information shared between modalities, uniqueness quantifies the information present in only one of the modalities, and synergy quantifies the emergence of new information not previously present in either modality.
A key aspect of these four measures is that they not only quantify interactions between modalities, but also how they relate to a downstream task. Figure~\ref{fig:info} shows a depiction of these four measures, which we refer to as \pid\ statistics.
Leveraging insights from neural representation learning, we propose two new estimators for \pid\ statistics that can scale to high-dimensional multimodal datasets and models.
The first estimator is exact, based on convex optimization, and is able to scale to features with discrete support, while the second estimator is an approximation based on sampling, which enables us to handle features with large discrete or even continuous supports. We validate our estimation of \pid\ in $2$ ways: (1) on synthetic datasets where \pid\ statistics are known from the nature of data generation, and (2) on real-world data where \pid\ is compared with human annotation.
Finally, we demonstrate that estimated \pid\ statistics can help in multimodal applications involving:
\begin{enumerate}[noitemsep,topsep=0pt,nosep,leftmargin=*,parsep=0pt,partopsep=0pt]
    \item \textbf{Dataset quantification}: We apply \pid\ to quantify large-scale multimodal datasets, showing that these estimates match common intuition for interpretable modalities (e.g., language, vision, and audio) and yield new insights in other domains (e.g, healthcare, HCI, and robotics).

    \item \textbf{Model quantification}: Across a suite of models, we apply \pid\ to interpret model predictions and find consistent patterns of interactions that different models capture.

    \item \textbf{Model selection}: Given our findings from dataset and model quantification, a new research question naturally arises: \textit{given a new multimodal task, can we quantify its \pid\ values to infer (a priori) what type of models are most suitable?} Our experiments show success in model selection for both existing benchmarks and completely new case studies engaging with domain experts in computational pathology, mood prediction, and robotics to select the best multimodal model.
\end{enumerate}

Finally, we release a suite of trained models across $10$ model families and $30$ datasets to accelerate future analysis of multimodal interactions at \url{https://github.com/pliang279/PID}.

\vspace{-2mm}
\section{\mbox{Background and Related Work}}

Let $\mathcal{X}_i$ and $\mathcal{Y}$ be sample spaces for features and labels.
Define $\Delta$ to be the set of joint distributions over $(\mathcal{X}_1, \mathcal{X}_2, \mathcal{Y})$.
We are concerned with features $X_1, X_2$ (with support $\mathcal{X}_i$) and labels $Y$ (with support $\mathcal{Y}$) drawn from some distribution $p \in \Delta$.
We denote the probability mass (or density) function by $p(x_1,x_2,y)$, where omitted parameters imply marginalization.
Key to our work is defining estimators that given $p$ or samples $\{(x_1,x_2,y): \mathcal{X}_1 \times \mathcal{X}_2 \times \mathcal{Y}\}$ thereof (i.e., dataset or model predictions), estimates the amount of redundant, unique, and synergistic interactions.

\subsection{\mbox{Partial information decomposition}}

Information theory formalizes the amount of information that one variable provides about another~\cite{shannon1948mathematical}. However, its extension to $3$ variables is an open question~\cite{garner1962uncertainty,mcgill1954multivariate,te1980multiple,watanabe1960information}.
In particular, the natural three-way mutual information $I(X_1; X_2; Y) = I(X_1;X_2) - I(X_1;X_2|Y)$~\cite{mcgill1954multivariate,te1980multiple} can be both positive and negative, which makes it difficult to interpret. In response, Partial information decomposition (PID)~\cite{williams2010nonnegative} generalizes information theory to multiple variables by decomposing $I_p(X_1,X_2; Y)$, the total information $2$ variables $X_1,X_2$ provide about a task $Y$ into 4 quantities (see Figure~\ref{fig:info}): redundancy $R$ between $X_1$ and $X_2$, uniqueness $U_1$ in $X_1$ and $U_2$ in $X_2$, and synergy $S$ that only emerges when both $X_1$ and $X_2$ are present.
We adopt the \pid\ definition proposed by~\citet{bertschinger2014quantifying}:
{
\begin{align}
R &= \max_{q \in \Delta_p} I_q(X_1; X_2; Y), \label{eqn:R-def}\\
U_1 &= \min_{q \in \Delta_p} I_q(X_1; Y | X_2), \quad U_2 = \min_{q \in \Delta_p} I_q(X_2; Y| X_1), \label{eqn:U2-def}\\
S &= I_p(X_1,X_2; Y) - \min_{q \in \Delta_p} I_q(X_1,X_2; Y), \label{eqn:S-def}
\end{align}
}where $\Delta_p = \{ q \in \Delta: q(x_i,y)=p(x_i,y) \ \forall y\in\mathcal{Y}, x_i \in \mathcal{X}_i, i \in [2] \}$ and the notation $I_p(\cdot)$ and $I_q(\cdot)$ disambiguates mutual information under $p$ and $q$ respectively. The key lies in optimizing $q \in \Delta_p$ to satisfy the marginals $q(x_i,y)=p(x_i,y)$, but relaxing the coupling between $x_1$ and $x_2$: $q(x_1,x_2)$ need not be equal to $p(x_1,x_2)$. The intuition behind this is that one should be able to infer redundancy and uniqueness given only access to $p(x_1,y)$ and $p(x_2,y)$, and therefore they should only depend on $q \in \Delta_p$. Synergy is the only term that should depend on the coupling $p(x_1,x_2)$, and this is reflected in (\ref{eqn:S-def}) depending on the full $p$ distribution. This definition enjoys several useful properties in line with intuition, as we will see in comparison with related frameworks for interactions below~\cite{bertschinger2014quantifying}.

\subsection{\mbox{Related frameworks for feature interactions}}

\textbf{Information-theoretic definitions}: Perhaps the first measure of redundancy in machine learning is co-training~\cite{blum1998combining,balcan2004co,christoudias2008multi}, where 2 variables are redundant if they are conditionally independent given the task: $I(X_1; X_2 | Y)=0$. As a result, redundancy can be measured by $I(X_1; X_2; Y)$. The same definition of redundancy is used in multi-view learning~\cite{tosh2021contrastive,tian2020makes,sridharan2008information} which further define $I(X_1; Y | X_2)$ and $I(X_2; Y | X_1)$ as unique information in $X_1,X_2$. However, $I(X_1; X_2; Y)$ can be both positive and negative~\cite{jakulin2003quantifying}. PID resolves this by separating $R$ and $S$ such that $R-S = I(X_1; X_2; Y)$, identifying that prior measures confound redundancy and synergy. This crucially provides an explanation for the distinction between \textit{mediation}, where one feature conveys the information already in another (i.e., $R>S$), versus \textit{moderation}, where one feature affects the relationship of other features (i.e., $S>R$)~\cite{baron1986moderator,ghassami2017interaction}. Furthermore, if $I(X_1; X_2; Y) = 0$ then existing frameworks are unable to distinguish between positive $R$ and $S$ canceling each other out.

\textbf{Statistical measures}: Other approaches have studied interaction measures via statistical measures, such as redundancy via distance between prediction logits using either feature~\cite{mazzetto21a}, statistical distribution tests on input features~\cite{yu2004efficient,auffarth2010comparison}, or via human annotations~\cite{ruiz2006examining}. However, it is unclear how to extend these definitions to uniqueness and synergy while remaining on the same standardized scale like PID provides. Also of interest are notions of redundant and synergistic interactions in human and animal communication~\cite{partan1999communication,partan2005issues,flom2007development,ruiz2006examining}, which we aim to formalize.

\textbf{Model-based methods}: Prior research has formalized definitions of non-additive interactions~\cite{friedman2008predictive} to quantify their presence~\cite{sorokina2008detecting,tsang2018detecting,tsang2019feature,hessel2020does} in trained models, or used Shapley values on trained features to measure interactions~\cite{ittner2021feature}. Parallel research has also focused on qualitative visualizations of real-world multimodal datasets and models, such as DIME~\cite{lyu2022dime}, M2Lens~\cite{wang2021m2lens}, and MultiViz~\cite{liang2023multiviz}.

\vspace{-2mm}
\section{Scalable Estimators for \pid}
\vspace{-1mm}

\textbf{\pid\ as a framework for multimodality}: Our core insight is that PID provides a formal framework to understand both the \textit{nature} and \textit{degree} of interactions involved when two features $X_1$ and $X_2$ are used for task $Y$. The nature of interactions is afforded by a precise decomposition into redundant, unique, and synergistic interactions, and the degree of interactions is afforded by a standardized unit of measure (bits).
However, computing \pid\ is a considerable challenge, since it involves optimization over $\Delta_p$ and estimating information-theoretic measures. Up to now, analytic approximations of these quantities were only possible for discrete and small support~\cite{bertschinger2014quantifying,griffith2014quantifying,wollstadt2019idtxl} or continuous but low-dimensional variables~\cite{pakman2021estimating,proca2022synergistic,wollstadt2021rigorous}. Leveraging ideas in representation learning, Sections~\ref{sec:est1} and~\ref{sec:est2} are our first technical contributions enabling scalable estimation of \pid\ for high-dimensional distributions. The first, \esta, is exact, based on convex optimization, and is able to scale to problems where $|\mathcal{X}_i|$ and $|\mathcal{Y}|$ are around $100$. The second, \estb, is an approximation based on sampling, which enables us to handle large or even continuous supports for $X_i$ and $Y$. Applying these estimators in Section~\ref{sec:applications}, we show that \pid\ provides a path towards understanding the nature of interactions in datasets and those learned by different models, and principled approaches for model selection.

\subsection{\esta: Dataset-level optimization}
\label{sec:est1}

Our first estimator, \esta, directly compute \pid\ from its definitions using convex programming. Crucially,~\citet{bertschinger2014quantifying} show that the solution to the max-entropy optimization problem: $q^* = \argmax_{q \in \Delta_p} H_q(Y | X_1, X_2)$ equivalently solves (\ref{eqn:R-def})-(\ref{eqn:S-def}).
When $\mathcal{X}_i$ and $\mathcal{Y}$ are small and discrete, we can represent all valid distributions $q(x_1,x_2,y)$ as a set of tensors $Q$ of shape $|\mathcal{X}_1| \times |\mathcal{X}_2| \times |\mathcal{Y}|$ with each entry representing $Q[i,j,k] = p(X_1=i,X_2=j,Y=k)$. The problem then boils down to optimizing over valid tensors $Q \in \Delta_p$ that match the marginals $p(x_i,y)$.

Given a tensor $Q$ representing $q$, our objective is the concave function $H_q(Y | X_1, X_2)$. While~\citet{bertschinger2014quantifying} report that direct optimization is numerically difficult as routines such as Mathematica's \textsc{FindMinimum} do not exploit convexity, we overcome this by rewriting conditional entropy as a KL-divergence~\cite{globerson2007approximate}, $H_q(Y|X_1, X_2) = \log |\mathcal{Y}| - KL(q||\tilde{q})$, where $\tilde{q}$ is an auxiliary product density of $q(x_1,x_2) \cdot \frac{1}{|\mathcal{Y}|}$ enforced using linear constraints: $ \tilde{q}(x_1, x_2, y) = q(x_1,x_2) / |\mathcal{Y}|$. Finally, optimizing over $Q \in \Delta_p$ that match the marginals can also be enforced through linear constraints: the 3D-tensor $Q$ summed over the second dimension gives $q(x_1,y)$ and summed over the first dimension gives $q(x_2,y)$, yielding the final optimization problem:
{\small
\begin{align}
    \argmax_{Q,\tilde{Q}} KL(Q||\tilde{Q}), \quad \textrm{s.t.} \quad &\tilde{Q}(x_1, x_2, y) = Q(x_1,x_2) / |\mathcal{Y}|, \\
    &\sum_{x_2} Q = p(x_1,y), \sum_{x_1} Q = p(x_2,y), Q \ge 0, \sum_{x_1,x_2,y} Q = 1.
    \label{eqn:cvx-optimizer}
\end{align}
}The KL-divergence objective is recognized as convex, allowing the use of conic solvers such as SCS \cite{ocpb:16}, ECOS \cite{domahidi2013ecos}, and MOSEK \cite{mosek}.
Plugging $q^*$ into (\ref{eqn:R-def})-(\ref{eqn:S-def}) yields the desired PID.

\textbf{Pre-processing via feature binning}: In practice, $X_1$ and $X_2$ often take continuous rather than discrete values. Thus, $Q$ is no longer a finite dimensional polytope. We work around this by histogramming each $X_i$, thereby estimating the continuous joint density by discrete distributions with finite support.
To make our discretization as data-independent as possible, we focus on a prespecified number of fixed-width bins (except for the first and last). For example, it is known that with a fixed number of samples, making the width of bins arbitrarily small will cause KL estimates to diverge. It is known that the number of bins should grow sub-linearly with the number of samples. For example,~\citet{rice2006mathematical} suggest setting the number of bins to be the cubed-root of number of samples.

\subsection{\mbox{\estb: Batch-level amortization}}
\label{sec:est2}

We now present \estb, our next estimator that is suitable for large datasets where $\mathcal{X}_i$ is high-dimensional and continuous ($|\mathcal{Y}|$ remains finite). To estimate \pid\ given a sampled dataset $\mathcal{D} = \{ (x_1^{(j)}, x_2^{(j)}, y^{(j)})\}$ of size $n$, we propose an end-to-end model parameterizing marginal-matching joint distributions in $\Delta_p$ and a training objective whose solution returns approximate \pid\ values.

\begin{figure*}[t]
\includegraphics[width=\textwidth]{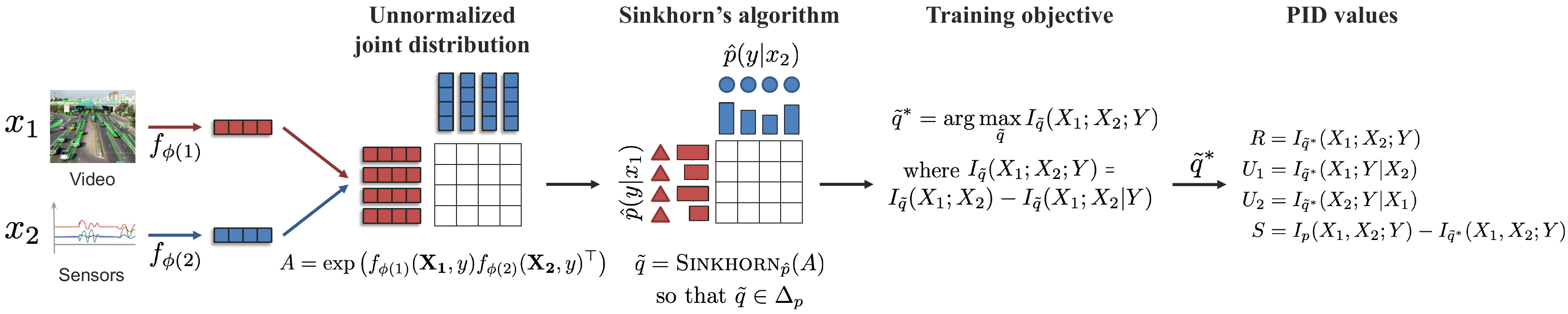}
\vspace{-2mm}
\caption{We propose \estb, a scalable estimator for \pid\ over high-dimensional continuous distributions. \estb\ parameterizes $\tilde{q}$ using a matrix $A$ learned by neural networks such that mutual information objectives over $\tilde{q}$ can be optimized via gradient-based approaches over minibatches. Marginal constraints $\tilde{q} \in {\Delta}_p$ are enforced through a variant of the Sinkhorn-Knopp algorithm on $A$.}
\label{fig:est2_main}
\vspace{-2mm}
\end{figure*}

\textbf{Simplified algorithm sketch}: Our goal, loosely speaking, is to optimize $\tilde{q} \in {\Delta}_p$ for objective (\ref{eqn:R-def}) through an approximation using neural networks instead of exact optimization. We show an overview in Figure~\ref{fig:est2_main}.
To explain our approach, we first describe (1) how we parameterize $\tilde{q}$ using neural networks such that it can be learned via gradient-based approaches, (2) how we ensure the marginal constraints $\tilde{q} \in {\Delta}_p$ through a variant of the Sinkhorn-Knopp algorithm, and finally (3) how to scale this up over small subsampled batches from large multimodal datasets.

\textbf{Parameterization using neural networks}: The space of joint distributions ${\Delta}$ is often too large to explicitly specify. To tackle this, we implicitly parameterize each distribution $\tilde{q} \in {\Delta}$ using a neural network $f_\phi$ that takes in batches of modalities $\mathbf{X_1} \in \widetilde{\mathcal{X}}_1^n, \mathbf{X_2} \in \widetilde{\mathcal{X}}_2^n$ and the label $\mathbf{Y} \in \mathcal{Y}^n$ before returning a matrix $A \in \mathbb{R}^{n \times n \times |\mathcal{Y}|}$ representing an (unnormalized) joint distribution $\tilde{q}$, i.e.,  we want $A[i][j][y] = \tilde{q}(\mathbf{X_1}[i], \mathbf{X_2}[j],y)$ for each $y \in \mathcal{Y}$. In practice, $f_\phi$ is implemented via a pair of encoders $f_{\phi(1)}$ and $f_{\phi(2)}$ that learn modality representations, before an outer product to learn joint relationships $A_y = \exp(f_{\phi(1)}(\mathbf{X_1},y) f_{\phi(2)}(\mathbf{X_2},y)^\top)$ for each $y$, yielding the desired $n \times n \times |\mathcal{Y}|$ joint distribution.
As a result, optimizing over $\tilde{q}$ can be performed via optimizing over parameters $\phi$.

\textbf{Respecting the marginal constraints}: How do we make sure the $\tilde{q}$'s learned by the network satisfies the marginal constraints (i.e., $\tilde{q} \in {\Delta}_p$)? We use an unrolled version of Sinkhorn's algorithm~\cite{cuturi2013sinkhorn} which projects $A$ onto ${\Delta}_p$ by iteratively normalizing $A$'s rows and columns to sum to $1$ and rescaling to satisfy the marginals $p(x_i,y)$. However, $p(x_i, y)$ is not easy to estimate for high-dimensional continuous $x_i$'s. In response, we first expand $p(x_i,y)$ into $p(y|x_i)$ and $p(x_i)$ using Bayes' rule. Since $A$ was constructed by samples $x_i$ from the dataset, the rows and columns of $A$ are already distributed according to $p(x_1)$ and $p(x_2)$ respectively. This means that it suffices to approximate $p(y|x_i)$ with unimodal classifiers $\hat{p}(y|x_i)$ parameterized by neural networks and trained separately, before using Sinkhorn's algorithm to normalize each row to $\hat p(y|x_1)$ and each column to $\hat p(y|x_2)$.

\textbf{Objective}: We choose the objective $I_q(X_1;X_2;Y)$, which equivalently solves the optimization problems in the other PID terms~\cite{bertschinger2014quantifying}. Given matrix $A$ representing $\tilde{q}(x_1,x_2,y)$, the objective can be computed in closed form through appropriate summation across dimensions in $A$ to obtain $\tilde{q}(x_i)$, $\tilde{q}(x_1,x_2)$, $\tilde{q}(x_i|y)$, and $\tilde{q}(x_1,x_2|y)$ and plugging into $I_{\tilde{q}}(X_1;X_2;Y) = I_{\tilde{q}}(X_1;X_2) - I_{\tilde{q}}(X_1;X_2|Y)$. We maximize $I_{\tilde{q}}(X_1;X_2;Y)$ by updating parameters $\phi$ via gradient-based methods.
Overall, each gradient step involves computing $\tilde{q} = \textsc{Sinkhorn}_{\hat{p}} (A)$, and updating $\phi$ to maximize (\ref{eqn:R-def}) under $\tilde{q}$.
Since Sinkhorn's algorithm is differentiable, gradients can be backpropagated end-to-end.

\textbf{Approximation with small subsampled batches}: Finally, to scale this up to large multimodal datasets where the full $\tilde{q}$ may be too large to store, we approximate $\tilde{q}$ with small subsampled batches: for each gradient iteration $t$, the network $f_\phi$ now takes in a batch of $m \ll n$ datapoints sampled from $\mathcal{D}$ and returns $A \in \mathbb{R}^{m \times m \times |\mathcal{Y}|}$ for the subsampled points. We perform Sinkhorn's algorithm on $A$ and a gradient step on $\phi$ as above, \textit{as if} $\mathcal{D}_t$ was the full dataset (i.e., mini-batch gradient descent).
While it is challenging to obtain full-batch gradients since computing the full $A$ is intractable, we found our approach to work in practice for large $m$.
Our approach can also be informally viewed as performing amortized optimization~\cite{amos2022tutorial} by using $\phi$ to implicitly share information about the full batch using subsampled batches.
Upon convergence of $\phi$, we extract \pid\ by plugging $\tilde{q}$ into (\ref{eqn:R-def})-(\ref{eqn:S-def}).

\textbf{Implementation details} such as the network architecture of $f$, approximation of objective (\ref{eqn:R-def}) via sampling from $\tilde{q}$, and estimation of $I_{\tilde{q}}(\{X_1,X_2\}; Y)$ from learned $\tilde{q}$ are included in the full version of the paper~\citep{liang2023quantifying}.

\begin{figure*}[t]
\includegraphics[width=.99\textwidth]{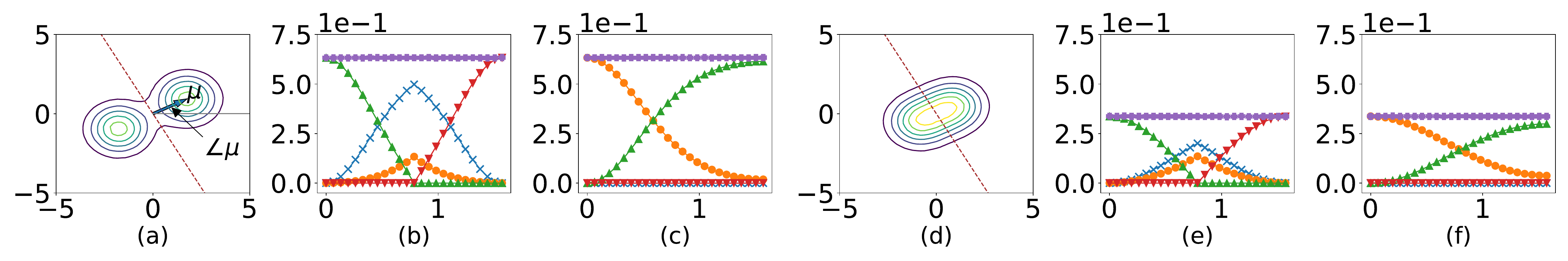}
\caption{Left to right: (a) Contour plots of the GMM's density for $||\mu||_2 = 2.0$. Red line denotes the optimal linear classifier. (b) \pid\ (\gmmcartesian) computed for varying $\angle \mu$ with respect to the $x$ axis. (c) \pid\ (\gmmpolar) for varying $\angle \mu$, with $U_1$ and $U_2$ corresponding to unique information from $(r, \theta)$. Plots (d)-(f) are similar to (a)-(c), but repeated for $||\mu||_2 = 1.0$.  
Legend: \gmmcolR{\fcross ($R$)}, \gmmcolUone{\FilledSmallTriangleUp ($U_1$)}, \gmmcolUtwo{\FilledSmallTriangleDown ($U_2$)}, \gmmcolS{\FilledSmallCircle ($S$)}, \gmmcolTot{\fplus (Sum)}. 
Observe how \pid\ changes with the change of variable from \gmmcartesian\ (b and e) to \gmmpolar\ (c and f), as well as how a change in $||\mu||_2$ can lead to a disproportionate change across \pid\ (b vs e).
}
\label{fig:gmm_intro_large}
\vspace{-2mm}
\end{figure*}

\vspace{-1mm}
\section{\mbox{Evaluation and Applications of \pid\ in Multimodal Learning}}
\label{sec:applications}

We design experiments to (1) understand \pid\ on synthetic data, (2) quantify real-world multimodal benchmarks, (3) understand the interactions captured by multimodal models, (4) perform model selection across different model families, and (5) applications on novel real-world tasks.

\subsection{\mbox{Validating \pid\ estimates on synthetic data}}

Our first goal is to evaluate the accuracy of our proposed estimators with respect to the ground truth (if it can be computed) or human judgment (for cases where the ground truth cannot be readily obtained).
We start with a suite of datasets spanning both synthetic and real-world distributions.

\begin{wraptable}{r}{8.6cm}
\centering
\fontsize{9}{11}\selectfont
\setlength\tabcolsep{2.0pt}
\vspace{-2mm}
\caption{Results on estimating \pid\ on synthetic bitwise datasets. Both our estimators exactly recover the correct \pid\ values as reported in~\citet{bertschinger2014quantifying}.}
\centering
\footnotesize

\begin{tabular}{l|cccc|cccc|cccc}
\hline \hline
Task & \multicolumn{4}{c|}{\textsc{OR}} & \multicolumn{4}{c|}{\textsc{AND}} & \multicolumn{4}{c}{\textsc{XOR}} \\
\hline
\pid & \red & \uone & \utwo & \syn & \red & \uone & \utwo & \syn & \red & \uone & \utwo & \syn \\ 
\hline
Exact & 0.31 & 0 & 0 & 0.5 & 0.31 & 0 & 0 & 0.5 & 0 & 0 & 0 & 1 \\
\esta & 0.31 & 0 & 0 & 0.5 & 0.31 & 0 & 0 & 0.5 & 0 & 0 & 0 & 1 \\
\estb & 0.31 & 0 & 0 & 0.5 & 0.31 & 0 & 0 & 0.5 & 0 & 0 & 0 & 1 \\
\hline \hline
\end{tabular}

\vspace{-2mm}
\label{tab:bits}
\end{wraptable}

\textbf{Synthetic bitwise features}: We sample from a binary bitwise distribution: $x_1, x_2 \sim \{0,1\}, y = x_1 \land x_2, y = x_1 \lor x_2, y = x_1 \oplus x_2, $. Each bitwise operator's \pid\ can be solved exactly when the $x_i$'s and labels are discrete and low-dimensional~\cite{bertschinger2014quantifying}.
Compared to the ground truth in~\citet{bertschinger2014quantifying}, both our estimators exactly recover the correct \pid\ values (Table~\ref{tab:bits}).

\textbf{Gaussian Mixture Models (GMM)}: Consider a GMM, where $X_1$, $X_2 \in \mathbb{R}$ and the label $Y \in \{-1,+1\}$, comprising two equally weighted standard multivariate Gaussians centered at $\pm\mu$, where $\mu \in \mathbb{R}^2$, i.e., $Y\sim \text{Bernoulli}(1/2)$, $(X_1, X_2)|Y=y \sim \mathcal{N}(y \cdot \mu, I)$.
\pid\ was estimated by sampling $1e6$ points, histogramming them into $50$ bins spanning $[-5,+5]$ to give $p$, and then applying the \esta\ estimator.
We term this \pid-\gmmcartesian. We also compute \pid-\gmmpolar, which are \pid\ computed using \textit{polar coordinates}, $(r, \theta)$. 
We use a variant where the angle $\theta$ is given by the arctangent with principal values $[0, \pi]$ and the length $r \in \mathbb{R}$ could be negative. $\theta$ specifies a line (through the origin), and $r$ tells us where along the line the datapoint lies on.

\textbf{Results:} We consider $||\mu||_2 \in \{1.0, 2.0 \}$, where for each $||\mu||_2$, we vary the angle $\angle \mu$ that $\mu$ makes with the horizontal axis. Our computed \pid\ is presented in Figure~\ref{fig:gmm_intro_large}. Overall, we find that the \pid\ matches what we expect from intuition. For \gmmcartesian, unique information dominates when the angle goes to $0$ or $\pi/2$ --- if centroids share a coordinate, then observing that coordinate yields no information about $y$. Conversely, synergy and redundancy peak at $\pi/4$. Interestingly, synergy seems to be independent of $||\mu||_2$. 
For \gmmpolar, redundancy is $0$. Furthermore, $\theta$ contains no unique information, since $\theta$ shows nothing about $y$ unless we know $r$ (in particular, its sign). When the angle goes to $\pi/2$, almost all information is unique in $r$. The distinctions between \gmmcartesian\ and \gmmpolar\ highlight how different representations of data can exhibit wildly different \pid\ values, even if total information is the same.

\begin{table*}[t]
\centering
\fontsize{9}{11}\selectfont
\setlength\tabcolsep{1.5pt}
\caption{Estimating \name\ on synthetic generative model datasets. Both \esta\ and \estb\ measures agree with each other on relative values and are consistent with ground truth interactions.}
\centering
\footnotesize

\begin{tabular}{l|cccc|cccc|cccc|cccc|cccc}
\hline \hline
Task & \multicolumn{4}{c|}{$\mathcal{D}_R$} & \multicolumn{4}{c|}{$\mathcal{D}_{U_1}$} & \multicolumn{4}{c|}{$\mathcal{D}_{U_2}$} & \multicolumn{4}{c|}{$\mathcal{D}_{S}$} & \multicolumn{4}{c}{$y=f(z_1^*,z_2^*,z_c^*)$} \\
\hline
\name & \red & \uone & \utwo & \syn & \red & \uone & \utwo & \syn & \red & \uone & \utwo & \syn & \red & \uone & \utwo & \syn & \red & \uone & \utwo & \syn \\ 
\hline
\esta & $\mathbf{0.16}$ & $0$ & $0$ & $0.05$ & $0$ & $\mathbf{0.16}$ & $0$ & $0.05$ & $0$ & $0$ & $\mathbf{0.17}$ & $0.05$ & $0.07$ & $0$ & $0.01$ & $\mathbf{0.14}$ & $\mathbf{0.04}$ & $0.01$ & $0$ & $\mathbf{0.07}$ \\
\estb & $\mathbf{0.29}$ & $0.02$ & $0.02$ & $0$ & $0$ & $\mathbf{0.30}$ & $0$ & $0$ & $0$ & $0$ & $\mathbf{0.30}$ & $0$ & $0.11$ & $0.02$ & $0.02$ & $\mathbf{0.15}$ & $\mathbf{0.06}$ & $0.01$ & $0.01$ & $\mathbf{0.06}$ \\
Truth & $0.58$ & $0$ & $0$ & $0$ & $0$ & $0.56$ & $0$ & $0$ & $0$ & $0$ & $0.54$ & $0$ & $0$ & $0$ & $0$ & $0.56$ & $0.13$ & $0$ & $0$ & $0.27$ \\
\hline \hline
\end{tabular}

\vspace{2mm}

\begin{tabular}{l|cccc|cccc|cccc|cccc|cccc}
\hline \hline
Task & \multicolumn{4}{c|}{$y=f(z_1,z_2^*,z_c^*)$} & \multicolumn{4}{c|}{$y=f(z_1,z_2,z_c^*)$} & \multicolumn{4}{c|}{$y=f(z_1^*,z_2^*,z_c)$} & \multicolumn{4}{c|}{$y=f(z_2^*,z_c^*)$} & \multicolumn{4}{c}{$y=f(z_2^*,z_c)$} \\
\hline
\name & \red & \uone & \utwo & \syn & \red & \uone & \utwo & \syn & \red & \uone & \utwo & \syn & \red & \uone & \utwo & \syn & \red & \uone & \utwo & \syn \\ 
\hline
\esta & $0.04$ & $\mathbf{0.06}$ & $0$ & $\mathbf{0.07}$ & $\mathbf{0.07}$ & $0$ & $0$ & $\mathbf{0.12}$ & $\mathbf{0.1}$ & $0$ & $0.01$ & $\mathbf{0.07}$ & $\mathbf{0.03}$ & $0$ & $\mathbf{0.04}$ & $0.05$ & $\mathbf{0.1}$ & $0$ & $0.04$ & $0.05$ \\
\estb & $0.04$ & $\mathbf{0.09}$ & $0$ & $\mathbf{0.06}$ & $\mathbf{0.11}$ & $0.02$ & $0.02$ & $\mathbf{0.10}$ & $\mathbf{0.11}$ & $0.02$ & $0.02$ & $\mathbf{0.05}$ & $\mathbf{0.07}$ & $0$ & $\mathbf{0.06}$ & $0$ & $\mathbf{0.19}$ & $0$ & $0.06$ & $0$ \\
Truth & $0$ & $0.25$ & $0$ & $0.25$ & $0.18$ & $0$ & $0$ & $0.36$ & $0.22$ & $0$ & $0$ & $0.22$ & $0.21$ & $0$ & $0.21$ & $0$ & $0.34$ & $0$ & $0.17$ & $0$ \\
\hline \hline
\end{tabular}

\vspace{-2mm}
\label{tab:medium}
\end{table*}

\textbf{Synthetic generative model}: We begin with a set of latent vectors $z_1, z_2, z_c \sim \mathcal{N}(0_d, \Sigma_d^2), d=50$ representing information unique to $X_1,X_2$ and common to both respectively. $[z_1,z_c]$ is transformed into high-dimensional $x_1$ using a fixed transformation $T_1$ and likewise $[z_2,z_c]$ to $x_2$ via $T_2$. The label $y$ is generated as a function of (1) only $z_c$, in which case we expect complete redundancy, (2) only $z_1$ or $z_2$ which represents complete uniqueness, (3) a combination of $z_1$ and $z_2$ representing complete synergy, or (4) arbitrary ratios of each of the above with $z_i^*$ representing half of the dimensions from $z_i$ and therefore half of each interaction.
In total, Table~\ref{tab:medium} shows the $10$ synthetic datasets we generated: $4$ specialized datasets $\mathcal{D}_I$, $I\in\{R,U_1,U_2,S\}$ where $y$ only depends on one interaction, and $6$ mixed datasets with varying interaction ratios. We also report the ground-truth interactions as defined by the label-generating process and the total capturable information using the bound in~\citet{feder1994relations}, which relates the accuracy of the best model on these datasets with the mutual information between the inputs to the label. Since the test accuracies for Table~\ref{tab:medium} datasets range from $67$-$75\%$, this corresponds to total MI of $0.42-0.59$ bits.

\textbf{Results}: From Table~\ref{tab:medium}, both \esta\ and \estb\ agree in relative \pid\ values, correctly assigning the predominant interaction type and interactions with minimal presence consistent with the ground-truth based on data generation. For example, $\mathcal{D}_R$ has the highest $R$ value, and when the ratio of $z_1$ increases, $U_1$ increases from $0.01$ on $y=f(z_1^*,z_2^*,z_c^*)$ to $0.06$ on $y=f(z_1,z_2^*,z_c^*)$.
We also note some interesting observations due to the random noise in label generation, such as the non-zero synergy measure of datasets such as $\mathcal{D}_R,\mathcal{D}_{U_1},\mathcal{D}_{U_2}$ whose labels do not depend on synergy.

\subsection{\mbox{Quantifying real-world multimodal benchmarks}}
\label{sec:datasets}

We now apply these estimators to quantify the interactions in real-world multimodal datasets.

\textbf{Real-world multimodal data setup}: We use a large collection of real-world datasets in MultiBench~\citep{liang2021multibench} which test \textit{multimodal fusion} of different input signals (including images, video, audio, text, time-series, sets, and tables) for different tasks (predicting humor, sentiment, emotions, mortality rate, ICD-$9$ codes, image-captions, human activities, digits, and design interfaces).
We also include experiments on \textit{question-answering} (Visual Question Answering 2.0~\cite{antol2015vqa,goyal2017making} and CLEVR~\cite{johnson2017clevr}) which test grounding of language into the visual domain. For the $4$ datasets (top row of Table~\ref{tab:datasets_small}) involving images and text where modality features are available and readily clustered, we apply the \esta\ estimator on top of discrete clusters. For the remaining $4$ datasets (bottom row of Table~\ref{tab:datasets_small}) with video, audio, and medical time-series modalities, clustering is not easy, so we use the end-to-end \estb\ estimator.

\textbf{Human judgment of interactions}: Real-world multimodal datasets do not have reference \pid\ values, and exact \pid\ computation is impossible due to continuous data. We therefore use human judgment as a reference. We design a new annotation scheme where we show both modalities and the label and ask each annotator to annotate the degree of redundancy, uniqueness, and synergy on a scale of $0$-$5$, alongside their confidence in their answers on a scale of $0$-$5$. We give $50$ datapoints from each dataset (except \textsc{MIMIC} and \textsc{ENRICO} which require specialized knowledge) to $3$ annotators each. We show a sample user interface and annotation procedures in the full paper~\citep{liang2023quantifying}, and also provide an in-depth study of how humans annotate multimodal interactions in a subsequent follow-up work~\citep{liang2023human}.

\textbf{Results on multimodal fusion}: From Table~\ref{tab:datasets_small}, we find that different datasets do require different interactions. Some interesting observations: (1) all pairs of modalities on \textsc{MUStARD} sarcasm detection show high synergy values, which aligns with intuition since sarcasm is often due to a contradiction between what is expressed in language and speech, (2) uniqueness values are strongly correlated with unimodal performance (e.g., modality $1$ in \textsc{AV-MNIST} and \textsc{MIMIC}), (3) datasets with high synergy do indeed benefit from interaction modeling as also seen in prior work (e.g., \textsc{MUStARD}, \textsc{UR-FUNNY})~\cite{castro2019towards,hasan2019ur}, and (4) conversely datasets with low synergy are those where unimodal performance is relatively strong (e.g., \textsc{MIMIC})~\cite{liang2021multibench}.

\textbf{Results on QA}: We observe very high synergy values as shown in Table~\ref{tab:datasets_small} consistent with prior work studying how these datasets were balanced (e.g., \textsc{VQA 2.0} having different images for the same question such that the answer can only be obtained through synergy)~\cite{goyal2017making} and that models trained on these datasets require non-additive interactions~\cite{hessel2020does}. \textsc{CLEVR} has a higher proportion of synergy than \textsc{VQA 2.0} ($83\%$ versus $75\%$): indeed, \textsc{CLEVR} is a more balanced dataset where the answer strictly depends on both the question and image with a lower likelihood of unimodal biases.

\begin{table*}[t]
\centering
\fontsize{9}{11}\selectfont
\setlength\tabcolsep{2.5pt}
\caption{Estimating \name\ on real-world MultiBench~\citep{liang2021multibench} datasets. Many of the estimated interactions align well with human judgement as well as unimodal performance.}
\centering
\footnotesize

\begin{tabular}{l|cccc|cccc|cccc|cccc}
\hline \hline
Task & \multicolumn{4}{c|}{\textsc{AV-MNIST}} & \multicolumn{4}{c|}{\textsc{ENRICO}} & \multicolumn{4}{c|}{\textsc{VQA 2.0}} & \multicolumn{4}{c}{\textsc{CLEVR}} \\
\hline
\name & \red & \uone & \utwo & \syn & \red & \uone & \utwo & \syn & \red & \uone & \utwo & \syn & \red & \uone & \utwo & \syn \\
\hline
\esta & $0.10$ & $\bm{0.97}$ & $0.03$ & $0.08$ & $\bm{0.73}$ & $0.38$ & $0.53$ & $0.34$ & $0.79$ & $0.87$ & $0$ & $\mathbf{4.92}$ & $0.55$ & $0.48$ & $0$ & $\mathbf{5.16}$\\
Human & $\bm{0.57}$ & $\bm{0.61}$ & $0$ & $0$ & - & - & - & - & $0$ & $0$ & $0$ & $\bm{6.58}$ & $0$ & $0$ & $0$ & $\bm{6.19}$ \\
\hline \hline
\end{tabular}

\vspace{2mm}

\begin{tabular}{l|cccc|cccc|cccc|cccc}
\hline \hline
Task & \multicolumn{4}{c|}{\textsc{MOSEI}} & \multicolumn{4}{c|}{\textsc{UR-FUNNY}} & \multicolumn{4}{c|}{\textsc{MUStARD}} & \multicolumn{4}{c}{\textsc{MIMIC}} \\
\hline
\name & \red & \uone & \utwo & \syn & \red & \uone & \utwo & \syn & \red & \uone & \utwo & \syn & \red & \uone & \utwo & \syn \\
\hline
\estb & $\bm{0.26}$ & $\bm{0.49}$ & $0.03$ & $0.04$ & $0.03$ & $0.04$ & $0.01$ & $\bm{0.08}$ & $0.14$ & $0.01$ & $0.01$ & $\bm{0.30}$ & $0.05$ & $\bm{0.17}$ & $0$ & $0.01$ \\
Human & $\bm{0.32}$ & $\bm{0.20}$ & $0.15$ & $0.15$ & $0.04$ & $\bm{0.05}$ & $0.03$ & $\bm{0.04}$ & $0.13$ & $\bm{0.17}$ & $0.04$ & $\bm{0.16}$ & - & - & - & - \\
\hline \hline
\end{tabular}

\vspace{-2mm}
\label{tab:datasets_small}
\end{table*}

\textbf{Comparisons with human judgment}: For human judgment, we cannot ask humans to give a score in bits, so it is on a completely different scale ($0$-$5$ scale). To put them on the same scale, we normalize the human ratings such that the sum of human interactions is equal to the sum of PID estimates. The resulting comparisons are in Table~\ref{tab:datasets_small}, and we find that the human-annotated interactions overall align with estimated PID: the highest values are the same for $4$ datasets: both explain highest synergy on \textsc{VQA} and \textsc{CLEVR}, image ($U_1$) being the dominant modality in \textsc{AV-MNIST}, and language ($U_1$) being the dominant modality in \textsc{MOSEI}. Overall, the Krippendorff's alpha for inter-annotator agreement is high ($0.72$ for $R$, $0.68$ for $U_1$, $0.70$ for $U_2$, $0.72$ for $S$) and the average confidence scores are also high ($4.36/5$ for $R$, $4.35/5$ for $U_1$, $4.27/5$ for $U_2$, $4.46/5$ for $S$), indicating that the human-annotated results are reliable. For the remaining two datasets (\textsc{UR-FUNNY} and \textsc{MUStARD}), estimated PID matches the second-highest human-annotated interaction. We believe this is because there is some annotator subjectivity in interpreting whether sentiment, humor, and sarcasm are present in language only ($U_1$) or when contextualizing both language and video ($S$), resulting in cases of low annotator agreement in $U_1$ and $S$: $-0.14$, $-0.03$ for \textsc{UR-FUNNY} and $-0.08$, $-0.04$ for \textsc{MUStARD}.

\textbf{Comparisons with other interaction measures}: Our framework allows for easy generalization to other interaction definitions: we also implemented $3$ information theoretic measures \textbf{I-min}~\citep{williams2010nonnegative}, \textbf{WMS}~\citep{chechik2001group}, and \textbf{CI}~\citep{nirenberg2001retinal}. These results are included in the full paper~\citep{liang2023quantifying}, where we explain the limitations of these methods as compared to PID, such as over- and under-estimation, and potential negative estimation~\citep{griffith2014quantifying}. These are critical problems with the application of information theory for shared $I(X_1; X_2; Y)$ and unique information $I(X_1; Y|X_2)$, $I(X_2; Y|X_1)$ often quoted in the co-training~\citep{balcan2004co,blum1998combining} and multi-view learning~\citep{sridharan2008information,tian2020makes,tosh2021contrastive} literature. We also tried $3$ non-info theory measures: Shapley values~\citep{lundberg2017unified}, Integrated gradients (IG)~\citep{sundararajan2017axiomatic}, and CCA~\citep{andrew2013deep}, which are based on quantifying interactions captured by a multimodal model. Our work is fundamentally different in that interactions are properties of data before training any models.

\subsection{Quantifying multimodal model predictions}
\label{sec:models}

We now shift our focus to quantifying multimodal models. \textit{Do different multimodal models learn different interactions?}
A better understanding of the types of interactions that our current models struggle to capture can provide new insights into improving these models.

\textbf{Setup}: For each dataset, we train a suite of models on the train set $\mathcal{D}_\textrm{train}$ and apply it to the validation set $\mathcal{D}_\textrm{val}$, yielding a predicted dataset $\mathcal{D}_\textrm{pred} = \{(x_1,x_2,\hat{y}) \in \mathcal{D}_\textrm{val} \}$. Running \pid\ on $\mathcal{D}_\textrm{pred}$ summarizes the interactions that the model captures. We categorize and implement a comprehensive suite of models (spanning representation fusion at different feature levels, types of interaction inductive biases, and training objectives) that have been previously motivated to capture redundant, unique, and synergistic interactions.

\begin{table*}[t]
\centering
\fontsize{9}{11}\selectfont
\setlength\tabcolsep{2.5pt}
\vspace{-0mm}
\caption{Average interactions ($R/U/S$) learned by models alongside their average performance on interaction-specialized datasets (${\mathcal{D}_R}/{\mathcal{D}_U}/{\mathcal{D}_S}$). Synergy is the hardest to capture and redundancy is relatively easier to capture by existing models.}
\centering
\footnotesize

\begin{tabular}{l|ccccccccccc}
\hline \hline
Model & \textsc{EF} & \textsc{Additive} & \textsc{Agree} & \textsc{Align} & \textsc{Elem} & \textsc{Tensor} & \textsc{MI} & \textsc{MulT} & \textsc{Lower} & \textsc{Rec} & \textsc{Average} \\
\hline
\red & $0.35$ & $\mathbf{0.48}$ & $\mathbf{0.44}$ & $\mathbf{0.47}$ & $0.27$ & $\mathbf{0.55}$ & $0.20$ & $0.40$ & $\mathbf{0.47}$ & $\mathbf{0.53}$ & $\mathbf{0.41\pm0.11}$ \\
\acc$({\mathcal{D}_R})$ & $0.71$ & $\mathbf{0.74}$ & $\mathbf{0.73}$ & $\mathbf{0.74}$ & $0.70$ & $\mathbf{0.75}$ & $0.67$ & $0.73$ & $\mathbf{0.74}$ & $\mathbf{0.75}$ & $0.73\pm0.02$ \\
\hline
$U$ & $0.29$ & $0.31$ & $0.19$ & $0.44$ & $0.20$ & $0.52$ & $0.18$ & $0.45$ & $\mathbf{0.55}$ & $\mathbf{0.55}$ & $\mathbf{0.37\pm0.14}$ \\
\acc$({\mathcal{D}_U})$ & $0.66$ & $0.55$ & $0.60$ & $0.73$ & $0.66$ & $0.73$ & $0.66$ & $0.72$ & $\mathbf{0.73}$ & $\mathbf{0.73}$ & $0.68\pm0.06$ \\
\hline
\syn & $0.13$ & $0.09$ & $0.08$ & $0.29$ & $0.14$ & $\mathbf{0.33}$ & $0.12$ & $\mathbf{0.29}$ & $\mathbf{0.31}$ & $\mathbf{0.32}$ & $\mathbf{0.21\pm0.10}$ \\
\acc$({\mathcal{D}_S})$ & $0.56$ & $0.66$ & $0.63$ & $0.72$ & $0.66$ & $\mathbf{0.74}$ & $0.65$ & $\mathbf{0.72}$ & $\mathbf{0.73}$ & $\mathbf{0.74}$ & $0.68\pm0.06$ \\
\hline \hline

\end{tabular}
\vspace{-2mm}
\label{tab:models_small}

\end{table*}

\textbf{Results}: We show results in Table~\ref{tab:models_small} and highlight the following observations:

\emph{General observations}: We first observe that model \pid\ values are consistently higher than dataset \pid. The sum of model \pid\ is also a good indicator of test performance, which agrees with their formal definition since their sum is equal to $I(\{X_1,X_2\}; Y)$, the total task-relevant information.

\emph{On redundancy}: Several methods succeed in capturing redundancy, with an overall average of $R=0.41 \pm 0.11$ and accuracy of $73.0 \pm 2.0\%$ on redundancy-specialized datasets. Additive, agreement, and alignment-based methods are particularly strong, and we do expect them to capture redundant shared information~\cite{ding2022cooperative,radford2021learning}. Methods based on tensor fusion (synergy-based), including lower-order interactions, and adding reconstruction objectives (unique-based) also capture redundancy.

\begin{wrapfigure}{R}{0.35\textwidth}
    \begin{minipage}{0.35\textwidth}
    \vspace{-2mm}
    \includegraphics[width=\linewidth]{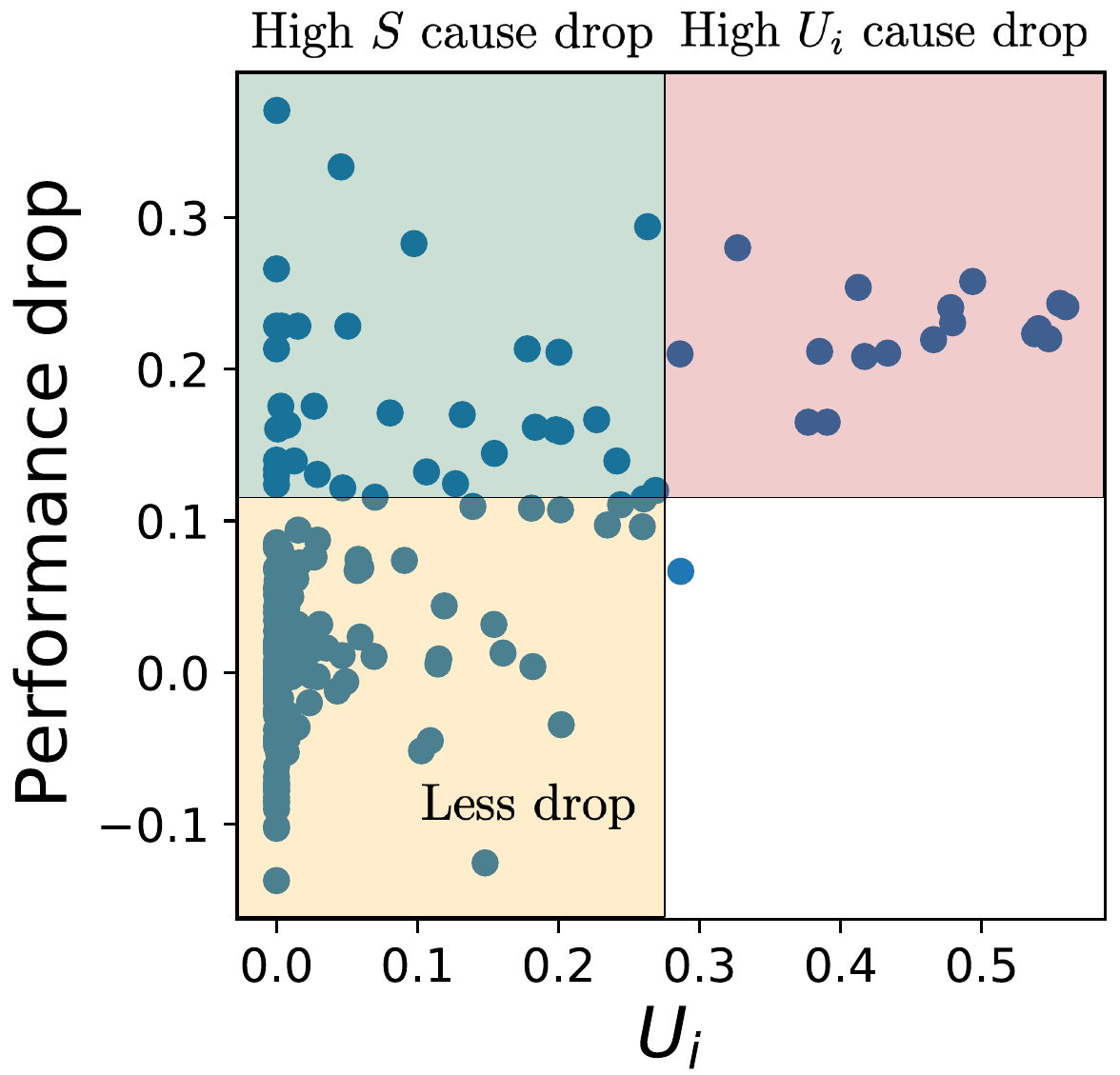}
    \vspace{-2mm}
    \caption{We find high correlation ($\rho=0.8$) between the performance drop when $X_i$ is missing and the model's $U_i$ value: high $U_i$ coincides with large performance drops ({\color{rr}red}), but low $U_i$ can also lead to performance drops. The latter can be further explained by large $S$ so $X_i$ is necessary ({\color{gg}green}).}
    \label{fig:noise}
\end{minipage}
\vspace{-2mm}
\end{wrapfigure}

\emph{On uniqueness}: Uniqueness is harder to capture than redundancy, with an average of $U=0.37 \pm 0.14$. Redundancy-based methods like additive and agreement do poorly on uniqueness, while those designed for uniqueness (lower-order interactions~\cite{zadeh2017tensor} and modality reconstruction objectives~\cite{tsai2019learning}) do well, with on average $U=0.55$ and $73.0\%$ accuracy on uniqueness datasets.

\emph{On synergy}: Synergy is the hardest to capture, with an average score of only $S=0.21 \pm 0.10$. Some of the strong methods are tensor fusion~\cite{fukui2016multimodal}, tensors with lower-order interactions~\cite{zadeh2017tensor}, modality reconstruction~\cite{tsai2019learning}, and multimodal transformer~\cite{xu2023multimodal}, which achieve around $S=0.30, \textrm{acc}=73.0\%$. Additive, agreement, and element-wise interactions do not seem to capture synergy well.

\emph{On robustness}: Finally, we also show connections between \pid\ and model performance in the presence of missing modalities. We find high correlation ($\rho=0.8$) between the performance drop when $X_i$ is missing and the model's $U_i$ value. Inspecting Figure~\ref{fig:noise}, we find that the implication only holds in one direction: high $U_i$ coincides with large performance drops ({\color{rr}in red}), but low $U_i$ can also lead to performance drops ({\color{gg}in green}). The latter can be further explained by the presence of large $S$ values: when $X_i$ is missing, synergy can no longer be learned which affects performance. For the subset of points when $U_i\le 0.05$, the correlation between $S$ and performance drop is $\rho=0.73$ (in contrast, the correlation for $R$ is $\rho=0.01$).

\subsection{\mbox{\pid\ agreement and model selection}}
\label{sec:link}

\begin{wrapfigure}{R}{0.35\textwidth}
    \begin{minipage}{0.35\textwidth}
    \vspace{-4mm}
    \includegraphics[width=\linewidth]{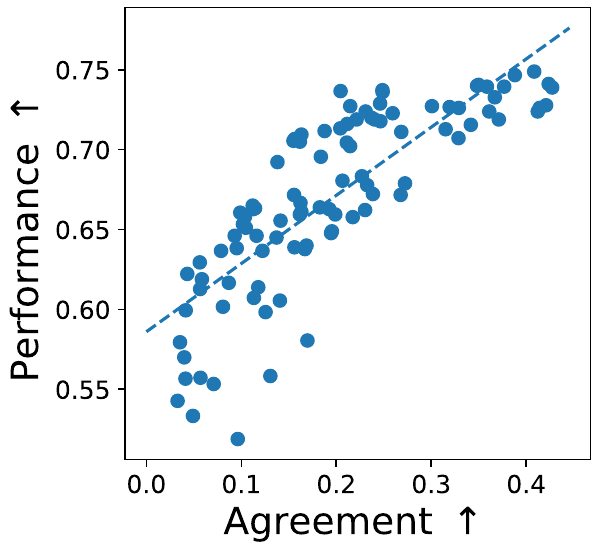}
    \vspace{-4mm}
    \caption{\name\ agreement $\alpha(f,\mathcal{D})$ between datasets and models strongly correlate with model accuracy ($\rho=0.81$).}
    \label{fig:agree}
\end{minipage}
\vspace{-6mm}
\end{wrapfigure}

Now that we have quantified datasets and models individually, the natural next question unifies both: \textit{what does the agreement between dataset and model \pid\ measures tell us about model performance?} We hypothesize that models able to capture the interactions necessary in a given dataset should also achieve high performance.
Given estimated interactions on dataset $\mathcal{D}$ and model $f(\mathcal{D})$ trained on $\mathcal{D}$, we define the agreement for each interaction $I \in \{R,U_1,U_2,S\}$ as:
\begin{equation}
    \label{eq:agreement}
    \alpha_I(f,\mathcal{D}) = \hat{I}_\mathcal{D} I_{f(\mathcal{D})},\quad \hat{I}_\mathcal{D}=\frac{I_\mathcal{D}}{\sum_{I' \in \{R,U_1,U_2,S\}}I'_\mathcal{D}},
\end{equation}
which summarizes the quantity of an interaction captured by a model ($I_{f(\mathcal{D})}$) weighted by its normalized importance in the dataset ($\hat{I}_\mathcal{D}$). The total agreement sums over $\alpha(f,\mathcal{D}) = \sum_I \alpha_I(f,\mathcal{D})$.

\textbf{Results}: Our key finding is that \pid\ agreement scores $\alpha(f,\mathcal{D})$ correlate ($\rho = 0.81$) with model accuracy across all $10$ synthetic datasets as illustrated in Figure~\ref{fig:agree}. This shows that \pid\ agreement can be a useful proxy for model performance. For the specialized datasets, we find that the correlation between $\alpha_I$ and $\mathcal{D}_I$ is $0.96$ for $R$, $0.86$ for $U$, and $0.91$ for $S$, and negatively correlated with other specialized datasets.
For mixed datasets with roughly equal ratios of each interaction, the measures that correlate most with performance are $\alpha_R$ ($\rho = 0.82$) and $\alpha_S$ ($\rho = 0.89$); datasets with relatively higher redundancy see $\rho = 0.89$ for $\alpha_R$; those with higher uniqueness have $\alpha_{U_1}$ and $\alpha_{U_2}$ correlate $\rho = 0.92$ and $\rho = 0.85$; those with higher synergy increases the correlation of $\alpha_S$ to $\rho = 0.97$.

Using these observations, our final experiment is model selection: \textit{can we choose the most appropriate model to tackle the interactions required for a dataset?}

\textbf{Setup}: Given a new dataset $\mathcal{D}$, we first compute its difference in normalized \pid\ values with respect to $\mathcal{D}'$ among our suite of $10$ synthetic datasets, $s(\mathcal{D}, \mathcal{D}')=\sum_{I \in \{R,U_1,U_2,S\}}|\hat{I}_\mathcal{D} - \hat{I}_{\mathcal{D}'}|$,
to rank the dataset $\mathcal{D}^*$ with the most similar interactions, and return the top-$3$ performing models on $\mathcal{D}^*$. In other words, we select models that best capture interactions that are of similar nature and degree as those in $\mathcal{D}$. We emphasize that even though we restrict dataset and model search to \textit{synthetic datasets}, we evaluate model selection on real-world datasets and find that it \textit{generalizes to the real world}.

\begin{table*}[t]
\centering
\fontsize{9}{11}\selectfont
\setlength\tabcolsep{3.0pt}
\vspace{-2mm}
\caption{\textbf{Model selection} results on unseen synthetic and real-world datasets. Given a new dataset $\mathcal{D}$, finding the closest synthetic dataset $\mathcal{D}'$ with similar \name\ values and recommending the best models on $\mathcal{D}'$ consistently achieves $95\%-100\%$ of the best-performing model on $\mathcal{D}$.}

\begin{tabular}{l|ccccccc}
\hline \hline
Dataset & $5$ Synthetic Datasets & \textsc{MIMIC} & \textsc{ENRICO} & \textsc{UR-FUNNY} & \textsc{MOSEI} & \textsc{MUStARD} & \textsc{MAPS} \\
\hline
$\%$ Performance & $99.91\%$ & $99.78\%$ & $100\%$ & $98.58\%$ & $99.35\%$ & $95.15\%$ & $100\%$ \\
\hline \hline
\end{tabular}
\label{tab:selection}
\vspace{-4mm}
\end{table*}

\textbf{Results}: We test our selected models on $5$ new synthetic datasets with different \pid\ ratios and $6$ real-world datasets, summarizing results in Table~\ref{tab:selection}. We find that the top $3$ chosen models achieve $95\%-100\%$ of the best-performing model accuracy, and $>98.5\%$ for all datasets except $95.2\%$ on \textsc{MUStARD}. For example, \textsc{UR-FUNNY} and \textsc{MUStARD} have the highest synergy ($S=0.13$, $S=0.3$) and indeed transformers and higher-order interactions are helpful (\textsc{MulT}: $65\%$, \textsc{MI}: $61\%$, \textsc{Tensor}: $60\%$). \textsc{Enrico} has the highest $R=0.73$ and $U_2=0.53$, and methods for redundant and unique interactions perform best (\textsc{Lower}: $52\%$, \textsc{Align}: $52\%$, \textsc{Agree}: $51\%$). \textsc{MIMIC} has the highest $U_1=0.17$, and unimodal models are mostly sufficient~\cite{liang2021multibench}.

\subsection{Real-world applications}
\label{sec:real_world}

Finally, we apply \pid\ to $3$ real-world case studies: pathology, mental health, and robotic perception.

\textbf{Case Study 1: Computational pathology.} Cancer prognostication is a challenging task in anatomic pathology that requires integration of whole-slide imaging (WSI) and molecular features for patient stratification~\cite{chen2021multimodal,lipkova2022artificial,mobadersany2018predicting}.
We use The Cancer Genome Atlas (TCGA), a large public data consortium of paired WSI, molecular, and survival information~\cite{weinstein2013cancer, tomczak2015review}, including modalities: (1) pre-extracted histology image features from diagnostic WSIs and (2) bulk gene mutation status, copy number variation, and RNA-Seq abundance values. We evaluate on two cancer datasets in TCGA, lower-grade glioma (LGG~\cite{cancer2015comprehensive}, $n=479$) and pancreatic adenocarcinoma (PAAD~\cite{raphael2017integrated}, $n=209$).

\textbf{Results}: In TCGA-LGG, most \pid\ measures were near-zero except $U_2=0.06$ for genomic features, which indicates that genomics is the only modality containing task-relevant information. This conclusion corroborates with the high performance of unimodal-genomic and multimodal models in~\citet{chen2022pan}, while unimodal-pathology performance was low. In TCGA-PAAD, the uniqueness in pathology and genomic features was less than synergy ($U_1=0.06$, and $U_2=0.08$ and $S=0.15$), which also match the improvement of using multimodal models that capture synergy.

\textbf{Case Study 2: Mental health.} Suicide is the second leading cause of death among adolescents~\citep{cdc}. Intensive monitoring of behaviors via adolescents' frequent use of smartphones may shed new light on the early risk of suicidal ideations~\citep{glenn2014improving,nahum2018just}, since smartphones provide rich behavioral markers~\citep{liang2021learning}. We used a dataset, \textsc{MAPS}, of mobile behaviors from high-risk consenting adolescent populations (approved by IRB). Passive sensing data is collected from each participant’s smartphone across $6$ months. The modalities include (1) \textit{text} entered by the user represented as a bag of top $1000$ words, (2) \textit{keystrokes} that record the exact timing and duration of each typed character, and (3) \textit{mobile applications} used per day as a bag of $137$ apps. Every morning, users self-report their daily mood, which we discretized into $-1,0,+1$. In total, \textsc{MAPS} has $844$ samples from $17$ participants.

\textbf{Results}: We first experiment with  MAPS$_{T,K}$ using text and keystroke features. \pid\ measures show that MAPS$_{T,K}$ has high synergy ($0.40$), some redundancy ($0.12$), and low uniqueness ($0.04$). We found the purely synergistic dataset $\mathcal{D}_S$ has the most similar interactions and the suggested models \textsc{Lower}, \textsc{Rec}, and \textsc{Tensor} that work best on $\mathcal{D}_S$ were indeed the top 3 best-performing models on MAPS$_{T,K}$, indicating that model selection is effective. Model selection also retrieves the best-performing model on MAPS$_{T,A}$ using text and app usage features.

\textbf{Case Study 3: Robotic Perception.} MuJoCo \textsc{Push}~\citep{lee2020multimodal} is a contact-rich planar pushing task in MuJoCo~\citep{todorov2012mujoco}, where a $7$-DoF Panda Franka robot is pushing a circular puck with its end-effector in simulation.
The dataset consists of $1000$ trajectories with $250$ steps sampled at $10$Hertz. The multimodal inputs are gray-scaled images from an RGB camera, force and binary contact information from a force/torque sensor, and the 3D position of the robot end-effector. We estimate the 2D position of the unknown object on a table surface while the robot intermittently interacts with it.

\textbf{Results}: We find that \estb\ predicts $U_1=1.79$ as the highest \pid\ value, which aligns with our observation that image is the best unimodal predictor. Comparing both estimators, \esta\ underestimates $U_1$ and $R$ since the high-dimensional time-series modality cannot be easily described by clusters without losing information. In addition, both estimators predict a low $U_2$ value but attribute high $R$, implying that a multimodal model with higher-order interactions would not be much better than unimodal models. Indeed, we observe no difference in performance between these two.

\vspace{-2mm}
\section{Conclusion}

Our work aims to quantify the nature and degree of feature interactions by proposing scalable estimators for redundancy, uniqueness, and synergy suitable for high-dimensional heterogeneous datasets. Through comprehensive experiments and real-world applications, we demonstrate the utility of our proposed framework in dataset quantification, model quantification, and model selection. We are aware of some potential \textbf{limitations}:
\begin{enumerate}[noitemsep,topsep=0pt,nosep,leftmargin=*,parsep=0pt,partopsep=0pt]
    \item These estimators only approximate real interactions due to cluster preprocessing or unimodal models, which naturally introduce optimization and generalization errors. We expect progress in density estimators, generative models, and unimodal classifiers to address these problems.

    \item It is harder to quantify interactions for certain datasets, such as \textsc{ENRICO} which displays all interactions which makes it difficult to distinguish between $R$ and $S$ or $U$ and $S$.

    \item Finally, there exist challenges in quantifying interactions since the data generation process is never known for real-world datasets, so we have to resort to human judgment, other automatic measures, and downstream tasks such as estimating model performance and model selection.
\end{enumerate}

\textbf{Future work} can leverage \pid\ for targeted dataset creation, representation learning optimized for \pid\ values, and applications of information theory to higher-dimensional data. More broadly, there are several exciting directions in investigating more applications of multivariate information theory in modeling feature interactions, predicting multimodal performance, and other tasks involving feature interactions such as privacy-preserving and fair representation learning from high-dimensional data~\citep{dutta2020information,hamman2023demystifying}.
Being able to provide guarantees for fairness and privacy-preserving learning can be particularly impactful.

\chapter{Factorized Learning of Multimodal Interactions}
\label{chap:foundations2}
\newcommand{\namel}{\textsc{Factorized Contrastive Learning}}
\newcommand{\names}{\textsc{FactorCL}}

\section{Introduction}

Using the mathematical foundation of multimodal interactions that we just presented, we now seek to learn representations from multimodal data that are suitable in capturing each of these interactions.
Learning representations from different modalities is a central paradigm in machine learning~\cite{liang2022foundations}.
Today, a popular learning method is to first pre-train general representations on unlabeled multimodal data before fine-tuning on task-specific labels~\cite{bugliarello2021multimodal,liang2022foundations,liang2022highmmt,lu2019vilbert}. These current multimodal pre-training approaches have largely been inherited from prior work in multi-view learning~\cite{chen2020simple,oord2018representation} that exploit a critical assumption of \textit{multi-view redundancy}: the property that shared information between modalities is almost exactly what is relevant for downstream tasks~\cite{sridharan2008information,tosh2021contrastive,tsai2020self}. 
When this assumption holds, approaches based on contrastive pre-training to capture shared information~\cite{chen2020simple,khosla2020supervised,radford2021learning,tian2020makes}, followed by fine-tuning to keep task-relevant shared information~\cite{tsai2020self}, have seen successful applications in learning from images and captions~\cite{radford2021learning}, video and audio~\cite{arandjelovic2017look}, speech and transcribed text~\cite{oord2018representation}, and instructions and actions~\cite{eysenbach2022contrastive}.
However, our paper studies two fundamental limitations in the application of contrastive learning (CL) to learn multimodal interactions in real-world settings
\begin{enumerate}[noitemsep,topsep=0pt,nosep,leftmargin=*,parsep=0pt,partopsep=0pt]
    \item \textbf{Low \textit{shared} information} relevant to tasks: There exists a wide range of multimodal tasks involving small amounts of shared information, such as between cartoon images and figurative captions (i.e., not literal but metaphoric or idiomatic descriptions of the images~\cite{marsh2003taxonomy, yosef2023irfl}).
    In these situations, standard multimodal CL will only receive a small percentage of information from the learned representations and struggle to learn the desired task-relevant information.
    \item \textbf{High \textit{unique} information} relevant to tasks: Many real-world modalities can provide unique information not present in other modalities. Examples include healthcare with medical sensors or robotics with force sensors~\cite{liang2021multibench,liang2023quantifying}. Standard CL will discard task-relevant unique information, leading to poor downstream performance.
\end{enumerate}
We refer the reader to Figure~\ref{fig:overview} for a visual depiction and experimental results showing the performance drop of CL in these two settings of low shared information and high unique information.

\begin{figure}[tbp]
\centering
\includegraphics[width=\linewidth]{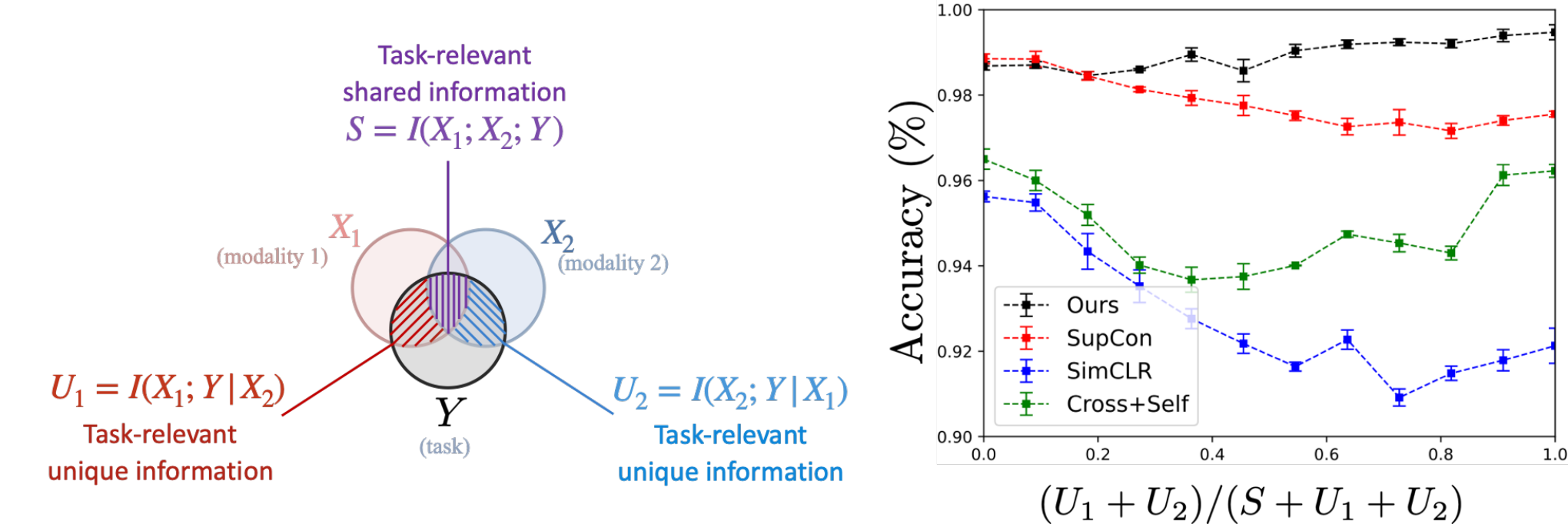}
\caption{\textbf{Left}: We define $S=I(X_1;X_2;Y)$ as task-relevant shared information and $U_1=I(X_1;Y|X_2)$, $U_2=I(X_2;Y|X_1)$ as task-relevant unique information. \textbf{Right}: On controllable datasets with varying ratios of $S$, $U_1$, and $U_2$, standard CL captures $S$ but struggles when there is more $U_1$ and $U_2$. Our \names\ approach maintains best performance, whereas SimCLR~\cite{chen2020simple} and SupCon~\cite{khosla2020supervised} see performance drops as unique information increases, and Cross+Self~\cite{huang2021multilingual,jain2021mural,lee2020parameter,yuan2021multimodal} recovers in fully unique settings but suffers at other ratios.}
\label{fig:overview}
\vspace{-4mm}
\end{figure}

In light of these limitations, how can we design suitable multimodal learning objectives that work beyond multi-view redundancy? In this paper, starting from the first principles in information theory, we provide formal definitions of shared and unique information via conditional mutual information and propose an approach, \namel\ (\names\ for short), to learn these multimodal representations beyond multi-view redundancy using three key ideas. The first idea is to explicitly \textit{factorize} shared and unique representations. The second idea is to \textit{capture task-relevant} information via maximizing lower bounds on MI and \textit{remove task-irrelevant} information via minimizing upper bounds on MI, resulting in representations with sufficient and necessary information content. Finally, a notion of task relevance without explicit labels in the self-supervised setting is achieved by leveraging \textit{multimodal augmentations}.
Experimentally, we evaluate the effectiveness of \names\ on a suite of synthetic datasets and large-scale real-world multimodal benchmarks involving images and figurative language~\cite{yosef2023irfl}, prediction of human sentiment~\cite{zadeh2016mosi}, emotions~\cite{zadeh2018multimodal}, humor~\cite{hasan2019ur}, and sarcasm~\cite{castro2019towards}, as well as patient disease and mortality prediction from health indicators and sensor readings~\cite{johnson2016mimic}, achieving new state-of-the-art performance on six datasets. Overall, we summarize our key technical contributions here:
\begin{enumerate}[noitemsep,topsep=0pt,nosep,leftmargin=*,parsep=0pt,partopsep=0pt]
    \item A new analysis of contrastive learning performance showing that standard multimodal CL fails to capture task-relevant unique information under low shared or high unique information cases.
    \item A new contrastive learning algorithm called \names:
    \begin{enumerate}[noitemsep,topsep=0pt,nosep,leftmargin=*,parsep=0pt,partopsep=0pt]
        \item \names\ factorizes task-relevant information into shared and unique information, expanding contrastive learning to better handle low shared or high unique information.
        \item \names\ optimizes shared and unique information separately, by removing task-irrelevant information via MI upper bounds and capturing task-relevant information via lower bounds, yielding optimal task-relevant representations.
        \item \names\ leverages multimodal augmentations to approximate task-relevant information, enabling self-supervised learning from our proposed \names.
    \end{enumerate}
\end{enumerate}

\vspace{-1mm}
\section{Analysis of Multi-view Contrastive Learning}
\vspace{-1mm}

We begin by formalizing definitions of four types of information: shared, unique, task-relevant, and task-irrelevant information in multimodal data. To formalize the learning setting, we assume there exist two modalities expressed as random variables $X_1$ and $X_2$ with outcomes $x_1$ and $x_2$, and a task with the random variable $Y$ and outcome $y$. We denote $X_{-i}$ as the other modality where appropriate.

\textbf{Shared and unique information}: We formalize shared and unique information by decomposing the total multimodal information $I (X_1,X_2; Y)$ into three conditional mutual information (MI) terms:
\begin{align}
\label{eq:sharedunique}
     I (X_1,X_2; Y) = \underbrace{I(X_1;X_2;Y)}_\text{$S=\textrm{shared}$} + \underbrace{I(X_1;Y|X_2)}_\text{$U_1=\textrm{uniqueness in } X_1$} + \underbrace{I(X_2;Y|X_1)}_\text{$U_2=\textrm{uniqueness in } X_2$},
\end{align}

where $I (X_1,X_2; Y) = \int p(x_1,x_2,y) \log \frac{p(x_1,x_2,y)}{p(x_1,x_2) p(y)} dx_1 dx_2 dy$ is the total MI between the joint random variable $X_1,X_2$ and the task $Y$, $S=I(X_1;X_2;Y) = I(X_1;X_2) - I(X_1;X_2|Y) = \int p(x_1,x_2) \log \frac{p(x_1,x_2)}{p(x_1) p(x_2)} dx_1 dx_2 - I(X_1;X_2|Y)$ is the task-relevant shared information, $I(X_1;X_2|Y)=\int p(x_1,x_2, y) \log \frac{p(x_1,x_2|y)}{p(x_1|y) p(x_2|y)} dx_1 dx_2 dy$ is the task-irrelevant shared information, and $U_1=I(X_1;Y|X_2)$, $U_2=I(X_2;Y|X_1)$ denote unique task-relevant information. 

\textbf{Limitations of CL}: Current approaches for CL maximize mutual information $I(X_1;X_2)$ (and subsequently task-relevant shared information $I(X_1;X_2;Y)$ during supervised fine-tuning), without modeling unique information. These methods generally learn a pair of representations \cite{tosh2021contrastive,tsai2020self}, 
\begin{align}
    Z_1 = \argmax_{Z_1 := f_\theta(X_1)} I(Z_1;X_2), \quad Z_2 = \argmax_{Z_2 := f_\theta(X_2)} I(X_1;Z_2). \label{eq:standard_cl_z}
\end{align}

For example, $Z_1$ could encode images $X_1$ and $Z_2$ encodes text $X_2$ via maximizing a lower bound on $I(X_1;X_2)$ using the NCE objective~\cite{oord2018representation}. The NCE objective falls into a broader class of contrastive learning methods~\cite{chen2020simple,he2020momentum,radford2021learning, chen2021empirical, khosla2020supervised} that model the ratio between joint densities $p(x_1,x_2)$ and product of marginal densities $p(x_1)p(x_2)$ using positive and negative samples~\cite{nguyen2010estimating, poole2019variational, tschannen2019mutual, wu2020mutual, ozair2019wasserstein} or probabilistic classifiers~\cite{mukherjee2020ccmi,tsai2020neural}. It has been shown that contrastive learning works well under the assumption of multi-view redundancy~\cite{sridharan2008information,hjelm2018learning,bachman2019learning, tian2020contrastive}:

\begin{definition}
\label{eq:multiview_redundancy_assump}
    (Multi-view redundancy) $\exists \epsilon >0$ such that $I(X_1;Y|X_2) \le \epsilon$ and $I(X_2;Y|X_1) \le \epsilon$.
\end{definition}

In other words, the task-relevant information in data is mostly shared across both views and the unique information is at most a small $\epsilon$. From a representation perspective, \citet{tian2020makes} further introduces the assumption that the optimal representation is minimal and sufficient, where all learned task-relevant information is shared information: $I(Z_1; Y | X_2)=I(Z_2; Y | X_1)=0$. While the multi-view redundancy is certainly true for particular types of multimodal distributions, it crucially ignores settings that display \textit{multi-view non-redundancy} and unique information can be important, such as when health indicators, medical sensors, and robotic visual or force sensors each provide unique information not present in other modalities~\cite{liang2021multibench,liang2023quantifying}.

\begin{definition}
\label{eq:multiview_nonredundancy_assump}
    (Multi-view non-redundancy) $\exists \epsilon >0$ such that $I(X_1;Y|X_2) > \epsilon$ or $I(X_2;Y|X_1) > \epsilon$.
\end{definition}

Under multi-view non-redundancy, we show that standard CL only receives a weak training signal since it can only maximize a lower bound on shared information $I(X_1;X_2)$, and struggles to learn task-relevant unique information. We formalize this intuition with the following statement:

\begin{theorem}
\label{thm:suboptimal_eq}
    (Suboptimality of standard CL) When there is multi-view non-redundancy as in Definition \ref{eq:multiview_nonredundancy_assump}, given optimal representations $\{Z_1,Z_2\}$ that satisfy Eq.(\ref{eq:standard_cl_z} and $I(Z_1; Y | X_2)=I(Z_2; Y | X_1)=0$~\citep{tian2020makes}, we have that
    {\small
    \begin{align}
        I(Z_1,Z_2;Y) = I(X_1,X_2;Y) - I(X_1;Y|X_2) - I(X_2;Y|X_1) = I(X_1;X_2) - I(X_1;X_2|Y)  < I(X_1,X_2;Y). \label{eq:suboptimal_eq}
    \end{align}}Correspondingly, the Bayes error rate $P_e(Z_1,Z_2):=1 - \mathbb{E}_{p(z_1, z_2)}\left[\max_{y\in Y} P\left(\hat{Y}=y \mid z_1, z_2 \right)\right]$ of contrastive representations $\{Z_1,Z_2\}$ for a downstream task $Y$ is given by:
    \begin{align}
        P_e &\le 1 - \exp \left[I(X_1, X_2; Y)- I(X_1;Y|X_2) - I(X_2;Y|X_1)  - H(Y) \right] \\
        &= 1 - \exp \left[I(X_1;X_2;Y) - H(Y) \right]
        \label{eq:bayes_error}
    \end{align}
\end{theorem}

We include proofs and a detailed discussion of the assumptions in the full paper~\citep{liang2023factorized}. Based on Eq.(\ref{eq:suboptimal_eq}), $I(Z_1,Z_2;Y)$ decreases with higher task-relevant unique information $I(X_1;Y|X_2)$ and $I(X_2;Y|X_1)$; we call this the difference $I(X_1, X_2 ; Y) - I(Z_1, Z_2; Y)$ the $\textit{uniqueness gap}$. The uniqueness gap measures the loss in task-relevant information between the input and encoded representation: as task-relevant unique information grows, the uniqueness gap increases. In addition, $I(Z_1,Z_2;Y)$ also drops with lower $I(X_1; X_2)$ (i.e., two modalities sharing little information to begin with), or with higher $I(X_1;X_2|Y)$ (i.e., when the shared information is mostly task-irrelevant). Similarly, in Eq.(\ref{eq:bayes_error}), the Bayes error rate of using $\{Z_1,Z_2\}$ for prediction is directly related to the task-relevant information in $\{Z_1,Z_2\}$: error on the downstream task increases with higher unique information and lower shared information.

\vspace{-1mm}
\section{\namel}
\vspace{-1mm}

\begin{figure}[tbp]
\centering
\includegraphics[width=\linewidth]{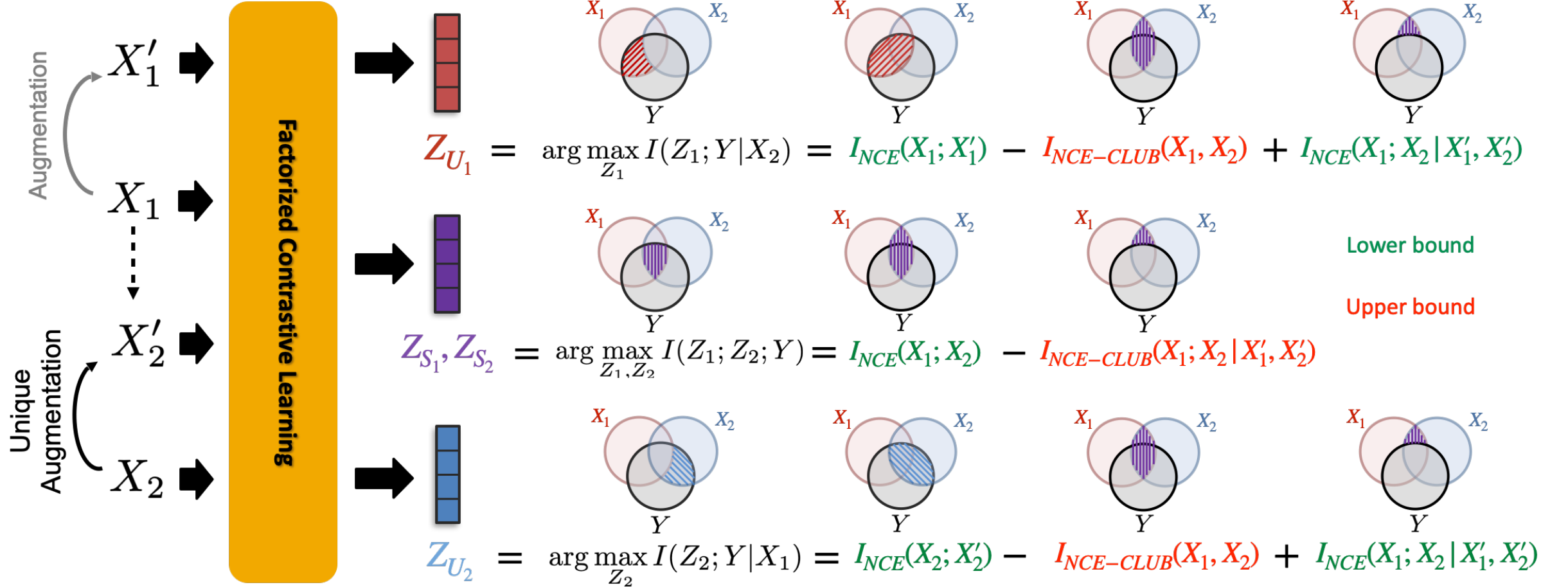}
\vspace{-2mm}
\caption{\names: We propose a self-supervised CL method to learn \textit{factorized} representations $Z_{S_1}$, $Z_{S_2}$, $Z_{U_1}$, and $Z_{U_2}$ to capture task-relevant information shared in both $X_1$ and $X_2$, unique to $X_1$, and unique to $X_2$. By starting with information-theoretic first principles of shared and unique information, we design contrastive estimators to both \textit{capture task-relevant} and \textit{remove task-irrelevant} information, where a notion of task-relevance without explicit labels is afforded by a new definition of \textit{multimodal augmentations} $X_1',X_2'$. Lower bounds are in {\color{gg}green} and upper bounds are in {\color{rr}red}.}
\label{fig:method}
\vspace{-2mm}
\end{figure}

We now present a suite of new CL objectives that alleviate the challenges above and work at all ranges of shared and unique information.
At a high level, we aim to learn a set of factorized representations $Z_{S_1},Z_{S_2},Z_{U_1},Z_{U_2}$ representing task-relevant information in $X_1$ shared with $X_2$, in $X_2$ shared with $X_1$, unique to $X_1$, and unique to $X_2$ respectively.
As common in practice~\cite{radford2021learning,tian2020makes}, we define neural networks $f_{\theta}$ with trainable parameters $\theta$ to extract representations from inputs $X_1$ and $X_2$. Learning these parameters requires optimizing differentiable and scalable training objectives to capture task-relevant shared and unique information (see overview in Figure~\ref{fig:method}):
\begin{align}
    Z_{S_1} &= \argmax_{Z_1 = f_\theta(X_1)} I(Z_1;X_2;Y), &&Z_{S_2} = \argmax_{Z_2 = f_\theta(X_2)} I(Z_2;X_1;Y), \label{eq:final_objectives1} \\
    Z_{U_1} &= \argmax_{Z_1 = f_\theta(X_1)} I(Z_1;Y|X_2), &&Z_{U_2} = \argmax_{Z_2 = f_\theta(X_2)} I(Z_2;Y|X_1). \label{eq:final_objectives2}
\end{align}
where $I(Z_1;X_2;Y) = I(Z_1; X_2) - I(Z_1; X_2 | Y)$ is the shared information and  $I(Z_2;X_1;Y) = I(Z_2; X_2) - I(Z_2; X_1 | Y)$ is the unique information. One important characteristic of our framework is that when unique information is zero: $I(X_1; Y|X_2) =0$ and $I(X_2; Y|X_1) = 0$, or all shared information is task-relevant: $I(X_1;X_2;Y) = I(X_1; X_2)$, our framework recovers standard CL as in Eq.(\ref{eq:standard_cl_z}). However, as we have previously indicated and will show empirically, these assumptions can easily be violated, and our framework enlarges Eq.(\ref{eq:standard_cl_z}) to cases where unique information is present.

The learned $Z$s can then be used as input to a linear classifier and fine-tuned to predict the label for multimodal classification or retrieval tasks. However, the shared and unique MI terms above are often intractable in practice. In the next section, we will build up our method step by step, eventually showing that each term in Eqs.(\ref{eq:final_objectives1}- \ref{eq:final_objectives2}) can be approximated as follows:
\begin{align}
    S = I(X_1; X_2; Y) &\ge I_\textsc{NCE}(X_1;X_2) - I_\textsc{NCE-CLUB}(X_1;X_2|X_1',X_2') \label{eq:shared}\\
    U_i = I(X_i; Y | X_{-i}) &\ge I_\textsc{NCE}(X_i;X_i') - I_\textsc{NCE-CLUB}(X_1;X_2) + I_\textsc{NCE}(X_1;X_2|X_1',X_2') \label{eq:unique}
\end{align}
where $I_\textsc{NCE}$ and $I_\textsc{NCE-CLUB}$ are scalable contrastive estimators (Section \ref{subsec:objectives}) and $X_1',X_2'$ are suitable data augmentations (Section \ref{subsec:unique_aug}) on each modality. Overall, these equations can be interpreted as both positive and negative signals to learn representations for $S$ and $U$. For shared information $S$, the estimator maximizes task-relevant shared information via $I_\textsc{NCE}(X_1;X_2)$ while removing task-irrelevant shared information via a novel upper bound $- I_\textsc{NCE-CLUB}(X_1;X_2|X_1',X_2')$.  For unique information $U_i$, we capture task-relevant uniqueness via $+I_\textsc{NCE}(X_i;X_i')$ while non-unique information is removed via $- (I_\textsc{NCE-CLUB}(X_1; X_2) - I_\textsc{NCE}(X_1;X_2|X_1',X_2'))$.  
In the following sections, we derive this final objective step-by-step: (1) approximating the MI objectives in $S$ and $U$ with CL estimators, (2) relaxing the dependence on labels $Y$ with self-supervised data augmentations, finally (3) discussing overall training and implementation details of end-to-end self-supervised learning.

\subsection{Supervised \names\ with shared and unique information}
\label{subsec:objectives}

To capture shared and unique information via an objective function, we will need to maximize lower bounds for all terms with a positive sign in Eq.(\ref{eq:shared}) and (\ref{eq:unique}) $\left( I\left(X_1; X_2\right), I\left(X_i; Y\right), I\left(X_1; X_2 | Y\right)\right)$ and minimize upper bounds for all terms with a negative sign $\left( I\left(X_1; X_2\right), I\left(X_1; X_2 | Y\right)\right)$. Our first theorem derives general lower and upper bounds for MI terms as variants of contrastive estimation: 

\begin{theorem}
    (Contrastive estimators for $I(X_1; X_2)$) Defining the NCE and NCE-CLUB estimators,
    \begin{align}
        I_\textsc{NCE}(X_1; X_2) &= \mathbb{E}_{\substack{x_1,x_2^+ \sim p(x_1,x_2)\\x_2^- \sim p(x_2)}} \left[ \log \frac{\exp f(x_1,x_2^+)}{\sum_k \exp f(x_1, x_2^-)} \right] \label{eq:nce_original}\\
        I_\textsc{NCE-CLUB}(X_1; X_2) &= \mathbb{E}_{x_1,x_2^+ \sim p(x_1,x_2)} \left[  f^*(x_1,x_2^+) \right] - \mathbb{E}_{\substack{x_1 \sim p(x_1)\\x_2^- \sim p(x_2)}} \left[f^*(x_1,x_2^-) \right] \label{eq:nceclub}
    \end{align}
    where $f^*(x_1,x_2)$ is the optimal critic from $I_\textsc{NCE}$ plugged into the $I_\textsc{CLUB}$ objective \cite{cheng2020club}. We call the proposed plug-in objective Eq.(\ref{eq:nceclub}) $I_\textsc{NCE-CLUB}$, and obtain lower and upper bounds on $I(X_1; X_2)$:
    \begin{align}
        I_\textsc{NCE}(X_1; X_2) \le I(X_1; X_2) \le I_\textsc{NCE-CLUB}(X_1; X_2).
    \end{align}
\end{theorem}

\begin{figure}[tbp]
\centering
\includegraphics[width=0.94\linewidth]{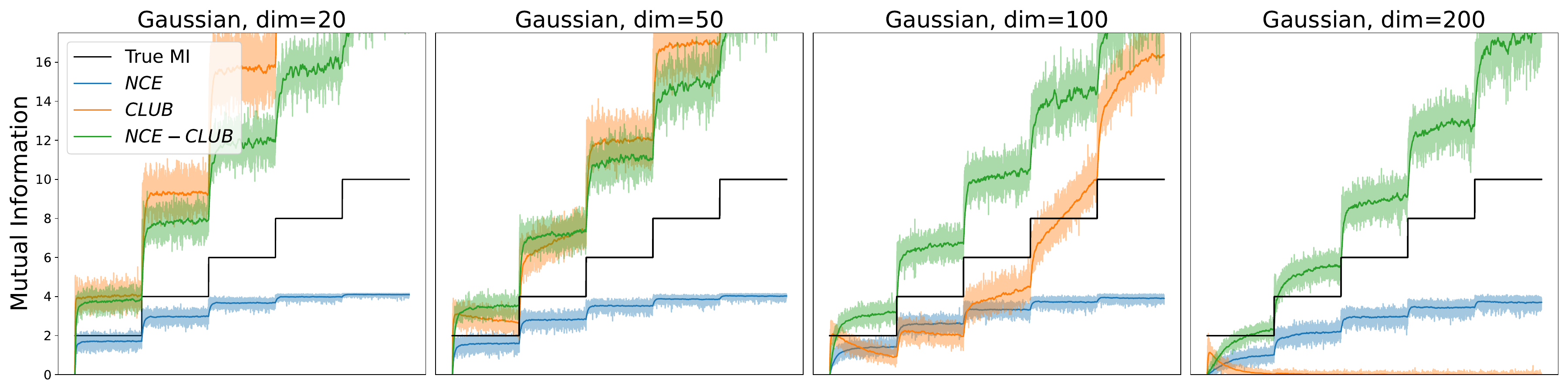}
\caption{Estimated $I_\textsc{NCE}$ lower bound~\cite{oord2018representation} and our proposed upper bound $I_\textsc{NCE-CLUB}$ on sample distributions with changing mutual information: our upper bound is tighter, more accurate, and more stable than $I_\textsc{CLUB}$ upper bound~\cite{cheng2020club}, and also comes for `free' via jointly estimating both lower and upper bounds simultaneously. We find that as dimension increases, the $I_\textsc{CLUB}$ estimator collapses to zero and no longer tracks true MI.}
\label{fig:bounds}
\vspace{-2mm}
\end{figure}

\begin{proof}
The lower bound $I_\textsc{NCE}(X_1; X_2) \le I(X_1; X_2)$ follows from~\citet{oord2018representation}: optimizing the objective leads to an optimal critic \cite{poole2019variational} $f^* = \log p(x_1 | x_2) + c(x_1)$, with a deterministic function $c(\cdot)$. Plugging optimal critic $f^*$ into $I_\textsc{NCE-CLUB}(X_1; X_2)$ cancels out the $c(x_1)$ term and yields $I_\textsc{NCE-CLUB}(X_1; X_2)$ and $I(X_1; X_2) \le I_\textsc{NCE-CLUB}$. We include a detailed proof in the full paper~\citep{liang2023factorized}.
\end{proof}
$I_\textsc{NCE-CLUB}(X_1; X_2)$ gives a desired upper bound of $I(X_1; X_2)$ ``for free'' while avoiding separately optimizing lower bound and upper bounds. In Figure~\ref{fig:bounds}, we show these two bounds in practice across two Gaussian distributions $X_1$ and $X_2$ with varying amounts of MI $I(X_1;X_2)$. We use the second formulation of $I_\textsc{CLUB}$ \cite{cheng2020club}, which assumes $p(x_1|x_2)$ to be unknown.
Our upper bound is empirically tighter (see Figure~\ref{fig:bounds}) and comes for ``free'' via jointly maximizing the lower bound $I_\textsc{NCE}$. These lower and upper bounds can be seen as new contrastive objectives over positive and negative $(x_1,x_2)$ pairs, enabling a close integration with existing pre-training paradigms.
Finally, we can similarly obtain bounds for the conditional MI $I_\textsc{NCE}(X_1;X_2|Y) \le I(X_1;X_2|Y) \le I_\textsc{NCE-CLUB}(X_1;X_2|Y)$:

\begin{align}
    I_\textsc{NCE}(X_1;X_2|Y) &= \mathbb{E}_{p(y)} \left[ \mathbb{E}_{\substack{x_1,x_2^+ \sim p(x_1,x_2|y)\\x_2^- \sim p(x_2|y)}} \left[ \log \frac{\exp f(x_1,x_2^+, y)}{\sum_k \exp f(x_1, x_2^-, y)} \right] \right] \label{eq:conditional_nce_supervised} \\
    I_\textsc{NCE-CLUB}(X_1;X_2|Y) &= \mathbb{E}_{p(y)} \left[ \mathbb{E}_{x_1,x_2^+ \sim p(x_1,x_2|y)} \left[ f^*(x_1,x_2^+, y) \right] - \mathbb{E}_{\substack{x_1 \sim p(x_1|y)\\x_2^- \sim p(x_2|y)}} \left[ f^*(x_1,x_2^-, y) \right] \right] \label{eq:conditional_club_supervised}
\end{align}

These two bounds result in \textit{conditional CL} objectives \cite{tsailearning, tsai2022conditional, ma2021conditional} - they differ critically from standard CL methods since they capture task-irrelevant shared information that remains between $X_1$ and $X_2$ after observing $Y$. This task-irrelevant shared information is removed by minimizing its upper bound. Note that $f(x_1, x_2, y)$ here denotes a different function from $f(x_1, x_2)$ in Eq.(\ref{eq:nce_original}), as the general forms are different (taking in $x_1, x_2$ versus $x_1, x_2, y$). $f(x_1, x_2, y)$ can be implemented in different ways, e.g., $g([x_1, y])^Th(x_2)$ where $g(), h()$ are trainable encoders and $[x_1, y]$ denotes concatenation \citep{sordoni2021decomposed}. 

\subsection{Self-supervised \names\ via multimodal augmentations}
\label{subsec:unique_aug}

The derivations above bring about supervised CL objectives with access to $Y$~\cite{khosla2020supervised}. For unsupervised CL~\cite{tian2020makes,oord2018representation}, we derive similar objectives without access to $Y$ by leveraging semantic augmentations on each modality. Denote $X'$ as some augmentation of $X$ (e.g., rotating, shifting, or cropping). Under the \textit{optimal augmentation} assumption from~\citet{tian2020makes} (restated below), replacing $Y$ with $X'$ in our formulations enables learning of task-relevant information without access to labels:
\begin{definition}
\label{def:optimal_single}
    (Optimal unimodal augmentation)~\cite{tian2020makes} $X_1'$ is an optimal unimodal augmentation for $X_1$ when $I(X; X') = I(X; Y)$, which implies that the only information shared between $X$ and $X'$ is task-relevant with no irrelevant noise.
\end{definition}
This assumption is satisfied when all information shared between $X$ and $X'$ is task-relevant, which implies that the augmentation keeps task-relevant information constant while changing task-irrelevant information. In the case of image classification, task-relevant information is the object in the picture, while task-irrelevant information is the background. 
By performing two separate unimodal augmentations giving $X_1'$ and $X_2'$, we can substitute contrastive estimators in Eqs.(\ref{eq:conditional_nce_supervised}) and (\ref{eq:conditional_club_supervised}), by replacing $I(X_i;Y)$ terms with $I(X_i;X_i')$ and replacing $I(X_1;X_2|Y)$ terms with  $I(X_1;X_2|X_1', X_2')$:
{\small
\begin{align}
    I_\textsc{NCE}(X_1;X_2|X_1',X_2') &=  \mathbb{E}_{p(x_1', x_2')} \left[ \mathbb{E}_{\substack{x_1,x_2^+ \sim p(x_1,x_2|x_1', x_2')\\x_2^- \sim p(x_2|x_1', x_2')}} \left[ \log \frac{\exp f(x_1,x_2^+, x_1', x_2')}{\sum_k \exp f(x_1, x_2^-, x_1', x_2')} \right] \right] \label{eq:nce_final_ssl} \\
    I_\textsc{NCE-CLUB}(X_1;X_2|X_1',X_2') &= \mathbb{E}_{p(x_1', x_2')} \Big[ \mathbb{E}_{x_1,x_2^+ \sim p(x_1,x_2|x_1', x_2')} [f^*(x_1,x_2^+, x_1', x_2') ] \nonumber \\
    &- \mathbb{E}_{\substack{x_1 \sim p(x_1|x_1', x_2')\\x_2^- \sim p(x_2|x_1', x_2')}} [f^*(x_1,x_2^-, x_1', x_2') ] \Big] \label{eq:nceclub_final_ssl}
\end{align}
}The objectives can be seen as conditional contrastive learning on augmentations $(X_1',X_2'$). Here again $f(x_1, x_2, x_1', x_2')$ is different from the critics in Eqs.(\ref{eq:conditional_nce_supervised} because of the different general forms. We implement $f()$ here as $g([x_1, x_1'])^Th([x_2, x_2'])$ where $g(), h()$ are trainable encoders specific for each modality and $[x_1, x_1']$ denotes concatenation. This concatenation is justified by the CMI estimators in~\citet{sordoni2021decomposed}, who show that concatenating the conditioning variable with the input in the critic $f(x_1, x_2, x_1', x_2')$ yields a Conditional InfoNCE estimator (Eq.(\ref{eq:nce_final_ssl})) that is a lower bound for CMI. However, the exact Conditional InfoNCE estimator learns a different conditional distribution $p(x_1, x_2 | x_1', x_2')$ for each augmented pair $x_1',x_2'$, which can be prohibitively expensive. We could approximate this by creating multiple augmentations of a single paired $x_1, x_2$. Our code uses one augmented pair $x_1', x_2'$ for each $x_1, x_2$ but could be extended to multiple pairs, and we find this simple approach yields consistent CMI lower and upper bounds that are empirically comparable to existing CMI estimators~\cite{mukherjee2020ccmi,sordoni2021decomposed}. We include full comparisons and implementation details in the full paper~\citep{liang2023factorized}.

\begin{wrapfigure}{L}{0.45\textwidth}
    \vspace{-4mm}
    \begin{minipage}{0.45\textwidth}
    \vspace{-0mm}
    \includegraphics[width=\linewidth]{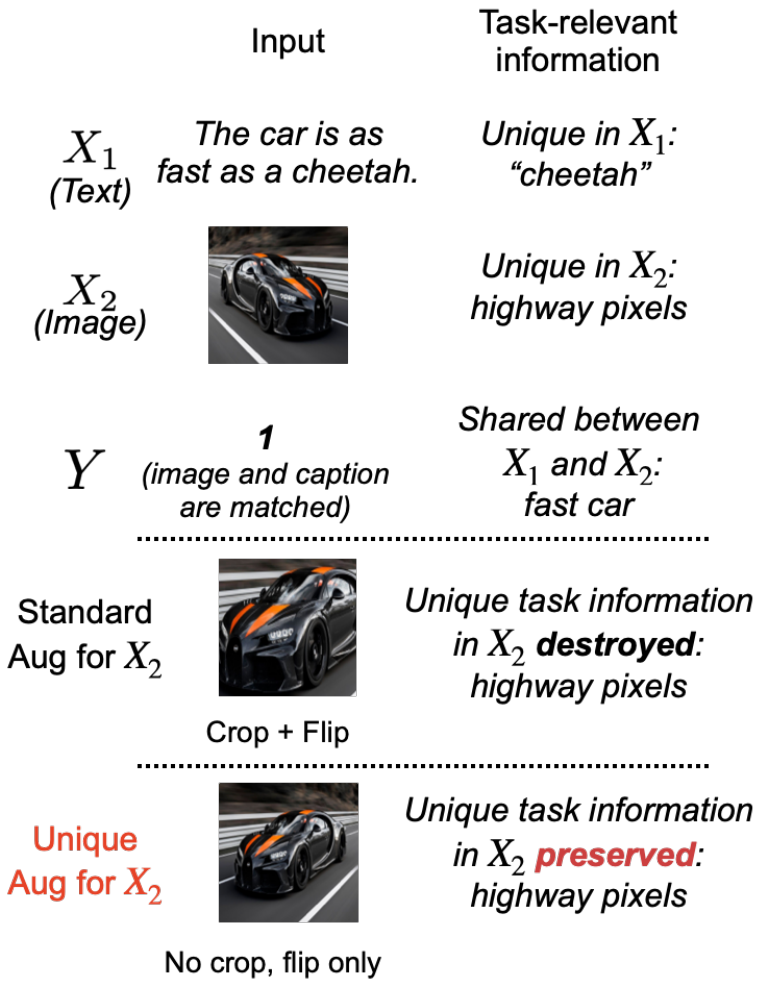}
    \vspace{-4mm}
    \caption{Standard vs. unique augmentations for the figurative language~\cite{yosef2023irfl} dataset. After augmenting text modality $X_1$ independently (same for both augmentation types), we illustrate their differences for image augmentation: unique augmentation on images should avoid removing information referred to by $X_1$ (the text). The text mentions that the car is fast so unique augmentation for images should \textit{not} remove the highway pixels of the image which can suggest the car is fast.}
    \label{fig:augs}
    \vspace{-4mm}

\end{minipage}
\vspace{-8mm}
\end{wrapfigure}

\vspace{1mm}
Although we find this method to work well in practice, a more careful analysis reveals that $2$ separate unimodal augmentations $X_1'$ and $X_2'$ each satisfying $I(X_i; X_i') = I(X_i; Y)$ do not together satisfy $I(X_1;X_2|Y)=I(X_1;X_2|X_1', X_2')$ needed for the substitution in Eqs.(\ref{eq:nce_final_ssl}) and (\ref{eq:nceclub_final_ssl}) to hold with equality. To satisfy this property exactly, we define optimal multimodal augmentations:
\begin{definition}
\label{def:optimal_multi}
    (Optimal multimodal augmentation) $X_1'$ and $X_2'$ are optimal multimodal augmentation for $X_1$ and $X_2$ when $I(X_1,X_2; X_1',X_2') = I(X_1,X_2; Y)$, which implies that the only information shared between $X_1,X_2$ and $X_1',X_2'$ is task-relevant with no irrelevant noise.
\end{definition}
We satisfy $I(X_1,X_2; X_1',X_2') = I(X_1,X_2; Y)$ using two steps:
\begin{align}
    \textit{Unimodal aug: } &X_1' \textrm{ s.t. } I(X_1; X_1') = I(X_1; Y), \label{assump:unimodal_aug}\\
    \textit{Unique aug: } &X_2' \textrm{ s.t. } I(X_2; X_2'|X_1) = I(X_2; Y|X_1). \label{assump:unique_aug}
\end{align}
We call the second step \textit{unique augmentation}: after observing $X_1$, we create augmented $X_2'$ from $X_2$ to keep task-relevant information not already in $X_1$. To empirically satisfy optimal multimodal augmentations, we avoid augmentations in one modality that will remove or strongly destroy information shared with the other modality. For example, in image captioning, we should avoid image augmentations such as cropping that destroy information from the caption (e.g., cropping object parts referred to by the caption), and instead, only augment images via flipping or color jittering which retains all caption information. Figure~\ref{fig:augs} shows an example of unique augmentation that satisfies these conditions. In our experiments, we will show that our augmentations consistently perform better than standard augmentations (Table \ref{tab:fig}), suggesting that approximately satisfying Eqs.(\ref{assump:unimodal_aug}) and (\ref{assump:unique_aug}) can be empirically sufficient, which is simple and straightforward to implement on real-world datasets.

\begin{figure}[tbp]
    \begin{minipage}{0.45\textwidth}
    \vspace{-22mm}
    \begin{algorithm}[H]
    \begin{algorithmic}
    \REQUIRE Multimodal dataset $\{\mathbf{X_1}, \mathbf{X_2}$\}.
    \vspace{.2em}\hrule\vspace{.5em}
    \STATE Initialize networks $f(\cdot)$.
    \WHILE{not converged}
    \FOR{sampled batch $\{\vx_1, \vx_2\}$}
    \STATE Estimate $I_{\textsc{NCE}}(X_1; X_2)$ from Eq. \ref{eq:nce_original}
    \STATE $\mathcal{L} = - I_{\textsc{NCE}}(X_1; X_2)$
    \STATE Update $f(\cdot)$ to minimize $\mathcal{L}$
    \ENDFOR
    \ENDWHILE
    \RETURN $f(\cdot)$
  \end{algorithmic}
  \caption{Standard multimodal CL.}
  \label{alg:standardcl}
\end{algorithm}
\end{minipage}
\begin{minipage}{0.48\textwidth}
    \vspace{-14.2mm}
    \begin{algorithm}[H]
    \begin{algorithmic}
    \REQUIRE Multimodal dataset $\{\mathbf{X_1}, \mathbf{X_2}$\}.
    \vspace{.2em}\hrule\vspace{.5em}
    \STATE Initialize networks $f(\cdot)$.
    \WHILE{not converged}
    \FOR{sampled batch $\{\vx_1, \vx_2\}$}
    \STATE $\vx_1' \leftarrow$ \textbf{{\color{rr}$\textrm{Augment}(\vx_1)$}}
    \STATE $\vx_2' \leftarrow$ \textbf{{\color{rr}$\textrm{Unique-Augment}(\vx_2| \vx_1)$}}
    \STATE Plug $\vx_1'$ and $\vx_2'$ into Eq. \ref{eq:nce_final_ssl} and \ref{eq:nceclub_final_ssl}
    \STATE Estimate \textbf{{\color{rr} $\mathbf{S}, \mathbf{U_1}, \mathbf{U_2}$}} from Eq. \ref{eq:shared} and \ref{eq:unique}
    \STATE $\mathbf{\color{rr} \mathcal{L} = -( S + U_1 + U_2)}$
    \STATE Update $f(\cdot)$ to minimize $\mathcal{L}$
    \ENDFOR
    \ENDWHILE
    \RETURN $f(\cdot)$
  \end{algorithmic}
  \caption{\textbf{\color{rr}\names}.}
  \label{alg:full}
\end{algorithm}
\vspace{-8mm}
\end{minipage}
\end{figure}

\vspace{-1mm}
\subsection{Overall method and implementation}
\vspace{-1mm}

The final algorithm sketch is in Algorithm~\ref{alg:full}, which we compare against standard CL in Algorithm~\ref{alg:standardcl}. It can be shown that \names\ learns all the task-relevant information from both modalities:
\begin{theorem}
    (Optimality of \names) If $Z_{S_1},Z_{S_2},Z_{U_1},Z_{U_2}$ perfectly maximize Eqs.(\ref{eq:final_objectives1}-\ref{eq:final_objectives2}) and the estimations in Eqs.(\ref{eq:shared}) and (\ref{eq:unique}) are tight, we obtain $I(X_1, X_2 ; Y) = I(Z_{S_1}; Z_{S_2}; Y) + I(Z_{U_1}; Y | Z_{S_2}) + I(Z_{U_2}; Y | Z_{S_1})$, suggesting that \names\  learns both shared and unique task-relevant information.
\end{theorem}
We include the full proof in the full paper~\citep{liang2023factorized}. In practice, while we do not expect perfect estimation of MI quantities and maximization with respect to MI objectives, we show that our method still improves empirical performance on several real-world datasets.

\textbf{Complexity}: Compared to heuristic combinations of cross-modal and single-modality CL~\cite{huang2021multilingual,jain2021mural,lee2020parameter,wang2022rethinking,yuan2021multimodal,shan2022ernievil, yang2022unified}, our approach does not significantly increase complexity: (1) upper bounds on MI can be estimated ``for free'' by directly plugging in the optimal critic from $I_\textsc{NCE}$, (2) removal of task-irrelevant information via $I(X_1;X_2|X_1',X_2')$ shares encoders with $I_\textsc{NCE}$, and (3) separate unimodal augmentations perform empirically well.

\vspace{-1mm}
\section{Experiments}
\vspace{-1mm}

We run comprehensive experiments on a suite of synthetic and large-scale real-world datasets with varying requirements of shared and unique task-relevant information, comparing our \names\ method to key baselines:
\begin{enumerate}[noitemsep,topsep=0pt,nosep,leftmargin=*,parsep=0pt,partopsep=0pt]
    \item SimCLR~\cite{chen2020simple}: the straightforward method of cross-modal $(X_1,X_2)$ contrastive learning.
    \item Cross+Self~\cite{yuan2021multimodal,huang2021multilingual,lee2020parameter,jain2021mural,shan2022ernievil, yang2022unified}: captures a range of methods combining cross-modal $(X_1,X_2)$ CL with additional unimodal $(X_i,X_i')$ CL objectives. This category also includes other ways of preserving unique information, such as through (variational) autoencoder reconstructions~\cite{wang2022rethinking}.
    \item Cross+Self+Fact~\cite{yuan2021multimodal,yang2022visionlanguage}: A factorized extension of Cross+Self, which is approximately done in prior work that adds separate (typically pre-trained) unimodal encoders for each modality.
    \item SupCon~\cite{khosla2020supervised}, which learns $I(X_1;X_2|Y)$ using CL conditioned on $Y$ from labeled data.
\end{enumerate}
We also carefully ablate each component of our method and investigate factors, including training data size and choice of augmentations. The intermediate ablations that emerge include:
\begin{enumerate}[noitemsep,topsep=0pt,nosep,leftmargin=*,parsep=0pt,partopsep=0pt]
    \item \names-SUP: The supervised CL version which uses labels $Y$ in Eqs.(\ref{eq:conditional_nce_supervised}) and (\ref{eq:conditional_club_supervised}).
    \item \names-SSL: The fully self-supervised version of our approach replacing $Y$ with multimodal augmentations $X_1'$ and $X_2'$ to approximate the task.
    \item OurCL-SUP: \names-SUP but removing the factorization so only two features $Z_1$ is optimized for both $I(X_1;X_2;Y)$ and $I(X_1;Y|X_2)$, $Z_2$ optimized for both $I(X_1;X_2;Y)$ and $I(X_2;Y|X_1)$.
    \item OurCL-SSL: \names-SSL but also removing the factorization in the self-supervised setting.
\end{enumerate}
The formulation of each ablation and implementation can be found in the full paper~\cite{liang2023factorized}.

\vspace{-1mm}
\subsection{Controlled experiments on synthetic datasets}
\vspace{-1mm}

\textbf{Synthetic data generation}: We begin by generating data with controllable ratios of task-relevant shared and unique information. Starting with a set of latent vectors $w_1, w_2, w_s \sim \mathcal{N}(0_d, \Sigma_d^2), d=50$ representing information unique to $X_1,X_2$ and common to both respectively, the concatenated vector $[w_1,w_s]$ is transformed into high-dimensional $x_1$ using a fixed transformation $T_1$ and likewise $[w_2,w_s]$ to $x_2$ via $T_2$. The label $y$ is generated as a function (with nonlinearity and noise) of varying ratios of $w_s$, $w_1$, and $w_2$ to represent shared and unique task-relevant information.

\textbf{Results}: In Figure~\ref{fig:overview}, we show our main result on synthetic data comparing \names\ with existing CL baselines. \names\ consistently maintains the best performance, whereas SimCLR~\cite{chen2020simple} and SupCon~\cite{khosla2020supervised} see performance drops as unique information increases. Cross+Self~\cite{yuan2021multimodal,huang2021multilingual,lee2020parameter,jain2021mural} recovers in fully unique settings (x-axis$=1.0$) but suffers at other ratios.




\begin{table*}[t]
\centering
\fontsize{9}{11}\selectfont
\setlength\tabcolsep{4pt}
\vspace{-8mm}
\caption{We probe whether contrastive representations learned by classic CL methods and \names\ contain shared $w_s$ or unique $w_1,w_2$ information. \names\ captures the most unique information.}
\centering

\begin{tabular}{l|cc|cc|cc|cccc}
\hline \hline
Model & \multicolumn{2}{c|}{SimCLR} & \multicolumn{2}{c|}{Cross+self} & \multicolumn{2}{c|}{SupCon} & \multicolumn{4}{c}{\names} \\
Representations & $Z_1$ & $Z_2$ & $Z_1$ & $Z_2$ & $Z_1$ & $Z_2$ & $Z_{U_1}$ & $Z_{U_2}$ & $Z_{S_1}$ & $Z_{S_2}$ \\
\hline
$I(Z;w_1)$ & 4.45 & 0.16 & 4.39 & 0.14 & 5.17 & 0.19 & \textbf{7.83} & 0.03 & 6.25 & 0.04\\
$I(Z;w_2)$ & 0.17 & 3.92 & 0.13 & 4.26 & 0.23 & 5.17 & 0.06 & \textbf{7.17} & 0.05 & 5.79\\
$I(Z;w_s)$ & 12.61 & 12.06 & 11.30 & 11.47 & 7.48 & 7.17 & 9.47 & 9.89 & 10.13 & 9.40\\
\hline \hline
\end{tabular}

\vspace{-4mm}
\label{tab:probing}
\end{table*}

\textbf{Representation probing information}: We run a probing experiment to compute how well different contrastive representations capture shared and unique information. In Table~\ref{tab:probing}, for the $Z_i$'s learned by each method, we approximately compute $I(Z_i;w_1)$, $I(Z_i;w_2)$, and $I(Z_i;w_s)$ with respect to ground truth generative variables $w_s$, $w_1$, and $w_2$. As expected, existing methods such as SimCLR capture smaller amounts of unique information (roughly $4$ bits in $I(Z_i;w_1)$ and $I(Z_i;w_2)$), focusing instead on learning $I(Z_i;w_s)$ (12 bits). Cross+self captures slightly larger $I(Z_i;w_2)=4.26$, and SupCon with labeled data captures up to $5$ bits of unique information. Our \names\ approach captures $7$ bits of unique information and maintains $10$ bits of shared information, with total information captured higher than the other approaches. Furthermore, $\{Z_{S_1},Z_{S_2}\}$ capture more information about $w_s$, $Z_{U_1}$ about $w_1$, and $Z_{U_2}$ about $w_2$, indicating that factorization in our approach is successful.

\vspace{-1mm}
\subsection{Self-supervised learning with low redundancy and high uniqueness}
\vspace{-1mm}

\textbf{Multimodal fusion datasets}: We use a large collection of real-world datasets provided in MultiBench~\citep{liang2021multibench}, where we expect varying ratios of shared and unique information important for the task, to compare \names\ with other CL baselines:
\begin{enumerate}[noitemsep,topsep=0pt,nosep,leftmargin=*,parsep=0pt,partopsep=0pt]
    \item \textsc{MIMIC}~\cite{johnson2016mimic}: mortality and disease prediction from $36,212$ medical records (tabular patient data and medical time-series sensors from ICU).

    \item \textsc{MOSEI}~\cite{zadeh2018multimodal}: multimodal sentiment and emotion benchmark with $23,000$ monologue videos.

    \item \textsc{MOSI}~\cite{zadeh2016mosi}: multimodal sentiment analysis from $2,199$ YouTube videos.

    \item \textsc{UR-FUNNY}~\citep{hasan2019ur}: a dataset of humor detection from more than $16,000$ TED talk videos.

    \item \textsc{MUsTARD}~\citep{castro2019towards}: a corpus of $690$ videos for research in sarcasm detection from TV shows.

    \item \textsc{IRFL}~\cite{yosef2023irfl}: $6,697$ matching images and figurative captions (rather than literal captions).
\end{enumerate}
Together, these datasets cover seven different modalities from the healthcare, affective computing, and multimedia research areas and total more than $84,000$ data points.
For \textsc{MIMIC} with tabular and medical sensor inputs, we train self-supervised CL models on top of raw modality inputs.
For \textsc{IRFL} with image and caption inputs, we start with a pretrained CLIP model~\cite{radford2021learning} and perform continued pre-training to update CLIP weights with our \names\ objectives, before linear classifier testing. For the remaining four video datasets, we train self-supervised CL models starting from standard pre-extracted text, video, and audio features~\cite{liang2021multibench}. Please refer to the full paper~\cite{liang2023factorized} for experimental details. We release our code and models at \url{https://github.com/pliang279/FactorCL}.

\textbf{Multimodal fusion results}: From Table~\ref{tab:fusion}, \names\ significantly outperforms the baselines that do not capture both shared and unique information in both supervised and self-supervised settings, particularly on \textsc{MuStARD} (where unique information expresses sarcasm, such as sardonic facial expressions or ironic tone of voice), and on \textsc{MIMIC} (with unique health indicators and sensor readings).
In Table~\ref{tab:fig}, we also show that \names\ substantially improves the state-of-the-art in classifying images and figurative captions which are not literally descriptive of the image on \textsc{IRFL}, outperforming zero-shot and fine-tuned CLIP~\cite{radford2021learning} as well as continued pre-training baselines on top of CLIP.

\begin{table*}[t]
\centering
\fontsize{8}{10}\selectfont
\setlength\tabcolsep{1pt}
\vspace{-1mm}
\caption{Results on MultiBench~\citep{liang2021multibench} datasets with varying shared and unique information: \names\ achieves strong results vs self-supervised (top $5$ rows) and supervised (bottom $3$ rows) baselines that do not have unique representations, factorization, upper-bounds to remove irrelevant information, and multimodal augmentations.}
\centering

\begin{tabular}{l|ccccc|ccccc}
\hline \hline
Model & $(X_1; X_2)$ & $(X_i; X_i')$ & $(X_1; X_2|Y)$ & $(X_2'')$ & Fact & \textsc{MIMIC} & \textsc{MOSEI} & \textsc{MOSI} & \textsc{UR-FUNNY} & \textsc{MUStARD} \\
\hline

SimCLR~\cite{chen2020simple} & \textcolor{gg}\cmark & \textcolor{rr}\xmark & \textcolor{rr}\xmark & \textcolor{rr}\xmark & \textcolor{rr}\xmark & 66.67\% & 71.03\% & 46.21\% & 50.09\% & 53.48\%\\
Cross+Self~\cite{wang2022rethinking} & \textcolor{gg}\cmark & \textcolor{gg}\cmark & \textcolor{rr}\xmark & \textcolor{rr}\xmark & \textcolor{rr}\xmark & 65.20\% & 71.04\% & 46.92\% & 56.52\% & 53.91\%\\

Cross+Self+Fact~\cite{yuan2021multimodal} & \textcolor{gg}\cmark & \textcolor{gg}\cmark & \textcolor{rr}\xmark & \textcolor{rr}\xmark & \textcolor{gg}\cmark & 65.49\% & 71.07\% & 52.37\% & 59.91\% & 53.91\% \\

OurCL-SSL & \textcolor{gg}\cmark & \textcolor{gg}\cmark & \textcolor{gg}\cmark & \textcolor{gg}\cmark & \textcolor{rr}\xmark & 65.22\% & 71.16\% & 48.98\% & 58.79\% & 53.98\%\\
\names-SSL & \textcolor{gg}\cmark & \textcolor{gg}\cmark & \textcolor{gg}\cmark & \textcolor{gg}\cmark & \textcolor{gg}\cmark & \textbf{67.34}\% & \textbf{74.88\%} & \textbf{52.91\%} & \textbf{60.50\%} & \textbf{55.80}\%\\
\hline
SupCon~\cite{khosla2020supervised} & \textcolor{rr}\xmark & \textcolor{rr}\xmark & \textcolor{gg}\cmark & \textcolor{rr}\xmark & \textcolor{rr}\xmark & 67.37\% & 72.71\% & 47.23\% & 50.98\% & 52.75\%\\
OurCL-SUP & \textcolor{gg}\cmark & \textcolor{gg}\cmark & \textcolor{gg}\cmark & \textcolor{rr}\xmark & \textcolor{rr}\xmark & 68.16\% & 71.15\% & 65.32\% & 58.32\% & 65.05\%\\
\names-SUP & \textcolor{gg}\cmark & \textcolor{gg}\cmark & \textcolor{gg}\cmark & \textcolor{rr}\xmark & \textcolor{gg}\cmark & \textbf{76.79\%} & \textbf{77.34\%} & \textbf{70.69\%} & \textbf{63.52\%} & \textbf{69.86}\%\\
\hline \hline
\end{tabular}

\vspace{-4mm}
\label{tab:fusion}
\end{table*}

\textbf{Modeling ablations}: 
In Table~\ref{tab:fusion}, we also carefully ablate each component in our method and indicate either existing baselines or newly-run ablation models.
\begin{enumerate}[noitemsep,topsep=0pt,nosep,leftmargin=*,parsep=0pt,partopsep=0pt]
    \item \textbf{Factorized representations}: In comparing \names-SSL with OurCL-SSL, and also \names-SUP with OurCL-SUP, we find that factorization is critical: without it, performance drops on average $6.1\%$, with performance drop as high as $8.6\%$ for \textsc{MIMIC}.
    \item \textbf{Information removal via upper bound}: By comparing \names\ with  SimCLR, Cross+Self, and Cross+Self+Fact, and SupCon that only seek to capture task-relevant information via contrastive lower bounds on MI, we find that separately modeling the task-relevant information (to be captured) and task-irrelevant information (to be removed) is helpful. Without removing task-irrelevant information via the upper-bound objective, performance drops on average $13.6\%$, with performance drops as high as $23.5\%$ for the \textsc{MOSI} dataset. We also found that training was more difficult without this objective, which is expected due to overwhelming superfluous information from the dataset \cite{zadeh2018multimodal}.
    \item \textbf{Multimodal augmentations}: Finally, we investigate the differences between separate unimodal augmentations (\names-IndAug in  Table~\ref{tab:fig}) versus a joint multimodal augmentation (\names-SSL) on the \textsc{IRFL} dataset. We choose this dataset since its images and captions are the easiest to visualize (see Figure~\ref{fig:augs} for augmentations from both strategies).
    In the self-supervised setting, we find that multimodal augmentations achieve $95\%$ performance, higher than the $92\%$ for separate unimodal augmentations, and both outperform baselines SimCLR and Cross+Self.
\end{enumerate}

\begin{wraptable}{r}{5cm}
\centering
\fontsize{9}{11}\selectfont
\setlength\tabcolsep{2pt}
\vspace{-4mm}
\caption{Continued pre-training on CLIP with our \names\ objectives on classifying images and figurative language.}
\centering
\begin{tabular}{l|c c}
\hline \hline
Task & \textsc{IRFL} \\
\hline
Zero-shot CLIP~\cite{radford2021learning} & 89.15\%\\
SimCLR~\cite{chen2020simple} & 91.57\%\\
Cross+Self~\cite{wang2022rethinking,yuan2021multimodal} & 95.18\%\\
\names-IndAug & 92.77\% \\
\names-SSL & \textbf{95.18}\%\\
\hline
Fine-tuned CLIP~\cite{radford2021learning} & 96.39\%\\
SupCon~\cite{khosla2020supervised} & 89.16\%\\
\names-SUP & \textbf{98.80}\%\\
\hline \hline
\end{tabular}
\label{tab:fig}
\vspace{-2mm}
\end{wraptable}

\textbf{Ablations on $S, U_1$ and $U_2$}:
In Table~\ref{tab:ablations}, we also test \names\ when training linear classifiers on top of only shared $\{Z_{S_1},Z_{S_2}\}$ and unique $Z_{U_1}$, $Z_{U_2}$ separately. We call these models \names-$S$, \names-$U_1$, and \names-$U_2$.
Immediately, we observe that performance drops as compared to the full \names\ model, indicating that both shared and unique information are critical in real-world multimodal tasks. 
As expected, the best-performing submodel is the one that captures the region with the largest amount of task-relevant information: \textsc{MOSEI} and \textsc{MOSI} are known to include a lot of redundancy and unique information since language is very important for detecting sentiment~\cite{zadeh2018multimodal}, so \names-$S$ and \names-$U_2$ perform best. For sarcasm detection on \textsc{MuStARD}, video information is most important with \names-$U_1$ performing best ($59.4\%$), and ablation models are also the furthest away from full multimodal performance ($69.9\%$). This is aligned with intuition where sarcasm is expressed through tone of voice and visual gestures (high $U_1$), as well as from contradictions between language and video (higher multimodal performance).

\begin{table*}[t]
\centering
\fontsize{9}{11}\selectfont
\setlength\tabcolsep{3pt}
\vspace{-0mm}
\caption{We ablate using only shared representations $\{Z_{S_1},Z_{S_2}\}$, unique representation $Z_{U_1}$, and $Z_{U_2}$ separately for prediction. Both shared and unique information are critical in real-world multimodal tasks.}
\centering

\begin{tabular}{l|ccccc}
\hline \hline
Model & \textsc{MIMIC} & \textsc{MOSEI} & \textsc{MOSI} & \textsc{UR-FUNNY} & \textsc{MUStARD} \\
\hline
\names-$S$ & 63.77\% & 77.17\% & 70.12\% & 63.42\% & 57.25\%\\
\names-$U_1$ & 55.90\% & 77.06\% & 70.11\% & 62.00\% & 59.42\%\\
\names-$U_2$ & 69.08\% & 71.01\% & 52.33\% & 54.35\% & 53.62\%\\
\hline
\names-SUP & \textbf{76.79\%} & \textbf{77.34\%} & \textbf{70.69\%} & \textbf{63.52\%} & \textbf{69.86\%} \\
\hline \hline
\end{tabular}

\vspace{-6mm}
\label{tab:ablations}
\end{table*}

\vspace{-1mm}
\section{Related Work}
\vspace{-1mm}

\textbf{Contrastive learning} is a successful self-supervised learning paradigm for computer vision~\cite{oord2018representation,chen2020simple,he2020momentum,grill2020bootstrap,chen2021exploring,caron2020unsupervised}, natural language~\cite{gao2021simcse, meng2021coco, neelakantan2022text}, speech~\cite{oord2018representation, schneider2019wav2vec, baevski2020wav2vec}, and multimodal tasks~\cite{radford2021learning, jia2021scaling, akbari2021vatt}. Its foundational underpinnings are inspired by work in multiview information theory~\cite{federici2020learning,khosla2020supervised,sridharan2008information,tian2020makes,tsai2020self} studying the shared information between two views and whether they are necessary or sufficient in predicting the label. Recently,~\citet{wang2022rethinking} and~\citet{kahana2022contrastive} discuss the limitations of assuming multiview redundancy and propose autoencoder reconstruction or unimodal contrastive learning to retain unique information, which resembles the Cross+self baselines in our experiments. We refer the reader to~\citet{shwartz2023compress} for a comprehensive review on multiview and contrastive learning. Our work also relates to conditional contrastive learning \cite{tsai2022conditional, ma2021conditional, ye2022contrastive, chi2022conditional}, where positive or negative pairs are supposed to sample from conditional distributions. 

\textbf{Multimodal contrastive learning} aims to align related data from different modalities, typically provided as positive pairs. This could be done via optimizing a contrastive objective for inter-modality pairs \cite{radford2021learning,alayrac2020self, akbari2021vatt, jia2021scaling}, or both intra- and inter-modality data pairs \cite{yuan2021multimodal, huang2021multilingual, lee2020parameter, jain2021mural, kim2022transferring}. Our work also relates to factorized representation learning, which primarily studies how to capture modality-specific information primarily in each modality and multimodal information redundant in both modalities~\cite{hsu2018disentangling,tsai2019learning}. Prior work has used disentangled latent variable models~\cite{Bengio:2013:RLR:2498740.2498889,higgins2016beta,hsu2018disentangling,tsai2019learning}, mixture-of-experts~\cite{shi2019variational}, or product-of-experts~\cite{wu2018multimodal} layer to explain factors in multimodal data.

\textbf{Information theory} \cite{cover1991information,shannon1948mathematical} has been used to study several phenomena in multimodal learning, including co-learning~\cite{zadeh2020foundations, rahate2022multimodal} and multi-view learning~\cite{tsai2020self, huang2021makes}. Due to its theoretical importance, several lower and upper bounds have been proposed for practical estimation~\cite{oord2018representation,wu2020mutual,poole2019variational,ozair2019wasserstein}. We build on the CLUB upper bound~\cite{cheng2020club} to create a more accurate and stable bound. 
Our characterizations of shared and unique information are also related to partial information decomposition \cite{williams2010nonnegative}, co-information \cite{bell2003co, vergara2014review}, interaction information \cite{mcgill1954multivariate}, and cross-domain disentanglement \cite{hwang2020variational} research.

\vspace{-1mm}
\section{Conclusion}
\vspace{-1mm}

This paper studied how standard CL methods suffer when task-relevant information lies in regions unique to each modality, which is extremely common in real-world applications such as sensor placement, medical testing, and multimodal interaction. In response, we proposed \names, a new method expanding CL techniques through the use of factorized representations, removing task-irrelevant information via upper bounds on MI, and multimodal data augmentations suitable for approximating the unobserved task. Based on \names's strong performance, there are several exciting directions in extending these ideas for masked and non-contrastive pre-training.

\chapter{Quantifying Multimodal Interactions in Trained Models}
\label{chap:foundations3}
\newcommand{\dataurl}{\url{https://github.com/pliang279/MultiViz}}
\renewcommand{\namel}{\textsc{MultiViz}}
\renewcommand{\name}{\textsc{MultiViz}}

\vspace{-3mm}
\section{Introduction}
\vspace{-1mm}

Using our foundation of multimodal interactions, we now present our work in \textit{model quantification}: visualizing and understanding the internal modeling of multimodal interactions in trained models. As multimodal models are increasingly deployed in real-world applications, it has become increasingly important to quantify and understand their internal mechanics~\citep{liang2022foundations,goyal2016towards,park2018multimodal} as a step towards accurately benchmarking their limitations for more reliable deployment~\citep{hendricks2018women,jabri2016revisiting}. However, modern multimodal models are typically black-box neural networks, such as pretrained transformers~\citep{li2019visualbert,lu2019vilbert}, which makes understanding what interactions they learn difficult.

As a step in interpreting multimodal models, this paper introduces an analysis and visualization method called \multiviz\ (see Figure~\ref{fig:analysis}). To tackle the challenges of visualizing model behavior, we scaffold the problem of interpretability into $4$ stages: (1) \textit{unimodal importance}: identifying the contributions of each modality towards downstream modeling and prediction, (2) \textit{cross-modal interactions}: uncovering the various ways in which different modalities can relate with each other and the types of new information possibly discovered as a result of these relationships, (3) \textit{multimodal representations}: how unimodal and cross-modal interactions are represented in decision-level features, and (4) \textit{multimodal prediction}: how decision-level features are composed to make a prediction for a given task.
In addition to including current approaches for unimodal importance~\citep{goyal2016towards,merrick2020explanation,lime} and cross-modal interactions~\citep{hessel2020does,lyu2022dime}, we additionally propose new methods for interpreting cross-modal interactions, multimodal representations, and prediction to complete these stages in \multiviz.
By viewing multimodal interpretability through the lens of these $4$ stages, \multiviz\ contributes a \textit{modular} and \textit{human-in-the-loop} visualization toolkit for the community to visualize popular multimodal datasets and models as well as compare with other interpretation perspectives, and for stakeholders to understand multimodal models in their research domains.

\multiviz\ is designed to support many modality inputs while also operating on diverse modalities, models, tasks, and research areas. Through experiments on $6$ real-world multimodal tasks (spanning fusion, retrieval, and question-answering), $6$ modalities, and $8$ models, we show that \multiviz\ helps users gain a deeper understanding of model behavior as measured via a proxy task of model simulation. We further demonstrate that \multiviz\ helps human users assign interpretable language concepts to previously uninterpretable features and perform error analysis on model misclassifications. Finally, using takeaways from error analysis, we present a case study of human-in-the-loop model debugging.
Overall, \multiviz\ provides a practical toolkit for interpreting multimodal models for human understanding and debugging. \multiviz\ datasets, models, and code are at \dataurl.

\vspace{-1mm}
\section{\mbox{\multiviz: Visualizing \& Understanding Multimodal Models}}
\label{sec:main}
\vspace{-1mm}

\begin{figure}[tbp]
\centering
\vspace{-0mm}
\includegraphics[width=\linewidth]{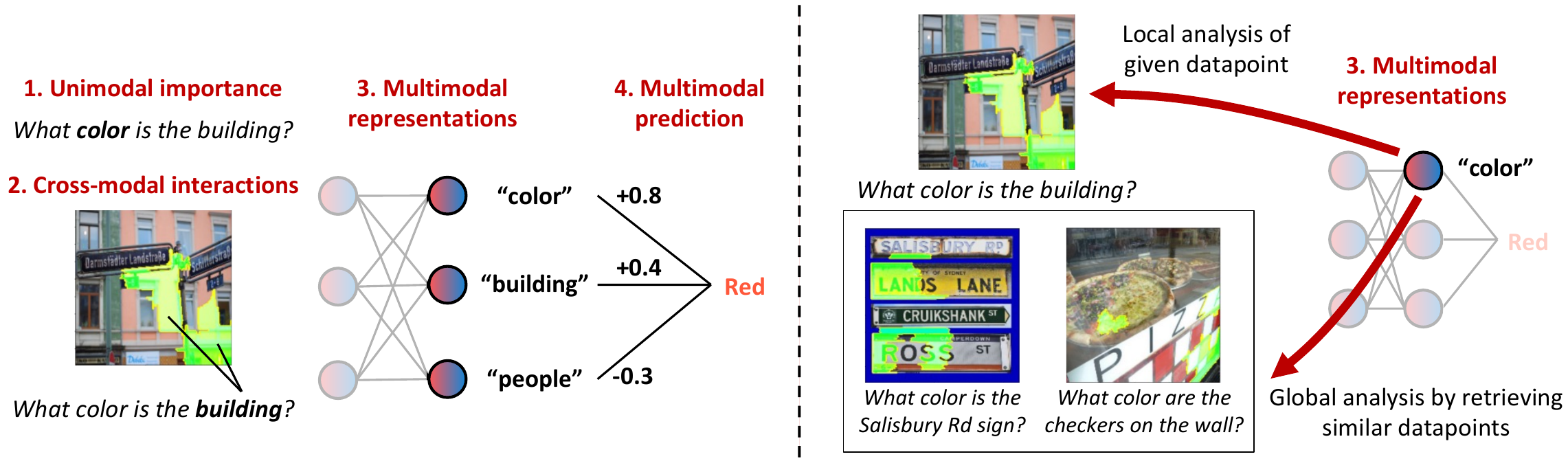}
\vspace{-2mm}
\caption{\textbf{Left}: We scaffold the problem of multimodal interpretability and propose \name, a comprehensive analysis method encompassing a set of fine-grained analysis stages: (1) \textbf{unimodal importance} identifies the contributions of each modality, (2) \textbf{cross-modal interactions} uncover how different modalities relate with each other and the types of new information possibly discovered as a result of these relationships, (3) \textbf{multimodal representations} study how unimodal and cross-modal interactions are represented in decision-level features, and (4) \textbf{multimodal prediction} studies how these features are composed to make a prediction. \textbf{Right}: We visualize multimodal representations through local and global analysis. Given an input datapoint, \textbf{local analysis} visualizes the unimodal and cross-modal interactions that activate a feature. \textbf{Global analysis} informs the user of similar datapoints that also maximally activate that feature, and is useful in assigning human-interpretable concepts to features by looking at similarly activated input regions (e.g., the concept of color).}
\label{fig:analysis}
\vspace{-2mm}
\end{figure}

This section presents \multiviz, our proposed analysis framework for analyzing the behavior of multimodal models. As a general setup, we assume multimodal datasets take the form $\mathcal{D} = \{ (\mathbf{x}_1, \mathbf{x}_2, y)_{i=1}^n \} = \{ ( x_1^{(1)}, x_1^{(2)}, ..., x_2^{(1)}, x_2^{(2)}, ..., y )_{i=1}^n \}$, with boldface $\mathbf{x}$ denoting the entire modality, each $x_1, x_2$ indicating modality atoms (i.e., fine-grained sub-parts of modalities that we would like to analyze, such as individual words in a sentence, object regions in an image, or time-steps in time-series data), and $y$ denoting the label. These datasets enable us to train a multimodal model $\hat{y} = f(\mathbf{x}_1, \mathbf{x}_2; \theta)$ which we are interested in visualizing.

Modern parameterizations of multimodal models $f$ are typically black-box neural networks, such as multimodal transformers~\citep{hendricks2021decoupling,tsai2019multimodal} and pretrained models~\citep{li2019visualbert,lu2019vilbert}. How can we visualize and understand the internal modeling of multimodal information and interactions in these models? Having an accurate understanding of their decision-making process would enable us to benchmark their opportunities and limitations for more reliable real-world deployment.
However, interpreting $f$ is difficult.
In many multimodal problems, it is useful to first scaffold the problem of interpreting $f$ into several intermediate stages from low-level unimodal inputs to high-level predictions, spanning \textit{unimodal importance}, \textit{cross-modal interactions}, \textit{multimodal representations}, and \textit{multimodal prediction}. Each of these stages provides complementary information on the decision-making process (see Figure~\ref{fig:analysis}). We now describe each step in detail and propose methods to analyze each step.

\vspace{-1mm}
\subsection{Unimodal importance (U)}
\label{analysis:unimodal}
\vspace{-1mm}

Unimodal importance aims to understand the contributions of each modality towards modeling and prediction. It builds upon ideas of gradients~\citep{simonyan2013deep,baehrens2010explain,erhan2009visualizing} and feature attributions (e.g., LIME~\citep{lime}, Shapley values~\citep{merrick2020explanation}).
We implement unimodal feature attribution methods as a module $\textsc{Uni}(f_\theta, y, \mathbf{x})$ taking in a trained model $f_\theta$, an output/feature $y$ which analysis is performed with respect to, and the modality of interest $\mathbf{x}$. $\textsc{Uni}$ returns importance weights across atoms $x$ of modality $\mathbf{x}$.

\vspace{-1mm}
\subsection{Cross-modal interactions (C)}
\label{analysis:crossmodal}
\vspace{-1mm}

\begin{figure}[tbp]
\centering
\vspace{-2mm}
\includegraphics[width=\linewidth]{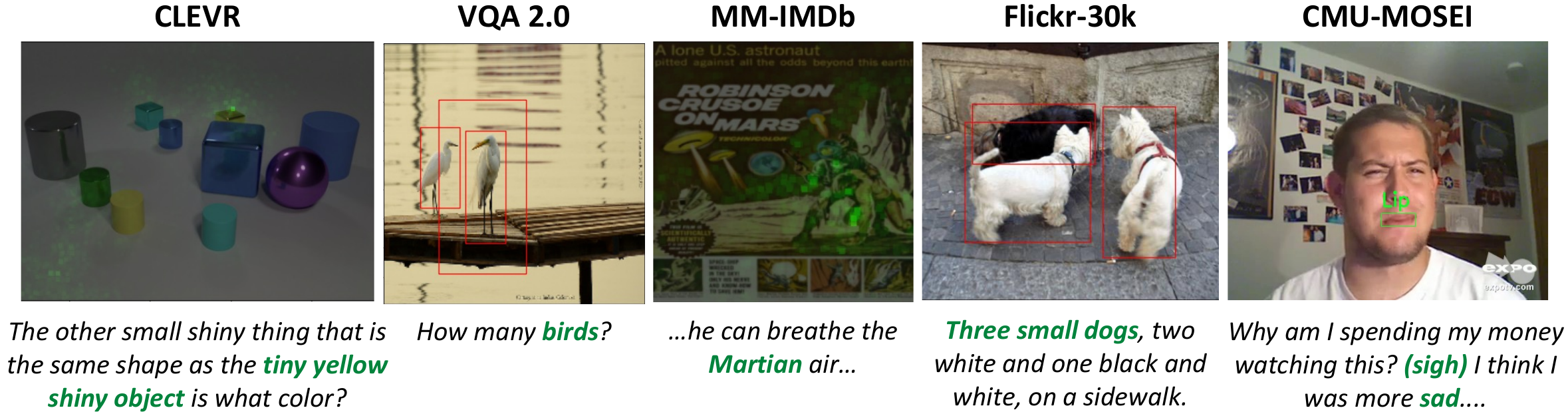}
\vspace{-4mm}
\caption{Examples of cross-modal interactions discovered by our proposed second-order gradient approach: first taking a gradient of model $f$ with respect to an input word (e.g., $x_1 = \textit{birds}$), before taking a second-order gradient with respect to all image pixels (highlighted in green) or bounding boxes (in red boxes) $\mathbf{x}_2$ indeed results in all birds in the image being highlighted.}
\label{fig:crossmodal}
\vspace{-4mm}
\end{figure}

Cross-modal interactions describe various ways in which atoms from different modalities can relate with each other and the types of new information possibly discovered as a result of these relationships.
Recent work~\citep{hessel2020does,lyu2022dime} has formalized a definition of cross-modal interactions by building upon literature in statistical non-additive interactions:

\textbf{Definition 1} (Statistical Non-Additive Interaction~\citep{friedman2008predictive,sorokina2008detecting,tsang2018detecting,tsang2019feature}). A function $f$ learns a feature interaction $\mathcal{I}$ between $2$ unimodal atoms $x_1$ and $x_2$ if and only if $f$ cannot be decomposed into a sum of unimodal subfunctions $g_1, g_2$ such that $f(x_1, x_2) = g_1(x_1) + g_2(x_2)$.

This definition of non-additive interactions is general enough to include different ways that interactions can happen, including multiplicative interactions from complementary views of the data (i.e., an interaction term $x_1 \mathbb{W} x_2$~\citep{jayakumar2020multiplicative}), or cooperative interactions from equivalent views (i.e., an interaction term $\textrm{majority} (f(x_1), f(x_2))$~\citep{ding2021cooperative}).
Using this definition, \multiviz\ first includes two recently proposed methods for understanding cross-modal interactions: EMAP~\citep{hessel2020does} decomposes $f(x_1,x_2) = g_1(x_1) + g_2(x_2) + g_{12}(x_1,x_2)$ into strictly unimodal representations $g_1, g_2$, and cross-modal representation $g_{12} = f - \mathbb{E}_{x_1} (f) - \mathbb{E}_{x_2} (f) + \mathbb{E}_{x_1,x_2} (f)$ to quantify the degree of global cross-modal interactions across an entire dataset. DIME~\citep{lyu2022dime} further extends EMAP using feature visualization on each disentangled representation locally (per datapoint).
However, these approaches require approximating expectations over modality subsets, which may not scale beyond $2$ modalities. To fill this gap, we propose an efficient approach for visualizing these cross-modal interactions by observing that the following gradient definition directly follows from Definition 1:

\textbf{Definition 2} (Gradient definition of statistical non-additive interaction). A function $f$ exhibits non-additive interactions among $2$ unimodal atoms $x_1$ and $x_2$ if $\mathbf{E}_{x_1,x_2} \left[ \frac{\partial^2 f(x_1,x_2)}{\partial x_1 \partial x_2} \right]^2> 0$.

Taking a second-order gradient of $f$ zeros out the unimodal terms $g_1(x_1)$ and $g_2(x_2)$ and isolates the interaction $g_{12}(x_1,x_2)$. Theoretically, second-order gradients are necessary and sufficient to recover cross-modal interactions: purely additive models will have strictly $0$ second-order gradients so $\mathbf{E}_{x_1,x_2} \left[ \frac{\partial^2 f(x_1,x_2)}{\partial x_1 \partial x_2} \right]^2 =0$, and any non-linear interaction term $g_{12}(x_1,x_2)$ has non-zero second-order gradients since $g$ cannot be a constant or unimodal function, so $\mathbf{E}_{x_1,x_2} \left[ \frac{\partial^2 f(x_1,x_2)}{\partial x_1 \partial x_2} \right]^2> 0$.

Definition 2 inspires us to extend first-order gradient and perturbation-based approaches~\citep{han2020explaining,lime,yosinski2015understanding} to the second order.
Our implementation first computes a gradient of $f$ with respect to a modality atom which the user is interested in querying cross-modal interactions for (e.g., $x_1 = \textit{birds}$), which results in a vector $\nabla_1 = \frac{\partial f}{\partial x_1}$ of the same dimension as $x_1$ (i.e., token embedding dimension). We aggregate the vector components of $\nabla_1$ via summation to produce a single scalar $\lVert \nabla_1 \rVert$, before taking a second-order gradient with respect to all atoms of the second modality $x_2 \in \mathbf{x}_2$ (e.g., all image pixels), which results in a vector $\nabla_{12} = \left[ \frac{\partial^2 f}{\partial x_1 \partial x_2^{(1)}}, ..., \frac{\partial^2 f}{\partial x_1 \partial x_2^{(|\mathbf{x}_2|)}} \right]$ of the same dimension as $\mathbf{x}_2$ (i.e., total number of pixels). Each scalar entry in $\nabla_{12}$ highlights atoms $x_2$ that have non-linear interactions with the original atom $x_1$, and we choose the $x_2$'s with the largest magnitude of interactions with $x_1$ (i.e., which highlights the birds in the image, see Figure~\ref{fig:crossmodal} for examples on real datasets). We implement a general module $\textsc{CM}(f_\theta, y, x_1, \mathbf{x}_2)$ for cross-modal visualizations, taking in a trained model $f_\theta$, an output/feature $y$, the first modality's atom of interest $x_1$, and the entire second modality of interest $\mathbf{x}_2$, before returning importance weights across atoms $x_2$ of modality $\mathbf{x}_2$.

\vspace{-1mm}
\subsection{Multimodal representations}
\label{analysis:representation}
\vspace{-1mm}

Given these highlighted unimodal and cross-modal interactions at the input level, the next stage aims to understand how these interactions are represented at the feature representation level. Specifically, given a trained multimodal model $f$, define the matrix $M_z \in \mathbb{R}^{N \times d}$ as the penultimate layer of $f$ representing (uninterpretable) deep feature representations implicitly containing information from both unimodal and cross-modal interactions. For the $i$th datapoint, $z = M_z (i)$ collects a set of individual feature representations $z_{1}, z_{2}, ..., z_{d} \in \mathbb{R}$. We aim to interpret these feature representations through both local and global analysis (see Figure~\ref{fig:analysis} (right) for an example):

\textbf{Local representation analysis ($\textrm{R}_\ell$)} informs the user on parts of the original datapoint that activate feature $z_{j}$. To do so, we run unimodal and cross-modal visualization methods with respect to feature $z_{j}$ (i.e., $\textsc{Uni}(f_\theta, z_{j}, \mathbf{x})$, $\textsc{CM}(f_\theta, z_{j}, x_1, \mathbf{x}_2)$) in order to explain the input unimodal and cross-modal interactions represented in feature $z_{j}$. Local analysis is useful in explaining model predictions on the original datapoint by studying the input regions activating feature $z_{j}$.

\textbf{Global representation analysis ($\textrm{R}_g$)} provides the user with the top $k$ datapoints $\mathcal{D}_k(z_{j}) = \{ (\mathbf{x}_1, \mathbf{x}_2, y)_{i=(1)}^k \}  $ that also maximally activate feature $z_{j}$. By further unimodal and cross-modal visualizations on datapoints in $\mathcal{D}_k(z_{j})$, global analysis is especially useful in helping humans assign interpretable language concepts to each feature by looking at similarly activated input regions across datapoints (e.g., the concept of color in Figure~\ref{fig:analysis}, right). Global analysis can also help to find related datapoints the model also struggles with for error analysis.

\vspace{-1mm}
\subsection{Multimodal prediction (P)}
\label{analysis:prediction}
\vspace{-1mm}

Finally, the prediction step takes the set of feature representations $z_{1}, z_{2}, ..., z_{d}$ and composes them to form higher-level abstract concepts suitable for a task.
We approximate the prediction process with a linear combination of penultimate layer features by integrating a sparse linear prediction model with neural network features~\citep{wong2021leveraging}. Given the penultimate layer $M_z \in \mathbb{R}^{N \times d}$, we fit a linear model $\mathbb{E}\left( Y|X=x \right) = M_z^\top \beta$ (bias $\beta_0$ omitted for simplicity) and solve for sparsity using:
\begin{equation}
\label{sparse_eqn}
    \hat{\beta} = \argmin_{\beta} \frac{1}{2N} \| M_z^\top \beta - y \|_2^2 + \lambda_1 \| \beta \|_1 + \lambda_2 \| \beta \|_2^2.
\end{equation}
The resulting understanding starts from the set of learned weights with the highest non-zero coefficients $\beta_{\textrm{top}} = \{ \beta_{(1)}, \beta_{(2)}, ... \}$ and corresponding ranked features $z_{\textrm{top}} = \{z_{(1)}, z_{(2)}, ...\}$. $\beta_{\textrm{top}}$ tells the user how features $z_{\textrm{top}}$ are composed to make a prediction, and $z_{\textrm{top}}$ can then be visualized with respect to unimodal and cross-modal interactions using the representation stage (Section~\ref{analysis:representation}).

\begin{figure}[tbp]
\centering
\vspace{-2mm}
\includegraphics[width=\linewidth]{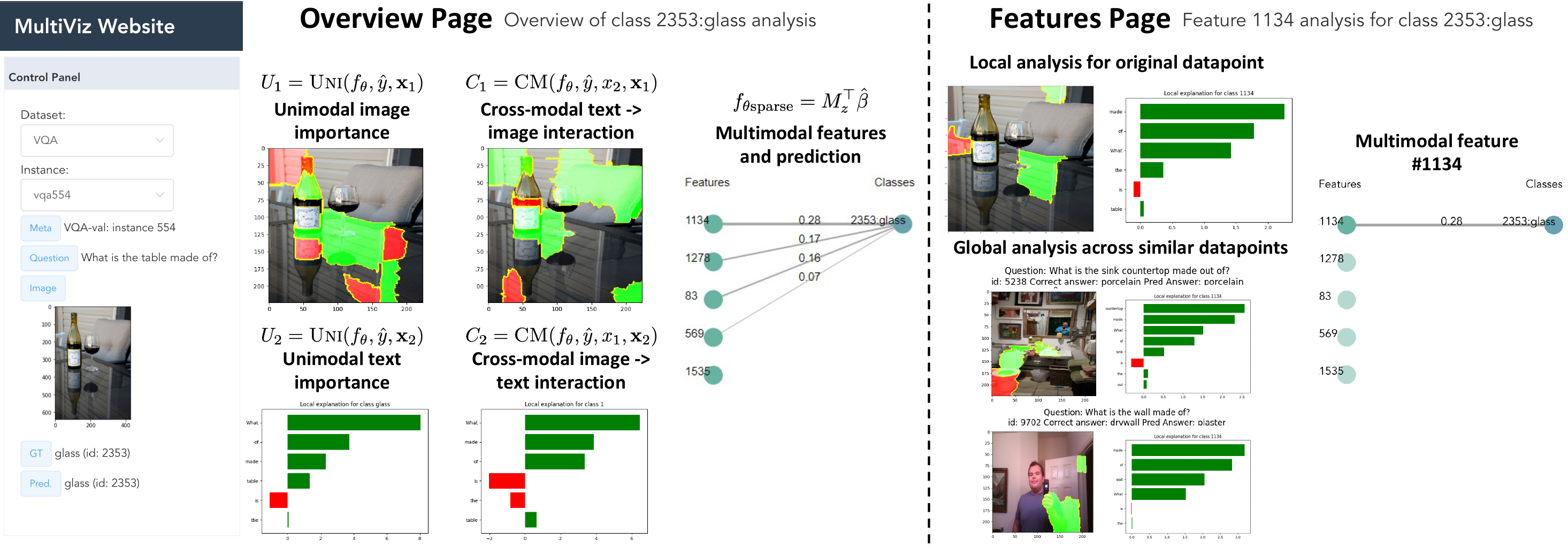}
\vspace{-4mm}
\caption{\name\ provides an interactive visualization API across multimodal datasets and models. The overview page shows general unimodal importance, cross-modal interactions, and prediction weights, while the features page enables local and global analysis of specific user-selected features.}
\label{fig:algo}
\vspace{-2mm}
\end{figure}

\vspace{-1mm}
\subsection{Putting everything together}
\vspace{-1mm}

We summarize these proposed approaches for understanding each step of the multimodal process and show the overall \multiviz\ user interface in Figure~\ref{fig:algo}. This interactive API enables users to choose multimodal datasets and models and be presented with a set of visualizations at each stage, with an \textbf{overview page} for general unimodal importance, cross-modal interactions, and prediction weights, as well as a \textbf{feature page} for local and global analysis of user-selected features (see full paper~\cite{liang2023multiviz} for details).

\begin{table*}[]
\fontsize{9}{11}\selectfont
\setlength\tabcolsep{3.0pt}
\vspace{-0mm}
\caption{\name\ enables fine-grained analysis across $6$ datasets spanning $3$ research areas, $6$ input modalities ($\ell$: language, $i$: image, $v$: video, $a$: audio, $t$: time-series, $ta$: tabular), and $8$ models.}
\centering
\footnotesize
\vspace{-0mm}
\begin{tabular}{l|cccccccc}
\hline \hline
\multicolumn{1}{l|}{Area} & Dataset & Model & Modalities & \# Samples & Prediction task \\ \hline
\multirow{3}{*}{Fusion} & \textsc{CMU-MOSEI} & \textsc{MulT} & $\{\ell,v,a\} \rightarrow y$ & $22,777$ & sentiment, emotions \\
& \textsc{MM-IMDb} & \textsc{LRTF} & $\{\ell,i\} \rightarrow y$ & $25,959$ & movie genre classification \\
& \textsc{MIMIC} & \textsc{LF} & $\{t,ta\} \rightarrow y$ & $36,212$ & mortality, ICD-$9$ codes \\ \hline \hline
\multirow{2}{*}{Retrieval} & \textsc{Flickr-30k} & \textsc{ViLT} & $\ell \leftrightarrow i$ & $158,000$ & image-caption retrieval \\
& \textsc{Flickr-30k} & \textsc{CLIP} & $\ell \leftrightarrow i$ & $158,000$ & image-caption retrieval \\ \hline \hline
\multirow{3}{*}{QA} & \textsc{CLEVR} &  \textsc{CNN-LSTM-SA} & $\{i,\ell\} \rightarrow y$ & $853,554$ & QA \\
& \textsc{CLEVR} & \textsc{MDETR} & $\{i,\ell\} \rightarrow y$ & $853,554$ & QA \\
& \textsc{VQA 2.0} & \textsc{LXMERT} & $\{i,\ell\} \rightarrow y$ & $1,100,000$ & QA \\ \hline \hline
\end{tabular}
\vspace{-4mm}
\label{tab:data}
\end{table*}


\vspace{-1mm}
\section{Experiments}
\vspace{-1mm}

Our experiments are designed to verify the usefulness and complementarity of the $4$ \multiviz\ stages. We start with a model simulation experiment to test the utility of each stage towards overall model understanding (Section~\ref{simulation}). We then dive deeper into the individual stages by testing how well \multiviz\ enables representation interpretation (Section~\ref{local}) and error analysis (Section~\ref{error}), before presenting a case study of model debugging from error analysis insights (Section~\ref{debugging}). We showcase the following selected experiments and defer results on other datasets to the full paper~\cite{liang2023multiviz}.

\textbf{Setup}: We use a large suite of datasets from MultiBench~\citep{liang2021multibench} which span real-world fusion~\citep{zadeh2018multimodal,arevalo2017gated,MIMIC}, retrieval~\citep{plummer2015flickr30k}, and QA~\citep{johnson2017clevr,goyal2017making} tasks. For each dataset, we test a corresponding state-of-the-art model: \textsc{MulT}~\citep{tsai2019multimodal}, \textsc{LRTF}~\citep{liu2018efficient}, \textsc{LF}~\citep{baltruvsaitis2018multimodal}, \textsc{ViLT}~\citep{kim2021vilt}, \textsc{CLIP}~\citep{radford2021learning}, \textsc{CNN-LSTM-SA}~\citep{johnson2017clevr}, \textsc{MDETR}~\citep{kamath2021mdetr}, and \textsc{LXMERT}~\citep{tan2019lxmert}. These cover models both pretrained and trained from scratch. We summarize all $6$ datasets and $8$ models tested in Table~\ref{tab:data}, and provide more details in the full paper~\cite{liang2023multiviz}.

\vspace{-1mm}
\subsection{Model simulation}
\label{simulation}
\vspace{-1mm}

We first design a model simulation experiment to determine if \multiviz\ helps users of multimodal models gain a deeper understanding of model behavior.
If \multiviz\ indeed generates human-understandable explanations, humans should be able to accurately simulate model predictions given these explanations only, as measured by correctness with respect to actual model predictions and annotator agreement (Krippendorff's alpha~\citep{krippendorff2011computing}). To investigate the utility of each stage in \multiviz, we design a human study to see how accurately $21$ humans users ($3$ users for each of the following $7$ local ablation settings) can simulate model predictions:

(1) \textbf{U}: Users are only shown the unimodal importance (U) of each modality towards label $y$.

(2) \textbf{U + C}: Users are also shown cross-modal interactions (C) highlighted towards label $y$.

(3) \textbf{U + C + $\textrm{R}_\ell$}: Users are also shown local analysis ($\textrm{R}_\ell$) of unimodal and cross-modal interactions of top features $z_{\textrm{top}} = \{z_{(1)}, z_{(2)}, ...\}$ maximally activating label $y$.

(4) \textbf{U + C + $\textrm{R}_\ell$ + $\textrm{R}_g$}: Users are additionally shown global analysis ($\textrm{R}_g$) through similar datapoints that also maximally activate top features $z_{\textrm{top}}$ for label $y$.

(5) \textbf{\multiviz\ (U + C + $\textrm{R}_\ell$ + $\textrm{R}_g$ + P)}: The entire \multiviz\ method by further including visualizations of the final prediction (P) stage: sorting top ranked feature neurons $z_{\textrm{top}} = \{z_{(1)}, z_{(2)}, ...\}$ with respect to their coefficients $\beta_{\textrm{top}} = \{ \beta_{(1)}, \beta_{(2)}, ... \}$ and showing these coefficients to the user.

\begin{table*}[]
\fontsize{9}{11}\selectfont
\setlength\tabcolsep{4.0pt}
\vspace{-0mm}
\caption{\textbf{Model simulation}: We tasked $15$ humans users ($3$ users for each of the following local ablation settings) to simulate model predictions based on visualized evidences from \name. Human annotators who have access to all stages visualized in \name\ are able to accurately and consistently simulate model predictions (regardless of whether the model made the correct prediction) with high accuracy and annotator agreement, representing a step towards model understanding.}
\centering
\footnotesize
\vspace{-0mm}
\begin{tabular}{l|cc|cc|cc}
\hline \hline
Research area & \multicolumn{2}{c|}{QA} & \multicolumn{2}{c|}{Fusion} & \multicolumn{2}{c}{Fusion} \\
Dataset & \multicolumn{2}{c|}{\textsc{VQA 2.0}} & \multicolumn{2}{c|}{\textsc{MM-IMDb}} & \multicolumn{2}{c}{\textsc{CMU-MOSEI}} \\
Model & \multicolumn{2}{c|}{\textsc{LXMERT}} & \multicolumn{2}{c|}{\textsc{LRTF}} & \multicolumn{2}{c}{\textsc{MulT}} \\
\hline
Metric & Correctness & Agreement & Correctness & Agreement & Correctness & Agreement \\ \hline
U & $55.0 \pm 0.0$ & $0.39$ & $50.0 \pm 13.2$ & $0.34$ & $71.7 \pm 17.6$ & $0.39$ \\
U + C & $65.0 \pm 5.0$ & $0.50$ & $53.7 \pm 7.6$ & $0.51$ & $76.7 \pm 10.4$ & $0.45$ \\
U + C + $\textrm{R}_\ell$ & $61.7\pm 7.6 $ & $0.57$ & $56.7 \pm 7.6$ & $0.59$ & $78.3 \pm 2.9$ & $0.42$ \\
U + C + $\textrm{R}_\ell$ + $\textrm{R}_g$ & $71.7 \pm 15.3$ & $0.61$ & $61.7 \pm 7.6$ & $0.43$ & $\mathbf{100.0\pm 0.0}$ & $\mathbf{1.00}$ \\
\name & $\mathbf{81.7 \pm 2.9}$ & $\mathbf{0.86}$ & $\mathbf{65.0 \pm 5.0}$ & $\mathbf{0.60}$ & $\mathbf{100.0 \pm 0.0}$ & $\mathbf{1.00}$ \\
\hline \hline
\end{tabular}
\vspace{-2mm}
\label{tab:simulation}
\end{table*}

Using $20$ datapoints per setting, these experiments with $15$ users on $3$ datasets and $3$ models involve $35$ total hours of users interacting with \multiviz, which is a significantly larger-scale study of model simulation compared to prior work~\citep{aflalo2022vl,lyu2022dime,wang2021m2lens}.

\textbf{Quantitative results}: We show these results in Table~\ref{tab:simulation} and find that having access to all stages in \multiviz\ leads to significantly highest accuracy of model simulation on \textsc{VQA 2.0}, along with lowest variance and most consistent agreement between annotators.
On fusion tasks with \textsc{MM-IMDb} and \textsc{CMU-MOSEI}, we also find that including each visualization stage consistently leads to higher correctness and agreement, despite the fact that fusion models may not require cross-modal interactions to solve the task~\citep{hessel2020does}. More importantly, humans are able to simulate model predictions, regardless of whether the model made the correct prediction or not.

To test additional intermediate ablations, we conducted user studies on (6) \textbf{$\textrm{R}_\ell$ + P} (local analysis on final-layer features along with their prediction weights) and (7) \textbf{$\textrm{R}_g$ + P} (global analysis on final-layer features along with their prediction weights), to ablate the effect of overall analysis (\textbf{U} and \textbf{C}) and feature analysis (\textbf{$\textrm{R}_\ell$} or \textbf{$\textrm{R}_g$} in isolation). \textbf{$\textrm{R}_\ell$ + P} results in an accuracy of $51.7 \pm 12.6$ with $0.40$ agreement, while \textbf{$\textrm{R}_g$ + P} gives $71.7 \pm 7.6$ with $0.53$ agreement. Indeed, these underperform as compared to including overall analysis (\textbf{U} and \textbf{C}) and feature analysis (\textbf{$\textrm{R}_\ell$} + \textbf{$\textrm{R}_g$}).

Finally, we also scaled to $100$ datapoints on \textsc{VQA 2.0}, representing upwards of $10$ hours of user interaction (for the full \multiviz\ setting), and obtain an overall correctness of $80\%$, reliably within the range of model simulation using $20$ points ($81.7 \pm 2.9$). Therefore, the sample size of $20$ points that makes all experiments feasible is still a reliable sample.

We also conducted \textbf{qualitative interviews} to determine what users found useful in \multiviz:

(1) Users reported that they found local and global representation analysis particularly useful: global analysis with other datapoints that also maximally activate feature representations were important for identifying similar concepts and assigning them to multimodal features.

(2) Between Overview (\textbf{U + C}) and Feature (\textbf{$\textrm{R}_\ell$ + $\textrm{R}_g$ + P}) visualizations, users found Feature visualizations more useful in $31.7\%$, $61.7\%$, and $80.0\%$ of the time under settings (3), (4), and (5) respectively, and found Overview more useful in the remaining points. This means that for each stage, there exists a significant fraction of data points where that stage is most needed.

(3) While it may be possible to determine the prediction of the model with a subset of stages, having more stages that confirm the same prediction makes them a lot more confident about their prediction, which is quantitatively substantiated by the higher accuracy, lower variance, and higher agreement in human predictions.
We also include additional experiments in the full paper~\cite{liang2023multiviz}.

\vspace{-1mm}
\subsection{Representation interpretation}
\label{local}
\vspace{-1mm}

\begin{table}[t]
    \fontsize{9}{11}\selectfont
    \setlength\tabcolsep{2.0pt}
    \vspace{-0mm}
    \caption{\textbf{Left}: Across $15$ human users ($5$ users for each of the following $3$ settings), we find that users are able to consistently assign concepts to previously uninterpretable multimodal features using both local and global representation analysis. \textbf{Right}: Across $10$ human users ($5$ users for each of the following $2$ settings), we find that users are also able to categorize model errors into one of $3$ stages they occur in when given full \name\ visualizations.}
    \vspace{-0mm}
    \begin{subtable}{.3\linewidth}
      \centering
        \begin{tabular}{l|cc}
        \hline \hline
        Research area & \multicolumn{2}{c}{QA} \\
        Dataset & \multicolumn{2}{c}{\textsc{VQA 2.0}} \\
        Model & \multicolumn{2}{c}{\textsc{LXMERT}} \\
        \hline
        Metric & Confidence & Agree. \\
        \hline
        $\textrm{R}_\ell$ & $1.74 \pm 0.52$ & $0.18$ \\
        $\textrm{R}_\ell$ + $\textrm{R}_g$ (no viz) & $3.67 \pm 0.45$ & $0.60$ \\
        $\textrm{R}_\ell$ + $\textrm{R}_g$ & $\mathbf{4.50 \pm 0.43}$ & $\mathbf{0.69}$ \\
        \hline \hline
        \end{tabular}
    \end{subtable}%
    \begin{subtable}{.8\linewidth}
      \centering
        \begin{tabular}{l|cc|cc}
        \hline \hline
        Research area & \multicolumn{2}{c|}{QA} & \multicolumn{2}{c}{QA} \\
        Dataset & \multicolumn{2}{c|}{\textsc{CLEVR}} & \multicolumn{2}{c}{\textsc{VQA 2.0}} \\
        Model & \multicolumn{2}{c|}{\textsc{CNN-LSTM-SA}} & \multicolumn{2}{c}{\textsc{LXMERT}} \\
        \hline
        Metric & Confidence & Agree. & Confidence & Agree. \\
        \hline
        No viz & $2.72 \pm 0.15$ & $0.05$ & $2.15 \pm 0.70$ & $0.14$ \\
        \name & $\mathbf{4.12 \pm 0.45}$ & $\mathbf{0.67}$ & $\mathbf{4.21 \pm 0.62}$ & $\mathbf{0.60}$ \\
        \hline \hline
        \end{tabular}
    \end{subtable}
    \vspace{-2mm}
    \label{tab:local_rep}
\end{table}

\begin{figure}[t]
\centering
\vspace{-0mm}
\includegraphics[width=\linewidth]{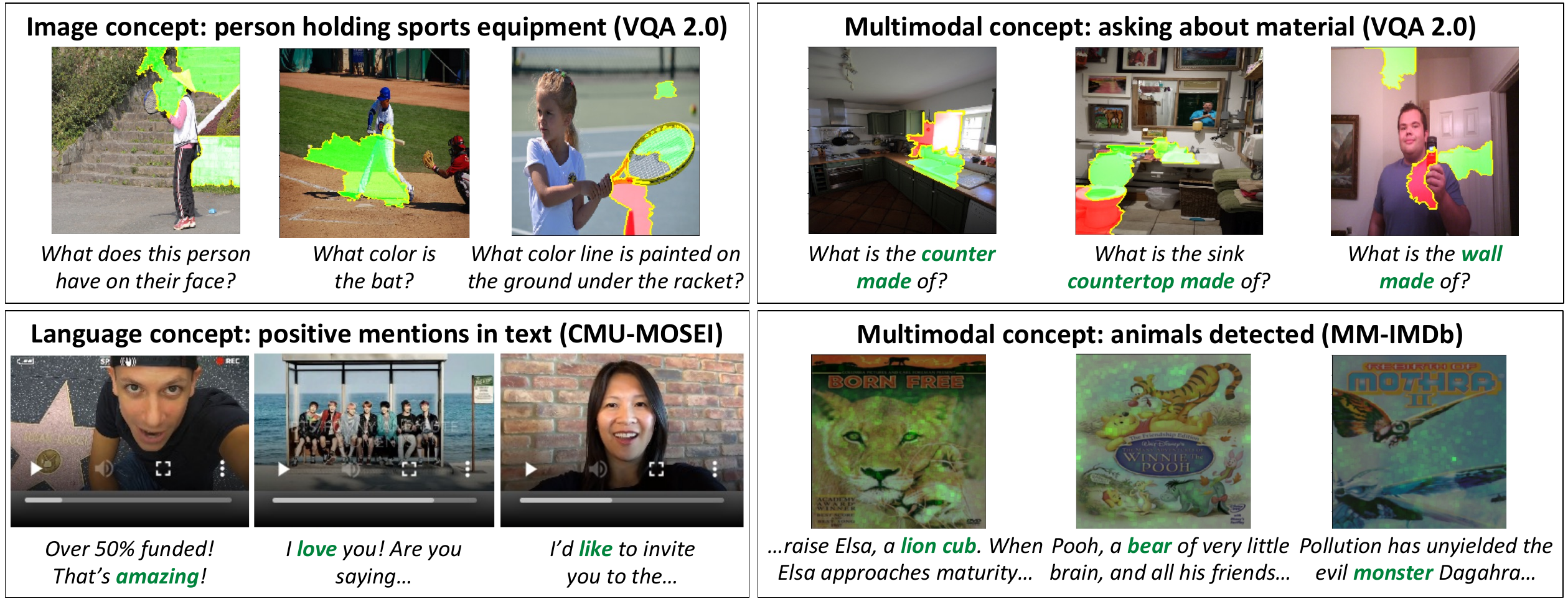}
\vspace{-2mm}
\caption{Examples of human-annotated \textbf{concepts} using \name\ on feature representations. We find that the features separately capture image-only, language-only, and multimodal concepts.}
\label{fig:concepts}
\vspace{-2mm}
\end{figure}

\begin{figure}[t]
\centering
\vspace{-0mm}
\includegraphics[width=\linewidth]{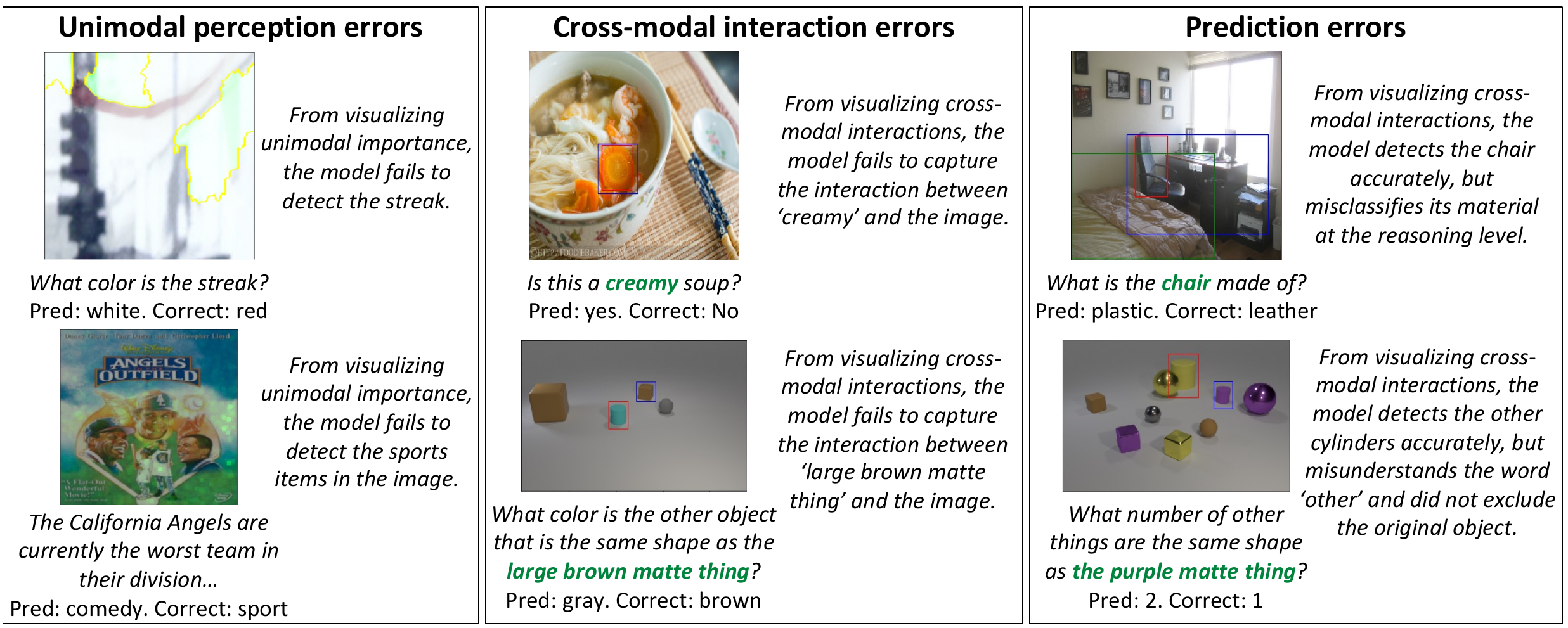}
\vspace{-4mm}
\caption{Examples of human-annotated \textbf{error analysis} using \name\ on multimodal models. Using all stages provided in \name\ enables fine-grained classification of model errors (e.g., errors in unimodal processing, cross-modal interactions, and predictions) for targeted debugging.}
\label{fig:error}
\vspace{-4mm}
\end{figure}

We now take a deeper look to check that \multiviz\ generates accurate explanations of multimodal representations.
Using local and global representation visualizations, can humans consistently assign interpretable concepts in natural language to previously uninterpretable features? We study this question by tasking $15$ human users ($5$ users for each of the following $3$ settings) to assign concepts to each feature $z$ when given access to visualizations of (1) \textbf{$\textrm{R}_\ell$} (local analysis of unimodal and cross-modal interactions in $z$), (2) \textbf{$\textrm{R}_\ell$ + $\textrm{R}_g$ (no viz)} (including global analysis through similar datapoints that also maximally activate feature $z$), and (3) \textbf{$\textrm{R}_\ell$ + $\textrm{R}_g$} (adding highlighted unimodal and cross-modal interactions of global datapoints). Using $20$ datapoints per setting, these experiments with $15$ users involve roughly $10$ total hours of users interacting with \multiviz.

\textbf{Quantitative results}: Since there are no ground-truth labels for feature concepts, we rely on annotator confidence ($1$-$5$ scale) and annotator agreement~\citep{krippendorff2011computing} as a proxy for accuracy. From Table~\ref{tab:local_rep} (left), we find that having access to both local and global visualizations are crucial towards interpreting multimodal features, as measured by higher confidence with low variance in confidence, as well as higher agreement among users.

\textbf{Qualitative interviews}: We show examples of human-assigned concepts in Figure~\ref{fig:concepts}. Note that the $3$ images in each box of Figure~\ref{fig:concepts} (even without feature highlighting) does constitute a visualization generated by \multiviz, as they belong to data instances that maximize the value of the feature neuron (i.e. $\textrm{R}_g$ in stage $3$ multimodal representations). Without \multiviz, it would not be possible to perform feature interpretation without combing through the entire dataset.
Participants also noted that feature visualizations make the decision a lot more confident if its highlights match the concept. 
Taking as example Figure~\ref{fig:concepts} top left, the visualizations serve to highlight what the model's feature neuron is learning (i.e., highlighting the person holding sports equipment), rather than what category of datapoint it is. If the visualization was different, such as highlighting the ground, then users would have to conclude that the feature neuron is capturing `\textit{outdoor ground}' rather than `\textit{sports equipment}'.
Similarly, for text highlights (Figure~\ref{fig:concepts} top right), without using \multiviz\ to highlight `\textit{counter}', `\textit{countertop}', and `\textit{wall}', along with the image crossmodal interactions corresponding to these entities, one would not be able to deduce that the feature asks about material - it could also represent `\textit{what}' questions, or `\textit{household objects}', and so on.
Therefore, these conclusions can only be reliably deduced with all MultiViz stages.

\vspace{-1mm}
\subsection{Error analysis}
\label{error}
\vspace{-1mm}

We examine a case study of error analysis on trained models. We task $10$ human users ($5$ users for each of the following $2$ settings) to use \multiviz\ and highlight the errors that a model exhibits by categorizing these errors into one of $3$ stages: failures in (1) unimodal perception, (2) capturing cross-modal interaction, and (3) prediction with perceived unimodal and cross-modal information.
Again, we rely on annotator confidence ($1$-$5$ scale) and agreement due to lack of ground-truth error categorization, and compare (1) \textbf{\multiviz} with (2) \textbf{No viz}, a baseline that does not provide any model visualizations to the user.
Using $20$ datapoints per setting, these experiments with $10$ users on $2$ datasets and $2$ models involve roughly $15$ total hours of users interacting with \multiviz.
From Table~\ref{tab:local_rep} (right), we find that \multiviz\ enables humans to consistently categorize model errors into one of $3$ stages. We show examples that human annotators classified into unimodal perception, cross-modal interaction, and prediction errors in Figure~\ref{fig:error}.

\vspace{-1mm}
\subsection{A case study in model debugging}
\label{debugging}
\vspace{-1mm}

\begin{figure}[t]
\centering
\vspace{-0mm}
\includegraphics[width=0.8\linewidth]{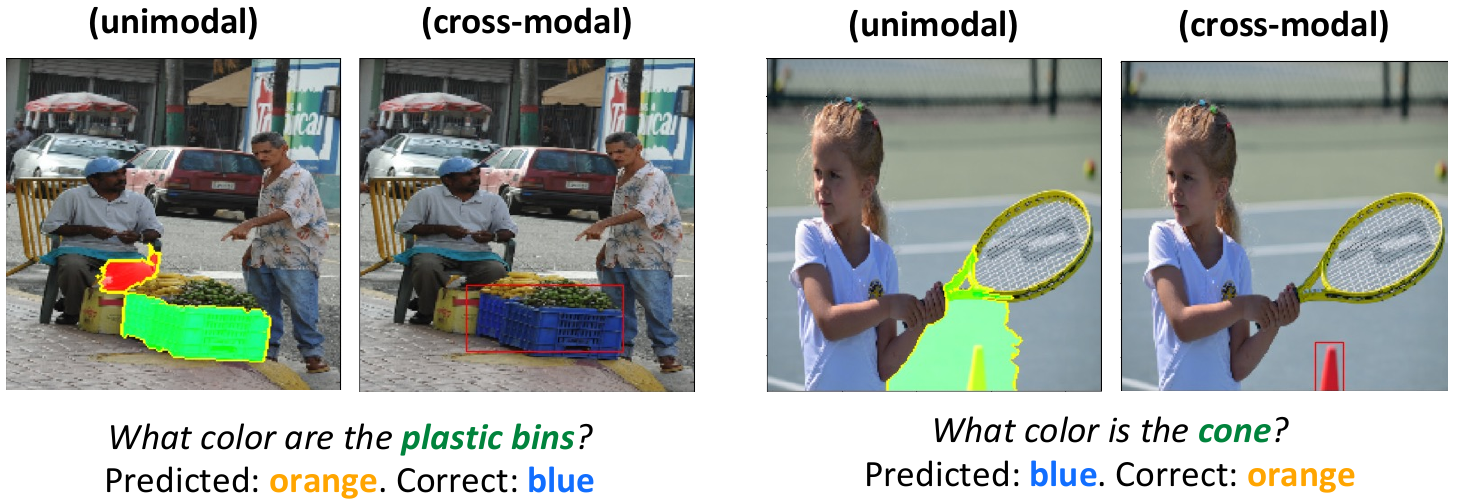}
\caption{A case study on \textbf{model debugging}: we task $3$ human users to use \name\ visualizations and highlight the errors that a pretrained \textsc{LXMERT} model fine-tuned on \textsc{VQA 2.0} exhibits, and find $2$ penultimate-layer neurons highlighting the model's failure to identify color (especially {\color{blue}blue}). Targeted localization of the error to this specific stage (prediction) and representation concept ({\color{blue}blue}) via \name\ enabled us to identify a bug in the popular Hugging Face \textsc{LXMERT} repository.}
\label{fig:debug}
\vspace{-2mm}
\end{figure}

Following error analysis, we take a deeper investigation into one of the errors on a pretrained \textsc{LXMERT} model fine-tuned on \textsc{VQA 2.0}. Specifically, we first found the top $5$ penultimate-layer neurons that are most activated on erroneous datapoints.
Inspecting these neurons carefully through \multiviz\ local and global representation analysis, human annotators found that $2$ of the $5$ neurons were consistently related to questions asking about color, which highlighted the model's failure to identify color correctly (especially {\color{blue}blue}). The model has an accuracy of only $5.5\%$ amongst all {\color{blue}blue}-related points (i.e., either have {\color{blue}blue} as correct answer or predicted answer), and these failures account for $8.8\%$ of all model errors. We show examples of such datapoints and their \multiviz\ visualizations in Figure~\ref{fig:debug}. Observe that the model is often able to capture unimodal and cross-modal interactions perfectly, but fails to identify color at prediction.

Curious as to the source of this error, we looked deeper into the source code for the entire pipeline of \textsc{LXMERT}, including that of its image encoder, Faster R-CNN~\citep{ren2015faster}\footnote{we used the popular Hugging Face implementation at \url{https://huggingface.co/unc-nlp/lxmert-vqa-uncased}}. We in fact uncovered a bug in data preprocessing for Faster R-CNN in the popular Hugging Face repository that swapped the image data storage format from RGB to BGR formats responsible for these errors. This presents a concrete use case of \multiviz: through visualizing each stage, we were able to (1) isolate the source of the bug (at prediction and not unimodal perception or cross-modal interactions), and (2) use representation analysis to localize the bug to the specific color concept.
In our full paper~\cite{liang2023multiviz}, we further detail our initial attempt at tackling this error by using \multiviz\ analysis to select additional targeted datapoints in an active learning scenario, which proved to be much more effective (higher improvement with fewer data) as compared to baselines that add data randomly or via uncertainty sampling~\citep{lewis1994heterogeneous}, which may be of independent interest.

\vspace{-1mm}
\subsection{Additional experiments and takeaways messages}
\label{sanity}
\vspace{-1mm}

\textbf{New models}: We included results on \textsc{ViLT}~\citep{kim2021vilt}, \textsc{CLIP}~\citep{radford2021learning}, and \textsc{MDETR}~\citep{kamath2021mdetr} in the full paper~\cite{liang2023multiviz}, showing that \multiviz\ is a general approach that can be quickly applied to new models. We also study the correlation between performance and cross-modal interactions across several older and recent models, and find that the ability to capture cross-modal alignment, as judged by \multiviz, correlates strongly with final task performance.

\textbf{Sanity checks}: In our full paper~\cite{liang2023multiviz}, we show that \multiviz\ passes the data randomization and model randomization sanity checks for interpretation approaches~\citep{gradientsanitycheck}.

\textbf{Intermediate-layer features}: Finally, we show that \multiviz\ can be extended to visualize any intermediate layer, not just the final layer of multimodal models. We showcase a few examples of $\textbf{R}_\ell$ and $\textbf{R}_g$ on intermediate-layer neurons and discuss several tradeoffs: while they reveal new visualization opportunities, they run the risk of overwhelming the user with the number of images they have to see multiplied by $d^L$ ($d$: dimension of each layer, $L$: number of layers).

\vspace{-1mm}
\section{Related Work}
\vspace{-1mm}

Interpretable ML aims to further our understanding and trust of ML models, enable model debugging, and use these insights for joint decision-making between stakeholders and AI~\citep{chen2022interpretable,gilpin2018explaining}. Interpretable ML is a critical area of research straddling machine learning~\citep{gradientsanitycheck}, language~\citep{tenney2020language}, vision~\citep{simonyan2013deep}, and HCI~\citep{chuang2012interpretation}.
We categorize related work in interpreting multimodal models into:

\textbf{Unimodal importance}: Several approaches have focused on building interpretable components for unimodal importance through soft~\citep{park2018multimodal} and hard attention mechanisms~\citep{chen2017multimodal}. When aiming to explain black-box multimodal models, related work rely primarily on gradient-based visualizations~\citep{simonyan2013deep,baehrens2010explain,erhan2009visualizing} and feature attributions (e.g., LIME~\citep{lime}, Shapley values~\citep{merrick2020explanation}) to highlight regions of the image which the model attends to.

\textbf{Cross-modal interactions}: Recent work investigates the activation patterns of pretrained transformers~\citep{cao2020behind,li2020does}, performs diagnostic experiments through specially curated inputs~\citep{frank2021vision,krojer2022image,parcalabescu2021seeing,thrush2022winoground}, or trains auxiliary explanation modules~\citep{kanehira2019multimodal,park2018multimodal}. Particularly related to our work is EMAP~\citep{hessel2020does} for disentangling the effects of unimodal (additive) contributions from cross-modal interactions in multimodal tasks, as well as M2Lens~\citep{wang2021m2lens}, an interactive visual analytics system to visualize multimodal models for sentiment analysis through both unimodal and cross-modal contributions.

\textbf{Multimodal representation and prediction}: Existing approaches have used language syntax (e.g., the question in VQA) for compositionality into higher-level features~\citep{amizadeh2020neuro,andreas2016neural,vedantam2019probabilistic}. Similarly, logical statements have been integrated with neural networks for interpretable logical reasoning~\citep{gokhale2020vqa,suzuki2019multimodal}. However, these are typically restricted to certain modalities or tasks. Finally, visualizations have also uncovered several biases in models and datasets (e.g., unimodal biases in VQA questions~\citep{anand2018blindfold,cadene2019rubi} or gender biases in image captioning~\citep{hendricks2018women}). We believe that \multiviz\ will enable the identification of biases across a wider range of modalities and tasks.

\vspace{-1mm}
\section{Conclusion}
\vspace{-1mm}

This paper proposes \multiviz\ for analyzing and visualizing multimodal models. \multiviz\ scaffolds the interpretation problem into unimodal importance, cross-modal interactions, multimodal representations, and multimodal prediction, before providing existing and newly proposed analysis tools in each stage. \multiviz\ is designed to be \textit{modular} (encompassing existing analysis tools and encouraging research towards understudied stages), \textit{general} (supporting diverse modalities, models, and tasks), and \textit{human-in-the-loop} (providing a visualization tool for human model interpretation, error analysis, and debugging), qualities which we strive to upkeep by ensuring its public access and regular updates from community feedback.

\chapter{Estimating Multimodal Performance and Modality Selection}
\label{chap:foundations4}
\newcommand{\lowa}{\underline{S}_\textrm{R}}
\newcommand{\lowb}{\underline{S}_\textrm{U}}
\newcommand{\high}{\overline{S}}

\vspace{-3mm}
\section{Introduction}
\vspace{-1mm}

To conclude the first part of this thesis, we provide a guideline for researchers to decide which modalities to collect that will lead to improved multimodal performance~\cite{liang2023multimodal}. Specifically, we study how to quantify interactions in a semi-supervised setting where there is only \textit{unlabeled multimodal data} $\mathcal{D}_M = \{(x_1,x_2)\}$ and some \textit{labeled unimodal data} $\mathcal{D}_i = \{(x_i,y)\}$ collected separately for each modality. This multimodal semi-supervised paradigm is reminiscent of many real-world settings with separate unimodal datasets like visual recognition~\cite{deng2009imagenet} and text classification~\cite{wang2018glue}, as well as naturally co-occurring multimodal data (e.g., news images and captions or video and audio), but when labeling them is time-consuming~\cite{hsu2018unsupervised,hu2019deep} or impossible due to partially observed modalities~\cite{liang2022highmmt} or privacy concerns~\cite{che2023multimodal}. We want to understand how the modalities can share, exchange, and create information to inform practitioners whether it is worth collecting multimodal data and trying multimodal models~\cite{jayakumar2020multiplicative,liang2023quantifying,zadeh2017tensor}.

Using a precise information-theoretic definition of interactions~\cite{bertschinger2014quantifying}, our key contributions are the derivations of lower and upper bounds to quantify multimodal interactions in this semi-supervised setting with only $\mathcal{D}_i$ and $\mathcal{D}_M$. We propose two lower bounds: the first relates interactions with the amount of \textit{shared information} between modalities, and the second is based on the \textit{disagreement} of classifiers trained separately on each modality.
Finally, we propose an upper bound through connections to approximate algorithms for \textit{min-entropy couplings}~\cite{cicalese2002supermodularity}.
To validate our bounds, we experiment on both synthetic and large real-world datasets with varying amounts of interactions.
In addition, these theoretical results naturally yield new guarantees regarding the performance of multimodal models. By analyzing the relationship between interaction estimates and downstream task performance assuming optimal multimodal classifiers are trained on labeled multimodal data, we can \textit{closely predict multimodal model performance, before even training the model itself}. These performance estimates also help develop new guidelines for deciding when to \textit{collect additional modality data} and \textit{select the appropriate multimodal fusion models}.
We believe these results shed light on the intriguing connections between multimodal interactions, modality disagreement, and model performance, and release our code and models at \url{https://github.com/pliang279/PID}.

\vspace{-1mm}
\section{Related Work and Technical Background}

\subsection{Semi-supervised multimodal learning}

Let $\mathcal{X}_i$ and $\mathcal{Y}$ be finite sample spaces for features and labels.
Define $\Delta$ to be the set of joint distributions over $(\mathcal{X}_1, \mathcal{X}_2, \mathcal{Y})$.
We are concerned with features $X_1, X_2$ (with support $\mathcal{X}_i$) and labels $Y$ (with support $\mathcal{Y}$) drawn from some distribution $p \in \Delta$. We denote the probability mass function by $p(x_1,x_2,y)$, where omitted parameters imply marginalization. 
Many real-world applications such as multimedia and healthcare naturally exhibit multimodal data (e.g., images and captions, video and audio, multimodal medical readings) which are difficult to label~\citep{liang2022highmmt,radford2021learning,singh2022flava,yu2004efficient,zellers2022merlot}. As such, rather than the full distribution from $p$, we only have partial datasets:
\begin{itemize}[noitemsep,topsep=0pt,nosep,leftmargin=*,parsep=0pt,partopsep=0pt]
    \item \textit{Labeled unimodal} data $\mathcal{D}_1 = \{(x_1,y): \mathcal{X}_1 \times \mathcal{Y}\}$, $\mathcal{D}_2 = \{(x_2,y): \mathcal{X}_2 \times \mathcal{Y}\}$.
    \item \textit{Unlabeled multimodal} data $\mathcal{D}_M = \{(x_1,x_2): \mathcal{X}_1 \times \mathcal{X}_2\}$.
\end{itemize}
$\mathcal{D}_1$, $\mathcal{D}_2$ and $\mathcal{D}_M$ follow the \textit{pairwise marginals} $p(x_1, y)$, $p(x_2, y)$ and $p(x_1, x_2)$. 
We define $\Delta_{p_{1,2}} = \{ q \in \Delta: q(x_i,y)=p(x_i,y) \ \forall y\in\mathcal{Y}, x_i \in \mathcal{X}_i, i \in [2] \}$ as the set of joint distributions which agree with the labeled unimodal data $\mathcal{D}_1$ and $\mathcal{D}_2$, and $\Delta_{p_{1,2,12}} = \{ r \in \Delta: r(x_1,x_2)=p(x_1,x_2), r(x_i,y)=p(x_i,y) \}$ as the set of joint distributions which agree with all $\mathcal{D}_1, \mathcal{D}_2$ and $\mathcal{D}_M$.

\subsection{Multimodal interactions and information theory}

The study of \textbf{multimodal interactions} aims to quantify the information shared between both modalities, in each modality alone, and how modalities can combine to form new information not present in either modality, eventually using these insights to design machine learning models to capture interactions from large-scale multimodal datasets~\citep{liang2022foundations}. Existing literature has primarily studied the interactions captured by trained models, such as using Shapley values~\citep{ittner2021feature} and Integrated gradients~\citep{sundararajan2017axiomatic,tsang2018detecting,liang2023multiviz} to measure the importance a model assigns to each modality, or approximating trained models with additive or non-additive functions to determine what functions are best suited to capture interactions~\citep{friedman2008predictive,sorokina2008detecting,hessel2020does}.
However, these measure interactions captured by a trained model - \textit{our work is fundamentally different in that interactions are properties of data}. Quantifying the interactions in data, independent of trained models, allows us to characterize datasets, predict model performance, and perform model selection, prior to choosing and training a model altogether. Prior work in understanding data interactions to design multimodal models is often driven by intuition, such as using contrastive learning~\citep{poklukar2022geometric,radford2021learning,tosh2021contrastive}, correlation analysis~\citep{andrew2013deep}, and agreement~\citep{ding2022cooperative} for shared information (e.g., images and descriptive captions), or using tensors and multiplicative interactions~\citep{zadeh2017tensor,jayakumar2020multiplicative} for higher-order interactions (e.g., in expressions of sarcasm from speech and gestures).

To fill the gap in data quantification, \textbf{information theory} has emerged as a theoretical foundation since it naturally formalizes information and its sharing as statistical properties of data distributions. 
Information theory studies the information that one random variable ($X_1$) provides about another ($X_2$), as quantified by Shannon's mutual information (MI) and conditional MI:
\begin{align*}
    I(X_1; X_2) = \int p(x_1,x_2) \log \frac{p(x_1,x_2)}{p(x_1) p(x_2)} d\bm{x}, \quad I(X_1;X_2|Y) = \int p(x_1,x_2,y) \log \frac{p(x_1,x_2|y)}{p(x_1|y) p(x_2|y)} d\bm{x} dy.
\end{align*}
$I(X_1; X_2)$ measures the amount of information (in bits) obtained about $X_1$ by observing $X_2$, and by extension, $I(X_1;X_2|Y)$ is the expected value of MI given the value of a third (e.g., task $Y$).

To generalize information theory for multimodal interactions, Partial information decomposition (PID)~\citep{williams2010nonnegative} decomposes the total information that two modalities $X_1,X_2$ provide about a task $Y$ into 4 quantities: $I_p(\{X_1,X_2\}; Y) = R + U_1 + U_2 + S$, where $I_p(\{X_1,X_2\}; Y)$ is the MI between the joint random variable $(X_1,X_2)$ and $Y$. These 4 quantities are: redundancy $R$ for the task-relevant information shared between $X_1$ and $X_2$, uniqueness $U_1$ and $U_2$ for the information present in only $X_1$ or $X_2$ respectively, and synergy $S$ for the emergence of new information only when both $X_1$ and $X_2$ are present~\citep{bertschinger2014quantifying,griffith2014quantifying}:
\begin{definition}
\label{def:pid}
    (Multimodal interactions) Given $X_1$, $X_2$, and a target $Y$, we define their redundant ($R$), unique ($U_1$ and $U_2$), and synergistic ($S$) interactions as:
    \begin{align}
        R &= \max_{q \in \Delta_{p_{1,2}}} I_q(X_1; X_2; Y), \quad U_1 = \min_{q \in \Delta_{p_{1,2}}} I_q(X_1; Y | X_2), \quad U_2 = \min_{q \in \Delta_{p_{1,2}}} I_q(X_2; Y| X_1), \label{eqn:U2-def}\\
        S &= I_p(\{X_1,X_2\}; Y) - \min_{q \in \Delta_{p_{1,2}}} I_q(\{X_1,X_2\}; Y), \label{eqn:S-def}
    \end{align}
    where the notation $I_p(\cdot)$ and $I_q(\cdot)$ disambiguates mutual information (MI) under $p$ and $q$ respectively.
\end{definition}
$I(X_1; X_2; Y) = I(X_1; X_2) - I(X_1;X_2|Y)$ is a multivariate extension of information theory~\citep{bell2003co,mcgill1954multivariate}. Most importantly, $R$, $U_1$, and $U_2$ can be computed exactly using convex programming over distributions $q \in \Delta_{p_{1,2}}$ with access only to the marginals $p(x_1,y)$ and $p(x_2,y)$ by solving a convex optimization problem with linear marginal-matching constraints $q^* = \argmax_{q \in \Delta_{p_{1,2}}} H_q(Y | X_1, X_2)$~\citep{bertschinger2014quantifying,liang2023quantifying}. This gives us an elegant interpretation that we need only labeled unimodal data in each feature from $\mathcal{D}_1$ and $\mathcal{D}_2$ to estimate redundant and unique interactions. Unfortunately, $S$ is impossible to compute via equation (\ref{eqn:S-def}) when we do not have access to the full joint distribution $p$, since the first term $I_p(\{X_1, X_2\};Y)$ is unknown.

It is worth noting that other valid information-theoretic definitions of multimodal interactions also exist, but are known to suffer from issues regarding over- and under-estimation, and may even be negative; these are critical problems with the application of information theory for shared $I(X_1; X_2; Y)$ and unique information $I(X_1; Y|X_2)$, $I(X_2; Y|X_1)$ often quoted in the co-training~\citep{blum1998combining,balcan2004co} and multi-view learning~\citep{tosh2021contrastive,tsai2020self,tian2020makes,sridharan2008information} literature.
We refer the reader to~\citet{griffith2014quantifying} for a full discussion. We choose the one in Definition~\ref{def:pid} above since it fulfills several desirable properties, but our results can be extended to other definitions as well.

\vspace{-1mm}
\section{Estimating Semi-supervised Multimodal Interactions}

Our goal is to estimate multimodal interactions $R$, $U_1$, $U_2$, and $S$ assuming access to only semi-supervised multimodal data $\mathcal{D}_1$, $\mathcal{D}_2$, and $\mathcal{D}_M$. Our first insight is that while $S$ cannot be computed exactly, $R$, $U_1$, and $U_2$ can be computed from equation~\ref{eqn:U2-def} with access to only semi-supervised data. Therefore, studying the relationships between $S$ and other multimodal interactions is key to its estimation. Using these relationships, we will then derive lower and upper bounds for synergy in the form $\underline{S} \leq S \leq \high$. Crucially, $\underline{S}$ and $\high$ depend \textit{only} on $\mathcal{D}_1$, $\mathcal{D}_2$, and $\mathcal{D}_M$.

\subsection{Understanding relationships between interactions}

We start by identifying two important relationships, between $S$ and $R$, and between $S$ and $U$.

\begin{figure*}[t]
\vspace{-0mm}
\centering
\includegraphics[width=\linewidth]{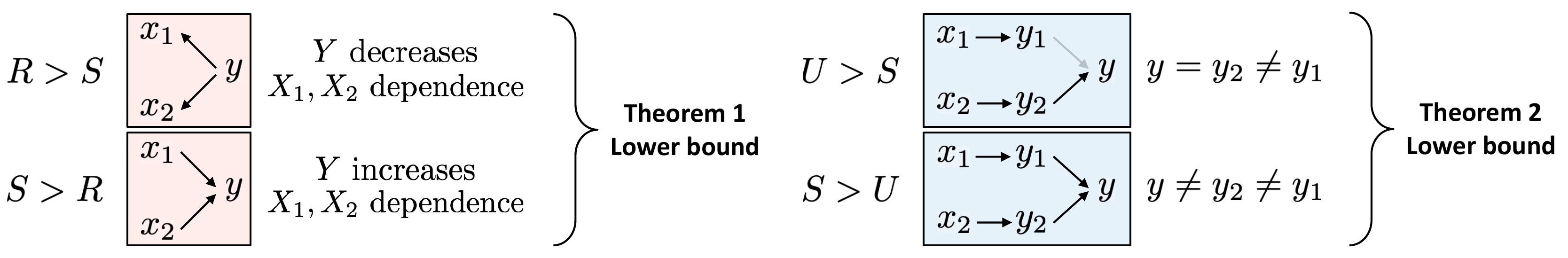}
\caption{We study the relationships between (left) \textit{synergy and redundancy} as a result of the task $Y$ either increasing or decreasing the shared information between $X_1$ and $X_2$ (i.e., common cause structures as opposed to redundancy in common effect), as well as (right) \textit{synergy and uniqueness} due to the disagreement between unimodal predictors resulting in a new prediction $y \neq y_1 \neq y_2$ (rather than uniqueness where $y = y_2 \neq y_1$).}
\label{fig:splits}
\vspace{-2mm}
\end{figure*}

\paragraph{Synergy and redundancy} Our first relationship stems from the case when two modalities contain shared information about the task. In studying these situations, a driving force for estimating $S$ is the amount of shared information $I(X_1;X_2)$ between modalities, with the intuition that more shared information naturally leads to redundancy which gives less opportunity for new synergistic interactions.
Mathematically, we formalize this by relating $S$ to $R$,
\begin{align}
\label{eq:SandR}
    S = R - I_p(X_1;X_2;Y) = R - I_p(X_1;X_2) + I_p(X_1;X_2|Y).
\end{align}
implying that synergy exists when there is high redundancy and low (or even negative) three-way MI $I_p(X_1;X_2;Y)$.
By comparing the difference in $X_1,X_2$ dependence with and without the task (i.e., $I_p(X_1;X_2)$ vs $I_p(X_1;X_2|Y)$), $2$ cases naturally emerge (see left side of Figure~\ref{fig:splits}):
\begin{enumerate}[noitemsep,topsep=0pt,nosep,leftmargin=*,parsep=0pt,partopsep=0pt]
    \item $\mathbf{S>R}$: When both modalities do not share a lot of information as measured by low $I(X_1;X_2)$, but conditioning on $Y$ \textit{increases} their dependence: $I(X_1;X_2|Y) > I(X_1;X_2)$, then there is synergy between modalities when combining them for task $Y$. This setting is reminiscent of common cause structures. Examples of these distributions in the real world are multimodal question answering, where the image and question are less dependent (some questions like `what is the color of the car' or `how many people are there' can be asked for many images), but the answer (e.g., `blue car') connects the two modalities, resulting in dependence given the label. As expected, $S = 4.92,R=0.79$ for the VQA 2.0 dataset~\citep{goyal2017making}.
    \item $\mathbf{R>S}$: Both modalities share a lot of information but conditioning on $Y$ \textit{reduces} their dependence: $I(X_1;X_2)>I(X_1;X_2|Y)$, which results in more redundant than synergistic information. This setting is reminiscent of common effect structures.
    A real-world example is in detecting sentiment from multimodal videos, where text and video are highly dependent since they are emitted by the same speaker, but the sentiment label explains away some of the dependencies between both modalities. Indeed, for multimodal sentiment analysis from text, video, and audio of monologue videos on \textsc{MOSEI}~\citep{zadeh2018multimodal}, $R=0.26$ and $S=0.04$.
\end{enumerate}

\paragraph{Synergy and uniqueness} The second relationship arises when two modalities contain disagreeing information about the task, and synergy arises due to this disagreement in information. To illustrate this, suppose $y_1=\argmax_y p(y|x_1)$ is the most likely prediction from the first modality, $y_2=\argmax_y p(y|x_2)$ for the second modality, and $y=\argmax_y p(y|x_1,x_2)$ is the true multimodal prediction. There are again $2$ cases (see right side of Figure~\ref{fig:splits}):
\begin{enumerate}[noitemsep,topsep=0pt,nosep,leftmargin=*,parsep=0pt,partopsep=0pt]
    \item $\mathbf{U>S}$: Multimodal prediction $y=\argmax_y p(y|x_1,x_2)$ is the same as one of the unimodal predictions (e.g., $y=y_2$), in which case unique information in modality $2$ leads to the outcome and there is no synergy.
    A real-world dataset is \textsc{MIMIC} involving mortality and disease prediction from tabular patient data and time-series medical sensors~\citep{johnson2016mimic} which primarily shows unique information in the tabular modality. The disagreement on \textsc{MIMIC} is high at $0.13$, but since disagreement is due to a lot of unique information, there is less synergy $S=0.01$.
    
    \item $\mathbf{S>U}$: Multimodal prediction $y$ is different from both $y_1$ and $y_2$, then both modalities interact synergistically to give rise to a final outcome different from both disagreeing unimodal predictions.
    This type of joint distribution is indicative of real-world expressions of sarcasm from language and speech - the presence of sarcasm is typically detected due to a contradiction between what is expressed in language and speech, as we observe from the experiments on \textsc{MUStARD}~\citep{castro2019towards} where $S=0.44$ and disagreement $=0.12$ are both large.
\end{enumerate}

\subsection{Lower and upper bounds on synergy}

Given these relationships between synergy and other interactions, we now derive bounds on $S$. We present two lower bounds $\lowa$ and $\lowb$, which are based on redundancy and uniqueness, as well as an upper bound $\high$. We also describe the computational complexity for computing each bound.

\textit{Remark on high dimensional, continuous modalities.} Our theoretical results are concerned with \textit{finite} spaces for features and labels. However, this may be restrictive when working with real-world datasets (e.g., images, video, text) which are often continuous and/or high-dimensional. In such situations, we preprocess by performing discretization of each modality via clustering to estimate $p(x_1,y)$, $p(x_2,y)$, $p(x_1,x_2)$, each with a small, finite support. These are subsequently used for the computation of $\lowa$, $\lowb$ and $\high$.
Discretization is a common way to approximate information theoretic quantities like mutual information~\citep{darbellay1999estimation,liang2023quantifying} and for learning representations over high-dimensional modalities~\citep{oord2018representation}. 

\paragraph{Lower bound using redundancy} Our first lower bound uses the relationship between synergy, redundancy, and dependence in equation~\ref{eq:SandR}. In semi-supervised settings, we can compute $R$ exactly from $p(x_1,y),p(x_2,y)$, as well as the shared information $I(X_1;X_2)$ from $p(x_1,x_2)$. However, $I_p(X_1;X_2|Y)$ cannot be computed without access to the full distribution $p$. In Theorem~\ref{thm:lower_connections}, we obtain a lower bound on $I_p(X_1;X_2|Y)$, resulting in a lower bound $\lowa$ for synergy.
\begin{theorem}
\label{thm:lower_connections}
    (Lower-bound on synergy via redundancy) We relate $S$ to modality dependence
    \begin{align}
    \label{eqn:lower_connections}
        \lowa = R - I_p(X_1;X_2) + \min_{r \in \Delta_{p_{1,2,12}}} I_r(X_1;X_2|Y) \le S
    \end{align}
\end{theorem}
We include a proof in the full paper~\cite{liang2023multimodal}. This bound compares $S$ to $R$ via the difference of their dependence $I_p(X_1;X_2)$ and their dependence given the task $I_p(X_1;X_2|Y)$. Since the full distribution $p$ is not available to compute $I_p(X_1;X_2|Y)$, we prove a lower bound using conditional MI computed with respect to a set of auxiliary distributions $r \in \Delta_{p_{1,2,12}}$ that are close to $p$, as measured by matching both unimodal marginals $r(x_i,y) = p(x_i,y)$ and modality marginals $r(x_1,x_2) = p(x_1,x_2)$. If conditioning on the task increases the dependence and $I_r(X_1;X_2|Y)$ is large relative to $I_p(X_1;X_2)$ then we obtain a larger value of $\lowa$, otherwise if conditioning on the task decreases the dependence and $I_r(X_1;X_2|Y)$ is small relative to $I_p(X_1;X_2)$ then we obtain a smaller value of $\lowa$.

\textit{Computational complexity.} $R$ and $\min_{r \in \Delta_{p_{1,2,12}}} I_r(X_1;X_2|Y)$ are convex optimization problems solvable in polynomial time with off-the-shelf solvers. $I_p(X_1;X_2)$ can be computed directly.

\paragraph{Lower bound using uniqueness} Our second bound formalizes the relationship between disagreement, uniqueness, and synergy. The key insight is that while labeled multimodal data is unavailable, the output of unimodal classifiers may be compared against each other.
Consider unimodal classifiers $f_i: \mathcal{X}_i \rightarrow \mathcal{Y}$ and multimodal classifiers $f_M: \mathcal{X}_1 \times \mathcal{X}_2 \rightarrow \mathcal{Y}$.
Define \textit{modality disagreement} as:
\begin{definition}
    (Modality disagreement) Given $X_1$, $X_2$, and a target $Y$, as well as unimodal classifiers $f_1$ and $f_2$, we define modality disagreement as $\alpha(f_1,f_2) = \mathbb{E}_{p(x_1,x_2)} [d(f_1,f_2)]$ where $d: \mathcal{Y} \times \mathcal{Y} \rightarrow \mathbb{R}^{\ge0}$ is a distance function in label space scoring the disagreement of $f_1$ and $f_2$'s predictions.
\end{definition}
Connecting \textit{modality disagreement} and synergy via Theorem~\ref{thm:lower_disagreement} yields a lower bound $\lowb$:
\begin{theorem}
\label{thm:lower_disagreement}
    (Lower-bound on synergy via uniqueness, informal) We can relate synergy $S$ and uniqueness $U$ to modality disagreement $\alpha(f_1,f_2)$ of optimal unimodal classifiers $f_1,f_2$ as follows:
    \begin{align}
    \label{eqn:lower_disagreement}
        \lowb = \alpha(f_1,f_2) \cdot c - \max(U_1,U_2) \le S
    \end{align}
    for some constant $c$ depending on the label dimension $|\mathcal{Y}|$ and choice of label distance function $d$.
\end{theorem}
Theorem~\ref{thm:lower_disagreement} implies that if there is substantial disagreement $\alpha(f_1,f_2)$ between unimodal classifiers, it must be due to the presence of unique or synergistic information. If uniqueness is small, then disagreement must be accounted for by synergy, thereby yielding a lower bound $\lowb$. Note that the optimality of unimodal classifiers is important: poorly trained unimodal classifiers could show high disagreement but would be uninformative about true interactions. We include the formal version of the theorem based on Bayes' optimality and a proof in the full paper~\cite{liang2023multimodal}.

\textit{Computational complexity.} Lower bound $\lowb$ can also be computed efficiently by estimating ${p}(y|x_1)$ and ${p}(y|x_2)$ over modality clusters or training unimodal classifiers ${f}_\theta(y|x_1)$ and ${f}_\theta(y|x_2)$. $U_1$ and $U_2$ can be computed using a convex solver in polynomial time.

Hence, the relationships between $S$, $R$, and $U$ yield two lower bounds $\lowa$ and $\lowb$. Note that these bounds \textit{always} hold, so we could take $\underline{S}=\max\{\lowa, \lowb \}$.

\paragraph{Upper bound on synergy} By definition, $S = I_p(\{X_1,X_2\}; Y) - R - U_1 - U_2$. However, $I_p(\{X_1,X_2\};Y)$ cannot be computed exactly without the full distribution $p$. Using the same idea as lower bound 1, we upper bound synergy by \textit{considering the worst-case maximum} $I_r(\{X_1,X_2\};Y)$ computed over a set of auxiliary distributions $r \in \Delta_{p_{1,2,12}}$ that match both unimodal marginals $r(x_i,y) = p(x_i,y)$ and modality marginals $r(x_1,x_2) = p(x_1,x_2)$:
\begin{align}
    \max_{r \in \Delta_{p_{1,2,12}}} I_r(\{X_1,X_2\}; Y) &= 
    \max_{r \in \Delta_{p_{1,2,12}}} \left\{ H_r(X_1, X_2) + H_r(Y) - H_r(X_1, X_2, Y) \right\} \\ &=
    H_p(X_1, X_2) + H_p(Y) - \min_{r \in \Delta_{p_{1,2,12}}} H_r(X_1, X_2, Y),
\end{align}
where the second line follows from the definition of $\Delta_{p_{1,2,12}}$. While the first two terms are easy to compute, the third may be difficult, as shown in the following theorem:
\begin{theorem}
\label{thm:min-entropy-np-hard}
Solving $r^* = \argmin_{r \in \Delta_{p_{1,2,12}}} H_r(X_1, X_2, Y)$ is NP-hard, even for a fixed $|\mathcal{Y}| \geq 4$. 
\end{theorem}
Theorem~\ref{thm:min-entropy-np-hard} suggests we cannot tractably find a joint distribution which tightly upper bounds synergy when the feature spaces are large. Fortunately, a relaxation of $r \in \Delta_{p_{1,2,12}}$ to $r \in \Delta_{p_{12,y}}$, where $r(x_1,x_2)=p(x_1,x_2)$ and $r(y)=p(y)$, recovers the classic \textit{min-entropy coupling} problem over $(X_1, X_2)$ and $Y$, which is still NP-hard but admits good approximations~\citep{cicalese2002supermodularity,cicalese2017find,kocaoglu2017entropic,compton2023minimum}. 
Our final upper bound $\high$ is:
\begin{theorem}
\label{thm:upper}
    (Upper-bound on synergy)
    \begin{align}
        S \leq H_p(X_1, X_2) + H_p(Y) - \min_{r \in \Delta_{p_{12,y}}} H_r(X_1, X_2, Y) - R - U_1 - U_2 = \high 
    \end{align}
\end{theorem}
Proofs of Theorem~\ref{thm:min-entropy-np-hard}, \ref{thm:upper}, and detailed approximation algorithms for min-entropy couplings are included in the full paper~\cite{liang2023multimodal}.

\textit{Computational complexity.} The upper bound $\high$ can be computed efficiently since solving the variant of the min-entropy problem in Theorem~\ref{thm:upper} admits approximations that can be computed in time $O(k \log k)$ where $k=\max( |\mathcal{X}_1|, |\mathcal{X}_2| )$. All other entropy and $R, U_1, U_2$ terms are easy to compute (or have been computed via convex optimization from the lower bounds).

Practically, calculating all three bounds is extremely simple, with just a few lines of code. The computation takes < 1 minute and < 180 MB memory space on average for our large datasets ($1,000$-$20,000$ datapoints), more efficient than training even the smallest multimodal prediction model which takes at least $3$x time and $15$x memory. As a result, \textit{these bounds scale to large and high-dimensional multimodal datasets found in the real world}, which we verify in the following experiments.

\section{Experiments}

We design comprehensive experiments to validate these estimated bounds and relationships between different multimodal interactions. Using these results, we describe applications in estimating optimal multimodal performance before training the model itself, which can be used to guide data collection and select appropriate multimodal models for various tasks.

\subsection{Verifying interaction estimation in semi-supervised learning}

\paragraph{Synthetic bitwise datasets} Let $\mathcal{X}_1=\mathcal{X}_2=\mathcal{Y}=\{0,1\}$. We generate joint distributions $\Delta$ by sampling $100,000$ vectors from the 8-dim probability simplex and assigning them to $p(x_1,x_2,y)$.

\begin{wrapfigure}{R}{0.3\textwidth}
    \centering
    \vspace{-2mm}
    \includegraphics[width=\linewidth]{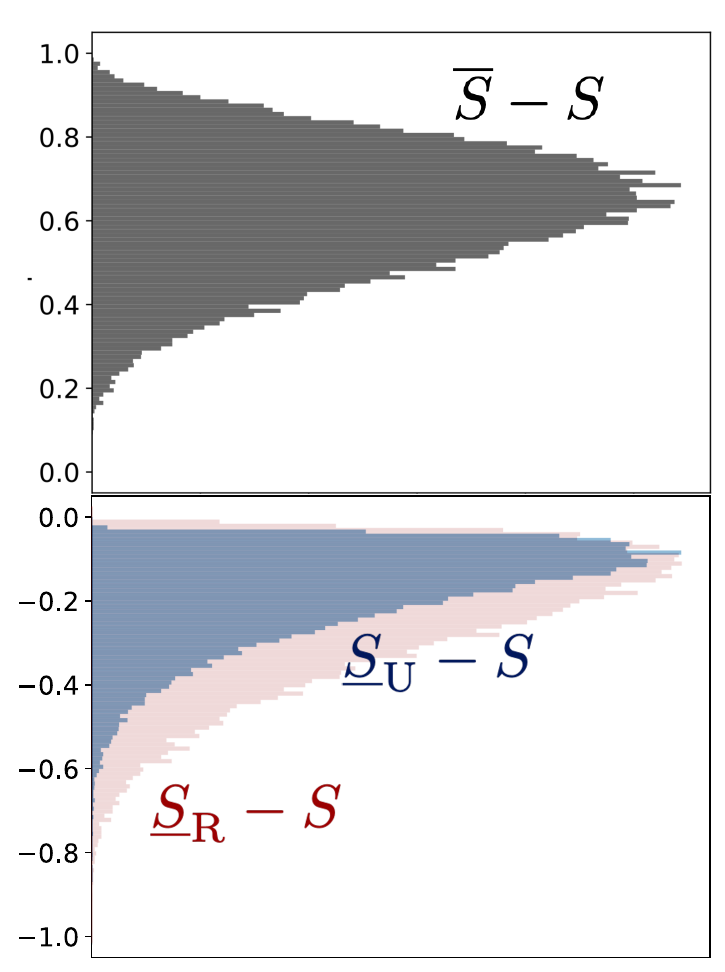}
    \vspace{-4mm}
    \caption{Our two lower bounds {\color{red}$\lowa$} and {\color{blue}$\lowb$} track actual synergy $S$ from below, and the upper bound $\high$ tracks $S$ from above. We find that $\lowa,\lowb$ tend to approximate $S$ better than $\high$.}
    \label{fig:gaps}
\vspace{-8mm}
\end{wrapfigure}

\paragraph{Large real-world multimodal datasets} We use a collection of $10$ real-world datasets from MultiBench~\citep{liang2021multibench} which add up to a size of more than $700,000$ datapoints.
\begin{enumerate}[noitemsep,topsep=0pt,nosep,leftmargin=*,parsep=0pt,partopsep=0pt]
    \item \textsc{MOSI}: $2,199$ videos for sentiment analysis~\citep{zadeh2016mosi},
    
    \item \textsc{MOSEI}: $23,000$ videos for sentiment and emotion analysis~\citep{zadeh2018multimodal},

    \item \textsc{MUStARD}: $690$ videos for sarcasm detection~\citep{castro2019towards}, 

    \item \textsc{UR-FUNNY}: a dataset of humor detection from $16,000$ TED talk videos~\citep{hasan2019ur},

    \item \textsc{MIMIC}: $36,212$ examples predicting patient mortality and diseases from tabular patient data and medical sensors~\citep{johnson2016mimic},

    \item \textsc{ENRICO}: $1,460$ examples classifying mobile user interfaces and screenshots~\citep{leiva2020enrico}.

    \item IRFL: $6,697$ images and figurative captions (e.g, ‘the car is as fast as a cheetah’ describing an image with a fast car in it)~\citep{yosef2023irfl}.

    \item NYCaps: $1,820$ New York Yimes cartoon images and humorous captions describing these images~\citep{hessel2022androids}.
    
    \item VQA: $614,000$ questions and answers about natural images~\citep{antol2015vqa}.

    \item ScienceQA: $21,000$ questions and answers about science problems with scientific diagrams~\citep{lu2022learn}.
\end{enumerate}

These high-dimensional and continuous modalities require approximating disagreement and mutual information: we train unimodal classifiers $\hat{f}_\theta(y|x_1)$ and $\hat{f}_\theta(y|x_2)$ to estimate disagreement, and we cluster modality features to approximate continuous modalities by discrete distributions with finite support to compute the lower and upper bounds. We summarize the following regarding the validity of each bound:

\paragraph{1. Overall trends} For the $100,000$ bitwise distributions, we compute $S$, the true value of synergy assuming oracle knowledge of the full multimodal distribution, and compute $\lowa-S$, $\lowb-S$, and $S-\high$ for each point. Plotting these points as a histogram in Figure~\ref{fig:gaps}, we find that the two lower bounds track synergy from below ($\lowa-S$ and $\lowb-S$ approaching $0$ from below), and the upper bound tracks synergy from above ($S-\high$ approaching $0$ from above). The two lower bounds are quite tight, as we see that for many points $\lowa-S$ and $\lowb-S$ are approaching close to $0$, with an average gap of $0.18$. $\lowb$ seems to be tighter empirically than $\lowa$: for half the points, $\lowb$ is within $0.14$ and $\lowa$ is within $0.2$ of $S$. For the upper bound, there is an average gap of $0.62$. However, it performs especially well on high synergy data: when $S > 0.6$, the average gap is $0.24$, with more than half of the points within $0.25$ of $S$.

\begin{table}[t]
\centering
\fontsize{9}{11}\selectfont
\setlength\tabcolsep{1.5pt}
\vspace{-0mm}
\caption{We compute lower bounds $\lowa$, $\lowb$, and upper bound $\high$ in semi-supervised multimodal settings and compare them to $S$ assuming knowledge of the full joint distribution $p$. The bounds always hold and track $S$ well on \textsc{MOSEI}, \textsc{UR-FUNNY}, \textsc{MOSI}, and \textsc{MUStARD}: true $S$ increases as estimated $\lowa$ and $\lowb$ increases.}
\centering
\footnotesize
\begin{tabular}{l|ccccccccccc}
\hline \hline
& \textsc{MOSEI} & \textsc{UR-FUNNY} & \textsc{MOSI} & \textsc{MUStARD} & \textsc{MIMIC} & \textsc{ENRICO} & \textsc{NYCaps} & \textsc{IRFL} & \textsc{VQA} & \textsc{ScienceQA} \\
\hline
$\high$ & $0.97$ & $0.97$ & $0.92$ & $0.79$ & $0.41$ & $2.09$ & $0.68$ & $0.01$ & $0.97$ & $1.67$\\
$S$ & $0.03$ & $0.18$ & $0.24$ & $0.44$ & $0.02$ & $1.02$ & $0.09$ & $0$ & $0.05$ & $0.16$\\
$\lowa$ & $0$ & $0$ & $0.01$ & $0.04$ & $0$ & $0.01$ & $0$ & $0$ & $0$ & $0.01$\\
$\lowb$ & $0.01$ & $0.01$ & $0.03$ & $0.11$ & $-0.12$ & $-0.55$ & $-0.03$ & $-0.01$ & $0$ & $0$\\
\hline \hline
\end{tabular}

\vspace{-2mm}
\label{tab:s}
\end{table}

On real-world MultiBench datasets, we show the estimated bounds and actual $S$ computed assuming knowledge of full $p$ in Table~\ref{tab:s}. The lower and upper bounds track true $S$: as estimated $\lowa$ and $\lowb$ increases from \textsc{MOSEI} to \textsc{UR-FUNNY} to \textsc{MOSI} to \textsc{MUStARD}, true $S$ also increases.
For datasets like \textsc{MIMIC} with disagreement but high uniqueness, $\lowb$ can be negative, but we can rely on $\lowa$ to give a tight estimate on low synergy. Unfortunately, our bounds do not track synergy well on \textsc{ENRICO}. We believe this is because \textsc{ENRICO} displays all interactions: $R=0.73,U_1=0.38,U_2=0.53,S=0.34$, which makes it difficult to distinguish between $R$ and $S$ using $\lowa$ or $U$ and $S$ using $\lowb$ since no interaction dominates over others, and $\high$ is also quite loose. Given these general observations, we now carefully analyze the relationships between redundancy, uniqueness, and synergy.

\begin{table}[t]
\vspace{-0mm}
\caption{Four representative examples: (a) disagreement XOR has high disagreement and high synergy, (b) agreement XOR has no disagreement and high synergy, (c) $y=x_1$ has high disagreement and uniqueness but no synergy, and (d) $y=x_1=x_2$ has high agreement and redundancy but no synergy.}
\begin{subtable}{0.24\textwidth}
\centering
\begin{tabular}{ccc|c}
\hline \hline
$x_1$ & $x_2$ & $y$ & $p$ \\
\hline
$0$ & $0$ & $0$ & $0$ \\
$0$ & $0$ & $1$ & $0.05$ \\
$0$ & $1$ & $0$ & $0.03$ \\
$0$ & $1$ & $1$ & $0.28$ \\
$1$ & $0$ & $0$ & $0.53$ \\
$1$ & $0$ & $1$ & $0.03$ \\
$1$ & $1$ & $0$ & $0.01$ \\
$1$ & $1$ & $1$ & $0.06$ \\
\hline \hline
\end{tabular}
\caption{Disagreement XOR}
\label{tab:dis_xor}
\end{subtable}
\begin{subtable}{0.24\textwidth}
\centering
\begin{tabular}{ccc|c}
\hline \hline
$x_1$ & $x_2$ & $y$ & $p$ \\
\hline
$0$ & $0$ & $0$ & $0.25$ \\
$0$ & $1$ & $1$ & $0.25$ \\
$1$ & $0$ & $1$ & $0.25$ \\
$1$ & $1$ & $0$ & $0.25$ \\
\hline \hline
\end{tabular}
\caption{Agreement XOR}
\label{tab:xor}
\end{subtable}
\begin{subtable}{0.24\textwidth}
\centering
\begin{tabular}{ccc|c}
\hline \hline
$x_1$ & $x_2$ & $y$ & $p$ \\
\hline
$0$ & $0$ & $0$ & $0.25$ \\
$0$ & $1$ & $0$ & $0.25$ \\
$1$ & $0$ & $1$ & $0.25$ \\
$1$ & $1$ & $1$ & $0.25$ \\
\hline \hline
\end{tabular}
\caption{$y=x_1$}
\label{tab:u}
\end{subtable}
\begin{subtable}{0.24\textwidth}
\centering
\begin{tabular}{ccc|c}
\hline \hline
$x_1$ & $x_2$ & $y$ & $p$ \\
\hline
$0$ & $0$ & $0$ & $0.5$ \\
$1$ & $1$ & $1$ & $0.5$ \\
\hline \hline
\end{tabular}
\caption{$y=x_1=x_2$}
\label{tab:r}
\end{subtable}
\vspace{-4mm}
\label{tab:lower_dis_ex}
\end{table}

\paragraph{2. Guidelines} We provide a guideline to decide whether a lower or upper bound on synergy can be considered `close enough'. It is close enough if the maximum interaction can be consistently estimated - often the exact value of synergy is not the most important (e.g, whether $S$ is $0.5$ or $0.6$) but rather synergy relative to other interactions (e.g., if we estimate $S \in [0.2, 0.5]$, and exactly compute $R = U_1 = U_2 = 0.1$, then we know for sure that $S$ is the most important interaction and can collect data or design models based on that). We find that our bounds accurately identify the same highest interaction on all 10 real-world datasets as the true synergy does. Furthermore, we observed that the estimated synergy correlates very well with true synergy: as high as $1.05$ on ENRICO (true $S = 1.02$) and as low as $0.21$ on MIMIC (true $S = 0.02$).

\paragraph{3. The relationship between $S$ and $R$} In Table~\ref{tab:xor} we show the classic \textsc{agreement XOR} distribution where $X_1$ and $X_2$ are independent, but $Y=1$ sets $X_1 \neq X_2$ to increase their dependence. $I(X_1;X_2;Y)$ is negative, and $\lowa = 1 \le 1=S$ is tight. On the other hand, Table~\ref{tab:r} is an extreme example where the probability mass is distributed uniformly only when $y=x_1=x_2$ and $0$ elsewhere. As a result, $X_1$ is always equal to $X_2$ (perfect dependence), and yet $Y$ perfectly explains away the dependence between $X_1$ and $X_2$ so $I(X_1;X_2|Y) = 0$: $\lowa = 0 \le 0=S$. A real-world example is multimodal sentiment analysis from text, video, and audio on \textsc{MOSEI}, $R=0.26$ and $S=0.03$, and as expected the lower bound is small $\lowa = 0 \le 0.03=S$ (Table~\ref{tab:s}).

\paragraph{4. The relationship between $S$ and $U$} In Table~\ref{tab:dis_xor} we show an example called \textsc{disagreement XOR}. There is maximum disagreement between $p(y|x_1)$ and $p(y|x_2)$: the likelihood for $y$ is high when $y$ is the opposite bit as $x_1$, but reversed for $x_2$. Given both $x_1$ and $x_2$: $y$ takes a `disagreement' XOR of the individual marginals, i.e. $p(y|x_1,x_2) = \argmax_y p(y|x_1) \ \textrm{XOR} \ \argmax_y p(y|x_2)$, which indicates synergy (note that an exact XOR would imply perfect agreement and high synergy). The actual disagreement is $0.15$, $S$ is $0.16$, and $U$ is $0.02$, indicating a very strong lower bound $\lowb=0.14 \le 0.16=S$. A real-world equivalent dataset is \textsc{MUStARD}, where the presence of sarcasm is often due to a contradiction between what is expressed in language and speech, so disagreement $\alpha=0.12$ is the highest out of all the video datasets, giving a lower bound $\lowb=0.11 \le 0.44 = S$.

The lower bound is low when all disagreement is explained by uniqueness (e.g., $y=x_1$, Table~\ref{tab:u}), which results in $\lowb = 0 \le 0 = S$ ($\alpha$ and $U$ cancel each other out). A real-world equivalent is \textsc{MIMIC}: from Table~\ref{tab:s}, disagreement is high $\alpha=0.13$ due to unique information $U_1=0.25$, so the lower bound informs us about the lack of synergy $\lowb = -0.12 \le 0.02 = S$.
Finally, the lower bound is loose when there is synergy without disagreement, such as \textsc{agreement XOR} ($y=x_1 \textrm{ XOR } x_2$, Table~\ref{tab:xor}) where the marginals $p(y|x_i)$ are both uniform, but there is full synergy: $\lowb = 0 \le 1 = S$. Real-world datasets include \textsc{UR-FUNNY} where there is low disagreement in predicting humor $\alpha=0.03$, and relatively high synergy $S=0.18$, which results in a loose lower bound $\lowb = 0.01 \le 0.18=S$.

\paragraph{5. On upper bounds for synergy} The upper bound for \textsc{MUStARD} is close to real synergy, $\high = 0.79 \ge 0.44=S$. On \textsc{MIMIC}, the upper bound is the lowest $\high = 0.41$, matching the lowest $S=0.02$. Some of the other examples in Table~\ref{tab:s} show weaker bounds.
This could be because (i) there exists high synergy distributions that match $\mathcal{D}_i$ and $\mathcal{D}_M$, but these are rare in the real world, or (ii) our approximation used in Theorem~\ref{thm:upper} is loose. We leave these as directions for future work.

\begin{table}[t]
\centering
\fontsize{9}{11}\selectfont
\setlength\tabcolsep{3.0pt}
\vspace{-0mm}
\caption{Estimated lower, upper, and average bounds on optimal multimodal performance in comparison with the actual best unimodal model, the best simple fusion model, and the best complex fusion model. Our performance estimates closely predict actual model performance, \textit{despite being computed only on semi-supervised data and never training the model itself.}}
\centering
\footnotesize
\begin{tabular}{l|cccccccccc}
\hline \hline
& \textsc{MOSEI} & \textsc{UR-FUNNY} & \textsc{MOSI} & \textsc{MUStARD} & \textsc{MIMIC} & \textsc{ENRICO} \\
\hline
Estimated upper bound & $1.07$ & $1.21$ & $1.29$ & $1.63$ & $1.27$ & $0.88$ \\
Best complex multimodal & $0.88$ & $0.77$ & $0.86$ & $0.79$ & $0.92$ & $0.51$ \\
Best simple multimodal & $0.85$ & $0.76$ & $0.84$ & $0.74$ & $0.92$ & $0.49$ \\
Best unimodal & $0.82$ & $0.74$ & $0.83$ & $0.74$ & $0.92$ & $0.47$  \\
Estimated lower bound & $0.52$ & $0.58$ & $0.62$ & $0.78$ & $0.76$ & $0.48$ \\
\hline
Estimated average & $0.80$ & $0.90$ & $0.96$ & $1.21$ & $1.02$ & $0.68$ \\
\hline \hline
\end{tabular}

\vspace{-4mm}
\label{tab:acc}
\end{table}

\paragraph{Additional results} In the full paper~\cite{liang2023multimodal}, we also study the effect of imperfect unimodal predictors and disagreement measurements on our derived bounds, by perturbing the label by various noise levels (from no noise to very noisy) and examining the changes in estimated upper and lower bounds. We found these bounds are quite robust to label noise, still giving close trends of $S$. We also include more discussions studying the relationships between various interactions, and how the relationship between disagreement and synergy can inspire new self-supervised learning methods.

\subsection{Implications towards performance, data collection, model selection}

Now that we have validated the accuracy of these bounds, we apply them to estimate multimodal performance in semi-supervised settings. This serves as a strong signal for deciding (1) whether to collect paired and labeled data from a second modality, and (2) what type of multimodal fusion method should be used. To estimate performance given $\mathcal{D}_1$, $\mathcal{D}_2$, and $\mathcal{D}_M$, we first compute our lower and upper bounds $\underline{S}$ and $\overline{S}$. Combined with the exact computation of $R$ and $U$, we obtain the total information $I_p(\{X_1,X_2\}; Y)$, and combine a result from~\citet{feder1994relations} with Fano's inequality~\citep{fano1968transmission} to yield tight bounds of performance as a function of total information.
\begin{theorem}
\label{thm:performance}
    Let $P_\textrm{acc}(f_M^*) = \mathbb{E}_p \left[ \mathbf{1} \left[ f_M^*(x_1,x_2) = y  \right] \right]$ denote the accuracy of the Bayes' optimal multimodal model $f_M^*$ (i.e., $P_\textrm{acc} (f_M^*) \ge P_\textrm{acc} (f'_M)$ for all $f'_M \in \mathcal{F}_M$). We have that
    \begin{align}
        2^{I_p(\{X_1,X_2\}; Y)-H(Y)} \leq P_\textrm{acc}(f_M^*) \leq \frac{I_p(\{X_1,X_2\}; Y) + 1}{\log |\mathcal{Y}|},
    \end{align}
    and we can plug in $R+U_1,U_2+\underline{S} \le I_p(\{X_1,X_2\}; Y) \le R+U_1,U_2+\high$ to obtain lower $\underline{P}_\textrm{acc}(f_M^*)$ and upper $\overline{P}_\textrm{acc}(f_M^*)$ bounds on optimal multimodal performance.
\end{theorem}
We show the proof in the full paper~\cite{liang2023multimodal}. Finally, we summarize estimated multimodal performance as the average $\hat{P}_M = (\underline{P}_\textrm{acc}(f_M^*) + \overline{P}_\textrm{acc}(f_M^*))/2$. A high $\hat{P}_M$ suggests the presence of important joint information from both modalities (not present in each) which could boost accuracy, so it is worthwhile to collect the full distribution $p$ and explore multimodal fusion.

\paragraph{Setup} For each MultiBench dataset, we implement a suite of unimodal and multimodel models spanning simple and complex fusion. Unimodal models are trained and evaluated separately on each modality. Simple fusion includes ensembling by taking an additive or majority vote between unimodal models~\citep{hastie1987generalized}. Complex fusion is designed to learn higher-order interactions as exemplified by bilinear pooling~\citep{fukui2016multimodal}, multiplicative interactions~\citep{jayakumar2020multiplicative}, tensor fusion~\citep{zadeh2017tensor}, and cross-modal self-attention~\citep{tsai2019multimodal}. See our full paper~\cite{liang2023multimodal} for models and training details. We include unimodal, simple and complex multimodal performance, as well as estimated lower and upper bounds on performance in Table~\ref{tab:acc}.

\paragraph{RQ1: Estimating multimodal fusion performance} \textit{How well could my multimodal model perform?} We find that estimating interactions enables us to \textit{closely predict multimodal model performance, before even training a model}. For example, on \textsc{MOSEI}, we estimate the performance to be $52\%$ based on the lower bound and $107\%$ based on the upper bound, for an average of $80\%$ which is very close to true model performance ranging from $82\%$ for the best unimodal model, and $85\%-88\%$ for various multimodal model. Estimated performances for \textsc{ENRICO}, \textsc{UR-FUNNY}, and \textsc{MOSI} are $68\%$, $90\%$, $96\%$, which track true performances $51\%$, $77\%$, $86\%$.

\begin{wrapfigure}{R}{0.6\textwidth}
\centering
\begin{subfigure}{.3\textwidth}
  \centering
  \vspace{-0mm}
  \includegraphics[width=\linewidth]{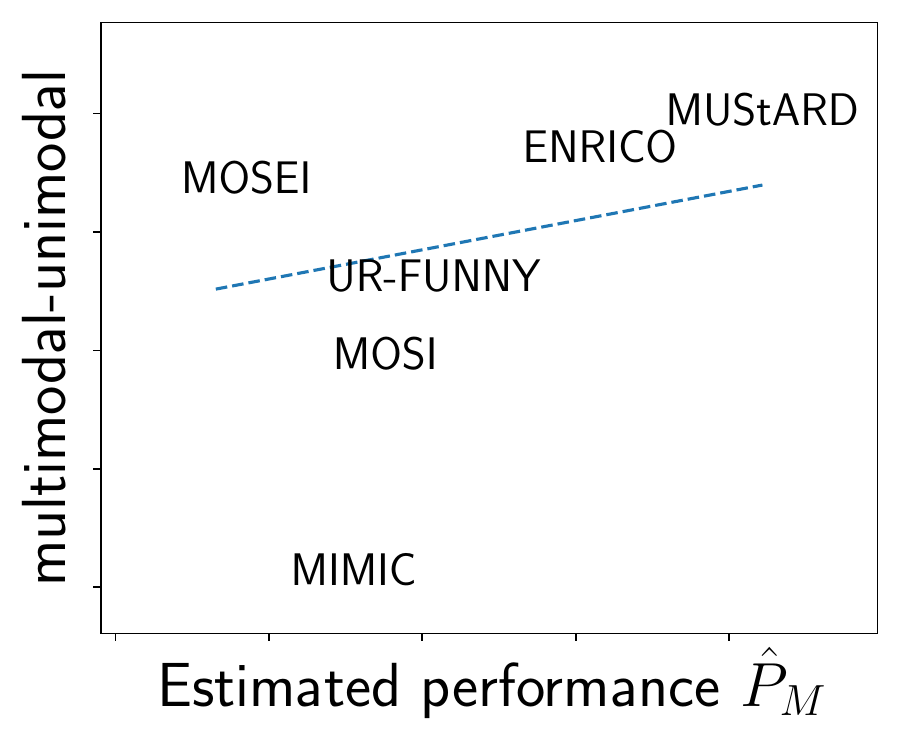}
  \vspace{-0mm}
\end{subfigure}%
\begin{subfigure}{.3\textwidth}
  \centering
  \vspace{-0mm}
  \includegraphics[width=\linewidth]{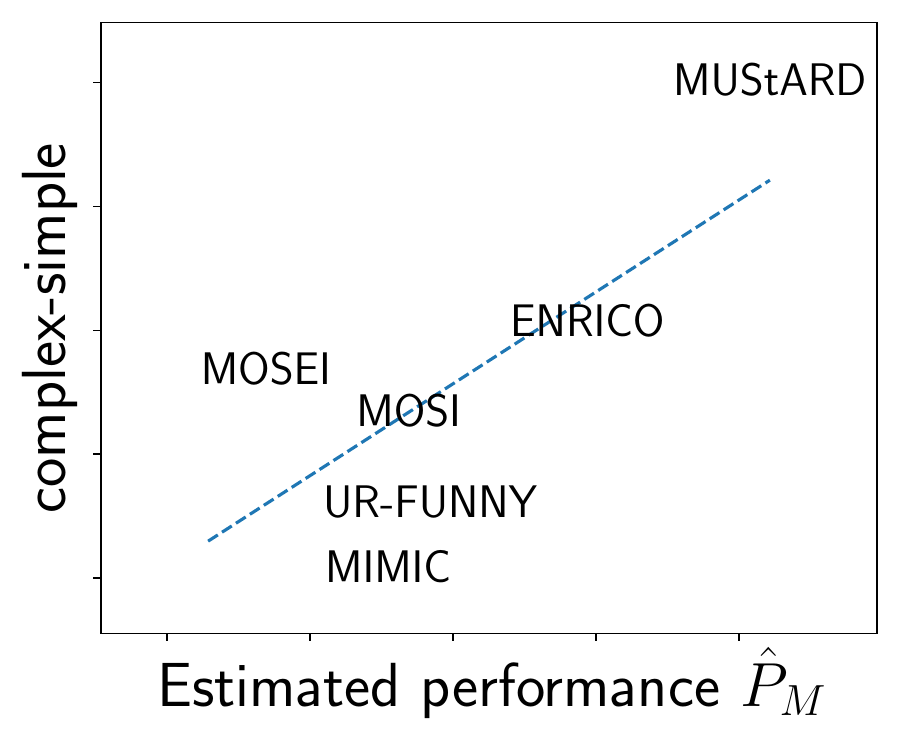}
  \vspace{-0mm}
\end{subfigure}
\vspace{-4mm}
\caption{Datasets with higher estimated multimodal performance $\hat{P}_M$ tend to show improvements from unimodal to multimodal (left) and from simple to complex multimodal fusion (right).}
\label{fig:plot2}
\vspace{-2mm}
\end{wrapfigure}

\paragraph{RQ2: Data collection} \textit{Should I collect multimodal data?} We compare estimated performance $\hat{P}_M$ with the actual difference between unimodal and best multimodal performance in Figure~\ref{fig:plot2} (left).
Higher estimated $\hat{P}_M$ correlates with a larger gain from unimodal to multimodal (correlation $\rho = 0.21$ and rises to $0.53$ if ignoring the outlier in MIMIC). \textsc{MUStARD} and \textsc{ENRICO} show the most opportunity for multimodal modeling. Therefore, a rough guideline is that if the estimated multimodal performance based on semi-supervised data is higher, then collecting the full labeled multimodal data is worth it.

\paragraph{RQ3: Model selection} \textit{What model should I choose for multimodal fusion?} We note strong relationships between estimated performance and the performance of different fusion methods.
From Table~\ref{tab:acc}, synergistic datasets like \textsc{MUStARD} and \textsc{ENRICO} show best multimodal performance only slightly above our estimated lower bound, indicating that there is a lot of room for improvement in better fusion methods. Indeed, more complex fusion methods such as multimodal transformer designed to capture synergy is the best on \textsc{MUStARD} which matches its high synergy ($72\%$ accuracy).
For datasets with less synergy like \textsc{MOSEI} and \textsc{MIMIC}, the best multimodal performance is much higher than the estimated lower bound, indicating that existing fusion methods may already be quite optimal. Indeed, simpler fusion methods such as feature alignment, designed to capture redudnancy, are the best on \textsc{MOSEI} which matches its high redundancy ($80\%$ accuracy).

Figure~\ref{fig:plot2} (right) shows a visual comparison, where plotting the performance gap between complex and simple fusion methods against estimated performance $\hat{P}_M$ shows a correlation coefficient of $0.77$.
We again observe positive trends between higher estimated performance and improvements with complex fusion, with large gains on \textsc{MUStARD} and \textsc{ENRICO}.
We expect new methods to further improve the state-of-the-art on these datasets due to their generally high interaction values and low multimodal performance relative to estimated lower bound $\underline{P}_\textrm{acc}(f_M^*)$. Therefore, a rough guideline is that if the estimated multimodal performance based on semi-supervised data is higher, then there is more potential for improvement by trying more complex multimodal fusion strategies.

\vspace{-1mm}
\section{Conclusion and Broader Impacts}
\label{sec:conclusion}
\vspace{-1mm}

We proposed estimators of multimodal interactions when observing only \textit{labeled unimodal data} and some \textit{unlabeled multimodal data}, a general semi-supervised setting that encompasses many real-world constraints involving partially observable modalities, limited labels, and privacy concerns. Our key results draw new connections between multimodal interactions, the disagreement of unimodal classifiers, and min-entropy couplings, which yield new insights for estimating multimodal model performance, data analysis, and model selection. We are aware of some potential \textbf{limitations}:
\begin{enumerate}[noitemsep,topsep=0pt,nosep,leftmargin=*,parsep=0pt,partopsep=0pt]
    \item These estimators only approximate real interactions due to cluster preprocessing or unimodal models, which naturally introduce optimization and generalization errors. We expect progress in density estimators, generative models, and unimodal classifiers to address these problems.

    \item It is harder to quantify interactions for certain datasets, such as \textsc{ENRICO} which displays all interactions which makes it difficult to distinguish between $R$ and $S$ or $U$ and $S$.

    \item Finally, there exist challenges in quantifying interactions since the data generation process is never known for real-world datasets, so we have to resort to human judgment, other automatic measures, and downstream tasks such as estimating model performance and model selection.
\end{enumerate}

\textbf{Future work} should investigate more applications of multivariate information theory in designing self-supervised models, predicting multimodal performance, and other tasks involving feature interactions such as privacy-preserving and fair representation learning from high-dimensional data~\citep{dutta2020information,hamman2023demystifying}. Being able to provide guarantees for fairness and privacy-preserving learning, especially for semi-supervised pretraining datasets, can be particularly impactful.

\chapter{MultiBench: Large-scale Resources for Multisensory Learning}
\label{chap:models1}
\section{Introduction}

Current multimodal research has led to impressive advances in benchmarking and modeling for specific domains such as language and vision~\cite{agrawal2017vqa,liang2018computational,lin2014microsoft,ramesh2021zero}. However, other domains, modalities, and tasks are relatively understudied. The future will lie in multisensory foundation models that are \textit{grounded in the world}: being able to simultaneously process a large number of modalities beyond language, to vision, audio~\cite{agrawal2017vqa,liang2018computational,lin2014microsoft,ramesh2021zero}, and leveraging advances in sensing technologies such as cellphones~\cite{liang2021learning}, wearable devices~\cite{hamisu2011accessible}, autonomous vehicles~\cite{yeong2021sensor}, healthcare technologies~\cite{MIMIC}, and robots~\cite{belpaeme2018social,kim2013social} that give a wealth of sensor data about the world.

\begin{figure*}[tbp]
\centering
\vspace{-0mm}
\includegraphics[width=\linewidth]{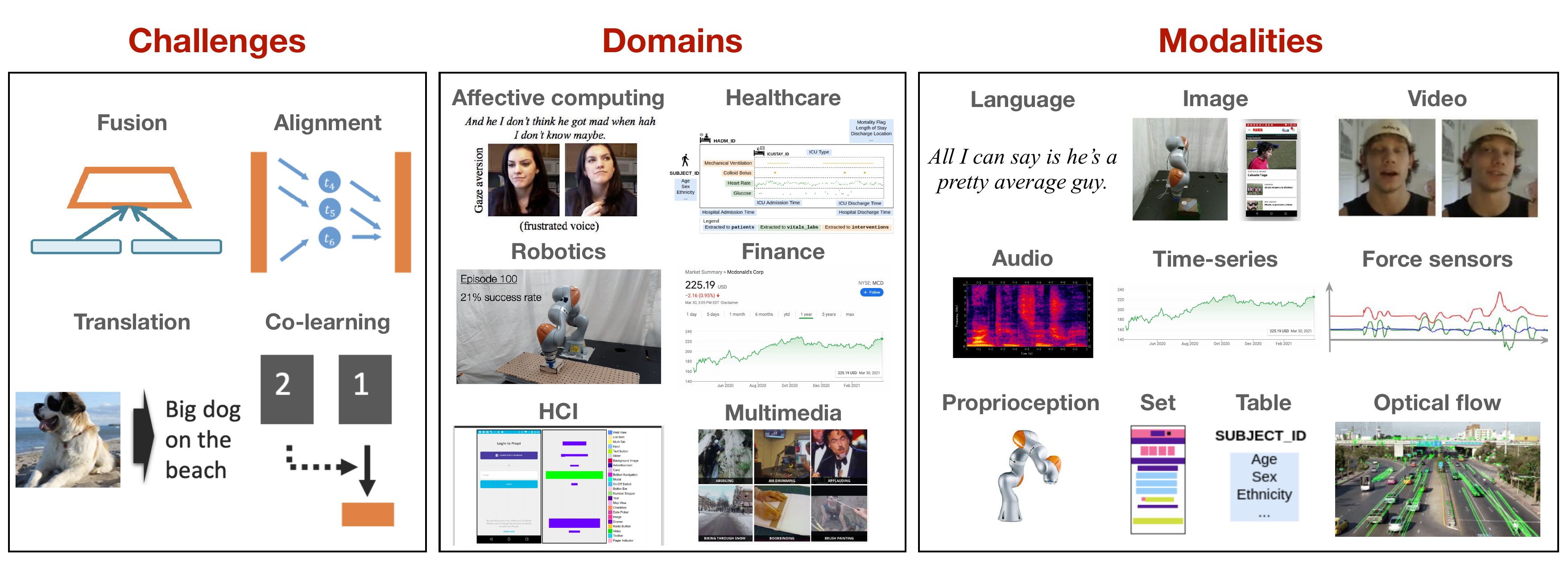}
\vspace{-0mm}
\caption{\multibench\ contains a diverse set of $28$ datasets spanning $14$ modalities and testing for more than $30$ prediction tasks across $6$ distinct research areas and $5$ technical challenges of multimodal machine learning, thereby enabling standardized, reliable, and reproducible large-scale benchmarking of multimodal models. To reflect real-world requirements, \multibench\ is designed to holistically evaluate generalization performance across domains and modalities.}
\vspace{-0mm}
\label{multibench:figs:overview}
\end{figure*}

\textbf{\multibench}: In order to accelerate research in building general-purpose multimodal foundation models, this chapter describes \multibench\ (Figure~\ref{multibench:figs:overview}), a systematic and unified large-scale benchmark that brings us closer to the requirements of real-world multimodal applications. \multibench\ is designed to comprehensively evaluate generalization across domains and modalities. To that end, \multibench\ contains a diverse set of $28$ datasets spanning $14$ modalities and testing for more than $30$ prediction tasks across $6$ distinct research areas and $5$ technical challenges of multimodal machine learning. These research areas include important tasks understudied from a multimodal learning perspective, such as healthcare, finance, and HCI. Building upon extensive data-collection efforts by domain experts, we worked with them to adapt datasets that reflect real-world relevance, present unique challenges to multimodal learning, and enable opportunities in algorithm design and evaluation.

Together, \multibench\ unifies efforts across separate research areas in multimodal learning to enable quick and accurate benchmarking across a wide range of datasets and metrics. 
To help the community accurately compare performance and ensure reproducibility, \multibench\ includes an end-to-end pipeline including data preprocessing, dataset splits, multimodal algorithms, evaluation metrics, and cross-validation protocols. This includes an implementation of $20$ core multimodal approaches spanning innovations in fusion paradigms, optimization objectives, and training approaches in a standard public toolkit called \multizoo.
We perform a systematic evaluation and show that directly applying these methods can improve the state-of-the-art performance on $9$ out of the $15$ datasets. Therefore, \multibench\ presents a step towards unifying disjoint efforts in multimodal research and paves a way towards a deeper understanding of multimodal models. Most importantly, our public zoo of multimodal benchmarks and models will ensure ease of use, accessibility, and reproducibility. Finally, we outline our plans to ensure the continual availability, maintenance, and expansion of \multibench, including using it as a theme for future workshops and competitions and to support the multimodal learning courses taught around the world.

\newcolumntype{L}[1]{>{\arraybackslash}p{#1}}
\newcolumntype{C}[1]{>{\centering\arraybackslash}p{#1}}

\begin{table*}[]
\setlength\tabcolsep{0.5pt}
\vspace{-0mm}
\caption{\multibench\ provides a comprehensive suite of $28$ multimodal datasets to benchmark current and proposed approaches in multimodal machine learning. It covers a diverse range of technical challenges, research areas, dataset sizes, input modalities (in the form of $a$: audio, $e$: embodied environment, $f$: force sensor, $g$: graph, $i$: image $\ell$: language, $o$: optical flow, $p$: proprioception sensor, $\pi$: policy/action, $q$: question (for question-answering tasks), $s$: set, $t$: time-series, $ta$: tabular, $v$: video), and prediction tasks. We provide a standardized data loader for datasets in \multibench, along with a set of state-of-the-art multimodal models.}
\centering
\footnotesize
\vspace{-0mm}
\begin{tabular}{l|l|cccccccc}
\hline \hline
\multicolumn{1}{l|}{Challenge} & \multicolumn{1}{l|}{Research Area} & Size & Dataset & Modalities & \# Samples & Prediction task \\ \hline
\multirow{17}{*}{Fusion} & \multirow{4}{*}{Affect} & \multirow{1}{*}{S} & \textsc{MUStARD}~\cite{castro2019towards} & $\{\ell,v,a\} \rightarrow y$ & $690$ & sarcasm \\
& & \multirow{1}{*}{M} & \textsc{CMU-MOSI}~\cite{zadeh2016mosi} & $\{\ell,v,a\} \rightarrow y$ & $2,199$ & sentiment \\
& & \multirow{1}{*}{L} & \textsc{UR-FUNNY}~\cite{hasan2019ur} & $\{\ell,v,a\} \rightarrow y$ & $16,514$ & humor \\
& & \multirow{1}{*}{L} & \textsc{CMU-MOSEI}~\cite{zadeh2018multimodal} & $\{\ell,v,a\} \rightarrow y$ & $22,777$ & sentiment, emotions \\
\cline{2-7}%
& \multirow{2}{*}{Healthcare} & L & \textsc{MIMIC}~\cite{MIMIC} & $\{t,ta\} \rightarrow y$ & $36,212$ & mortality, ICD-$9$ codes\\
& & L & \textsc{MIMIC-CXR}~\cite{johnson2019mimic} & $\{\ell,i\} \rightarrow y$ & $377,110$ & mortality, ICD-$9$ codes\\
\cline{2-7}%
& \multirow{2}{*}{Robotics} & M & \textsc{MuJoCo Push}~\cite{lee2020multimodal} & $\{i,f,p\} \rightarrow y$ & $37,990$ & object pose \\ 
& & L & \textsc{Vision\&Touch}~\cite{lee2019making} & $\{i,f,p\} \rightarrow y$ & $147,000$ & contact, robot pose \\
\cline{2-7}%
& \multirow{3}{*}{Finance} & M & \textsc{Stocks-F\&B} & $\{t \times 18\} \rightarrow y$ & $5,218$ & stock price, volatility \\
& & M & \textsc{Stocks-Health} & $\{t \times 63\} \rightarrow y$ & $5,218$ & stock price, volatility \\
& & M & \textsc{Stocks-Tech} & $\{t \times 100\} \rightarrow y$ & $5,218$ & stock price, volatility \\
\cline{2-7}%
& \multirow{1}{*}{HCI} & S & \textsc{ENRICO}~\cite{leiva2020enrico} & $\{i,s\} \rightarrow y$ & $1,460$ & design interface \\
\cline{2-7}%
& \multirow{4}{*}{Multimedia} & \multirow{1}{*}{M} & \textsc{Hateful Memes}~\cite{kiela2020hateful} & $\{\ell,i\} \rightarrow y$ & $10,000$ & hate speech\\
& & \multirow{1}{*}{M} & \textsc{MM-IMDb}~\cite{arevalo2017gated} & $\{\ell,i\} \rightarrow y$ & $25,959$ & movie genre \\
& & \multirow{1}{*}{M} & \textsc{AV-MNIST}~\cite{vielzeuf2018centralnet} & $\{i,a\} \rightarrow y$ & $70,000$ & digit \\
& & \multirow{1}{*}{L} & \textsc{Kinetics400}~\cite{kay2017kinetics} & $\{v,a,o\} \rightarrow y$ & $306,245$ & human action \\
\hline \hline
\multirow{3}{*}{\begin{tabular}[c]{@{}l@{}}Question\\Answering\end{tabular}} & \multirow{1}{*}{Affect} & M & \textsc{Social IQ}~\cite{zadeh2019social} & $\{v,a,\ell,q\} \rightarrow y$ & $7,500$ & QA \\
\cline{2-7}%
& \multirow{2}{*}{Multimedia} & L & \textsc{CLEVR}~\cite{johnson2017clevr} & $\{i,q\} \rightarrow y$ & $853,554$ & QA \\
& & L & \textsc{VQA 2.0}~\cite{goyal2017making} & $\{i,q\} \rightarrow y$ & $1,100,000$ & QA \\ \hline \hline
\multirow{4}{*}{Retrieval} & \multirow{4}{*}{Multimedia} & S & \textsc{CIFAR-ESC}~\cite{liang2021cross} & $i \leftrightarrow a$ & $2,080$ & image-audio retrieval \\
& & M & \textsc{Clotho}~\cite{drossos2020clotho} & $a \leftrightarrow \ell$ & $24,905$ & audio-caption retrieval \\
& & M & \textsc{Yummly-28K}~\cite{min2016being} & $i \leftrightarrow \ell$ & $27,638$ & image-caption retrieval \\
& & L & \textsc{Flickr-30k}~\cite{plummer2015flickr30k} & $i \leftrightarrow \ell$ & $158,000$ & image-caption retrieval \\
\hline \hline
\multirow{1}{*}{RL} & \multirow{1}{*}{Simulation} & L & \textsc{RTFM}~\cite{zhong2019rtfm} & $\{e, \ell \rightarrow \pi \}$ & - & multimodal RL \\
\hline \hline
\multirow{4}{*}{Co-learning} & \multirow{2}{*}{Affect} & M & \textsc{CMU-MOSI} $\rightarrow$ \textsc{SST}~\cite{zadeh2020foundations} & $\{\ell,v,a\} \rightarrow \ell$ & $11,855$ & video $\rightarrow$ text \\
& & L & \textsc{CMU-MOSEI} $\rightarrow$ \textsc{SST}~\cite{zadeh2020foundations} & $\{\ell,v,a\} \rightarrow \ell$ & $11,855$ & video $\rightarrow$ text \\
\cline{2-7}%
& \multirow{2}{*}{Multimedia} & M & \textsc{GloVe} $\rightarrow$ \textsc{CIFAR10}~\cite{socher2013zero} & $\{i,\ell\} \rightarrow i$ & $60,000$ & text $\rightarrow$ image \\
& & L & Visual Genome~\cite{krishna2017visual,marino2016more} & $\{i,g\} \rightarrow i$ & $100,000$ & knowledge graph $\rightarrow$ image \\
\hline \hline
\end{tabular}
\vspace{-4mm}
\label{multibench:data:overview}
\end{table*}

\section{\multibench: The Multiscale Multimodal Benchmark}
\label{multibench:dataset}

\textbf{Background}: We define a modality as a single particular mode in which a signal is expressed or experienced. Multiple modalities then refer to a combination of multiple heterogeneous signals~\citep{baltruvsaitis2018multimodal}.
The first version of \multibench\ focuses on benchmarking algorithms for \textit{multimodal fusion}, where the main challenge is to join information from two or more modalities to perform a prediction (e.g., classification, regression). Classic examples for multimodal fusion include audio-visual speech recognition where visual lip motion is fused with speech signals to predict spoken words~\citep{dupont2000audio}. Multimodal fusion can be contrasted with multimodal translation where the goal is to generate a new and different modality~\citep{vinyals2016show}, grounding and question answering where one modality is used to query information in another (e.g., visual question answering~\citep{agrawal2017vqa}), and unsupervised or self-supervised multimodal representation learning~\citep{lu2019vilbert,Su2020VLBERT}. We plan future versions of \multibench\ to study these important topics in multimodal research.

Each of the following $15$ datasets in \multibench\ contributes a unique perspective to the various technical challenges in multimodal learning involving learning and aligning complementary information, scalability to a large number of modalities, and robustness to realistic real-world imperfections.

\begin{figure*}[]
\centering
\vspace{-0mm}
\includegraphics[width=\linewidth]{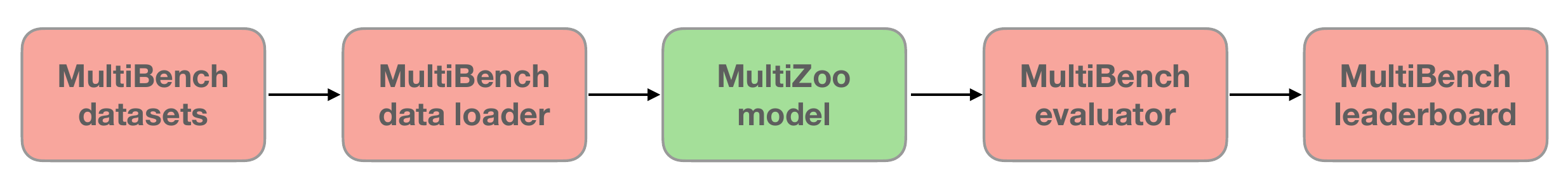}
\vspace{-0mm}
\caption{\multibench\ provides a standardized machine learning pipeline across data processing, data loading, multimodal models, evaluation metrics, and a public leaderboard to encourage future research in multimodal representation learning. \multibench\ aims to present a milestone in unifying disjoint efforts in multimodal machine learning research and paves the way towards a better understanding of the capabilities and limitations of multimodal models, all the while ensuring ease of use, accessibility, and reproducibility.}
\vspace{-0mm}
\label{multibench:figs:pipeline}
\end{figure*}

\multibench\ provides a standardized machine learning pipeline that starts from data loading to running multimodal models, providing evaluation metrics, and a public leaderboard to encourage future research in multimodal representation learning (see Figure~\ref{multibench:figs:pipeline}).
Table~\ref{multibench:data:overview} shows an overview of these datasets. We provide a brief overview of the research areas, modalities, and tasks for each of these datasets.

\subsection{Research areas}

\textbf{Affective computing} studies the perception of human affective states (emotions, sentiment, and personalities) from our display of multimodal signals spanning language (spoken words), visual (facial expressions, gestures), and acoustic (prosody, speech tone)~\cite{picard2000affective}. It has impacts towards building emotionally intelligent computers, human behavior analysis, and AI-assisted education.

\textbf{Healthcare}: Modern medical decision-making often involves integrating complementary information and signals from several sources such as lab tests, imaging reports, and patient-doctor conversations. Multimodal models can help doctors make sense of high-dimensional data and assist them in the diagnosis process~\cite{amisha2019overview}.

\textbf{Robotics}: Modern robot systems are equipped with multiple sensors to aid in their decision-making. Some systems also have a large number of heterogeneous sensors deployed in the real world with realistic noise and imperfections. These present scalability and robustness challenges for multimodal machine learning.

\textbf{Finance}: The field of machine learning for finance studies the use of algorithms to make better automatic trading decisions through historical data, news and document understanding, social media analytics, and other multimodal signals. This field presents challenges in time-series analysis on high-frequency multimodal signals, a dynamic and large number of possible modalities, as well as robustness and compute efficiency for real-world deployment.

\textbf{Human Computer Interaction (HCI)} studies the design of computer technology and interactive interfaces between humans and computers~\cite{dix2000human}. Many real-world human-centric problems involve multimodal inputs such as language, visual, and audio interfaces. Designing multimodal models that actively interact with humans further necessitates guarantees on their fairness and robustness in real-world scenarios.

\textbf{Multimedia}: A significant body of research in multimodal learning has been fueled by the large availability of multimedia data (language, image, video, and audio) on the internet. Multimedia research is exemplified by the research tasks of media description, multimodal question answering, and cross-modal retrieval.

\textbf{Simulated environments}: Finally, simulated interactive environments such as Atari games~\cite{bellemare2013arcade}, Minecraft~\cite{guss2019minerl}, and NetHack~\cite{kuttler2020nethack} present scalable opportunities for research in reinforcement learning while also enabling rich programming of multimodal environments involving text~\cite{luketina2019survey}, audio~\cite{dean2020see}, and video~\cite{chaplot2017gated}. By way of their flexible design, these environments can often provide richer interactions between text and embodied environments, more difficult planning and exploration challenges, and procedurally generated tasks of increasing difficulty.

\subsection{Fusion datasets}

In multimodal fusion, the main challenge is to join information from two or more modalities to perform a prediction. Classic examples include audio-visual speech recognition where visual lip motion is fused with speech signals to predict spoken words~\cite{dupont2000audio}. Information coming from different modalities have varying predictive power by themselves and also when complemented by each other (i.e., higher-order interactions). In order to capture higher-order interactions, there is also a need to identify the relations between granular units from two or more different modalities (i.e., alignment). When dealing with temporal data, it also requires capturing possible long-range dependencies across time (i.e., temporal alignment). \multibench\ contains the following datasets for multimodal fusion spanning several research areas:

\textbf{Affective computing}: \multibench\ contains $4$ datasets involving fusing \textit{language}, \textit{video}, and \textit{audio} time-series data to predict sentiment (\textsc{CMU-MOSI}~\cite{zadeh2016mosi}), emotions (\textsc{CMU-MOSEI}~\cite{zadeh2018multimodal}), humor (\textsc{UR-FUNNY}~\cite{hasan2019ur}), and sarcasm (\textsc{MUStARD}~\cite{castro2019towards}). Complementary information may occurs at different moments, requiring models to address the multimodal challenges of grounding and alignment.

\textbf{Healthcare}: \multibench\ includes the large-scale \textsc{MIMIC} dataset~\cite{MIMIC} which records ICU patient data including \textit{time-series} data measured every hour and other demographic variables (e.g., age, gender, ethnicity in the form of \textit{tabular numerical} data). These are used to predict the disease ICD-$9$ code and mortality rate. \textsc{MIMIC} poses unique challenges in integrating time-varying and static modalities, reinforcing the need of aligning multimodal information at correct granularities. Extending \textsc{MIMIC}, we also include the \textsc{MIMIC-CXR}~\cite{johnson2019mimic} datasets of de-identified publicly available chest radiographs and free-text reports

\textbf{Robotics}: We include \textsc{MuJoCo Push}~\cite{lee2020multimodal} and \textsc{Vision\&Touch}~\cite{lee2019making} which record the manipulation of simulated and real robotic arms equipped with \textit{visual} (RGB and depth), \textit{force}, and \textit{proprioception} sensors. In \textsc{MuJoCo Push}, the goal is to predict the pose of the object being pushed by the robot end-effector. In \textsc{Vision\&Touch}, the goal is to predict action-conditional learning objectives that capture forward dynamics of contact prediction and robot end-effector pose.
Robustness is important due to the risk of real-world sensor failures~\cite{lee2020detect}.

\textbf{Finance}: We gathered historical stock data from the internet to create our own dataset for financial time-series prediction across $3$ groups of correlated stocks: \textsc{Stocks-F\&B}, \textsc{Stocks-Health}, and \textsc{Stocks-Tech}. Within each group, the previous stock prices of a set of stocks are used as multimodal \textit{time-series} inputs to predict the price and volatility of a related stock (e.g., using Apple, Google, and Microsoft data to predict future Microsoft prices). Multimodal stock prediction~\cite{sardelich2018multimodal} presents scalability issues due to a large number of modalities ($18/63/100$ vs $2/3$ in most datasets), as well as robustness challenges arising from real-world data with an inherently low signal-to-noise ratio.

\textbf{HCI}: We use the \textsc{Enrico} (Enhanced Rico) dataset~\cite{deka2017rico,leiva2020enrico} of Android app screens (consisting of an \textit{image} as well as a \textit{set} of apps and their locations) categorized by their design motifs and collected for data-driven design applications such as design search, user interface (UI) layout generation, UI code generation, and user interaction modeling.

\textbf{Multimedia}: \multibench\ includes $4$ popular large-scale multimedia datasets with varying sizes and levels of difficulty: (1) the hateful memes challenge~\cite{kiela2020hateful} as a core challenge in multimedia to ensure safer learning from ubiquitous text and images from the internet, (2) \textsc{AV-MNIST}~\cite{vielzeuf2018centralnet} is assembled from \textit{images} of handwritten digits~\cite{mnist} and \textit{audio} samples of spoken digits~\cite{tidigits}, (3) \textsc{MM-IMDb}~\cite{arevalo2017gated} uses movie \textit{titles}, \textit{metadata}, and movie \textit{posters} to perform multi-label classification of movie genres, and (4) \textsc{Kinetics}~\cite{kay2017kinetics} contains \textit{video}, \textit{audio}, and \textit{optical flow} of $306,245$ video clips annotated for $400$ human actions.

\subsection{Question answering datasets}

Within the domain of language and vision, there has been growing interest in language-based question answering (i.e., ``query'' modality) of entities in the visual, video, or embodied domain (i.e., ``queried'' modality). Datasets such as Visual Question Answering~\cite{agrawal2017vqa}, Social IQ~\cite{zadeh2019social}, and Embodied Question Answering~\cite{das2018embodied} have been proposed to benchmark the performance of multimodal models in these settings. A core challenge lies in aligning words asked in the question with entities in the queried modalities, which typically take the form of visual entities in images or videos (i.e., alignment). \multibench\ contains the following datasets for multimodal question answering spanning several research areas:

\textbf{Affective computing}: \textsc{Social IQ}~\cite{zadeh2019social} is an unconstrained benchmark
specifically designed to train and evaluate socially intelligent AI through a rich source of open-ended questions and answers. It contains $1,250$ videos of natural social situations, $7,500$ questions and $52,500$ correct and incorrect answers

\textbf{Multimedia}: \textsc{CLEVR}~\cite{johnson2017clevr} is a diagnostic dataset for studying the ability of VQA systems to perform visual reasoning. It contains $100,000$ rendered images and about $853,000$ unique automatically generated questions that test visual reasoning abilities such as counting, comparing, logical reasoning, and storing information in memory. \textsc{VQA 2.0}~\cite{goyal2017making} is a balanced version of the popular VQA~\cite{agrawal2017vqa} dataset by collecting complementary images such that
every question is associated with not just a single image, but rather a pair of similar images that result in two different answers to the question. The reduces the occurrence of spurious correlations in the dataset and enables training of more robust models.

\subsection{Retrieval datasets}

Another area of great interest lies in cross-modal retrieval~\cite{liang2021cross,zhen2019deep}, where the goal is to retrieve semantically similar data from a new modality using a modality as a query (e.g., given a phrase, retrieve the closest image describing that phrase). The core challenge is to perform alignment of representations across both modalities. \multibench\ contains the following datasets for multimodal retrieval and grounding:

\textbf{Multimedia}: \textsc{CIFAR-ESC}~\cite{liang2021cross} is an image-audio retrieval dataset constructed by combining CIFAR-$100$, CIFAR-$10$~\cite{krizhevsky2009learning}, and ESC-$50$~\cite{piczak2015esc} into $17$ shared classes using concept ontologies from WordNet~\cite{miller1995wordnet}. \textsc{Clotho}~\cite{drossos2020clotho} is a dataset for audio captioning with $4981$ audio samples of $15$ to $30$ seconds duration and $24,905$ captions of $8$ to $20$ words length. \textsc{Yummly-28K}~\cite{min2016being} contains parallel text descriptions and images of recipes with $27,638$ recipes in total. Each recipe contains one recipe image, the ingredients, the cuisine and the course information. \textsc{Flickr-30k}~\cite{plummer2015flickr30k} contains $32,000$ images collected from Flickr, together with $5$ reference sentences provided by human annotators enabling the tasks of text-to-image reference resolution, localizing textual entity mentions in an image, and bidirectional image-caption retrieval.

\subsection{Reinforcement learning environments}

Learning from multiple modalities in an interactive setting is an area of interest towards building more intelligent embodied agents that can perceive the visual world, language instructions, auditory feedback, and other sensor modalities~\cite{luketina2019survey}. Recent work has also explored audio as a modality in an agent's multisensory interaction with the world~\cite{dean2020see}.
These multimodal problems are fundamentally different from those that are concerned with prediction tasks. Alongside the core challenges in learning complementary information and aligning entities in language instructions to those in the visual environment, there also lies the core challenge of learning \textit{actionable} representations that link to the set of actions that can be taken and their associated long-term rewards~\cite{luketina2019survey}. \multibench\ contains the following datasets for multimodal reinforcement learning in both real-world and simulated environments:

\textbf{Simulated environments}: We choose the \textsc{RTFM}~\cite{zhong2019rtfm} (Reading to Fight Monsters) simulated text and visual environment. \textsc{RTFM} requires an agent to jointly reason over a language goal, a document that specifies environment dynamics, and environment observations. It can also be procedurally generated for increasing difficult interactions between environment dynamics and natural language. \textsc{RTFM} is also part of the larger \textsc{SILG} benchmark~\cite{zhongsilg} of $5$ similar diverse grounded language learning environments under a common interface, so it enables generalization to these other environments as well.

\subsection{Co-learning datasets}

Co-learning aims to transfer knowledge between modalities and their representations. Exemplified by algorithms of fine-tuning, co-training, and contrastive learning, how can knowledge learned from an additional secondary modality (e.g., predicted labels or representation) help a computational model trained on a primary modality? This challenge is particularly relevant when the primary modality has limited resources such as lack of annotated data, noisy input, and unreliable labels.

\textbf{Affective computing}: In affective computing, we investigate transferring information from \textsc{CMU-MOSI} to \textsc{SST}, as well as the larger \textsc{CMU-MOSEI} to \textsc{SST}~\cite{zadeh2020foundations}. The former $2$ are multimodal (language + vision + audio) datasets annotated for sentiment, while \textsc{SST} is a language-only sentiment analysis dataset.

\textbf{Multimedia}: In multimedia, we transfer information from \textsc{GloVe} word embeddings for \textsc{CIFAR10} image classification~\cite{socher2013zero}. We also transfer information from knowledge graphs to image classification by providing the Visual Genome dataset~\cite{krishna2017visual,marino2016more}.

\section{Evaluation protocol}
\label{multibench:eval}

\multibench\ provides standardized evaluation using metrics designed for each dataset, ranging from MSE and MAE for regression to accuracy, micro \& macro F1-score, and AUPRC for classification on each dataset. To assess for generalization, we compute the variance of a particular model's performance across all datasets in \multibench\ on which it is tested. We split these results on multiple datasets into \textit{in-domain} datasets and \textit{out-domain} datasets. \textit{In-domain} datasets refer to model performance on datasets that it was initially proposed and tested on, while \textit{out-domain} datasets refer to model performance on the remaining datasets. Comparing out-domain vs in-domain performance, as well as variance in performance across datasets as a whole, allow us to summarize the generalization statistics of each multimodal model.

\section{\multizoo: A Zoo of Multimodal Algorithms}
\label{multibench:algorithms}

\begin{figure*}[tbp]
\centering
\vspace{-0mm}
\includegraphics[width=\linewidth]{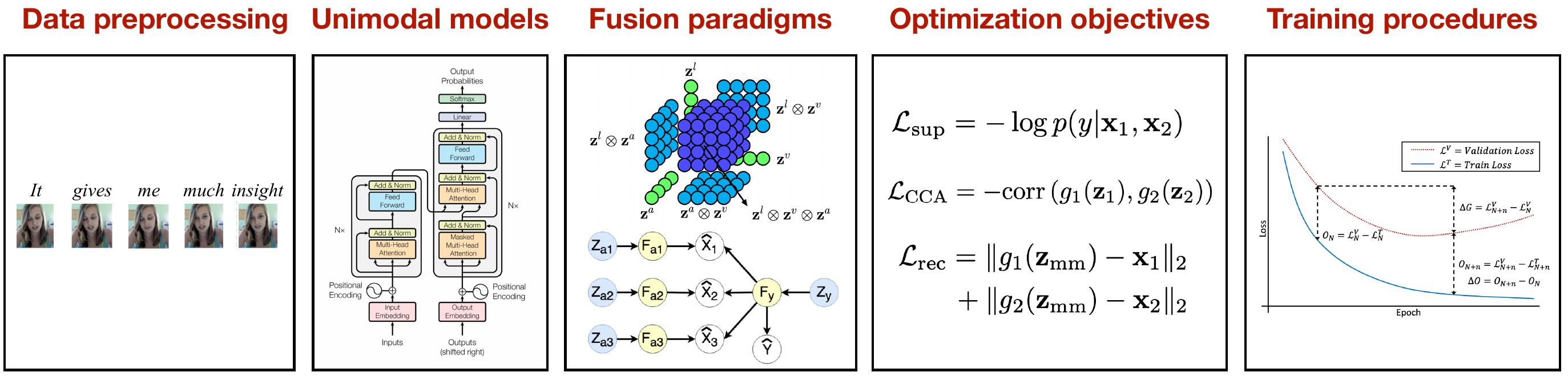}
\vspace{-0mm}
\caption{\multizoo\ provides a standardized implementation of a suite of multimodal methods in a modular fashion to enable accessibility for new researchers, compositionality of approaches, and reproducibility of results.}
\vspace{-0mm}
\label{multibench:figs:multizoo}
\end{figure*}

To complement \multibench, we release a comprehensive toolkit, \multizoo, as starter code for multimodal algorithms which implements $20$ methods spanning different methodological innovations in (1) data preprocessing, (2) fusion paradigms, (3) optimization objectives, and (4) training procedures (see Figure~\ref{multibench:figs:multizoo}). To introduce these algorithms, we use the simple setting with $2$ modalities for notational convenience. We use $\mathbf{x}_1, \mathbf{x}_2$ for input modalities, $\mathbf{z}_1, \mathbf{z}_2$ for unimodal representations, $\mathbf{z}_\textrm{mm}$ for the multimodal representation, and $\hat{y}$ for the predicted label.

\subsection{Data preprocessing}

\textbf{Temporal alignment}~\cite{chen2017multimodal} has been shown to help tackle the multimodal alignment problem for time-series data. This approach assumes a temporal granularity of the modalities (e.g., at the level of words for text) and aligns information from the remaining modalities to the same granularity. We call this approach \textsc{WordAlign}~\cite{chen2017multimodal} for temporal data where text is one of the modalities.

\subsection{Fusion paradigms}
\label{multibench:model_design}

\textbf{Early and late fusion} have been the de-facto first-approach when tackling new multimodal problems. Early fusion performs concatenation at the input data level before using a suitable prediction model (i.e., $\mathbf{z}_\textrm{mm} = \left[ \mathbf{x}_1, \mathbf{x}_2\right]$) and late fusion applies suitable unimodal models to each modality to obtain their feature representations, concatenates these features, and defines a classifier to the label (i.e., $\mathbf{z}_\textrm{mm} = \left[ \mathbf{z}_1, \mathbf{z}_2\right]$)~\citep{baltruvsaitis2018multimodal}. \multizoo\ includes their implementations denoted as \textsc{EF} and \textsc{LF} respectively.

\textbf{Tensors} are specifically designed to tackle the multimodal complementarity challenge by explicitly capturing higher-order interactions across modalities~\cite{zadeh2017tensor}. Given unimodal representations $\mathbf{z}_1, \mathbf{z}_2$, a multimodal tensor representation is defined as $\mathbf{z}_\textrm{mm} = \begin{bmatrix} \mathbf{z}_{1} \\ 1 \end{bmatrix} \otimes \begin{bmatrix} \mathbf{z}_{2} \\ 1 \end{bmatrix}$ where $\otimes$ denotes an outer product. However, computing tensor products is expensive since their dimension scales exponentially with the number of modalities. Several efficient variants have been proposed to approximate expensive full tensor products with cheaper variants while maintaining performance~\cite{hou2019deep,liang2019tensor,liu2018efficient}. \multizoo\ includes Tensor Fusion (\textsc{TF})~\citep{zadeh2017tensor} as well as approximate Low-rank Tensor Fusion (\textsc{LRTF})~\citep{liu2018efficient}.
As future work, we also plan to include more expressive higher-order tensor fusion methods~\cite{hou2019deep}.

\textbf{Multiplicative Interactions (MI)} further generalize tensor products to include learnable parameters that capture the interactions between streams of information~\citep{mipaper}. In its most general form, MI defines a bilinear product $\mathbf{z}_\textrm{mm} = \mathbf{z}_1 \mathbb{W} \mathbf{z}_2 + \mathbf{z}_1^\top \mathbf{U} + \mathbf{V} \mathbf{z}_2 + \mathbf{b}$ where $\mathbb{W}, \mathbf{U}, \mathbf{Z}$, and $\mathbf{b}$ are trainable parameters. By appropriately constraining the rank and structure of these parameters, MI recovers HyperNetworks~\citep{ha2016hypernetworks} (unconstrained parameters resulting in a matrix output), Feature-wise linear modulation (FiLM)~\cite{perez2018film,zhong2019rtfm} (diagonal parameters resulting in vector output), and 
Sigmoid units~\cite{dauphin2017language} (scalar parameters resulting in scalar output). \multizoo\ includes all $3$ as \textsc{MI-Matrix}, \textsc{MI-Vector}, and \textsc{MI-Scalar} respectively.

We also referred to the implementation of Feature-wise linear modulation (FiLM)~\cite{perez2018film} and added it as a module in \multibench, which we call \textsc{FiLM}. While \textsc{MI-Vector} (i.e., diagonal parameters in a MI layer which results in a vector output) corresponds to the most basic implementation of \textsc{FiLM}, the original \textsc{FiLM} layer uses multiple non-linear layers instead of a single linear transformation in \textsc{MI-Vector} which has been shown to improve performance~\cite{perez2018film}.

\textbf{Multimodal gated units} are prevalent in learning combinations of two representations that dynamically change for every input~\cite{chaplot2017gated,wang2020makes,xu2015show}. Its general form can be written as $\mathbf{z}_\textrm{mm} = \mathbf{z}_1 \odot h(\mathbf{z}_2)$, where $h$ represents a function with sigmoid activation and $\odot$ denotes the element-wise product. The output $h(\mathbf{z}_2)$ is commonly referred to as ``attention weights'' learned from $\mathbf{z}_2$ used to attend on $\mathbf{z}_1$.
We implement the Query-Key-Value mechanism as \textsc{NL Gate} as proposed in~\cite{wang2020makes}. This attention mechanism is conceptually similar to the \textsc{MI-Vector} case above but recent work has explored more expressive forms of $h$ such as using a Query-Key-Value mechanism~\cite{wang2020makes} or several fully-connected layers~\citep{chaplot2017gated} rather than a linear transformation in \textsc{MI-Vector}.

\textbf{Multimodal transformers} are useful in tackling the challenge of multimodal alignment and complementarity. Transformer models~\citep{vaswani2017attention} have been shown to be useful for temporal multimodal data by automatically aligning and capturing complementary features at different time-steps~\cite{tsai2019multimodal,yao2020multimodal}. We include the Multimodal Transformer (\textsc{MulT})~\citep{tsai2019multimodal} which uses a Crossmodal Transformer block that uses $\mathbf{z}_1$ to attend to $\mathbf{z}_2$ (and vice-versa), before concatenating both representations to obtain $\mathbf{z}_\textrm{mm} = \left[ \mathbf{z}_{1 \rightarrow 2}, \mathbf{z}_{2 \rightarrow 1} \right] = \left[ \textsc{CM}(\mathbf{z}_1, \mathbf{z}_2), \textsc{CM}(\mathbf{z}_2, \mathbf{z}_1) \right]$.

To extend this to $3$ modalities, the crossmodal transformer block is repeated across all $3$ sets of modality pairs (i.e., $\mathbf{z}_\textrm{mm} = \left[ \mathbf{z}_{1 \rightarrow 2}, \mathbf{z}_{2 \rightarrow 1}, \mathbf{z}_{1 \rightarrow 3}, \mathbf{z}_{3 \rightarrow 1}, \mathbf{z}_{2 \rightarrow 3}, \mathbf{z}_{3 \rightarrow 2} \right]$). While this is still computationally feasible for $3$ modalities such as the language, video, and audio datasets that \textsc{MulT} was originally designed for, this quickly becomes intractable for problems involving more than $3$ modalities. To adapt \textsc{MulT} for the financial prediction task involving more than $10$ modalities, we cluster all modalities into $3$ groups based on similarities in their data and perform early fusion on the data within each cluster before applying \textsc{MulT} only on the $3$ clusters of modalities. While \textsc{MulT} is a strong model based on performance, it may pose scalability issues since the number of cross-modal attention blocks grows quadratically with the number of modalities.

\textbf{Architecture search}: Finally, instead of hand-designing multimodal architectures, several approaches define a set of atomic neural operations (e.g., linear transformation, activation, attention, etc.) and use architecture search to automatically learn the best order of these operations for a given multimodal task~\cite{perez2019mfas,xu2021mufasa}. We focus on the more general approach, \textsc{MFAS}~\citep{perez2019mfas}, designed for language and vision datasets.
While this approach is categorized under innovations in model architecture (since it primarily targets better architectures for multimodal fusion), its code in the \multizoo\ toolkit is implemented under training structures, since architecture search requires an outer loop to learn model architectures over multiple inner supervised learning loops that train an individual model architecture. Therefore, we are unable to integrate \textsc{MFAS} directly with the basic supervised learning training loops like we do for the other fusion paradigms described above.

\subsection{Optimization objectives}

In addition to the standard supervised losses (e.g., cross entropy for classification, MSE/MAE for regression), several proposed methods have proposed new objective functions based on:

\textbf{Prediction-level alignment}: There has been extensive research in defining objective functions to tackle the challenge of multimodal alignment: capturing a representation space where semantically similar concepts from different modalities are close together. While primarily useful for cross-modal retrieval~\cite{liang2021cross,zhen2019deep}, recent work has also shown its utility in learning representations for prediction~\cite{bachman2019learning,cui2020unsupervised,lee2019making,tian2020contrastive}. These alignment objectives have been applied at both prediction and feature levels. In the former, we implement Canonical Correlation Analysis (\textsc{CCA})~\cite{andrew2013deep,wang2015deep}, which computes $\mathcal{L}_\textrm{CCA} = \textrm{corr} \left( g_1(\mathbf{z}_{1}), g_2(\mathbf{z}_{2}) \right)$ where $g_1,g_2$ are auxiliary classifiers mapping each unimodal representation to the label. This method corresponds to prediction-level alignment since they aim to learn representations of each modality that agree on the label, as measured by the correlation of label predictions made by each modality across a batch of samples.
We refer to the paper that most closely implements CCA-based alignment for multimodal data (specifically directly testing on the CMU-MOSI dataset)~\citep{sun2020learning}.

\textbf{Feature-level alignment}: In the latter, contrastive learning has emerged as a popular approach that brings similar concepts close in feature space and different concepts far away~\cite{cui2020unsupervised,lee2019making,tian2020contrastive}. \multizoo\ includes \textsc{ReFNet}~\cite{sankaran2021multimodal} which includes a self-supervised contrastive loss between unimodal representations $\mathbf{z}_{1}, \mathbf{z}_{2}$ and the multimodal representation $\mathbf{z}_\textrm{mm}$, i.e., $\mathcal{L}_\textrm{contrast} = 1 - \textrm{cos} (\mathbf{z}_\textrm{mm}, g_1(\mathbf{z}_{1})) + 1 - \textrm{cos} (\mathbf{z}_\textrm{mm}, g_2(\mathbf{z}_{2})) $ where $g_1,g_2$ is an auxiliary layer mapping each modality's representation into the joint multimodal space. The intuition here is that the unimodal representations $\mathbf{z}_{1}, \mathbf{z}_{2}$ and the multimodal representation $\mathbf{z}_\textrm{mm}$ should be aligned in the multimodal feature space as measured by cosine similarity. While the original \textsc{ReFNet} method does not use negative samples, closely related work in multi-view contrastive learning has extended this idea to use negative samples which is more closely in line with recent work in contrastive learning~\cite{tian2020contrastive}.

\textbf{Reconstruction objectives}: Methods based on generative-discriminative models (e.g., VAEs) include an objective to reconstruct the input (or some part of the input)~\cite{lee2019making,tsai2019learning}. These have been shown to better preserve task-relevant information learned in the representation, especially in settings with sparse supervised signals such as robotics~\cite{lee2019making} and long videos~\cite{tsai2019learning}. We include the Multimodal Factorized Model (\textsc{MFM})~\citep{tsai2019learning} which is a general approach that learns a representation $\mathbf{z}_\textrm{mm}$ that can reconstruct input data $\mathbf{x}_{1}, \mathbf{x}_{2}$ while also predicting the label. The multimodal representation is a concatenation of factorized representations $\mathbf{z}_1$, $\mathbf{z}_2$, ..., $\mathbf{z}_M$, and $\mathbf{z}_y$.

Since \textsc{MFM} optimizes a variational lower-bound to the log likelihood, the overall objective consists of $3$ terms - generative, discriminative, and prior regularization:
\begin{equation}
\label{eq:approxmulti}
\begin{split}
&\underset{f_i, f_\textrm{mm}, g_i, g_y}{\mathrm{min}}\,\,\mathbf{E}_{{P}_{\mathbf{x}_{1:M},\mathbf{y}}}\mathbf{E}_{{f_1}({\mathbf{z}_{1}|\mathbf{x}_{1}})} \cdots \mathbf{E}_{{f_M}({\mathbf{z}_{M}| \mathbf{x}_{M}})} \mathbf{E}_{{f_{\textrm{mm}}}({\mathbf{z}_y|\mathbf{x}_{1:M}})}\\
&\left[\sum_{i=1}^M \norm{ \mathbf{x}_i, g_{i} (\mathbf{z}_{i}, \mathbf{z_y}) }_2 + \ell \left(\mathbf{y}, g_y( \mathbf{z_y}) \right) \right]  + \lambda \textrm{MMD}({Q}_{\mathbf{z}}, {P}_{\mathbf{z}}),
\end{split}
\end{equation}
where $f_i$'s are encoders from each modality to representations, $f_\textrm{mm}$ is a multimodal encoder to the joint representation $\mathbf{z}_y$, $g_i$'s are decoders from latent representations back into input data, and $g_y$ is a classification head to the label. The final $\textrm{MMD}$ term is a regularizer to bring the representations close to a unit Gaussian prior. The multimodal encoder $f_\textrm{mm}$ in \textsc{MFM} can be instantiated with any multimodal model (e.g., learning $\mathbf{z}_y$ via tensors and adding a term to reconstruct input data). We use the public implementation in \url{https://github.com/pliang279/factorized}, which uses a temporal attention model as $f_\textrm{mm}$ for multimodal time-series data. For the remaining experiments we replace $f_\textrm{mm}$ with a simple late fusion but also run some experiments with multimodal methods that are state-of-the-art in each domain.

\newlength{\textfloatsepsave} \setlength{\textfloatsepsave}{\textfloatsep} \setlength{\textfloatsep}{0pt}

\begin{algorithm}[tb]
    \caption{PyTorch code integrating \multibench\ datasets and \multizoo\ models.}
    \label{multibench:alg:code}
   
    \definecolor{codeblue}{rgb}{0.25,0.5,0.5}
    \lstset{
      basicstyle=\fontsize{7.2pt}{7.2pt}\ttfamily\bfseries,
      commentstyle=\fontsize{7.2pt}{7.2pt}\color{codeblue},
      keywordstyle=\fontsize{7.2pt}{7.2pt},
    }
\begin{lstlisting}[language=python]
from datasets.get_data import get_dataloader
from unimodals.common_models import ResNet, Transformer
from fusions.common_fusions import MultInteractions
from training_structures.gradient_blend import train, test

# loading Multimodal IMDB dataset
traindata, validdata, testdata = get_dataloader('multimodal_imdb')
out_channels = 3
# defining ResNet and Transformer unimodal encoders
encoders = [ResNet(in_channels=1, out_channels, layers=5),
            Transformer(in_channels=1, out_channels, layers=3)]
# defining a Multiplicative Interactions fusion layer
fusion = MultInteractions([out_channels*8, out_channels*32], out_channels*32, 'matrix')
classifier = MLP(out_channels*32, 100, labels=23)
# training using Gradient Blend algorithm
model = train(encoders, fusion, classifier, traindata, validdata, 
        epochs=100, optimtype=torch.optim.SGD, lr=0.01, weight_decay=0.0001)
# testing
performance, complexity, robustness = test(model, testdata)
\end{lstlisting}
\end{algorithm}

\setlength{\textfloatsep}{\textfloatsepsave}

\textbf{Improving robustness}: These approaches modify the objective function to account for robustness to noisy~\cite{liang2019tensor} or missing~\cite{lee2020detect,ma2021smil,pham2019found} modalities. \multizoo\ includes \textsc{MCTN}~\cite{pham2019found} which uses cycle-consistent translation to predict the noisy/missing modality from present ones. The key insight is that a joint representation between modalities $\mathbf{x}_{1}$ and $\mathbf{x}_{2}$ can be learned by using $\mathbf{x}_{1}$ to predict $\mathbf{x}_{2}$, in a vein similar to machine translation or image/text style transfer. \textsc{MCTN} defines a cyclic translation path $\mathbf{x}_{1} \rightarrow \mathbf{z}_\textrm{mm} \rightarrow \hat{\mathbf{x}}_{2} \rightarrow \mathbf{z}_\textrm{mm} \rightarrow \hat{\mathbf{x}}_{1}$ and adds additional reconstruction losses $\mathcal{L}_\textrm{rec} = \norm{ \mathbf{x}_{1} - \hat{\mathbf{x}}_{1} }_2 + \norm{ \mathbf{x}_{2} - \hat{\mathbf{x}}_{2} }_2$ on top of the supervised learning loss. The representations $\mathbf{z}_\textrm{mm}$ learned via translation are then used to predict the label. Surprisingly, the model needs to take in only $\mathbf{x}_{1}$ at test time and is therefore robust to all levels of noise or missingness in $\mathbf{x}_{2}$.

\subsection{Training procedures}

\textbf{Improving generalization}: Recent work has found that directly training a multimodal model with all modalities using supervised learning is sub-optimal since different modalities overfit and generalize at different rates. \multizoo\ includes an approach to solve this, called Gradient Blending (\textsc{GradBlend}), that computes generalization statistics for each modality to determine their weights during multimodal fusion~\citep{wang2020makes}.
We also include a similar work, Regularization by Maximizing Functional Entropies (\textsc{RMFE}), which uses functional entropy to balance the contribution of each modality to the classification result~\cite{gat2020removing}.

\subsection{Putting everything together}

In Algorithm~\ref{multibench:alg:code}, we show a sample code snippet in Python that loads a dataset from \multibench, defines the unimodal and multimodal architectures, optimization objective, and training procedures, before running the evaluation protocol. Our \multizoo\ toolkit is easy to use and trains entire multimodal models in less than $10$ lines of code. By standardizing the implementation of each module and disentangling the individual effects of models, optimizations, and training, \multizoo\ ensures both accessibility and reproducibility of its algorithms.

\section{Experiments and Discussion}

\textbf{Setup}: Using \multibench, we load each of the datasets and test the multimodal approaches in \multizoo. We only vary the contributed method of interest and keep all other possibly confounding factors constant (i.e., using the exact same training loop when testing a new multimodal fusion paradigm), a practice unfortunately not consistent in previous work. Our code is available at \url{https://github.com/pliang279/MultiBench}. \multibench\ allows for careful analysis of multimodal models and we summarize the main take-away messages below.

\begin{table*}[]
\fontsize{9}{11}\selectfont
\setlength\tabcolsep{2.0pt}
\caption{\textbf{Standardizing methods and datasets} enables quick application of methods from different research areas which achieves stronger performance on $9/15$ datasets in \multibench, especially in healthcare, HCI, robotics, and finance. \textit{In-domain} refers to the best performance across methods previously proposed on that dataset and \textit{out-domain} shows best performance across remaining methods. $\uparrow$ indicates metrics where higher is better (Acc, AUPRC), $\downarrow$ indicates lower is better (MSE).}
\centering
\footnotesize
\vspace{-0mm}
\begin{tabular}{l|c|c|c|c|c}
\hline \hline
Dataset & \textsc{MUStARD} $\uparrow$ & \textsc{CMU-MOSI} $\uparrow$ & \textsc{UR-FUNNY} $\uparrow$ & \textsc{CMU-MOSEI} $\uparrow$ & \textsc{MIMIC} $\uparrow$ \\
\hline
Unimodal & $68.6 \pm 0.4 $ & $74.2 \pm 0.5$ & $58.3 \pm 0.2$ & $78.8 \pm 1.5$ & $76.7 \pm 0.3$ \\
\hline
In-domain & $66.3 \pm 0.3$ & $\mathbf{83.0 \pm 0.1}$ & $62.9 \pm 0.2$ & $\mathbf{82.1 \pm 0.5}$ & $77.9 \pm 0.3$ \\
Out-domain & $\mathbf{71.8 \pm 0.3}$ & $75.5 \pm 0.5$ & $\mathbf{66.7 \pm 0.3}$ & $ 78.1 \pm 0.3$ & $\mathbf{78.2 \pm 0.2}$ \\
Improvement & \textcolor{gg}{$\mathbf{4.7\%}$} & - & \textcolor{gg}{$\mathbf{6.0\%}$} & - & \textcolor{gg}{$\mathbf{0.4\%}$} \\
\hline \hline
\end{tabular}

\vspace{4mm}

\begin{tabular}{l|c|c|c|c|c}
\hline \hline
Dataset & \textsc{MuJoCo Push} $\downarrow$ & \textsc{V\&T EE} $\downarrow$ & \textsc{Stocks-F\&B} $\downarrow$ & \textsc{Stocks-Health} $\downarrow$ & \textsc{Stocks-Tech} $\downarrow$ \\
\hline
Unimodal & $0.334 \pm 0.034$ & $0.202 \pm 0.022$ & $1.856 \pm 0.093$ & $0.541 \pm 0.010$ & $0.125 \pm 0.004$ \\
\hline
In-domain & $\mathbf{0.290 \pm 0.018}$ & $0.258 \pm 0.011$ & $1.856 \pm 0.093$ & $0.541 \pm 0.010$ & $0.125 \pm 0.004$ \\
Out-domain & $0.402 \pm 0.026$ & $\mathbf{0.185 \pm 0.011}$ & $\mathbf{1.820 \pm 0.138}$ & $\mathbf{0.526 \pm 0.017}$ & $\mathbf{0.120 \pm 0.008}$ \\
Improvement & - & \textcolor{gg}{$\mathbf{8.4\%}$} & \textcolor{gg}{$\mathbf{1.9\%}$} & \textcolor{gg}{$\mathbf{2.8\%}$} & \textcolor{gg}{$\mathbf{4.0\%}$}\\
\hline \hline
\end{tabular}

\vspace{4mm}

\begin{tabular}{l|c|c|c|c|c}
\hline \hline
Dataset & \textsc{ENRICO} $\uparrow$ & \textsc{MM-IMDb} $\uparrow$ & \textsc{AV-MNIST} $\uparrow$ & \textsc{Kinetics-S} $\uparrow$ & \textsc{Kinetics-L} $\uparrow$ \\
\hline
Unimodal & $47.0 \pm 1.6$ & $45.6 \pm 4.5$ & $65.1 \pm 0.2$ & $\mathbf{56.5}$ & $72.6$ \\
\hline
In-domain & $47.0 \pm 1.6$ & $49.8 \pm 1.7$ & $\mathbf{72.8 \pm 0.2}$ & $56.1$ & $\mathbf{74.7}$ \\
Out-domain & $\mathbf{51.0 \pm 1.4}$ & $\mathbf{50.2 \pm 0.9}$ & $72.3 \pm 0.2$ & $23.7$ & $71.7$ \\
Improvement & \textcolor{gg}{$\mathbf{8.5\%}$} & \textcolor{gg}{$\mathbf{0.8\%}$} & - & - & - \\
\hline \hline
\end{tabular}

\vspace{-4mm}
\label{multibench:results:overall}
\end{table*}

\subsection{Benefits of standardization}

From Table~\ref{multibench:results:overall}, simply applying methods in a research different area achieves state-of-the-art performance on $9$ out of the $15$ fusion tasks. We find that this is especially true for domains and modalities that have been relatively less studied in multimodal research (i.e., healthcare, finance, HCI). Performance gains can be obtained using multimodal methods \textit{outside} of that research area. Therefore, this motivates the benefits of standardizing and unifying areas of research in multimodal machine learning. We believe that the ever-expanding diversity of datasets in \multibench\ can greatly accelerate research in multimodal learning.

\subsection{Generalization across domains and modalities}

\begin{figure}
\centering
\includegraphics[width=\textwidth]{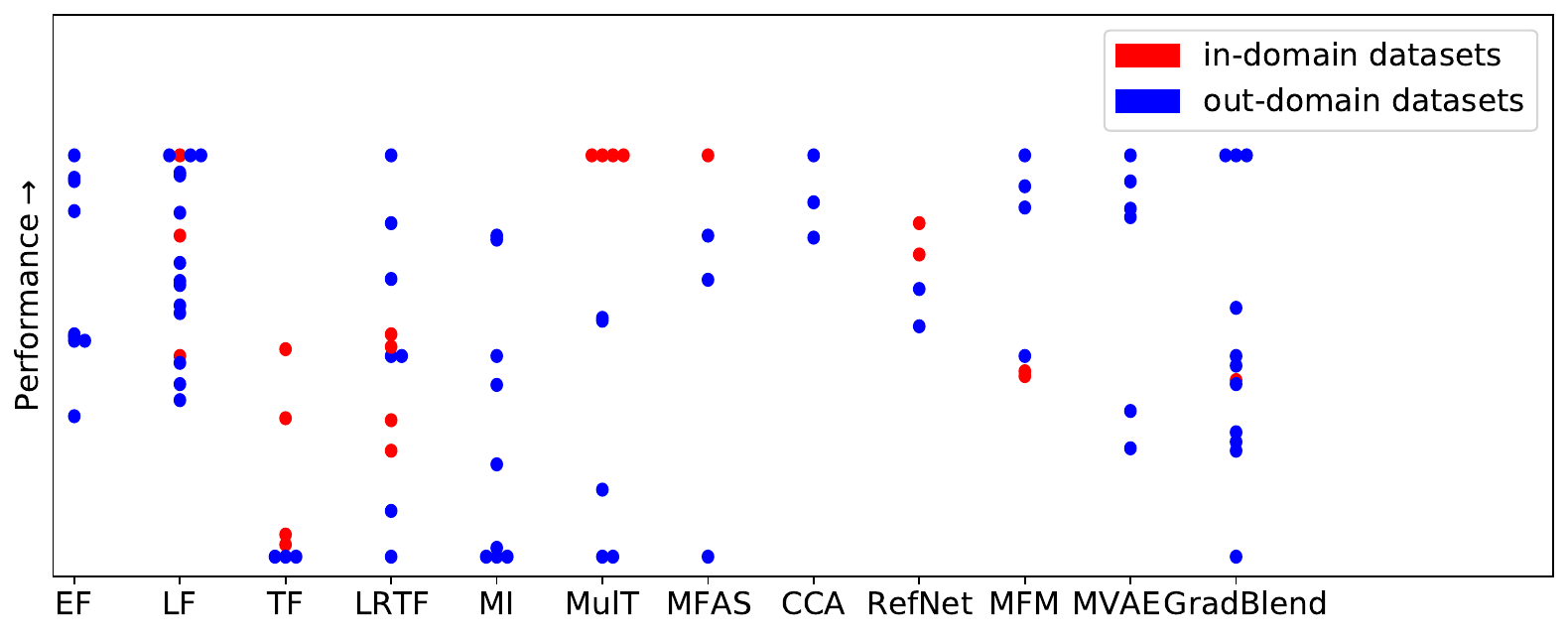}
\caption{Relative performance of each model across in-domain ({\color{red}{red}} dots) and out-domain datasets ({\color{blue}{blue}} dots). \textit{In-domain} refers to the performance on datasets that the method was previously proposed for and \textit{out-domain} shows performance on the remaining datasets. We find that many methods show strongest performance on in-domain datasets which drop when tested on different domains, modalities, and tasks. In general, we also observe high variance in the performance of multimodal methods across datasets in \multibench, which suggest open questions in building more generalizable models.}
\vspace{-0mm}
\label{multibench:fig:perfonly1}
\end{figure}

\begin{figure}
\centering
\includegraphics[width=\textwidth]{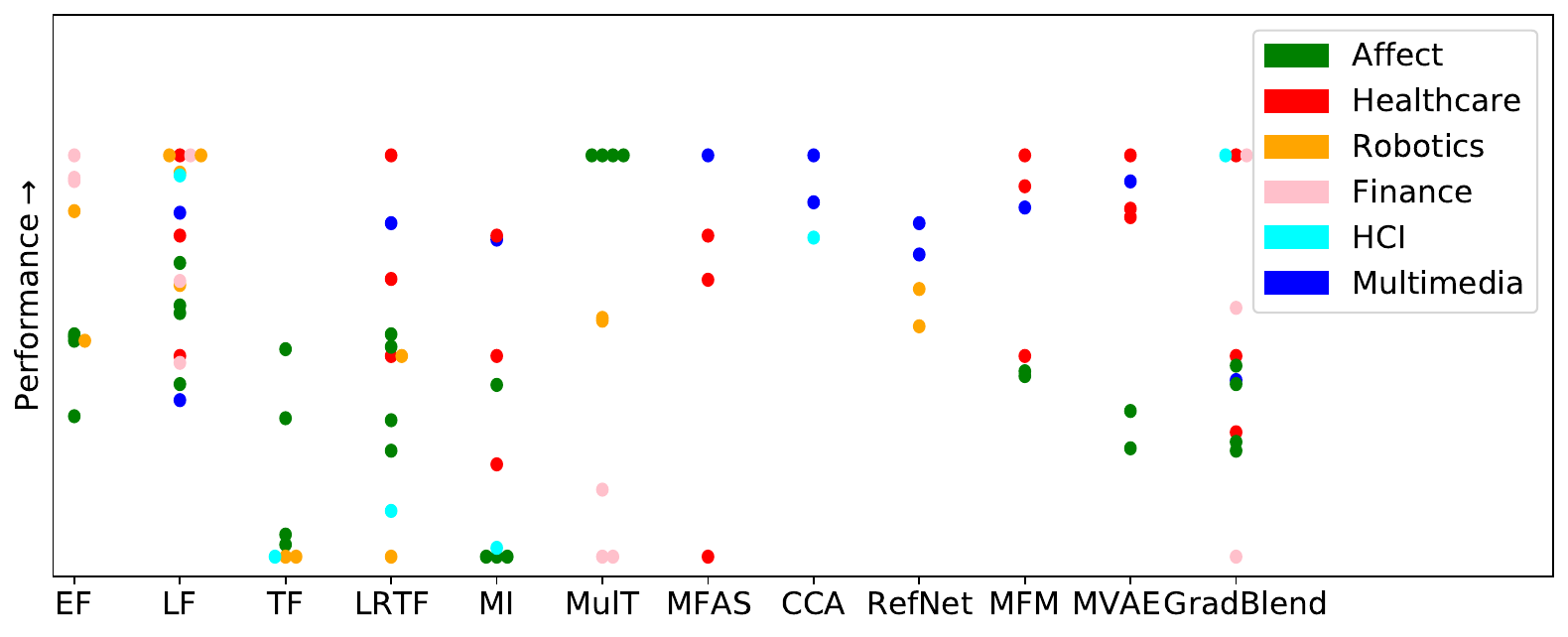}
\caption{Relative performance of each model across different domains. We find that the performance of multimodal models varies significantly across datasets spanning different research areas and modalities. Similarly, the best-performing methods on each domain are also different. Therefore, there still does not exist a one-size-fits-all model, especially for understudied modalities and tasks.}
\vspace{-0mm}
\label{multibench:fig:perfonly2}
\end{figure}

\multibench\ offers an opportunity to analyze algorithmic developments across a large suite of modalities, domains, and tasks. We illustrate these observations through $2$ summary plots of the generalization performance of multimodal models. Firstly, in Figure~\ref{multibench:fig:perfonly1}, we plot the performance of each multimodal method across all datasets that it is tested on, using the color {\color{red}{red}} to indicate performance on datasets that it was initially proposed and tested on (which we label as \textit{in-domain}), and {\color{blue}{blue}} to indicate its performance on the remaining datasets (which we label as \textit{out-domain}). Secondly, in Figure~\ref{multibench:fig:perfonly2}, we color-code the performance on each dataset depending on which research area the dataset belongs to (one of the $6$ research areas covered in \multibench).

We summarize several observations regarding generalization across modalities and tasks:
\begin{enumerate}
    \item Many multimodal methods still do not generalize across domains and datasets. For examples, \textsc{MFAS}~\citep{perez2019mfas} works well on domains it was designed for (\textsc{AV-MNIST} and \textsc{MM-IMDb} in the multimedia domain), but does not generalize to other domains such as healthcare (\textsc{MIMIC}). Similarly, the method designed for robustness, \textsc{MCTN}~\citep{pham2019found}, does not generalize to datasets within the affective computing domain (\textsc{UR-FUNNY} and \textsc{MUStARD}). Finally, \textsc{GradBlend}~\citep{wang2020makes}, an approach specifically designed to improve generalization in multimodal learning and tested on video and audio datasets (e.g., Kinetics), does not perform well on other datasets. Therefore, there still does not exist a one-size-fits-all model, especially on understudied modalities and tasks.
    \item From Figure~\ref{multibench:fig:perfonly1}, many methods show strongest performance on in-domain datasets, and their performance drops when tested on different domains, modalities, and tasks. \textsc{MulT} performs extremely well on the affect recognition datasets it was designed for but struggles on other multimodal time-series in the finance and robotics domains. On the other hand, \textsc{MFM} shows an impressive performance in generalizing to new domains, although its in-domain performance has been exceeded by several other methods.
    \item From Figure~\ref{multibench:fig:perfonly1}, there is high variance in multimodal performance across datasets in \multibench, which suggest open questions in building more generalizable models. We find that \textsc{LF} is quite stable and always achieves above-average performance.
    \item There are methods that are quite generalizable - typically general modality-agnostic methods such as \textsc{LF}. While simple, it is a strong method that balances simplicity, performance, and low complexity. However, it does not achieve the best performance on any dataset, which suggests that it is a good starting point but perhaps not the best eventual method.
    \item From Figure~\ref{multibench:fig:perfonly2}, we find that performance also varies significantly across research areas.
    \item Several methods such as \textsc{MFAS} and \textsc{CCA} are designed for only $2$ modalities (usually image and text), and \textsc{TF} and \textsc{MI} do not scale efficiently beyond $2/3$ modalities. Therefore, we were unable to directly adapt these approaches to other datasets. We encourage the community to generalize these approaches across datasets and modalities on \multibench.
\end{enumerate}

\begin{figure*}[tbp]
\centering
    \vspace{-6mm}
    \begin{minipage}{0.4\textwidth}
        \centering
        \subfloat[\centering All datasets]{{\includegraphics[width=\textwidth]{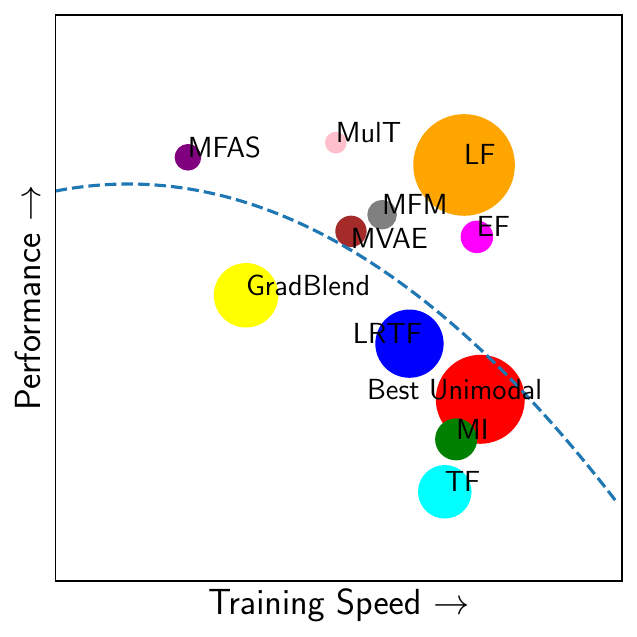}}}
    \end{minipage}%
    \begin{minipage}{0.4\textwidth}
        \centering
        \subfloat[\centering Datasets with $>6$ approaches]{{\includegraphics[width=\textwidth]{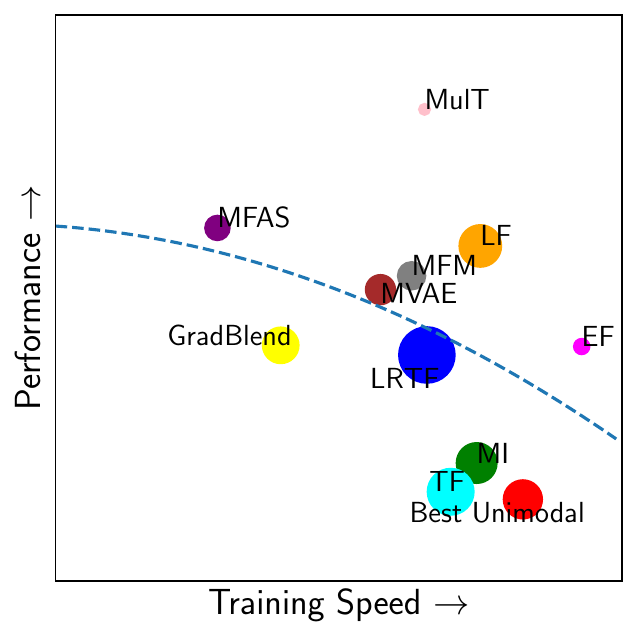}}}
    \end{minipage}
\caption{\textbf{Tradeoff between performance and complexity}. Size of circles shows variance in performance across (a) all datasets and (b) datasets on which we tested $>6$ approaches. We plot a dotted {\color{blue}{blue line}} of best quadratic fit to show the Pareto frontier. These strong tradeoffs should encourage future work in lightweight multimodal models that generalize across datasets, as well as in adapting several possibly well-performing methods (such as \textsc{MFAS} or \textsc{MulT}) to new datasets and domains.}
\label{multibench:figs:tradeoff_complexity}
\end{figure*}

\subsection{Tradeoffs between modalities}

How far can we go with unimodal methods? Surprisingly far! From Table~\ref{multibench:results:overall}, we observe that decent performance can be obtained with the best-performing modality. Further improvement via multimodal models may come at the expense of around $2-3\times$ the parameters.

\subsection{Tradeoffs between performance and complexity}

\begin{figure*}[tbp]
\centering
    \vspace{-6mm}
    \begin{minipage}{0.5\textwidth}
        \centering
        \subfloat[\centering Relative robustness]{{\includegraphics[width=\textwidth]{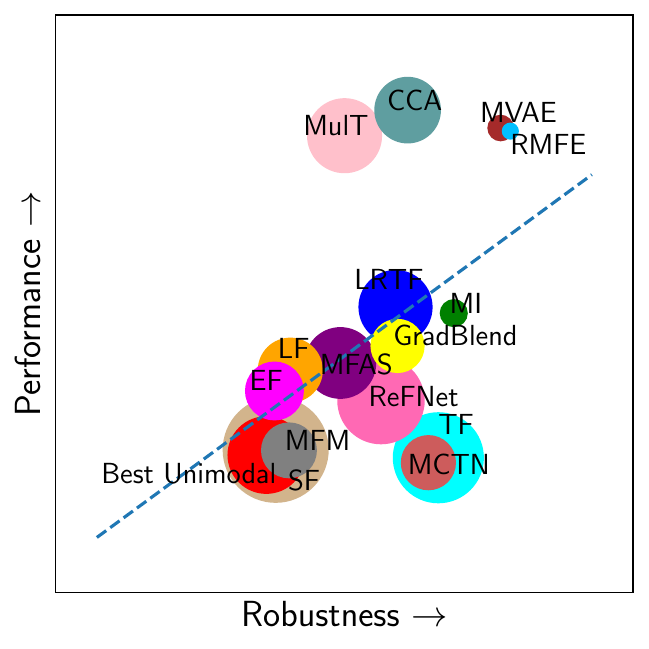}}}
    \end{minipage}%
    \begin{minipage}{0.5\textwidth}
        \centering
        \subfloat[\centering Effective robustness]{{\includegraphics[width=\textwidth]{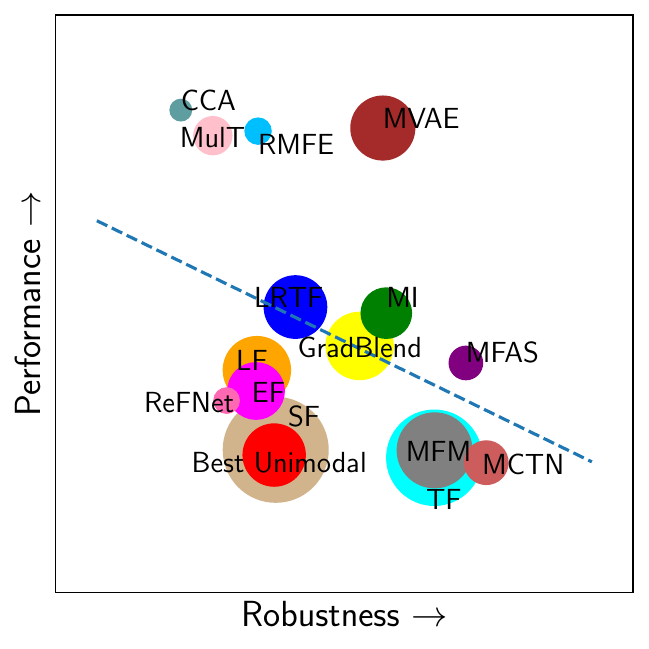}}}
    \end{minipage}
\caption{\textbf{Tradeoff between performance and robustness}. Size of circles shows variance in robustness across datasets. We show the line of best linear fit in dotted {\color{blue}{blue}}. While better performing methods show better \textit{relative} robustness (a), some suffer in \textit{effective} robustness since performance \textit{drops off faster} (b). Few models currently achieve both relative and effective robustness, suggesting directions for future research.\vspace{-4mm}}
\label{multibench:figs:tradeoff_robustness}
\end{figure*}

In Figure~\ref{multibench:figs:tradeoff_complexity}(a), we summarize the performance of all methods in terms of performance and complexity. We find a strong tradeoff between these two desiderata: simple fusion techniques (e.g., \textsc{LF}) are actually appealing choices which score high on both metrics, especially when compared to complex (but slightly better performing) methods such as architecture search (\textsc{MFAS}) or Multimodal Transformers (\textsc{MulT}). While \textsc{LF} is the easiest to adapt to new datasets and domains, we encountered difficulties in adapting several possibly well-performing methods (such as \textsc{MFAS} or \textsc{MulT}) to new datasets and domains. Therefore, while their average performance is only slightly better than \textsc{LF} on all datasets (see Figure~\ref{multibench:figs:tradeoff_complexity}(a)), they perform much better on well-studied datasets (see Figure~\ref{multibench:figs:tradeoff_complexity}(b)). We hope that the release of \names\ will greatly accelerate research in adapting complex methods on new datasets.

\subsection{Tradeoffs between performance and robustness}

In Figure~\ref{multibench:figs:tradeoff_robustness}, we plot a similar tradeoff plot between accuracy and (relative \& effective) robustness. As a reminder, relative robustness directly measures accuracy under imperfections while effective robustness measures the rate at which accuracy drops after equalizing for initial accuracy on clean test data. We observe a positive correlation between performance and relative robustness (see Figure~\ref{multibench:figs:tradeoff_robustness}(a)), implying that models starting off with higher accuracy tend to stay above other models on the performance-imperfection curve. However, we observe a negative best fit between performance and effective robustness (see Figure~\ref{multibench:figs:tradeoff_robustness}(b)) because several well-performing methods such as \textsc{MulT}, \textsc{CCA}, and \textsc{MVAE} tend to \textit{drop off faster} after equalizing for initial accuracy on clean test data. Furthermore, very few models currently achieve both positive relative and effective robustness, which is a crucial area for future multimodal research.

\section{Related Work}

We review related work on standardizing datasets and methods in multimodal learning.

\textbf{Comparisons with related benchmarks}: To the best of our knowledge, \multibench\ is the first multimodal benchmark with such a large number of datasets, modalities, and tasks. Most previous multimodal benchmarks have focused on a single research area such as within affective computing~\citep{gkoumas2021makes}, human multimodal language~\citep{liang2018computational}, language and vision-based question answering~\citep{ferraro-etal-2015-survey,sharif2020vision}, text classification with external multimodal information~\citep{multimodaltoolkit}, and multimodal learning for education~\citep{datacollection2021}. \multibench\ is specifically designed to go beyond the commonly studied language, vision, and audio modalities to encourage the research community to explore relatively understudied modalities (e.g., tabular data, time-series, sensors, graph and set data) and build general multimodal methods that can handle a diverse set of modalities.

Our work is also inspired by recent progress in better evaluation benchmarks for a suite of important tasks in ML such as language representation learning~\citep{wang2019superglue,wang2018glue}, long-range sequence modeling~\citep{tay2021long}, multilingual representation learning~\citep{hu2020xtreme}, graph representation learning~\citep{hu2020open}, and robustness to distribution shift~\citep{koh2020wilds}. These well-crafted benchmarks have accelerated progress in new algorithms, evaluation, and analysis in their respective research areas.

\textbf{Standardizing multimodal learning}: There have also been several attempts to build a single model that works well on a suite of multimodal tasks~\citep{li2019visualbert,lu2019vilbert,Su2020VLBERT}. However, these are limited to the language and vision space, and multimodal training is highly tailored for text and images. Transformer architectures have emerged as a popular choice due to their suitability for both language and image data~\citep{chen2020uniter,hu2021transformer} and a recent public toolkit was released for incorporating multimodal data on top of text-based Transformers for prediction tasks~\citep{multimodaltoolkit}. By going beyond Transformers and text data, \multibench\ opens the door to important research questions involving a much more diverse set of modalities and tasks while holistically evaluating performance, complexity, and robustness.

\textbf{Analysis of multimodal representations}: Recent work has carefully analyzed and challenged long-standing assumptions in multimodal learning. They have shown that certain models do not actually learn cross-modal interactions but rather rely on ensembles of unimodal statistics~\citep{hessel2020does} and that certain datasets and models are biased to the most dominant modality~\cite{cadene2019rubi,goyal2017making}, sometimes ignoring others completely~\citep{agrawal2016analyzing}. These observations are currently only conducted on specific datasets and models without testing their generalization to others, a shortcoming we hope to solve using \multibench\ which enables scalable analysis over modalities, tasks, and models.

\section{Conclusion}

\textbf{Limitations}: While \multibench\ can help to accelerate research in multimodal ML, we are aware of the following possible limitations:

1. \textit{Tradeoffs between generality and specificity}: While it is desirable to build models that work across modalities and tasks, there is undoubtedly merit in building modality and task-specific models that can often utilize domain knowledge to improve performance and interpretability (e.g., see neuro-symbolic VQA~\citep{vedantam2019probabilistic}, or syntax models for the language modality~\citep{cirik2018using}). By easing access to data, models, and evaluation, we hope that \multibench\ will challenge researchers to design interpretable models leveraging domain knowledge for many multimodal tasks. It remains an open question to define ``interpretability'' for other modalities beyond image and text, a question we hope \multibench\ will drive research in.

2. \textit{Scale of datasets, models, and metrics}: We plan for \multibench\ to be a continuously-growing community effort with regular maintenance and expansion. While \multibench\ currently does not include several important research areas outside of multimodal fusion (e.g., question answering~\citep{agrawal2017vqa,hannan2020manymodalqa}, retrieval~\citep{zhen2019deep}, grounding~\citep{cirik2018visual}, and reinforcement learning~\citep{luketina2019survey}), and is also limited by the models and metrics it supports, we have plans to expand \multibench\ towards a wider scale of datasets, models, and metrics.

\textbf{Projected expansions of \multibench}: In this subsection, we describe concrete ongoing and future work towards expanding \multibench:

1. \textit{Other multimodal research problems}: We are genuinely committed to building a community around these resources and continue improving it over time. While we chose to focus on multimodal fusion by design for this first version to have a more coherent way to standardize and evaluate methods across datasets, we acknowledge the breadth of multimodal learning and are looking forward to expanding it in other directions in collaboration with domain experts. We have already included $2$ datasets in captioning (and more generally for non-language outputs, retrieval): (1) Yummly-28K of paired videos and text descriptions of food recipes~\citep{min2016being}, and (2) Clotho dataset for audio-captioning~\citep{drossos2020clotho} as well as a language-guided RL environment Read to Fight Monsters (RTFM)~\citep{zhong2019rtfm} and are also working towards more datasets in QA, retrieval, and multimodal RL.

To help in scalable expansion, we plan for an open call to the community for suggestions and feedback about domains, datasets, and metrics. As a step in this direction, we have concrete plans to use \multibench\ as a theme for future workshops and competitions (building on top of the multimodal workshops we have been organizing at \href[pdfnewwindow=true]{http://multicomp.cs.cmu.edu/naacl2021multimodalworkshop}{NAACL 2021}, \href[pdfnewwindow=true]{http://multicomp.cs.cmu.edu/acl2020multimodalworkshop}{ACL 2020}, and \href[pdfnewwindow=true]{http://multicomp.cs.cmu.edu/acl2018multimodalchallenge}{ACL 2019}, and in multimodal learning courses (starting with the \href[pdfnewwindow=true]{https://cmu-multicomp-lab.github.io/mmml-course/fall2020}{course taught annually at CMU}). Since \multibench\ is public and will be regularly maintained, the existing benchmark, code, evaluation, and experimental protocols can greatly accelerate any dataset and modeling innovations added in the future. In our public GitHub, we have included a section on contributing through task proposals or additions of datasets and algorithms. The authors will regularly monitor new proposals through this channel.

2. \textit{New evaluation metrics}: We also plan to include evaluation for distribution shift, uncertainty estimation, tests for fairness and social biases, as well as labels/metrics for interpretable multimodal learning. In the latter, we plan to include the EMAP score~\citep{hessel2020does} as an interpretability metric assessing whether cross-modal interactions improve performance.

3. \textit{Multimodal transfer learning and co-learning}: Can training data in one dataset help learning on other datasets? \multibench\ enables easy experimentation of such research questions: our initial experiments on transfer learning found that pre-training on larger datasets in the same domain can improve performance on smaller datasets when fine-tuned on a smaller dataset: performance on the smaller \textsc{CMU-MOSI} dataset improved from $75.2$ to $75.8$ using the same late fusion model with transfer learning from the larger \textsc{UR-FUNNY} and \textsc{CMU-MOSEI} datasets. Furthermore, recent work has shown that multimodal training can help improve unimodal performance as well~\citep{socher2013zero,NIPS2019_8731,zadeh2020foundations}. While previous experiments were on a small scale and limited to a single domain, we plan to expand significantly on this phenomenon (multimodal co-learning) in future versions of \multibench.

4. \textit{Multitask learning across modalities}: Multitask learning across multimodal tasks with a shared set of input modalities is a promising direction that can enable statistical strength sharing across datasets and efficiency in training a single model. Using \multibench, we also ran an extra experiment on multi-dataset multitask learning. We used the $4$ datasets in the affective computing domain and trained a single model across all $4$ of them with adjustable input embedding layers if the input features were different and separate classification heads for each dataset’s task. We found promising initial results with performance on the largest \textsc{CMU-MOSEI} dataset improving from $79.2$ to $80.9$ for a late fusion model and from $82.1$ to $82.9$ using a multimodal transformer model, although performance on the smaller \textsc{CMU-MOSI} dataset decreased from $75.2$ to $70.8$. We believe that these potential future studies in multitask and transfer learning are strengths of \multibench\ since it shows the potential of interesting experiments and usage.

\textbf{In conclusion}, we present \multibench, a large-scale benchmark unifying previously disjoint efforts in multimodal research with a focus on ease of use, accessibility, and reproducibility, thereby paving the way towards a deeper understanding of multimodal models. Through its unprecedented range of research areas, datasets, modalities, tasks, and evaluation metrics, \multibench\ highlights several future directions in building more generalizable, lightweight, and robust multimodal models.

\chapter{Neural Architectures for Multisensory Foundation Models}
\label{chap:models2}
\newcommand{\pipelinel}{\textsc{Recurrent Multistage Fusion Network}}
\newcommand{\pipelines}{\textsc{RMFN}}
\newcommand{\ourl}{Multistage Fusion Process}
\newcommand{\ours}{\textsc{MFP}}

\section{Introduction}

\begin{figure}[t!]
\centering{
\includegraphics[width=0.7\linewidth]{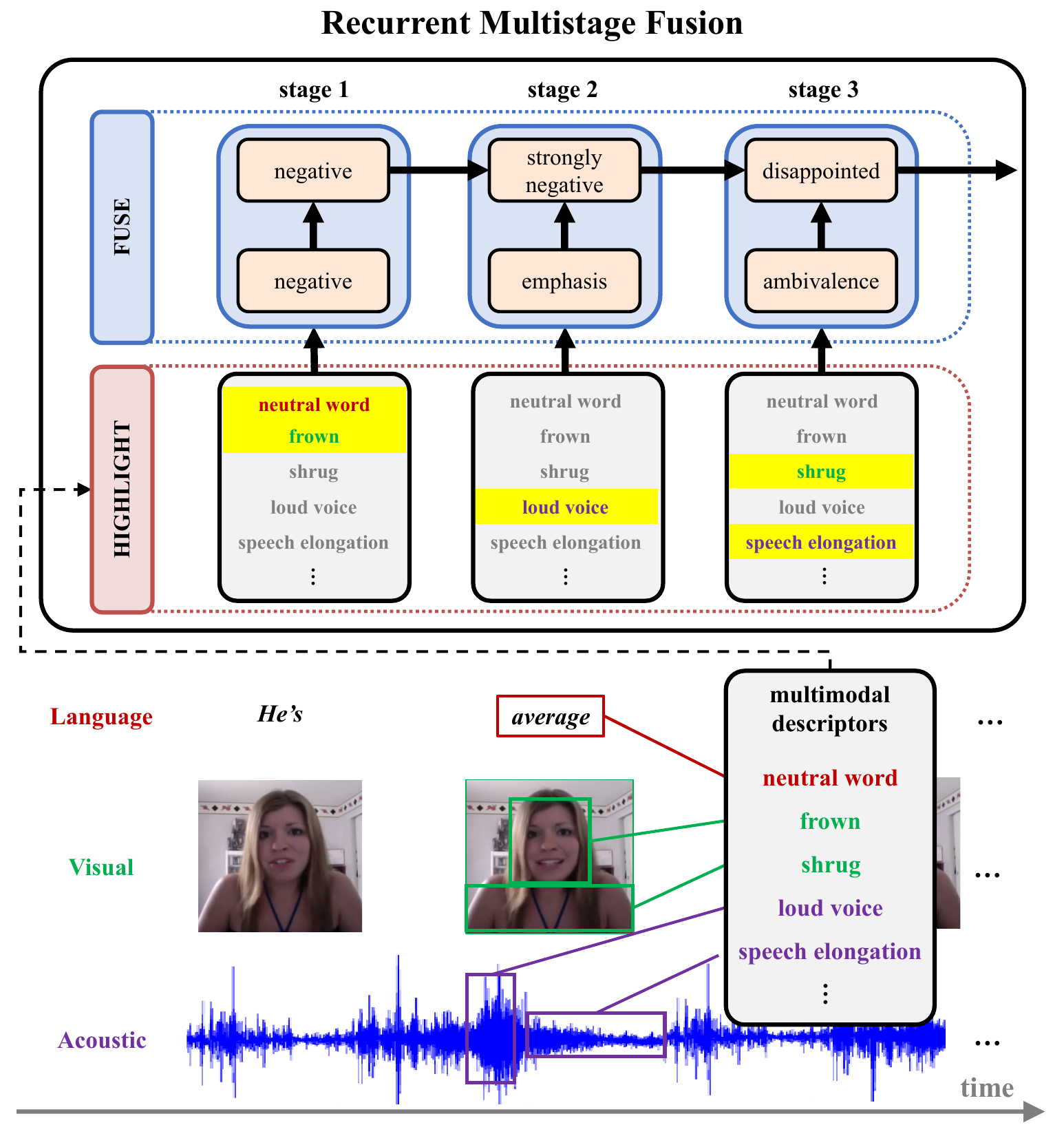}}
\caption{An illustrative example for Recurrent Multistage Fusion. At each recursive stage, a subset of multimodal signals is highlighted and then fused with previous fusion representations. The first fusion stage selects the neutral word and frowning behaviors which create an intermediate representation reflecting negative emotion when fused together. The second stage selects the loud voice behavior which is locally interpreted as emphasis before being fused with previous stages into a strongly negative representation. Finally, the third stage selects the shrugging and speech elongation behaviors that reflect ambivalence and when fused with previous stages is interpreted as a representation for the disappointed emotion.}
\vspace{-2mm}
\label{fig:overview}
\end{figure}    
    

To build general multisensory foundation models that work across the diverse modalities and tasks in \multibench, this chapter of the thesis presents two architectures that are broadly generalizable across diverse modalities. The large number of heterogeneous modalities creates challenges in building multisensory foundation models. For example, the healthcare domain typically collects tabular data and high-frequency sensors~\cite{MIMIC}, and it remains an open question how to best combine large language models with tabular data and sensors~\cite{shwartz2022tabular}. To tackle the heterogeneity across many different modalities, we treat modalities in their most general form as sequences of elements, and study how to learn interactions between multiple elements across modalities. As motivated in the first part of the thesis, these local interactions between two elements can be \textit{redundant}, \textit{unique}, and \textit{synergistic}: redundancy quantifies information shared between modalities, uniqueness quantifies the information present in only one of the modalities, and synergy quantifies the emergence of new information not previously present in either modality.

Treating modalities as sequences of elements now introduces a new challenge due to asynchrony in time: for example, the simultaneous co-occurrence between a smile and a positive word, or the delayed occurrence of laughter after the end of a sentence. Modeling these interactions lie at the heart of analyzing human communication, audio-video data, sensor fusion, and medical modalities. We now present two approaches to learn interactions from heterogeneous modality elements across sequences: the \textit{cross-modal attention}~\cite{liang2018multimodal,chen2017multimodal} and \textit{multimodal transformer}~\cite{tsai2019multimodal} architectures.

The first architecture is called \pipelinel, or \pipelines\ for short. This method automatically decomposes the multimodal fusion problem into multiple recursive stages across the sequence. At each stage, a subset of multimodal signals is highlighted and fused with previous fusion representations (see Figure~\ref{fig:overview}). This divide-and-conquer approach decreases the burden on each fusion stage, allowing each stage to be performed in a more specialized and effective way. This is in contrast with conventional fusion approaches which usually model interactions over multimodal sequences altogether in one iteration (e.g., early or late fusion~\citep{baltruvsaitis2017multimodal}). In \pipelines, multimodal interactions are modeled by integrating our new multistage fusion process with a system of recurrent neural networks. Overall, \pipelines\ recursively models all forms of redundant, unique, and synergistic multimodal interactions across the sequence and is differentiable end-to-end.

\begin{figure}[t!]
    \centering
    \includegraphics[width=0.5\textwidth]{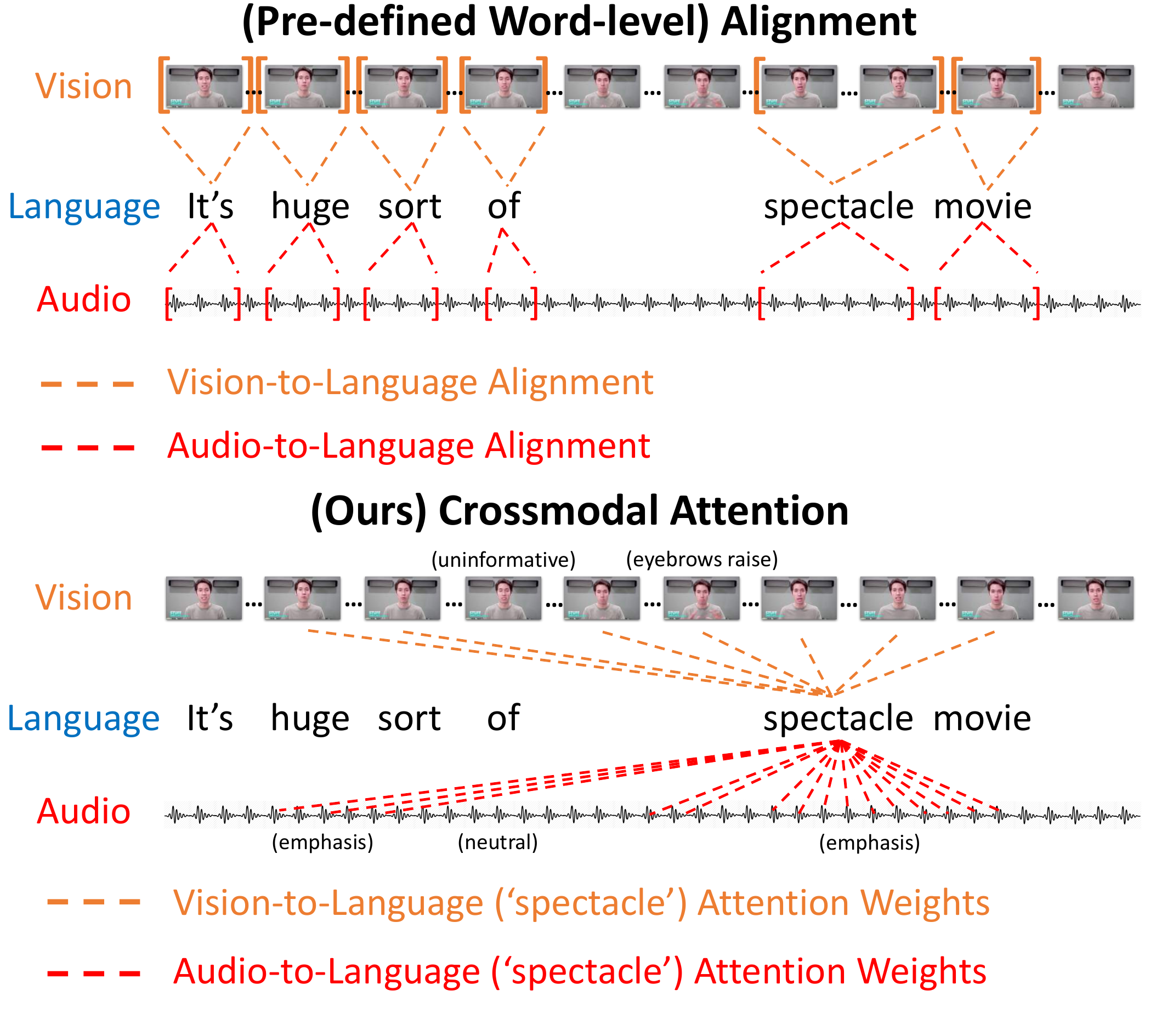}
    \caption{Example video clip from movie reviews. [Top]: Illustration of word-level alignment where video and audio features are averaged across the time interval of each spoken word. [Bottom] Illustration of crossmodal attention weights between text (``spectacle'') and vision/audio.}
    \label{fig:fig1}
\end{figure}

The second architecture we propose is the \textsc{Multimodal Transformer} (\mult), an end-to-end model that extends the standard Transformer network~\cite{vaswani2017attention} to learn representations directly from unaligned multimodal sequences. At the heart of \mult\ is the crossmodal attention module, which learns multimodal interactions between all elements in the first modality with all elements in the second modality. As a result, all multimodal interactions across the entire sequence are learned simultaneously, and can be parallelized efficiently over GPUs as compared to the first recurrent fusion approach. This makes \mult\ extremely scalable and effective, especially in settings where modality elements are asynchronous and where obtaining alignment information is difficult (e.g., by forced word-aligning before training~\cite{zadeh2018multimodal,pham2019found}, see Figure~\ref{fig:fig1} for a comparison).

We evaluate \pipelines\ and \mult\ on three different tasks related to human multimodal language: sentiment analysis, emotion recognition, and speaker traits recognition across three public multimodal datasets. \pipelines \ achieves state-of-the-art performance in all three tasks. Through a comprehensive set of ablation experiments and visualizations, we demonstrate the advantages of explicitly defining multiple recursive stages for multimodal fusion.

\section{Related Work}
\label{Related Work}

Previous approaches in human multimodal language modeling can be categorized as follows:

\textbf{Non-temporal Models}: These models simplify the problem by using feature-summarizing temporal observations~\citep{contextmultimodalacl2017}. Each modality is represented by averaging temporal information through time, as shown for language-based sentiment analysis~\citep{Iyyer2015,DBLP:journals/corr/ChenASWC16} and multimodal sentiment analysis~\citep{abburi2016multimodal,Nojavanasghari:2016:DMF:2993148.2993176,zadeh2016multimodal,morency2011towards}. Conventional supervised learning methods are utilized to discover intra-modal and cross-modal interactions without specific model design~\cite{wang2016select,poria2016convolutional}. These approaches have trouble modeling long sequences since the average statistics do not properly capture the temporal intra-modal and cross-modal dynamics~\cite{xu2013survey}.

\textbf{Multimodal Temporal Graphical Models}: The application of graphical models in sequence modeling has been an important research problem. Hidden Markov Models (HMMs)~\cite{baum1966statistical}, Conditional Random Fields (CRFs)~\citep{Lafferty:2001:CRF:645530.655813}, and Hidden Conditional Random Fields (HCRFs)~\cite{Quattoni:2007:HCR:1313053.1313265} were shown to work well on modeling sequential data from the language~\citep{W17-4114,ma2016endtoend,journals/corr/HuangXY15} and acoustic~\citep{P2FA} modalities. These temporal graphical models have also been extended for modeling multimodal data. Several methods have been proposed including multi-view HCRFs where the potentials of the HCRF are designed to model data from multiple views~\cite{song2012multi}, multi-layered CRFs with latent variables to learn hidden spatio-temporal dynamics from multi-view data~\cite{song2012multi}, and multi-view Hierarchical Sequence Summarization models that recursively build up hierarchical representations~\citep{song2013action}.

\textbf{Multimodal Temporal Neural Networks}: More recently, with the advent of deep learning, Recurrent Neural Networks~\citep{elman1990finding,Jain:1999:RNN:553011} have been used extensively for language and speech based sequence modeling~\citep{zilly2016recurrent,DBLP:journals/corr/SoltauLS16}, sentiment analysis~\citep{Socher-etal:2013,DBLP:conf/coling/SantosG14,Glorot:2011:DAL:3104482.3104547,7435182}, and emotion recognition~\citep{speech-emotion-recognition-using-deep-neural-network-and-extreme-learning-machine,unknown,DBLP:journals/corr/abs-1803-11508}. Long-short Term Memory (LSTM) networks~\cite{hochreiter1997long} have also been extended for multimodal settings~\cite{rajagopalan2016extending} and by learning binary gating mechanisms to remove noisy modalities~\cite{chen2017multimodal}. Recently, more advanced models were proposed to model both intra-modal and cross-modal interactions. These use Bayesian ranking algorithms~\citep{NIPS2006_3079} to model both person-independent and person-dependent features~\citep{liang2018ranking}, generative-discriminative objectives to learn either joint~\citep{seq2seq} or factorized multimodal representations~\citep{tsai2019learning}, external memory mechanisms to synchronize multimodal data~\citep{zadeh2018memory}, or low-rank tensors to approximate expensive tensor products~\citep{liu2018efficient}. All these methods assume that cross-modal interactions should be discovered all at once rather than across multiple stages, where each stage solves a simpler fusion problem. Our empirical evaluations show the advantages of the multistage fusion approach.

\textbf{Transformer Network}: The Transformer network~\cite{vaswani2017attention} was first introduced for neural machine translation, where the encoder and decoder side each leverages a \emph{self-attention}~\cite{parikh2016decomposable, lin2017structured, vaswani2017attention} transformer. After each layer of self-attention, the encoder and decoder are connected by an additional decoder sublayer where the decoder attends to each element of the source text for each element of the target text. In addition to translation, transformer networks have also been successfully applied to other tasks, including language modeling~\citep{dai2018transformer,baevski2018adaptive}, semantic role labeling~\citep{D18-1548}, word sense disambiguation~\citep{tang2018self}, learning sentence representations~\citep{devlin2019bert}, and video activity recognition~\citep{wang2018non}.

\section{\pipelinel}

We first describe the \pipelinel\ (\pipelines\ for short) for multimodal language analysis (Figure \ref{fig:model}). Given a set of modalities $\{ l(anguage), v(isual), a(coustic) \}$, the signal from each modality $m \in \{l,v,a\}$ is represented as a temporal sequence $\mathbf{X}^m = \{\textbf{x}_{1}^m, \textbf{x}_{2}^m, \textbf{x}_{3}^m, \cdots, \textbf{x}_{T}^m \}$, where $\textbf{x}^m_{t}$ is the input at time $t$. Each sequence $\mathbf{X}^m$ is modeled with an intra-modal recurrent neural network (see subsection \ref{sec:LSTHM} for details). At time $t$, each intra-modal recurrent network will output a unimodal representation $\mathbf{h}^m_t$. The \ourl \ uses a recursive approach to fuse all unimodal representations $\mathbf{h}^m_t$ into a cross-modal representation $\mathbf{z}_t$ which is then fed back into each intra-modal recurrent network. 

\begin{figure*}[t!]
\centering{
\includegraphics[width=0.8\linewidth]{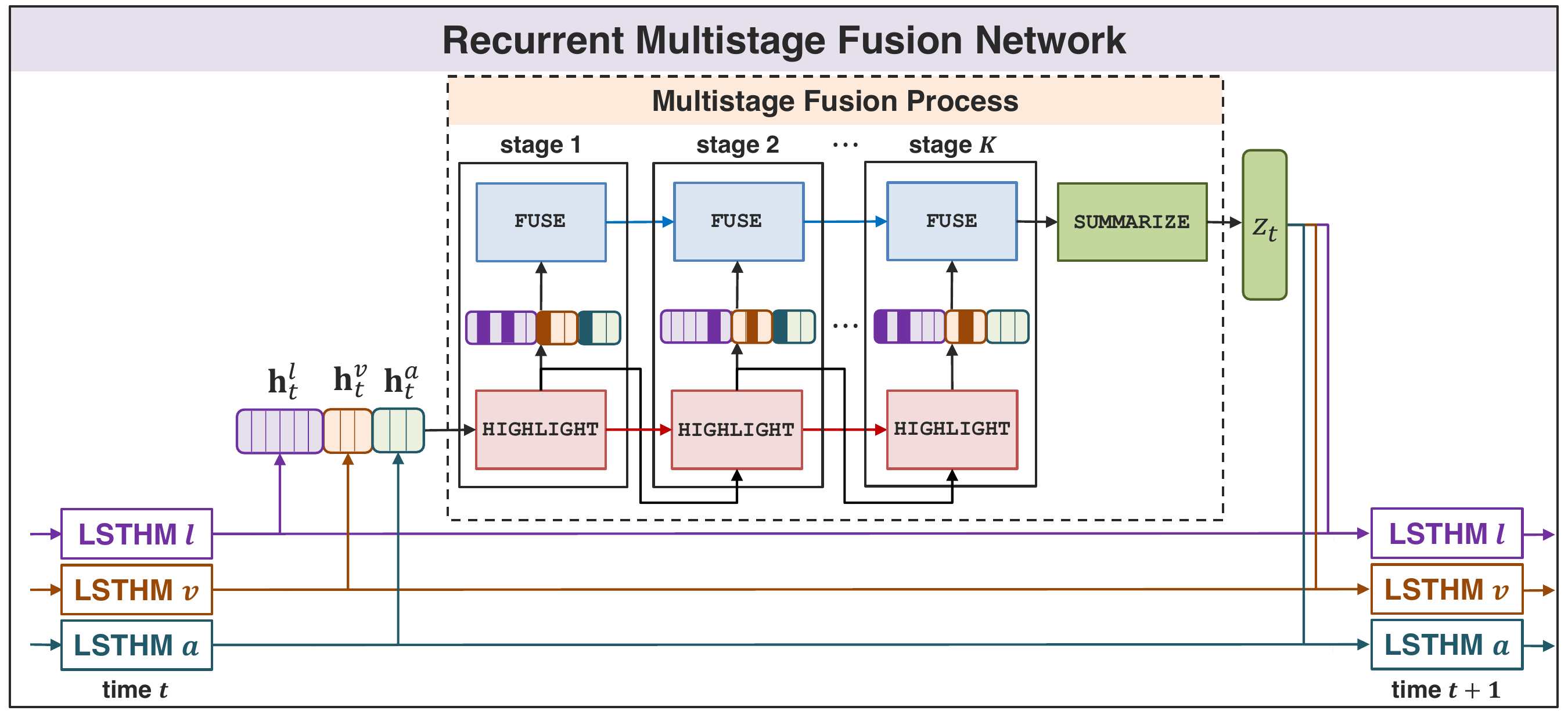}}
    \caption{The \pipelinel \ for multimodal language analysis. The \ourl \ has three modules: \texttt{HIGHLIGHT}, \texttt{FUSE} and \texttt{SUMMARIZE}. Multistage fusion begins with the concatenated intra-modal network outputs $\mathbf{h}^l_t, \mathbf{h}^v_t, \mathbf{h}^a_t$. At each stage, the \texttt{HIGHLIGHT} module identifies a subset of multimodal signals and the \texttt{FUSE} module performs local fusion before integration with previous fusion representations. The \texttt{SUMMARIZE} module translates the representation at the final stage into a cross-modal representation $\mathbf{z}_t$ to be fed back into the intra-modal recurrent networks.}
 \vspace{-2mm}
\label{fig:model}
\end{figure*}

\subsection{Multistage fusion process}

The \ourl \ (\ours) is a modular neural approach that performs multistage fusion to model cross-modal interactions. Multistage fusion is a divide-and-conquer approach which decreases the burden on each stage of multimodal fusion, allowing each stage to be performed in a more specialized and effective way. The \ours \ has three main modules: \texttt{HIGHLIGHT}, \texttt{FUSE} and \texttt{SUMMARIZE}. 

Two modules are repeated at each stage: \texttt{HIGHLIGHT} and \texttt{FUSE}. The \texttt{HIGHLIGHT} module identifies a subset of multimodal signals from $[\mathbf{h}^l_t,\mathbf{h}^v_t,\mathbf{h}^a_t]$ that will be used for that stage of fusion. The \texttt{FUSE} module then performs two subtasks simultaneously: a local fusion of the highlighted features and integration with representations from previous stages. Both \texttt{HIGHLIGHT} and \texttt{FUSE} modules are realized using memory-based neural networks which enable coherence between stages and storage of previously modeled cross-modal interactions. As a final step, the \texttt{SUMMARIZE} module takes the multimodal representation of the final stage and translates it into a cross-modal representation $\mathbf{z}_t$.

Figure~\ref{fig:overview} shows an illustrative example for multistage fusion. The \texttt{HIGHLIGHT} module selects ``neutral words'' and ``frowning'' expression for the first stage. The local and integrated fusion at this stage creates a representation reflecting negative emotion. For stage 2, the \texttt{HIGHLIGHT} module identifies the acoustic feature ``loud voice''. The local fusion at this stage interprets it as an expression of emphasis and is fused with the previous fusion results to represent a strong negative emotion. Finally, the highlighted features of ``shrug'' and ``speech elongation'' are selected and are locally interpreted as ``ambivalence''. The integration with previous stages then gives a representation closer to ``disappointed''.

\subsection{Module descriptions}
\label{multistage}

In this section, we present the details of the three multistage fusion modules: \texttt{HIGHLIGHT}, \texttt{FUSE} and \texttt{SUMMARIZE}. Multistage fusion begins with the concatenation of intra-modal network outputs $\mathbf{h}_t = \bigoplus_{m \in M} \mathbf{h}^m_t$. We use superscript $^{[k]}$ to denote the indices of each stage $k = 1, \cdots, K$ during $K$ total stages of multistage fusion. Let $\Theta$ denote the neural network parameters across all modules.

\texttt{HIGHLIGHT}: At each stage $k$, a subset of the multimodal signals represented in $\mathbf{h}_t$ will be automatically highlighted for fusion. Formally, this module is defined by the process function $f_{H}$: 
\begin{equation}\label{eq:highlight}
\mathbf{a}_t^{[k]} = f_{H} (\mathbf{h}_t \ ; \ \mathbf{a}_t^{[1:k-1]}, \Theta)
\end{equation}
where at stage $k$, $\mathbf{a}_t^{[k]}$ is a set of attention weights which are inferred based on the previously assigned attention weights $\mathbf{a}_t^{[1:k-1]}$. As a result, the highlights at a specific stage $k$ will be dependent on previous highlights. To fully encapsulate these dependencies, the attention assignment process is performed in a recurrent manner using a LSTM which we call the \texttt{HIGHLIGHT} LSTM. The initial \texttt{HIGHLIGHT} LSTM memory at stage 0, $\mathbf{c}_t^{\texttt{HIGHLIGHT}[0]}$, is initialized using a network $\mathcal{M}$ that maps $\mathbf{h}_t$ into LSTM memory space:
\begin{equation}
\mathbf{c}_t^{\texttt{HIGHLIGHT}[0]} = \mathcal{M} (\mathbf{h}_t \ ; \ \Theta)
\end{equation}
This allows the memory mechanism of the \texttt{HIGHLIGHT} LSTM to dynamically adjust to the intra-modal representations $\mathbf{h}_t$. The output of the \texttt{HIGHLIGHT} LSTM $\mathbf{h}_t^{\texttt{HIGHLIGHT}[k]}$ is softmax activated to produce attention weights $\mathbf{a}_t^{[k]}$ at every stage $k$ of the multistage fusion process:
\begin{equation}
{\mathbf{a}_t^{[k]}}_j = \frac{\exp \ ( {\mathbf{h}_t^{\texttt{HIGHLIGHT}[k]}}_j) }{\sum_{d=1}^{|\mathbf{h}_t^{\texttt{HIGHLIGHT}[k]}|} \exp \ ( {\mathbf{h}_t^{\texttt{HIGHLIGHT}[k]}}_d) } 
\end{equation}
and ${\mathbf{a}_t^{[k]}}$ is fed as input into the \texttt{HIGHLIGHT} LSTM at stage $k+1$. Therefore, the \texttt{HIGHLIGHT} LSTM functions as a decoder LSTM~\citep{DBLP:journals/corr/SutskeverVL14,DBLP:journals/corr/ChoMGBSB14} in order to capture the dependencies on previous attention assignments. Highlighting is performed by element-wise multiplying the attention weights $\mathbf{a}_t^{[k]}$ with the concatenated intra-modal representations $\mathbf{h}_t$:
\begin{equation}\label{eq:discover}
\tilde{\mathbf{h}}_t^{[k]} = \mathbf{h}_t \odot \mathbf{a}_t^{[k]}
\end{equation}
where $\odot$ denotes the Hadamard product and $\tilde{\mathbf{h}}_t^{[k]}$ are the attended multimodal signals that will be used for the fusion at stage $k$.

\texttt{FUSE}: The highlighted multimodal signals are simultaneously fused in a local fusion and then integrated with fusion representations from previous stages. Formally, this module is defined by the process function $f_{F}$:
\begin{equation}\label{eq:fuse}
\mathbf{s}_t^{[k]} = f_{F} (\tilde{\mathbf{h}}_t^{[k]} \ ; \ \mathbf{s}_t^{[1:k-1]}, \Theta)
\end{equation}
where $\mathbf{s}_t^{[k]}$ denotes the integrated fusion representations at stage $k$. We employ a \texttt{FUSE} LSTM to simultaneously perform the local fusion and the integration with previous fusion representations. The \texttt{FUSE} LSTM input gate enables a local fusion while the \texttt{FUSE} LSTM forget and output gates enable integration with previous fusion results. The initial \texttt{FUSE} LSTM memory at stage 0, $\mathbf{c}_t^{\texttt{FUSE}[0]}$, is initialized using random orthogonal matrices~\citep{DBLP:journals/corr/ArjovskySB15,DBLP:journals/corr/LeJH15}.

\texttt{SUMMARIZE}: After completing $K$ recursive stages of \texttt{HIGHLIGHT} and \texttt{FUSE}, the \texttt{SUMMARIZE} operation generates a cross-modal representation using all final fusion representations $\mathbf{s}_t^{[1:K]}$.
Formally, this operation is defined as:
\begin{equation}
\mathbf{z}_t = \mathcal{S}(\mathbf{s}_t^{[1:K]} \ ; \ \Theta)
\end{equation}
where $\mathbf{z}_t$ is the final output of the multistage fusion process and represents all cross-modal interactions discovered at time $t$. The summarized cross-modal representation is then fed into the intra-modal recurrent networks as described in the subsection~\ref{sec:LSTHM}.

\subsection{System of long short-term hybrid memories}
\label{sec:LSTHM}

To integrate the cross-modal representations $\mathbf{z}_{t}$ with the temporal intra-modal representations, we employ a system of Long Short-term Hybrid Memories (LSTHMs)~\citep{zadeh2018multi}. The LSTHM extends the LSTM formulation to include the cross-modal representation $\mathbf{z}_{t}$ in a hybrid memory component:
\par\nobreak
\vspace{-4mm}
{\small
\begin{align}
\mathbf{i}_{t+1}^m &= \sigma(\mathbf{W}_i^m\ \mathbf{x}^m_{t+1}+\mathbf{U}^m_i\ \mathbf{h}^m_{t}+\mathbf{V}^m_i\ \mathbf{z}_{t}+\mathbf{b}^m_{i}) \\
\mathbf{f}^m_{t+1} &= \sigma(\mathbf{W}^m_{f}\ \mathbf{x}^m_{t+1} + \mathbf{U}^m_{f}\ \mathbf{h}^m_{t} + \mathbf{V}^m_f\ \mathbf{z}_{t}+\mathbf{b}^m_{f}) \\
\mathbf{o}^m_{t+1} &= \sigma(\mathbf{W}^m_{o}\ \mathbf{x}^m_{t+1} + \mathbf{U}^m_{o}\ \mathbf{h}^m_{t} + \mathbf{V}^m_o\ \mathbf{z}_{t}+\mathbf{b}^m_{o}) \\
\bar{\mathbf{c}}_{t+1}^m &= \mathbf{W}_{\bar{c}}^m\ \mathbf{x}^m_{t+1} + \mathbf{U}_{\bar{c}}^m\ \mathbf{h}^m_{t} + \mathbf{V}_{\bar{c}}^m\ \mathbf{z}_{t} + \mathbf{b}^m_{\bar{c}} \\
\mathbf{c}^m_{t+1} &= \mathbf{f}^m_t \odot \mathbf{\mathbf{c}}^m_{t} + \mathbf{i}^m_t \odot tanh(\bar{\mathbf{c}}_{t+1}^m) \\
\mathbf{h}^m_{t+1} &= \mathbf{o}^m_{t+1} \odot tanh(\mathbf{c}^m_{t+1})
\end{align}
}%
where $\sigma$ is the (hard-)sigmoid activation function, $tanh$ is the tangent hyperbolic activation function, $\odot$ denotes the Hadamard product. $\mathbf{i},\mathbf{f}$ and $\mathbf{o}$ are the input, forget and output gates respectively. $\bar{\mathbf{c}}_{t+1}^m$ is the proposed update to the hybrid memory $\mathbf{c}^m_t$ at time $t+1$ and $\mathbf{h}^m_t$ is the time distributed output of each modality. The cross-modal representation $\mathbf{z}_t$ is modeled by the \ourl \ as discussed in subsection~\ref{multistage}. The hybrid memory $\mathbf{c}^m_t$ contains both intra-modal interactions from individual modalities $\mathbf{x}^m_t$ as well as the cross-modal interactions captured in $\mathbf{z}_t$.

\subsection{Optimization}
The multimodal prediction task is performed using a final representation $\mathcal{E}$ which integrate (1) the last outputs from the LSTHMs and (2) the last cross-modal representation $\mathbf{z}_T$. Formally, $\mathcal{E}$ is defined as: 
\begin{equation}
\mathcal{E} = (\bigoplus_{m \in M} \mathbf{h}^m_T) \bigoplus \mathbf{z}_T
\end{equation}
where $\bigoplus$ denotes vector concatenation. $\mathcal{E}$ can then be used as a multimodal representation for supervised or unsupervised analysis of multimodal language. It summarizes all modeled intra-modal and cross-modal representations from the multimodal sequences. \pipelines \ is differentiable end-to-end which allows the network parameters $\Theta$ to be learned using gradient descent approaches.

\section{\textsc{Multimodal Transformer}}
\label{sec:method}

We next describe our second proposed architecture, the \textsc{Multimodal Transformer} (\mult) (Figure \ref{fig:overall}) for modeling unaligned multimodal language sequences. At the high level, \mult\ merges multimodal time-series via a feed-forward fusion process from multiple directional pairwise crossmodal transformers. Specifically, each crossmodal transformer (introduced in Section \ref{subsec:overall}) serves to repeatedly reinforce a \emph{target modality} with the low-level features from another \emph{source modality} by learning the attention across the two modalities' features. A \mult\ architecture hence models all pairs of modalities with such crossmodal transformers, followed by sequence models (e.g., self-attention transformer) that predicts using the fused features.

The core of our proposed model is crossmodal attention module, which we first introduce in Section~\ref{subsec:cm-attn}.
Then, in Section~\ref{subsec:overall} and~\ref{subsec:advantage}, we present in details the various ingredients of the \mult\ architecture (see Figure~\ref{fig:overall}) and discuss the difference between crossmodal attention and classical multimodal alignment.

\begin{figure}[t!]
    \centering
    \includegraphics[width=0.5\textwidth]{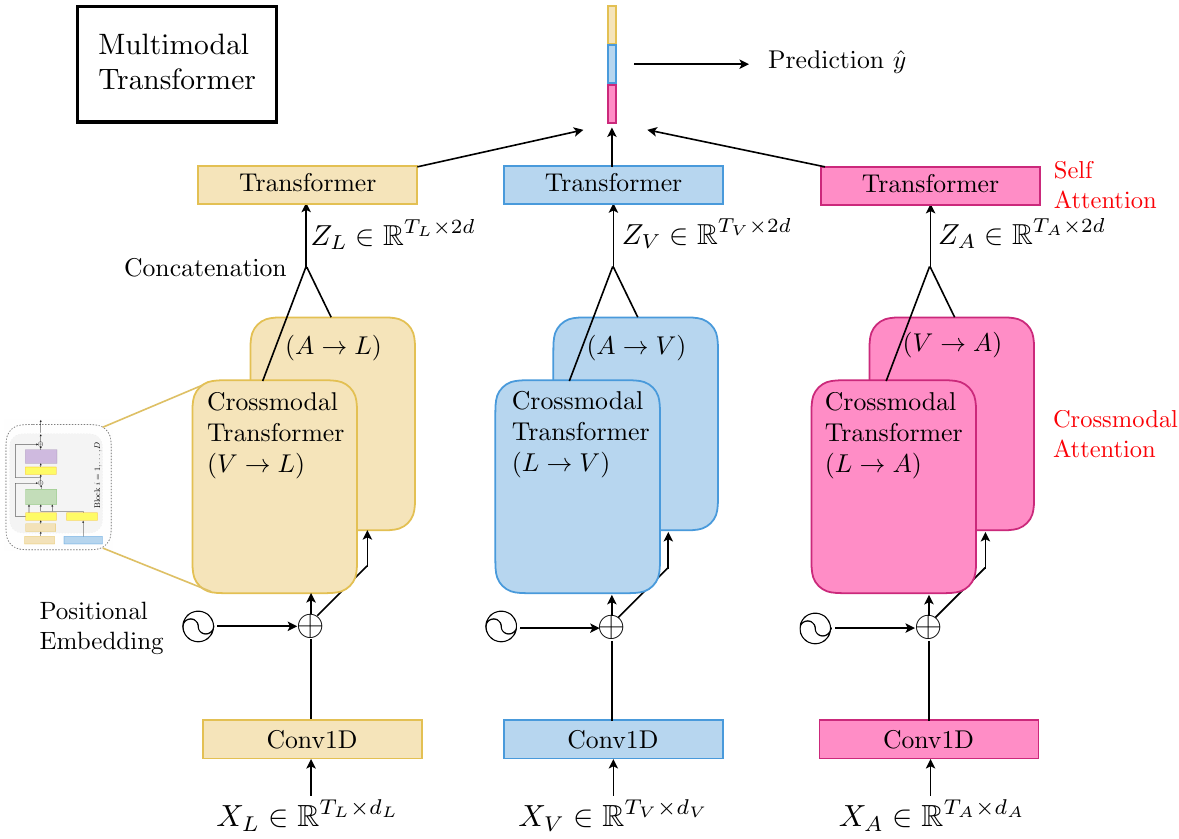}
    \caption{Overall architecture for \mult\ on modalities ($L, V, A$). The crossmodal transformers, which suggests latent crossmodal adaptations, are the core components of \mult\ for multimodal fusion.}
    \label{fig:overall}
\end{figure}

\begin{figure*}[t]
    \centering
    \begin{subfigure}[b]{0.6\textwidth}
        \centering
        \includegraphics[width=0.85\textwidth]{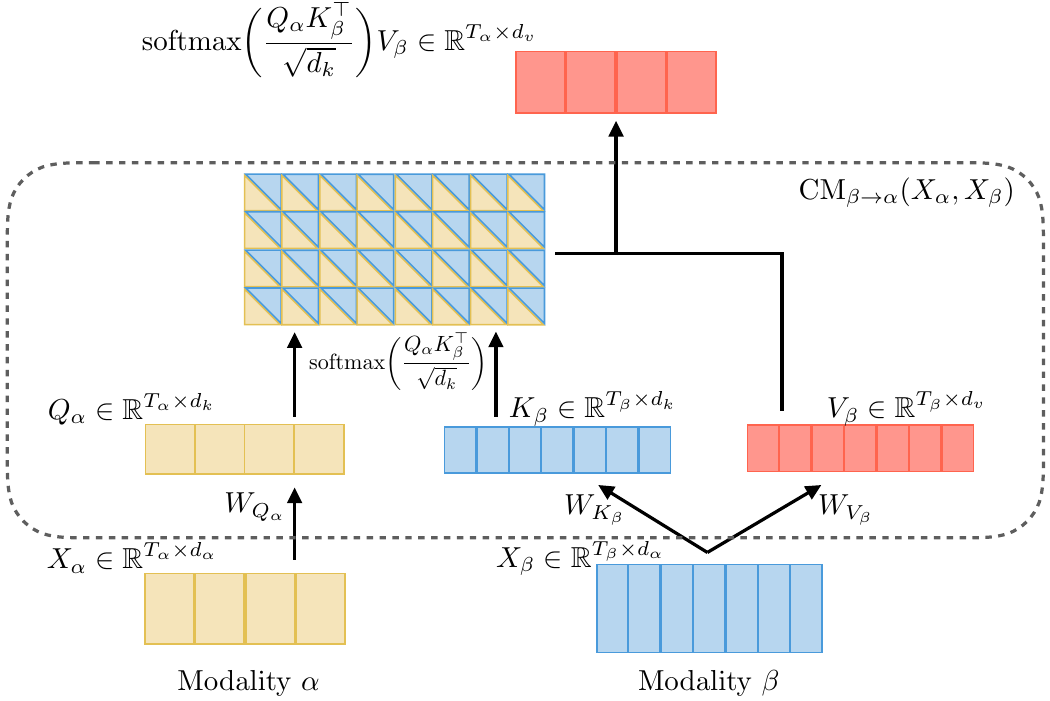} 
        \caption{Crossmodal attention $\text{CM}_{\beta \rightarrow \alpha}(X_\alpha, X_\beta)$ between sequences $X_\alpha$, $X_\beta$ from distinct modalities.}
        \label{fig:components-cm}
    \end{subfigure}
    ~
    \begin{subfigure}[b]{0.36\textwidth}
        \centering
        \includegraphics[width=\textwidth]{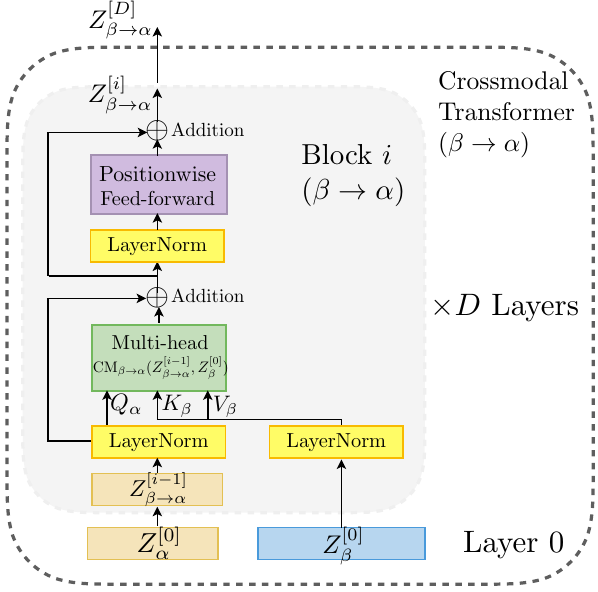}
        \caption{A crossmodal transformer is a deep stacking of several crossmodal attention blocks.}
        \label{fig:components-ct}
    \end{subfigure}
    \caption{Architectural elements of a crossmodal transformer between two time-series from modality $\alpha$ and $\beta$.}
    \label{fig:components}
\end{figure*}

\subsection{Crossmodal attention}
\label{subsec:cm-attn}

We consider two modalities $\alpha$ and $\beta$, with two (potentially non-aligned) sequences from each of them denoted $X_\alpha \in \mathbb{R}^{T_\alpha \times d_\alpha}$ and $X_\beta \in \mathbb{R}^{T_\beta \times d_\beta}$, respectively. For the rest of the paper, $T_{(\cdot)}$ and $d_{(\cdot)}$ are used to represent sequence length and feature dimension, respectively. Inspired by the decoder transformer in NMT~\citep{vaswani2017attention} that translates one language to another, we hypothesize a good way to fuse crossmodal information is providing a latent adaptation across modalities; i.e., $\beta$ to $\alpha$. Note that the modalities consider in our paper may span very different domains such as facial attributes and spoken words.

We define the $\mathrm{Query}$s as $Q_\alpha = X_\alpha W_{Q_\alpha}$, $\mathrm{Key}$s as $K_\beta = X_\beta W_{K_\beta}$, and $\mathrm{Value}$s as $V_\beta = X_\beta W_{V_\beta}$, where $W_{Q_\alpha} \in \mathbb{R}^{d_\alpha \times d_k}, W_{K_\beta} \in \mathbb{R}^{d_\beta \times d_k}$ and $W_{V_\beta} \in \mathbb{R}^{d_\beta \times d_v}$ are weights. The latent adaptation from $\beta$ to $\alpha$ is presented as the crossmodal attention $Y_\alpha := \text{CM}_{\beta \rightarrow \alpha}(X_\alpha, X_B) \in \mathbb{R}^{T_\alpha \times d_v}$:
\begin{equation}
\label{eq:crossmodalBA}
\begin{split}
    Y_\alpha &= \text{CM}_{\beta \rightarrow \alpha}(X_\alpha, X_\beta) \\
    &= \text{softmax}\left(\frac{Q_\alpha K_\beta^\top}{\sqrt{d_k}}\right) V_\beta
    \\
    &= \text{softmax}\left(\frac{X_\alpha W_{Q_\alpha} W_{K_\beta}^\top X_\beta^\top}{\sqrt{d_k}}\right) X_\beta W_{V_\beta}.
\end{split}
\end{equation}
Note that $Y_\alpha$ has the same length as $Q_\alpha$ (i.e., $T_\alpha$), but is meanwhile represented in the feature space of $V_\beta$. Specifically, the scaled (by $\sqrt{d_k}$) softmax in Equation~(\ref{eq:crossmodalBA}) computes a score matrix $\mathrm{softmax\,}(
\cdot) \in \mathbb{R}^{T_\alpha \times T_\beta}$, whose $(i,j)$-th entry measures the attention given by the $i$-th time step of modality $\alpha$ to the $j$-th time step of modality $\beta$. Hence, the $i$-th time step of $Y_\alpha$ is a weighted summary of $V_\beta$, with the weight determined by $i$-th row in $\mathrm{softmax}(\cdot)$. We call Equation~\eqref{eq:crossmodalBA} a \emph{single-head} crossmodal attention, which is illustrated in Figure \ref{fig:components-cm}. 

Following prior works on transformers~\citep{vaswani2017attention,chen2018thebest,devlin2019bert,dai2018transformer}, we add a residual connection to the crossmodal attention computation. Then, another positionwise feed-forward sublayer is injected to complete a \emph{crossmodal attention
block} (see Figure \ref{fig:components-ct}). Each crossmodal attention block adapts directly from the low-level feature sequence (i.e., $Z_\beta^{[0]}$ in Figure \ref{fig:components-ct}) and does not rely on self-attention, which makes it different from the NMT encoder-decoder architecture~\citep{vaswani2017attention, shaw2018self} (i.e., taking intermediate-level features).
We argue that performing adaptation from low-level feature benefits our model to preserve the low-level information for each modality.

\subsection{Overall architecture}
\label{subsec:overall}

Three major modalities are typically involved in multimodal language sequences: language ($L$), video ($V$), and audio ($A$) modalities. We denote with ${X}_{\{L,V,A\}}\in \mathbb{R}^{T_{\{L,V,A\}} \times d_{\{L,V,A\}}}$ the input feature sequences (and the dimensions thereof) from these 3 modalities. With these notations, in this subsection, we describe in greater details the components of Multimodal Transformer and how crossmodal attention modules are applied.

\paragraph{Temporal Convolutions.} To ensure that each element of the input sequences has sufficient awareness of its neighborhood elements, we pass the input sequences through a 1D temporal convolutional layer:
\begin{equation}
\hat{X}_{\{L,V,A\}} = \text{Conv1D}(X_{\{L,V,A\}}, k_{\{L,V,A\}}) \in \mathbb{R}^{T_{\{L,V,A\}} \times d}
\end{equation}
where $k_{\{L,V,A\}}$ are the sizes of the convolutional kernels for modalities $\{L,V,A\}$, and $d$ is a common dimension. The convolved sequences are expected to contain the local structure of the sequence, which is important since the sequences are collected at different sampling rates. Moreover, since the temporal convolutions project the features of different modalities to the same dimension $d$, the dot-products are admittable in the crossmodal attention module.

\paragraph{Positional Embedding.} To enable the sequences to carry temporal information, following prior work~\cite{vaswani2017attention}, we augment positional embedding (PE) to $\hat{X}_{\{L,V,A\}}$:
\begin{equation}
    Z_{\{L,V,A\}}^{[0]} = \hat{X}_{\{L,V,A\}} + \text{PE}(T_{\{L,V,A\}}, d)
\end{equation}
where PE$(T_{\{L,V,A\}}, d) \in \mathbb{R}^{T_{\{L,V,A\}} \times d}$ computes the (fixed) embeddings for each position index, and $Z_{\{L,V,A\}}^{[0]}$ are the resulting low-level position-aware features for different modalities. We leave more details of the positional embedding to the full paper~\cite{tsai2019multimodal}.

\paragraph{Crossmodal Transformers.} Based on the crossmodal attention blocks, we design the crossmodal transformer that enables one modality for receiving information from another modality. In the following, we use the example for passing vision ($V$) information to language ($L$), which is denoted by ``$V\rightarrow L$''. We fix all the dimensions ($d_{\{\alpha,\beta,k,v \}}$) for each crossmodal attention block as $d$.

Each crossmodal transformer consists of $D$ layers of crossmodal attention blocks (see Figure \ref{fig:components-ct}). Formally, a crossmodal transformer computes feed-forwardly for $i=1, \dots, D$ layers:
\begin{equation}
\begin{split}
    Z_{V \rightarrow L}^{[0]} &= Z_L^{[0]} \\
    \hat{Z}_{V \rightarrow L}^{[i]} &= \text{CM}_{V \rightarrow L}^{[i], \text{mul}}(\text{LN}(Z_{V \rightarrow L}^{[i-1]}), \text{LN}(Z_V^{[0]})) + \text{LN}(Z_{V \rightarrow L}^{[i-1]}) \\
    Z_{V \rightarrow L}^{[i]} &= f_{\theta_{V \rightarrow L}^{[i]}}(\text{LN}(\hat{Z}_{V \rightarrow L}^{[i]})) + \text{LN}(\hat{Z}_{V \rightarrow L}^{[i]})
\end{split}
\end{equation}
where $f_\theta$ is a positionwise feed-forward sublayer parametrized by $\theta$, and $\text{CM}_{V \rightarrow L}^{[i], \text{mul}}$ means a multi-head (see prior work~\cite{vaswani2017attention} for more details) version of $\text{CM}_{V \rightarrow L}$ at layer $i$ (note: $d$ should be divisible by the number of heads). LN means layer normalization~\citep{ba2016layer}.

In this process, each modality keeps updating its sequence via low-level external information from the multi-head crossmodal attention module. At every level of the crossmodal attention block, the low-level signals from source modality are transformed to a different set of $\mathrm{Key}$/$\mathrm{Value}$ pairs to interact with the target modality. Empirically, we find that the crossmodal transformer learns to correlate meaningful elements across modalities. The eventual \mult\ is based on modeling every pair of crossmodal interactions. Therefore, with $3$ modalities (i.e., $L,V,A$) in consideration, we have $6$ crossmodal transformers in total (see Figure \ref{fig:overall}).

\paragraph{Self-Attention Transformers and Prediction.} As a final step, we concatenate the outputs from the crossmodal transformers that share the same target modality to yield $Z_{\{L,V,A\}} \in \mathbb{R}^{T_{\{L,V,A\}} \times 2d}$. For example, $Z_L = [Z_{V\rightarrow L}^{[D]}; Z_{A\rightarrow L}^{[D]}]$. Each of them is then passed through a sequence model to collect temporal information to make predictions. We choose the self-attention transformer~\citep{vaswani2017attention}. Eventually, the last elements of the sequences models are extracted to pass through fully-connected layers to make predictions. 

\begin{figure}[t!]
    \centering
    \includegraphics[width=0.49\textwidth]{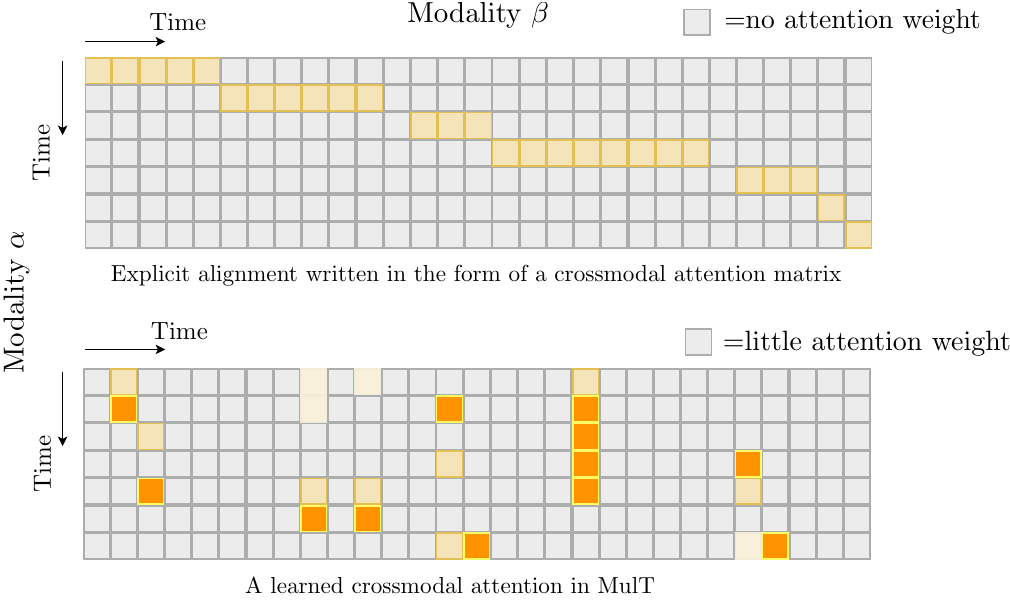}
    \caption{An example of visualizing alignment using attention matrix from modality $\beta$ to $\alpha$. Multimodal alignment is a special (monotonic) case for crossmodal attention.
    }
    \label{fig:align-attn}
\end{figure}

\subsection{Discussion about attention \& alignment}
\label{subsec:advantage}

When modeling unaligned multimodal language sequences, \mult\ relies on crossmodal attention blocks to merge signals across modalities. While the multimodal sequences were (manually) aligned to the same length in prior works before training~\citep{zadeh2018multimodal, liang2018multimodal, tsai2019learning, pham2019found, wang2018words}, we note that \mult\ looks at the non-alignment issue through a completely different lens. Specifically, for \mult, the correlations between elements of multiple modalities are purely based on attention. In other words, \mult\ does not handle modality non-alignment by (simply) aligning them; instead, the crossmodal attention encourages the model to directly attend to elements in other modalities where strong signals or relevant information is present. As a result, \mult\ can capture long-range crossmodal contingencies in a way that conventional alignment could not easily reveal. Classical crossmodal alignment, on the other hand, can be expressed as a special (step diagonal) crossmodal attention matrix  (i.e., monotonic attention~\cite{yu2016online}). We illustrate their differences in Figure~\ref{fig:align-attn}.

\section{Experimental Setup}

To evaluate the performance and generalization of \pipelines\ and \mult, three domains of human multimodal language were selected: multimodal sentiment analysis, emotion recognition, and speaker traits recognition. Our goal is to compare \mult\ with prior competitive approaches on both \emph{word-aligned} (by word, which almost all prior works employ) and \emph{unaligned} (which is more challenging, and which \mult\ is generically designed for) multimodal language sequences.

\subsection{Datasets}

All datasets consist of monologue videos. The speaker's intentions are conveyed through three modalities: language, visual and acoustic.

\textbf{Multimodal Sentiment Analysis} involves analyzing speaker sentiment based on video content. Multimodal sentiment analysis extends conventional language-based sentiment analysis to a multimodal setup where both verbal and non-verbal signals contribute to the expression of sentiment. We use \textbf{CMU-MOSI}~\cite{zadeh2016multimodal} which consists of 2199 opinion segments from online videos each annotated with sentiment in the range [-3,3]. 

\textbf{Multimodal Emotion Recognition} involves identifying speaker emotions based on both verbal and nonverbal behaviors. We perform experiments on the \textbf{IEMOCAP} dataset~\cite{Busso2008IEMOCAP:Interactiveemotionaldyadic} which consists of 7318 segments of recorded dyadic dialogues annotated for the presence of human emotions happiness, sadness, anger and neutral.

\textbf{Multimodal Speaker Traits Recognition} involves recognizing speaker traits based on multimodal communicative behaviors. \textbf{POM}~\cite{Park:2014:CAP:2663204.2663260} contains 903 movie review videos each annotated for 12 speaker traits: confident (con), passionate (pas), voice pleasant (voi), credible (cre), vivid (viv), expertise (exp), reserved (res), trusting (tru), relaxed (rel), thorough (tho), nervous (ner), persuasive (per) and humorous (hum).

Each task consists of a \emph{word-aligned} (processed in the same way as in prior works) and an \emph{unaligned} version. For both versions, the multimodal features are extracted from the textual (GloVe word embeddings~\cite{pennington2014glove}), visual (Facet~\cite{emotient}), and acoustic (COVAREP~\cite{degottex2014covarep}) data modalities. A more detailed introduction to the features is included in the full paper~\cite{liang2018multimodal}.

For the word-aligned version, following~\cite{zadeh2018memory, tsai2019learning, pham2019found}, we first use P2FA~\cite{P2FA} to obtain the aligned timesteps (segmented w.r.t. words) for audio and vision streams, and we then perform averaging on the audio and vision features within these time ranges. All sequences in the word-aligned case have length 50. The process remains the same across all the datasets. On the other hand, for the unaligned version, we keep the original audio and visual features as extracted, without any word-segmented alignment or manual subsampling. As a result, the lengths of each modality vary significantly, where audio and vision sequences may contain up to $>1,000$ time steps. We elaborate on the three tasks below.

\subsection{Multimodal features and alignment}

GloVe word embeddings~\cite{pennington2014glove}, Facet~\cite{emotient} and COVAREP~\cite{degottex2014covarep} are extracted for the language, visual and acoustic modalities respectively \footnote{Details on feature extraction are in supplementary.}. Forced alignment is performed using P2FA~\cite{P2FA} to obtain the exact utterance times of each word. We obtain the aligned video and audio features by computing the expectation of their modality feature values over each word utterance time interval~\cite{tsai2019learning}.

\subsection{Baseline models}
\label{sec:base}

\newcolumntype{K}[1]{>{\centering\arraybackslash}p{#1}}
\begin{table}[t!]
\fontsize{9}{11}\selectfont
\centering
\setlength\tabcolsep{4.0pt}
\caption{Results for multimodal sentiment analysis on CMU-MOSI with aligned and non-aligned multimodal sequences. ${}^h$ means higher is better and ${}^\ell$ means lower is better. EF stands for early fusion, and LF stands for late fusion.}
\begin{tabular}{c||*{5}{K{1cm}}}
\hline \hline
Metric         & Acc$_7$ $\uparrow $ & Acc$_2$ $\uparrow $ & F1 $\uparrow $ & MAE $\downarrow $ & Corr $\uparrow $  \\ \hline \hline
\multicolumn{6}{c}{\textcolor{blue}{(Word Aligned)} \textbf{CMU-MOSI Sentiment}}    \\ \hline \hline
EF-LSTM        & 33.7  & 75.3  & 75.2  & 1.023 & 0.608 \\
LF-LSTM        & 35.3     & 76.8  & 76.7  & 1.015 & 0.625 \\
RMFN~\cite{liang2018multimodal}     & 38.3  & 78.4  & 78.0  & 0.922 & 0.681 \\
MFM~\cite{tsai2019learning}         & 36.2  & 78.1  & 78.1  & 0.951 & 0.662 \\
RAVEN~\cite{wang2018words}          & 33.2     & 78.0  & 76.6    & 0.915 & \textbf{0.691} \\
MCTN~\cite{pham2019found}           & 35.6     & 79.3  & 79.1  & 0.909 & 0.676 \\ \hline \hline
MulT (ours)    & \bf{40.0}     & \bf{83.0}  & \bf{82.8}  & \bf{0.871} & \bf{0.698} \\ \hline \hline
\multicolumn{6}{c}{\textcolor{red}{(Unaligned)} \textbf{CMU-MOSI Sentiment}} \\ \hline \hline
CTC~\citep{graves2006connectionist} + EF-LSTM  & 31.0     & 73.6  & 74.5  & 1.078 & 0.542 \\
LF-LSTM        & 33.7     & 77.6  & 77.8  & 0.988 & 0.624 \\
CTC + MCTN~\cite{pham2019found}     & 32.7     & 75.9  & 76.4  & 0.991 & 0.613 \\
CTC + RAVEN~\cite{wang2018words}    & 31.7     & 72.7  & 73.1  & 1.076 & 0.544 \\
\hline \hline
MulT (ours)    & \bf{39.1}     & \bf{81.1}  & \bf{81.0}  & \bf{0.889} & \bf{0.686}  \\
\hline \hline
\end{tabular}
\label{tbl:mosi}
\end{table}
\newcolumntype{K}[1]{>{\centering\arraybackslash}p{#1}}
\begin{table}[t!]
\fontsize{9}{11}\selectfont
\centering
\setlength\tabcolsep{4.0pt}
\caption{Results for multimodal sentiment analysis on (relatively large scale) CMU-MOSEI with aligned and non-aligned multimodal sequences.}
\begin{tabular}{c||*{5}{K{1.4cm}}}
\hline \hline
Metric         & Acc$_7$ $\uparrow $ & Acc$_2$ $\uparrow $ & F1 $\uparrow $ & MAE $\downarrow $ & Corr $\uparrow $  \\ \hline \hline
\multicolumn{6}{c}{\textcolor{blue}{(Word Aligned)} \textbf{CMU-MOSEI Sentiment}}     \\ \hline \hline
EF-LSTM        & 47.4  & 78.2  & 77.9  & 0.642 & 0.616 \\
LF-LSTM        & 48.8    & 80.6  & 80.6  & 0.619 & 0.659 \\
Graph-MFN~\cite{zadeh2018multimodal}           & 45.0  & 76.9  & 77.0  & 0.71 & 0.54 \\
RAVEN~\cite{wang2018words}          & 50.0     & 79.1  & 79.5     & 0.614 & 0.662 \\ 
MCTN~\cite{pham2019found}           & 49.6     & 79.8  & 80.6  & 0.609 & 0.670 \\ \hline \hline
MulT (ours)           & \bf{51.8}     & \bf{82.5}  & \bf{82.3}  & \bf{0.580} & \bf{0.703} \\ \hline \hline
\multicolumn{6}{c}{\textcolor{red}{(Unaligned)} \textbf{CMU-MOSEI Sentiment}} \\ \hline \hline
CTC~\citep{graves2006connectionist} + EF-LSTM  & 46.3  &  76.1  & 75.9  & 0.680 & 0.585   \\
LF-LSTM        & 48.8     & 77.5  & 78.2    &  0.624 & 0.656 \\
CTC + RAVEN~\cite{wang2018words}    & 45.5     & 75.4  & 75.7  & 0.664 & 0.599 \\ 
CTC + MCTN~\cite{pham2019found}     & 48.2     & 79.3  & 79.7  & 0.631 &  0.645 \\ \hline \hline
MulT (ours)           & \textbf{50.7}     & \textbf{81.6}  & \textbf{81.6}  & \textbf{0.591} & \textbf{0.694} \\
\hline \hline
\end{tabular}
\label{tbl:mosei}
\end{table}
\newcolumntype{K}[1]{>{\centering\arraybackslash}p{#1}}
\begin{table*}[t!]
\fontsize{9}{11}\selectfont
\centering
\setlength\tabcolsep{1pt}
\caption{Results for multimodal emotions analysis on IEMOCAP with aligned and non-aligned multimodal sequences.}
\begin{tabular}{c||*{8}{K{1cm}}}
\hline \hline
Task          & \multicolumn{2}{c}{Happy} & \multicolumn{2}{c}{Sad} & \multicolumn{2}{c}{Angry} & \multicolumn{2}{c}{Neutral} \\
Metric        & Acc$_2$ $\uparrow $        & F1$\uparrow $          & Acc$_2$ $\uparrow $       & F1 $\uparrow $       & Acc$_2$ $\uparrow $       & F1 $\uparrow $      & Acc$_2$ $\uparrow $         & F1 $\uparrow $         \\ \hline \hline
\multicolumn{9}{c}{\textcolor{blue}{(Word Aligned)} \textbf{IEMOCAP Emotions}} \\ \hline  \hline
EF-LSTM       & 86.0        & 84.2        & 80.2       & 80.5       & 85.2        & 84.5        & 67.8          & 67.1         \\
LF-LSTM       & 85.1        & 86.3        & 78.9       & 81.7       & 84.7        & 83.0        & 67.1          & 67.6         \\
RMFN~\cite{liang2018multimodal}          & 87.5        & 85.8        & 83.8       & 82.9       & 85.1        & 84.6        & 69.5          & 69.1         \\
MFM~\cite{tsai2019learning}           & 90.2        & 85.8        & \textbf{88.4}    & \textbf{86.1}       & \textbf{87.5}        & 86.7        & 72.1          & 68.1         \\
RAVEN~\cite{wang2018words}         & 87.3        & 85.8        & 83.4       & 83.1     &  \textbf{87.3}        & 86.7        & 69.7          & 69.3         \\
MCTN~\cite{pham2019found}          & 84.9          & 83.1           & 80.5        & 79.6          & 79.7          & 80.4           & 62.3            & 57.0            \\ \hline  \hline
MulT (ours)          & \bf{90.7}        & \bf{88.6}        & 86.7       & \textbf{86.0}       & \textbf{87.4}        & \textbf{87.0}        & \textbf{72.4}          & \bf{70.7}         \\ \hline  \hline
\multicolumn{9}{c}{\textcolor{red}{(Unaligned)} \textbf{IEMOCAP Emotions}} \\ \hline  \hline
CTC~\cite{graves2006connectionist} + EF-LSTM  &  76.2  &  75.7  & 70.2  & 70.5  & 72.7  &  67.1  &  58.1 &  57.4 \\
LF-LSTM       & 72.5  &  71.8  &  72.9  &  70.4   &  68.6  &  67.9  &   59.6   &   56.2     \\
CTC + RAVEN~\cite{wang2018words}  &  77.0  &  76.8  & 67.6  &  65.6  & 65.0  & 64.1  &  \bf{62.0}  &  \bf{59.5} \\
CTC + MCTN~\cite{pham2019found}   &  80.5  & 77.5  & 72.0  & 71.7 &   64.9  &  65.6  &  49.4  &  49.3 \\  \hline \hline
MulT (ours)         & \bf{84.8}        & \bf{81.9}        & \bf{77.7}       & \bf{74.1}       & \bf{73.9}        & \bf{70.2}        & \bf{62.5}          & \bf{59.7}      \\
\hline \hline
\end{tabular}
\label{tbl:iemocap}
\end{table*}
\newcolumntype{K}[1]{>{\centering\arraybackslash}p{#1}}
\definecolor{gg}{RGB}{45,190,45}

\begin{table*}[!htbp]
\fontsize{9}{11}\selectfont
\centering
\setlength\tabcolsep{1.3pt}
\caption{Results for personality trait recognition on POM. Best results are highlighted in bold and $\Delta_{SOTA}$ shows improvement over previous SOTA. Symbols denote baseline model which achieves the reported performance: MFN: $\star$, MARN: $\S$, BC-LSTM: $\bullet$, TFN: $\dagger$, MV-LSTM: $\#$, EF-LSTM: $\flat$, RF: $\heartsuit$, SVM: $\times$. The \ours \ outperforms the current SOTA across all evaluation metrics except the $\Delta_{SOTA}$ entries highlighted in gray. Improvements are highlighted in green.}
\begin{tabular}{l | *{16}{K{1.15cm}}}
\hline \hline
Dataset & \multicolumn{16}{c}{\textbf{POM Speaker Personality Traits}} \\  \hline \hline
Task & \multicolumn{1}{c}{Con} & \multicolumn{1}{c}{Pas} & \multicolumn{1}{c}{Voi} & \multicolumn{1}{c}{Cre} & \multicolumn{1}{c}{Viv} & \multicolumn{1}{c}{Exp} & \multicolumn{1}{c}{Res} & \multicolumn{1}{c}{Rel} & \multicolumn{1}{c}{Tho} & \multicolumn{1}{c}{Ner} & \multicolumn{1}{c}{Per} & \multicolumn{1}{c}{Hum}\\
Metric & Acc$_7$ $\uparrow $ & Acc$_7$ $\uparrow $ & Acc$_7$ $\uparrow $ & Acc$_7$ $\uparrow $ & Acc$_7$ $\uparrow $ & Acc$_7$ $\uparrow $ & Acc$_5$ $\uparrow $ & Acc$_5$ $\uparrow $ & Acc$_5$ $\uparrow $ & Acc$_5$ $\uparrow $ & Acc$_7$ $\uparrow $ & Acc$_6$ $\uparrow $ \\ 
\hline
EF-LSTM	& 25.1& 30.5 & 34.0 & 36.9 & 29.6 & 32.5 & 31.0 & 48.3 & 42.4 & 44.8 & 25.6 & 39.4 \\
MV-LSTM 		& 25.6&28.6&28.1&25.6&32.5&29.6&33.0&50.7&37.9&42.4&26.1&38.9\\
BC-LSTM    		& 26.6&26.6&31.0&27.6&36.5&30.5&33.0&47.3&45.8&36.0&27.1&36.5\\ 
TFN      		& 24.1&31.0&31.5&24.6&25.6&27.6&30.5&35.5&33.0&42.4&27.6&33.0\\ 
MFN			& {34.5}&{35.5}&\textbf{37.4}&{34.5}&{36.9}&{36.0}&{38.4}&{53.2}&{47.3}&{47.8}&{34.0}&\textbf{47.3} \\
\hline 
{\pipelines} (ours)  &  \textbf{37.4} & \textbf{38.4} &\textbf{37.4} &\textbf{37.4} &\textbf{38.9} &\textbf{38.9} &\textbf{39.4} &\textbf{53.7} &\textbf{48.3} &\textbf{48.3} &\textbf{35.0} &{46.8}\\
\hline \hline 
\end{tabular}
\label{pom}
\vspace{-4mm}
\end{table*}

We compare to the following models for multimodal machine learning: {MFN} \citep{zadeh2018memory} synchronizes multimodal sequences using a multi-view gated memory. It is the current state of the art on CMU-MOSI and POM. {MARN} \citep{zadeh2018multi} models intra-modal and cross-modal interactions using multiple attention coefficients and hybrid LSTM memory components. {GME-LSTM(A)} \cite{chen2017multimodal} learns binary gating mechanisms to remove noisy modalities that are contradictory or redundant for prediction. {TFN} \cite{zadeh2017tensor} models unimodal, bimodal and trimodal interactions using tensor products. {BC-LSTM} \cite{contextmultimodalacl2017} performs context-dependent sentiment analysis and emotion recognition, currently state of the art on IEMOCAP. {EF-LSTM} concatenates the multimodal inputs and uses that as input to a single LSTM \citep{Hochreiter:1997:LSM:1246443.1246450}. We also implement the Stacked, ({EF-SLSTM}) \citep{6638947} Bidirectional ({EF-BLSTM}) \citep{Schuster:1997:BRN:2198065.2205129} and Stacked Bidirectional ({EF-SBLSTM}) LSTMs.

\subsection{Evaluation metrics}
For classification, we report accuracy A$c$ where $c$ denotes the number of classes and F1 score. For regression, we report Mean Absolute Error MAE and Pearson's correlation $r$. For MAE lower values indicate stronger performance. For all remaining metrics, higher values indicate stronger performance.

\section{Results and Discussion}

\subsection{Overall performance on multimodal language}
\label{results}

\paragraph{Word-Aligned Experiments.}
We first evaluate \pipelines\ and \mult\ on the \emph{word-aligned sequences}--- the ``home turf'' of prior approaches modeling human multimodal language~\cite{sheikh2018sentiment,tsai2019learning,pham2019found,wang2018words}. The upper part of the Table~\ref{tbl:mosi},~\ref{tbl:mosei},~\ref{tbl:iemocap}, and~\ref{pom} show the results of our proposed approaches and previous baselines on the word-aligned task. With similar model sizes (around 200K parameters), \mult\ outperforms the other competitive approaches on different metrics on all tasks, with the exception of the ``sad'' class results on IEMOCAP. We also observe that \pipelines\ does not improve results on IEMOCAP neutral emotion and the model outperforming RMFN is a memory-based fusion baseline~\citep{zadeh2018memory}. We believe that this is because neutral expressions are quite idiosyncratic. Some people may always look angry given their facial configuration (e.g., natural eyebrow raises of actor Jack Nicholson). In these situations, it becomes useful to compare the current image with a memorized or aggregated representation of the speaker's face. Our proposed multistage fusion approach can easily be extended to memory-based fusion methods.

\begin{figure}[t!]
    \centering
    \includegraphics[width=0.49\textwidth]{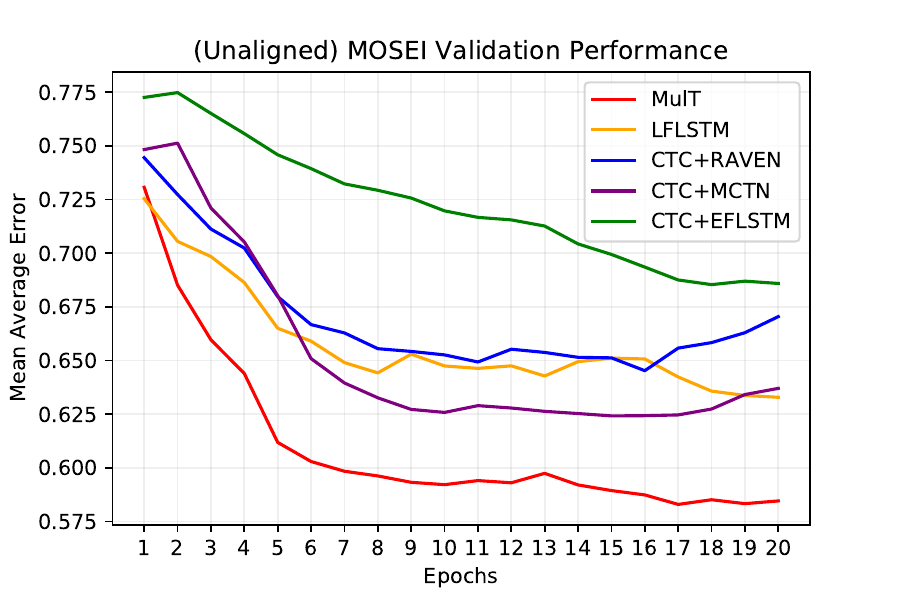}
    \caption{Validation set convergence of \mult\ when compared to other baselines on the \textcolor{red}{unaligned} CMU-MOSEI task.}
    \label{fig:cmu-mosei}
\end{figure}

\begin{figure*}[t!]
\centering
\includegraphics[width=.86\textwidth]{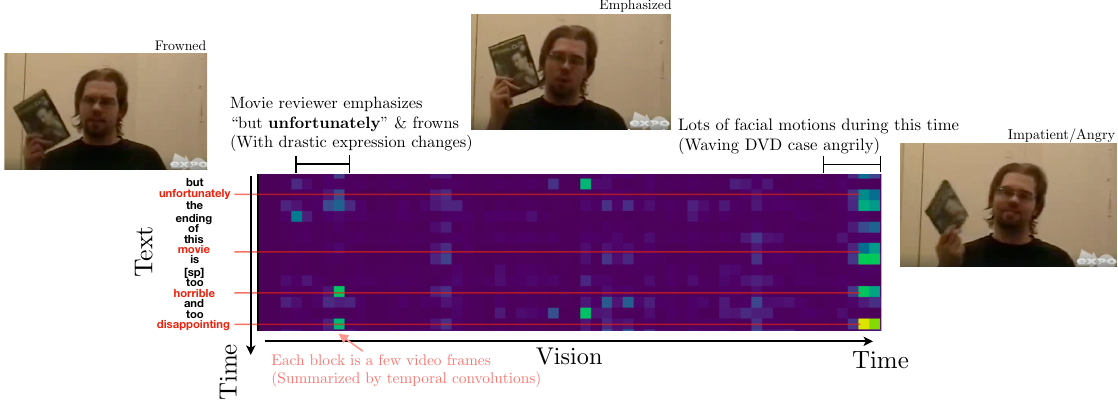} \\
\caption{Visualization of sample crossmodal attention weights from layer 3 of $[V \rightarrow L]$ crossmodal transformer on CMU-MOSEI. We found that the crossmodal attention has learned to correlate certain meaningful words (e.g., ``movie'', ``disappointing'') with segments of stronger visual signals (typically stronger facial motions or expression change), despite the lack of alignment between original $L/V$ sequences. Note that due to temporal convolution, each textual/visual feature contains the representation of nearby elements.}
\label{fig:attn}
\end{figure*}

\paragraph{Unaligned Experiments.} 
Next, we evaluate \mult\ on the same set of datasets in the unaligned setting. Note that \mult\ can be directly applied to unaligned multimodal stream, while the baseline models (except for LF-LSTM) require the need of additional alignment module (e.g., CTC module).

The results are shown in the bottom part of Table~\ref{tbl:mosi},~\ref{tbl:mosei}, and~\ref{tbl:iemocap}. On the three benchmark datasets, \mult\ improves upon the prior methods (some with CTC) by 10\%-15\% on most attributes. Empirically, we find that \mult\ converges faster to better results at training when compared to other competitive approaches (see Figure~\ref{fig:cmu-mosei}). In addition, while we note that in general there is a performance drop on all models when we shift from the word-aligned to unaligned multimodal time-series, the impact \mult\ takes is much smaller than the other approaches. We hypothesize such performance drop occurs because the asynchronous (and much longer) data streams introduce more difficulty in recognizing important features and computing the appropriate attention.

\subsection{Deeper analysis of \pipelines}

\newcolumntype{K}[1]{>{\centering\arraybackslash}p{#1}}
\definecolor{gg}{RGB}{45,190,45}

\begin{table}[t!]
\fontsize{9}{11}\selectfont
\centering
\setlength\tabcolsep{2.0pt}
\caption{Effect of varying the number of stages on CMU-MOSI sentiment analysis performance. Multistage fusion improves performance as compared to single stage fusion.}
\begin{tabular}{l : *{16}{K{1.2cm}}}
\hline \hline
Dataset & \multicolumn{5}{c}{\textbf{CMU-MOSI}} \\
Task & \multicolumn{5}{c}{Sentiment} \\
Metric       & A$2$ $\uparrow $ & F1 $\uparrow $ & A$7$ $\uparrow $ & MAE $\downarrow $ & Corr $\uparrow $\\ 
\hline
\pipelines-R1 \ \ \ \ \ \ \ \ \ \ \	& {75.5} & {75.5} & {35.1} & {0.997} & {0.653} \\
\pipelines-R2 \ \ \ \ \ \ \ \ \ \ \	& {76.4} & {76.4} & {34.5} & {0.967} & {0.642} \\
\pipelines-R3 \ \ \ \ \ \ \ \ \ \ \	& \textbf{78.4} & \textbf{78.0} & \textbf{38.3} & \textbf{0.922} & \textbf{0.681} \\
\pipelines-R4 \ \ \ \ \ \ \ \ \ \ \	& {76.0} & {76.0} & {36.0} & {0.999} & {0.640} \\
\pipelines-R5 \ \ \ \ \ \ \ \ \ \ \	& {75.5} & {75.5} & {30.9} & {1.009} & {0.617} \\
\pipelines-R6 \ \ \ \ \ \ \ \ \ \ \	& {70.4} & {70.5} & {30.8} & {1.109} & {0.560} \\
\hline
\pipelines	& \textbf{78.4} & \textbf{78.0} & \textbf{38.3} & \textbf{0.922} & \textbf{0.681} \\
\hline \hline
\end{tabular}
\label{length}
\vspace{-2mm}
\end{table}
\newcolumntype{K}[1]{>{\centering\arraybackslash}p{#1}}
\definecolor{gg}{RGB}{45,190,45}

\begin{table}[t!]
\fontsize{9}{11}\selectfont
\centering
\setlength\tabcolsep{2.0pt}
\caption{Comparison studies of \pipelines \ on CMU-MOSI. Modeling cross-modal interactions using multistage fusion and attention weights are crucial in multimodal language analysis.}
\begin{tabular}{l : *{16}{K{1cm}}}
\hline \hline
Dataset & \multicolumn{5}{c}{\textbf{CMU-MOSI}} \\
Task & \multicolumn{5}{c}{Sentiment} \\
Metric       & A$2$ $\uparrow $ & F1 $\uparrow $ & A$7$ $\uparrow $ & MAE $\downarrow $ & Corr $\uparrow $\\ 
\hline
MARN	& 77.1	& 77.0 & 34.7 & 0.968  & 0.625 \\
{\pipelines } (no \ours)  & {76.5} & {76.5} &  30.8	&  {0.998} & 0.582 \\ 
\pipelines \ (no \texttt{HIGHLIGHT})   & {77.9} & {77.9} & {35.9} & {0.952} & {0.666} \\
\hline
{\pipelines}      		& \textbf{78.4} & \textbf{78.0} & \textbf{38.3} & \textbf{0.922} & \textbf{0.681} \\
\hline \hline
\end{tabular}
\label{ablation}
\vspace{-4mm}
\end{table}

\textbf{Ablation studies}: To achieve a deeper understanding of the multistage fusion process, we study five research questions. (Q1): whether modeling cross-modal interactions across multiple stages is beneficial. (Q2): the effect of the number of stages $K$ during multistage fusion on performance. (Q3): the comparison between multistage and independent modeling of cross-modal interactions. (Q4): whether modeling cross-modal interactions are helpful. (Q5): whether attention weights from the \texttt{HIGHLIGHT} module are required for modeling cross-modal interactions. 

\textbf{Q1}: To study the effectiveness of the multistage fusion process, we test the baseline \pipelines-R$1$ which performs fusion in only one stage instead of across multiple stages. This model makes the strong assumption that all cross-modal interactions can be modeled during only one stage. From Table~\ref{length}, \pipelines-R$1$ underperforms as compared to \pipelines \ which performs multistage fusion.

\textbf{Q2}: We test baselines \pipelines-R$K$ which perform $K$ stages of fusion. From Table~\ref{length}, we observe that increasing the number of stages $K$ increases the model's capability to model cross-modal interactions up to a certain point ($K=3$) in our experiments. Further increases led to decreases in performance and we hypothesize this is due to overfitting on the dataset.

\textbf{Q3}: To compare multistage against independent modeling of cross-modal interactions, we pay close attention to the performance comparison with respect to MARN which models multiple cross-modal interactions all at once (see Table~\ref{ablation}). \pipelines \ shows improved performance, indicating that multistage fusion is both effective and efficient for human multimodal language modeling.

\textbf{Q4}: \pipelines \ (no \ours) represents a system of LSTHMs without the integration of $\mathbf{z}_t$ from the \ours \ to model cross-modal interactions. From Table~\ref{ablation}, \pipelines \ (no \ours) is outperformed by \pipelines, confirming that modeling cross-modal interactions is crucial in analyzing human multimodal language. 

\textbf{Q5}: \pipelines \ (no \texttt{HIGHLIGHT}) removes the \texttt{HIGHLIGHT} module from \ours \ during multistage fusion. From Table \ref{ablation}, \pipelines \ (no \texttt{HIGHLIGHT}) underperforms, indicating that highlighting multimodal representations using attention weights are important for modeling cross-modal interactions.

\begin{figure*}[t!]
\centering{
\includegraphics[width=0.9\linewidth]{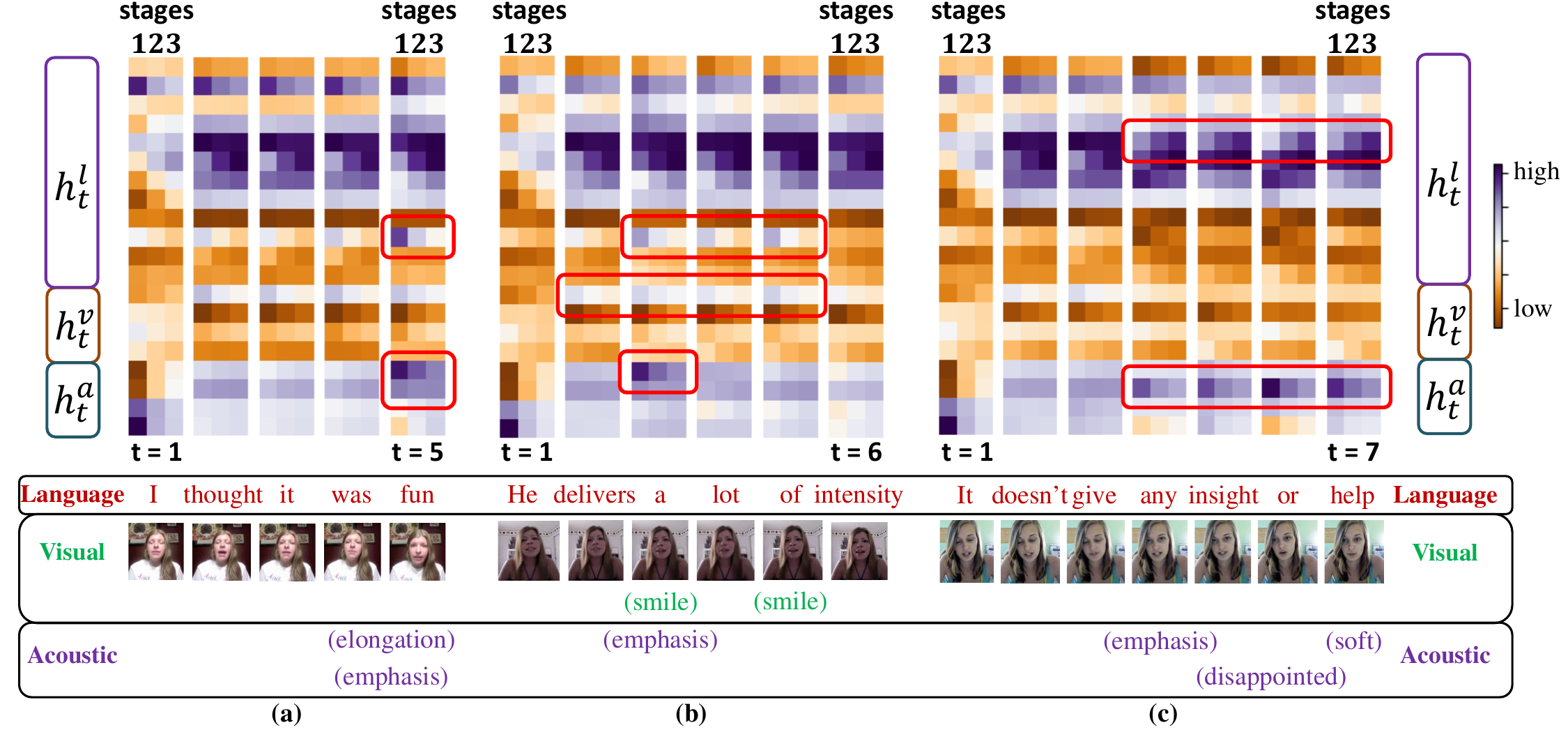}}
    \caption{Visualization of learned attention weights across stages 1,2 and 3 of the multistage fusion process and across time of the multimodal sequence. We observe that the attention weights are diverse and evolve across stages and time. In these three examples, the red boxes emphasize specific moments of interest. (a) Synchronized interactions: the positive word ``fun'' and the acoustic behaviors of emphasis and elongation ($t=5$) are synchronized in both attention weights for language and acoustic features. (b) Asynchronous trimodal interactions: the asynchronous presence of a smile ($t=2:5$) and emphasis ($t=3$) help to disambiguate the language modality. (c) Bimodal interactions: the interactions between the language and acoustic modalities are highlighted by alternating stages of fusion ($t=4:7$).}
    \vspace{-2mm}
     \label{fig:vis}
\end{figure*}

\textbf{Visualizations of learned fusion patterns}: Using an attention assignment mechanism during the \texttt{HIGHLIGHT} process gives more interpretability to the model since it allows us to visualize the attended multimodal signals at each stage and time step (see Figure~\ref{fig:vis}). Using \pipelines \ trained on the CMU-MOSI dataset, we plot the attention weights across the multistage fusion process for three videos in CMU-MOSI. Based on these visualizations we first draw the following general observations on multistage fusion:

\textbf{Across stages}: Attention weights change their behaviors across the multiple stages of fusion. Some features are highlighted by earlier stages while other features are used in later stages. This supports our hypothesis that \pipelines \ learns to specialize in different stages of the fusion process.

\textbf{Across time}: Attention weights vary over time and adapt to the multimodal inputs. We observe that the attention weights are similar if the input contains no new information. As soon as new multimodal information comes in, the highlighting mechanism in \pipelines \ adapts to these new inputs.

\textbf{Priors}: Based on the distribution of attention weights, we observe that the language and acoustic modalities seem the most commonly highlighted. This represents a prior over the expression of sentiment in human multimodal language and is closely related to the strong connections between language and speech in human communication~\citep{Kuhl11850}.

\textbf{Inactivity}: Some attention coefficients are not active (always orange) throughout time. We hypothesize that these corresponding dimensions carry only intra-modal dynamics and are not involved in the formation of cross-modal interactions.

In addition to the general observations above, Figure~\ref{fig:vis} shows three examples where multistage fusion learns cross-modal representations across three different scenarios.

\textbf{Synchronized Interactions}: In Figure~\ref{fig:vis}(a), the language features are highlighted corresponding to the utterance of the word ``fun'' that is highly indicative of sentiment ($t=5$). This sudden change is also accompanied by a synchronized highlighting of the acoustic features. We also notice that the highlighting of the acoustic features lasts longer across the 3 stages since it may take multiple stages to interpret all the new acoustic behaviors (elongated tone of voice and phonological emphasis).

\textbf{Asynchronous Trimodal Interactions}: In Figure~\ref{fig:vis}(b), the language modality displays ambiguous sentiment: ``delivers a lot of intensity'' can be inferred as both positive or negative. We observe that the circled attention units in the visual and acoustic features correspond to the asynchronous presence of a smile ($t=2:5$) and phonological emphasis ($t=3$) respectively. These nonverbal behaviors resolve ambiguity in language and result in an overall display of positive sentiment. We further note the coupling of attention weights that highlight the language, visual and acoustic features across stages ($t=3:5$), further emphasizing the coordination of all three modalities during multistage fusion despite their asynchronous occurrences.

\textbf{Bimodal Interactions}: In Figure~\ref{fig:vis}(c), the language modality is better interpreted in the context of acoustic behaviors. The disappointed tone and soft voice provide the nonverbal information useful for sentiment inference. This example highlights the bimodal interactions ($t=4:7$) in alternating stages: the acoustic features are highlighted more in earlier stages while the language features are highlighted increasingly in later stages.

\subsection{Deeper analysis of \mult}

\begin{table}[t!]
\fontsize{9}{11}\selectfont
\caption{An ablation study on the benefit of MulT's crossmodal transformers using CMU-MOSEI).}
\centering
\begin{tabular}{c|cccccc}
\hline \hline
                                 & \multicolumn{5}{c}{\textcolor{red}{(Unaligned)} CMU-MOSEI} \\
 Description                     & \multicolumn{5}{c}{Sentiment}           \\
                                 & Acc$_7^h$   & Acc$_2^h$  & F1$^h$  & MAE$^\ell$  & Corr$^h$  \\ \hline \hline
\multicolumn{6}{c}{Unimodal Transformers}       \\ \hline \hline
Language only                   &  46.5  &   77.4  &   78.2  &   0.653  &  0.631  \\
Audio only                      &  41.4  &   65.6   &   68.8  &   0.764  &  0.310  \\
Vision only                     &  43.5  &   66.4   &   69.3  &   0.759  &  0.343  \\ \hline \hline
\multicolumn{6}{c}{Late Fusion by using Multiple Unimodal Transformers}   \\ \hline \hline
LF-Transformer                     &  47.9  &   78.6   &   78.5  &   0.636  &  0.658 
\\ \hline \hline
\multicolumn{6}{c}{Temporally Concatenated Early Fusion Transformer}   \\ \hline \hline
EF-Transformer                     &  47.8  &   78.9   &   78.8  &   0.648  &  0.647 
\\ \hline \hline
\multicolumn{6}{c}{Multimodal Transfomers} \\ \hline \hline
Only $[V,A \rightarrow L]$ (ours)     &  \textbf{50.5}   &   80.1   &   80.4   &  0.605  & 0.670         \\
Only $[L,A \rightarrow V]$ (ours)     &  48.2  &  79.7  &  80.2   &  0.611    &  0.651  \\
Only $[L,V \rightarrow A]$ (ours)     &  47.5  &  79.2  &  79.7   &  0.620    &  0.648         \\
\multirow{2}{*}{\shortstack[c]{MulT mixing intermediate-\\level features (ours)}}       &  \multirow{2}{*}{50.3}   &   \multirow{2}{*}{80.5}   &   \multirow{2}{*}{80.6}   &  \multirow{2}{*}{0.602}   &  \multirow{2}{*}{0.674}  \\
& & & & & \\
MulT (ours)                         &  \textbf{50.7}   &   \textbf{81.6}   &   \textbf{81.6}   &  \textbf{0.591}   &  \textbf{0.691}  \\ \hline \hline
\end{tabular}
\label{tbl:abla}
\end{table}

\textbf{Ablation studies}: To further study the influence of the individual components in \mult, we perform comprehensive ablation analysis using the unaligned version of CMU-MOSEI. The results are shown in Table~\ref{tbl:abla}. 

First, we consider the performance for only using unimodal transformers (i.e., language, audio or vision only). We find that the language transformer outperforms the other two by a large margin. For example, for the Acc$_2^h$ metric, the model improves from $65.6$ to $77.4$ when comparing audio only to language only unimodal transformer. This fact aligns with the observations in prior work~\cite{pham2019found}, where the authors found that a good language network could already achieve good performance at inference time. 

Second, we consider 1) a late-fusion transformer that feature-wise concatenates the last elements of three self-attention transformers; and 2) an early-fusion self-attention transformer that takes in a temporal concatenation of three asynchronous sequences $[\hat{X}_L, \hat{X}_V, \hat{X}_A] \in \mathbb{R}^{(T_L + T_V + T_A) \times d_q}$ (see Section~\ref{subsec:overall}). Empirically, we find that both EF- and LF-Transformer (which fuse multimodal signals) outperform unimodal transformers. 

Finally, we study the importance of individual crossmodal transformers according to the target modalities (i.e., using $[V, A \rightarrow L]$, $[L, A \rightarrow V]$, or $[L, V \rightarrow A]$ network). As shown in Table \ref{tbl:abla}, we find crossmodal attention modules consistently improve over the late- and early-fusion transformer models in most metrics on unaligned CMU-MOSEI. In particular, among the three crossmodal transformers, the one where language($L$) is the target modality works best. We also additionally study the effect of adapting intermediate-level instead of the low-level features from source modality in crossmodal attention blocks (similar to the NMT encoder-decoder architecture but without self-attention; see Section \ref{subsec:cm-attn}). While \mult\ leveraging intermediate-level features still outperform models in other ablative settings, we empirically find adapting from low-level features works best. The ablations suggest that crossmodal attention concretely benefits \mult\ with better representation learning.

\textbf{Qualitative analysis of learned cross-modal attention}: To understand how crossmodal attention works while modeling unaligned multimodal data, we empirically inspect what kind of signals \mult\ picks up by visualizing the attention activations. Figure \ref{fig:attn} shows an example of a section of the crossmodal attention matrix on layer 3 of the $V \rightarrow L$ network of \mult\ (the original matrix has dimension $T_L \times T_V$; the figure shows the attention corresponding to approximately a 6-sec short window of that matrix). We find that crossmodal attention has learned to attend to meaningful signals across the two modalities. For example, stronger attention is given to the intersection of words that tend to suggest emotions (e.g., ``movie'', ``disappointing'') and drastic facial expression changes in the video (start and end of the above vision sequence). This observation advocates one of the aforementioned advantage of \mult\ over conventional alignment (see Section \ref{subsec:advantage}): crossmodal attention enables \mult\ to directly capture potentially long-range signals, including those off-diagonals on the attention matrix.

\section{Conclusion}

This chapter proposed the \pipelinel\ (\pipelines) and Multimodal Transformer (\mult) architectures or analyzing human multimodal language. \pipelines\ which recursively decomposes the multimodal fusion problem into multiple stages, each focused on learning interactions from a subset of attended multimodal signals. \mult\ uses the crossmodal attention module to learn multimodal interactions between all elements in the first modality with all elements in the second modality. As a result, all multimodal interactions across the entire sequence are learned simultaneously, and can be parallelized efficiently over GPUs.

Both methods show strong results on multiple datasets with multimodal temporal data (e.g., human communication), displaying capabilities to capture long-range multimodal interactions, handling unaligned multimodal data, and learning redundant, unique, and synergistic interactions.

\chapter{\mbox{Training High-modality Foundation Models}}
\label{chap:models3}
\renewcommand{\namel}{\textsc{High-Modality Multimodal Transformer}}
\renewcommand{\names}{\textsc{HighMMT}}

\vspace{-4mm}
\section{Introduction}
\vspace{-2mm}

\begin{wrapfigure}{r}{0.6\textwidth}
\centering
\includegraphics[width=\linewidth]{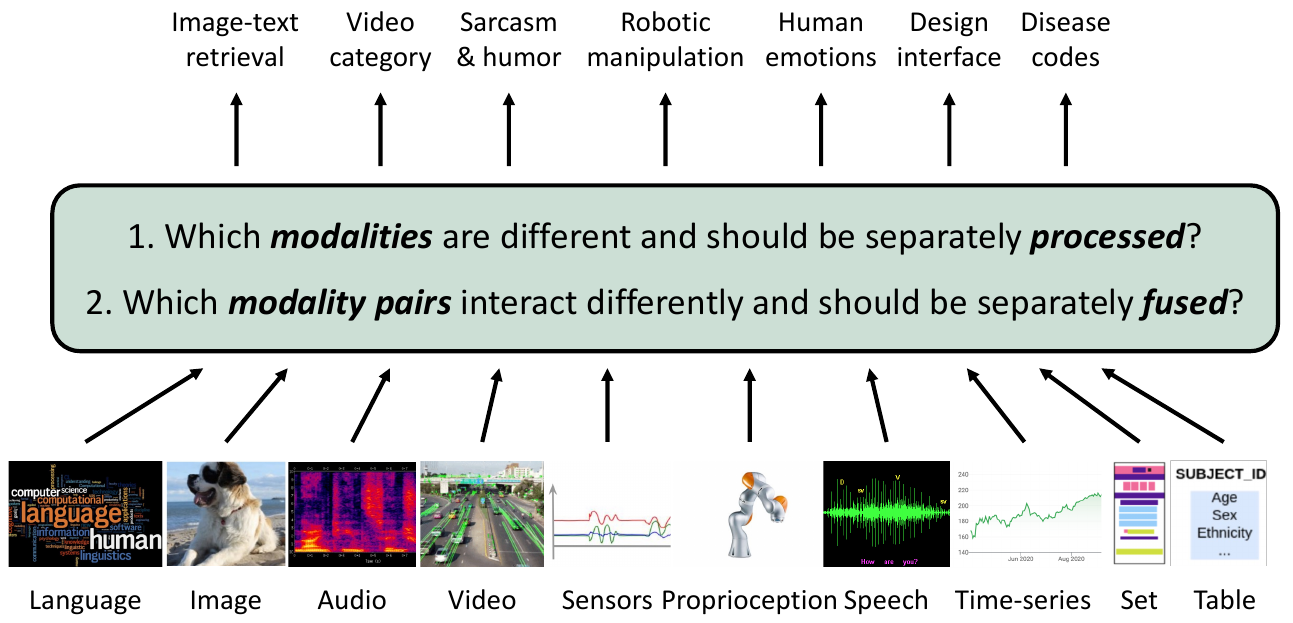}
\caption{\textbf{Heterogeneity quantification}: Efficiently learning from many modalities requires measuring (1) \textit{modality heterogeneity}: which modalities are different and should be separately processed, and (2) \textit{interaction heterogeneity}: which modality pairs interact differently and should be separately fused. \names\ uses these measurements to dynamically group parameters balancing performance and efficiency.}
\label{fig:overview}
\end{wrapfigure}

Finally, using \multibench, we scale multimodal transformers to the high-modality setting where there are a large number of modalities partially observed for different tasks~\cite{liang2022highmmt}. While there have been impressive advances in modeling language, vision, and audio~\citep{agrawal2017vqa,ramesh2021zero}, advances in sensing technologies have resulted in many real-world platforms such as cellphones, smart devices, self-driving cars, healthcare technologies, and robots now integrating a much larger number of sensors such as time-series, proprioception, sets, tables, and high-frequency sensors~\citep{medical,lee2019making,leiva2020enrico,liang2021learning,belpaeme2018social,yeong2021sensor}.
This new setting of \textit{high-modality learning} involves learning representations over many diverse modality inputs. As more modalities are introduced, adding new model parameters for every new modality or task~\citep{tsai2019multimodal,jayakumar2020multiplicative,lu2019vilbert} becomes prohibitively expensive and not scalable~\citep{liang2022foundations}. A critical technical challenge for efficient high-modality learning, therefore, is \textit{heterogeneity quantification}: how can we measure which modalities encode \textit{similar information} and \textit{similar interactions} in order to permit parameter sharing with previous modalities (see Figure~\ref{fig:overview})? For example, how can one determine whether the same modality encoder can be shared when processing language and speech, or that the same fusion network can be shared when fusing human speech and gestures as well as robot visual and force sensors?

In this paper, we propose a principled approach for heterogeneity quantification via modality information transfer, an approach that measures the amount of transferable information from one modality to another. Our first proposed metric, (1) \textit{modality heterogeneity} studies how similar $2$ modalities $\{X_1,X_2\}$ are by measuring how much usable information can be transferred from $X_1$ to $X_2$, and our second metric, (2) \textit{interaction heterogeneity} studies how similarly $2$ modality pairs $\{X_1,X_2\}, \{X_3,X_4\}$ interact by measuring how much usable interaction information can be transferred from $\{X_1,X_2\}$ to $\{X_3,X_4\}$.
We show the importance of these $2$ proposed metrics in high-modality scenarios as a way to automatically prioritize the fusion of modalities that contain unique information or unique interactions, and otherwise sharing parameters across similar modalities displaying similar information or interactions.

Operationalizing these ideas on a suite of $10$ modalities, $15$ prediction tasks, and $5$ research areas, we show how to train a single model, \names, that (1) improves the tradeoff between performance and efficiency over task-specific state-of-the-art models~\citep{liang2021multibench,jayakumar2020multiplicative}, and general multimodal models with full parameter sharing~\citep{jaegle2021perceiver,hu2021transformer,akbari2021vatt,reed2022generalist}, (2) enables cross-modal transfer by pretraining on source tasks before transferring to new target modalities and tasks, and (3) is especially beneficial for low-resource scenarios (less training data and partially-observable modalities).
Beyond these empirical results, we believe that our insights on quantifying heterogeneity and information sharing in multimodal models are independently useful for future work.

\vspace{-2mm}
\section{\namel}
\label{sec:model}
\vspace{-2mm}

In this section, we describe our overall approach for high-modality representation learning (see Figure~\ref{fig:model}).
In \S\ref{sec:hetero}, we formalize modality and interaction heterogeneity to understand whether modalities should be processed similarly or differently. Using these insights, \S\ref{sec:design} describes our proposed \names\ model with dynamic parameter sharing based on heterogeneity measurements.

\subsection{Measuring heterogeneity via modality information transfer}
\label{sec:hetero}

We begin our motivation by formalizing two important sources of heterogeneity in multimodal tasks. Firstly, \textit{modality heterogeneity} occurs because the information present in different modalities often shows diverse qualities, structures, and representations. Secondly, \textit{interaction heterogeneity} occurs because different modalities interact differently to give rise to new information when used for task inference. Formalizing and measuring these two sources of heterogeneity results in actionable insights for building multimodal models: measuring modality heterogeneity enables us to answer: should I use the same unimodal model to encode $X_1$ and $X_2$? Measuring interaction heterogeneity enables us to answer: should I use the same fusion model to fuse $\{X_1,X_2\}$ and $\{X_3,X_4\}$? We will formalize heterogeneity via \textit{modality transfer}, an approach that measures the amount of transferable information from one modality to another.

\begin{figure*}[tbp]
\centering
\includegraphics[width=0.9\linewidth]{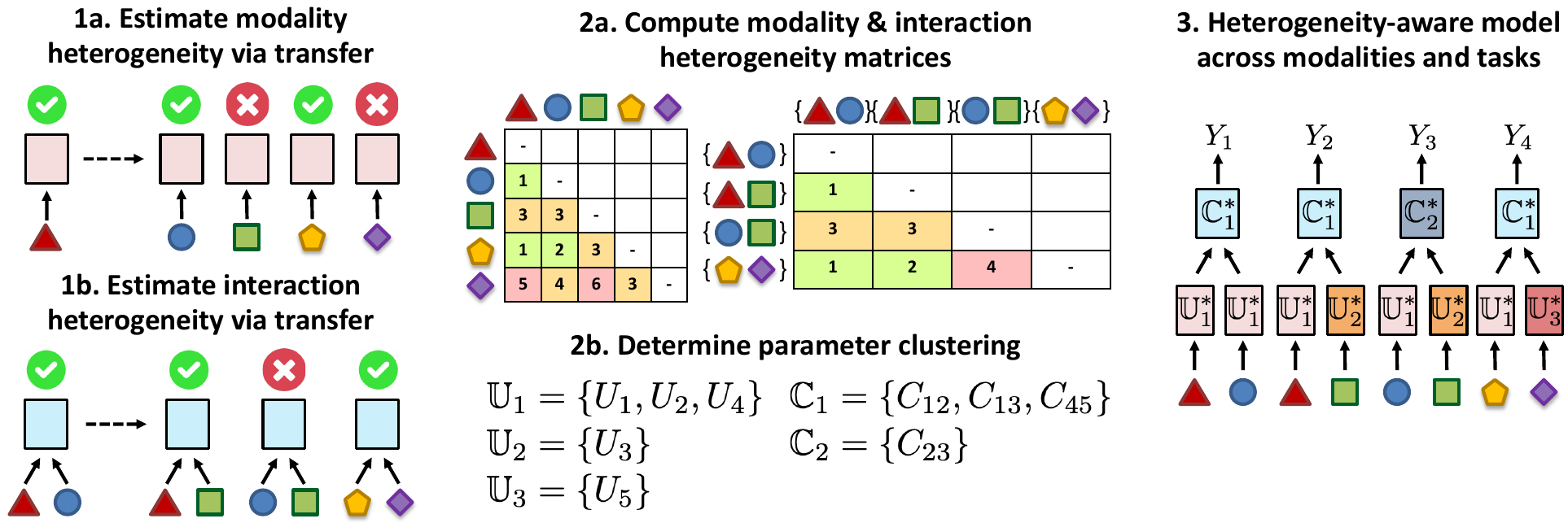}
\caption{\textbf{\names\ workflow}: (1) We estimate modality and interaction heterogeneity via modality transfer to determine which modalities should be processed and fused differently. (2) Using the inferred heterogeneity, we determine the optimal grouping of parameters balancing both total performance and parameter efficiency, which (3) informs our design of a heterogeneity-aware model with dynamic parameter sharing across many modalities and tasks. \names\ enables statistical strength sharing, efficiency, and generalization to new modalities and tasks.}
\label{fig:model}
\vspace{-2mm}
\end{figure*}

\textbf{Estimating modality heterogeneity via unimodal information transfer.} We propose to measure heterogeneity between modalities $X_1$ and $X_2$ via unimodal transfer. Given a task $Y$ defined over $X_1$ and $X_2$, how well does an unimodal model trained on the task $(X_1;Y)$ transfer to $(X_2;Y)$? We choose model transfer as our focus of heterogeneity since it is captured at the level of features extracted via representation learning, rather than at the data-level. Even though the input data may be very different (e.g., images from different cameras or paraphrased sentences), effective feature extractors may be able to learn similar representations from them. Furthermore, it directly models task-relevance: the degree of heterogeneity depends on the end task, which enables using these heterogeneity measures subsequently for end-task optimization.

We formalize unimodal transfer as the difference in performance between unimodal models trained on $X_1$ before transfer to $X_2$, versus those trained directly on $X_2$. Specifically, we represent an unimodal model using modality $X_2$ with parameters $\theta$ as $\hat{y} = f(y|x_2;\theta)$. For a suitably chosen loss function $\ell(\hat{y},y)$, define the loss of a model as $\mathbb{E}_{p(x_2,y)} \ell(f(y|x_2;\theta), y)$ which measures the expected error over the joint distribution $p(x_2,y)$. To measure transfer, we train $2$ models to obtain an approximation of task performance: the first randomly initialized and trained on the target task giving loss $\mathcal{L}_2^*$,
\begin{align}
    \mathcal{L}_2^* &= \min_\theta \mathbb{E}_{p(x_2,y)} \ell(f(y|x_2;\theta), y),
\end{align}
and the second using initialization from model parameters $\theta_{1}$ trained on the source task $(X_1;Y)$ before fine-training on the target task giving loss $\mathcal{L}_{1 \rightarrow 2}^*$.
\begin{align}
    \theta_1 &= \argmin_\theta \mathbb{E}_{p(x_1,y)} \ell(f(y|x_1;\theta), y), \\
    \mathcal{L}_{1 \rightarrow 2}^* &= \min_\theta \mathbb{E}_{p(x_2,y)} \ell(f(y|x_2;\theta \leftarrow \theta_1), y),
\end{align}
where $\theta \leftarrow \theta_1$ denotes parameter initialization with $\theta_{1}$. Intuitively, $\mathcal{L}_2^*$ measures the (baseline) task-relevant information in $X_2$, while $\mathcal{L}_{1 \rightarrow 2}^*$ measures the task-relevant information transferable from $X_1$ to $X_2$. The differences between these $2$ losses,
\begin{align}
    T(X_1 \rightarrow X_2; Y) = \mathcal{L}_{1 \rightarrow 2}^* - \mathcal{L}_{2}^*,
\end{align}
therefore measures the difficulty of transferring a model trained on the source task $(X_1;Y)$ to a target task $(X_2;Y)$. Note that computing $T(X_1 \rightarrow X_2; Y)$ only requires the training or fine-tuning of $2$ models across the source and target modalities, which is efficient. In practice, the expectations over $p(x_1,y)$ and $p(x_2,y)$ are approximated using empirical samples from the training set (for model fine-tuning) and validation dataset (for final evaluation of performance).

What are some properties of $T(X_1 \rightarrow X_2; Y)$? For very different modalities $X_1$ and $X_2$, we typically expect a source task $(X_1,Y)$ to contain less usable information for a target task $(X_2;Y)$, which would imply that $\mathcal{L}_{1 \rightarrow 2}^* \ge \mathcal{L}_{2}^*$ and therefore $T(X_1 \rightarrow X_2; Y) \ge 0$ (i.e., positive distance). This is consistent with work demonstrating negative transfer across different modalities~\citep{liang2021multibench,liang2021cross,tulving1974negative,wang2019characterizing}. Under these scenarios, the larger the positive magnitude of $T(X_1 \rightarrow X_2; Y)$, the more different modalities $X_1$ and $X_2$ are in the context of task $Y$ (more difficult to transfer). However, there can also be cases of zero or even positive transfer (i.e., $T(X_1 \rightarrow X_2; Y) \le 0$), even in the surprising case of very different modalities~\citep{lu2021pretrained}. These cases reinforce the benefits of feature-based approaches to measure heterogeneity: while the raw modalities themselves seem very different, they can still be processed by similar models resulting in positive transfer, and should be assigned a difference of $0$. Our final heterogeneity measure $d(X_1; X_2)$ aggregates the non-negative value (to account for positive transfer) of transfer difficulty statistics across both transfer directions $X_1 \rightarrow X_2$ and $X_2 \rightarrow X_1$:
\begin{align}
    d(X_1; X_2) = T(X_1 \rightarrow X_2; Y)_{\ge0} + T(X_2 \rightarrow X_1; Y)_{\ge0}.
\end{align}
where $x_{\ge0} = \max(x,0)$. Under certain assumptions on the modalities and tasks, our modality heterogeneity measure $d(X_1; X_2)$ is a metric: it satisfies \textit{non-negativity}: $d(X_1; X_2) \ge 0$, with $d(X_1; X_2) = 0$ when $X_1=X_2$, and \textit{symmetry}: $d(X_1; X_2) = d(X_2; X_1)$, \textit{positivity}, $X_1 \neq X_2$ implies that $d(X_1; X_2) > 0$, and a relaxed version of the \textit{triangle inequality}: $d(X_1; X_3) \le d(X_1; X_2) + d(X_2; X_3)$. However, in the most general case, there may be settings where positivity and the triangle inequality are not satisfied since the exact dynamics of transfer learning is still not well understood for general deep networks: positive transfer can happen (which would imply cases of $X_1 \neq X_2$ but $d(X_1; X_2) = 0$), and in practice, the relaxed triangle inequality is satisfied $96\%$ of the time from a real heterogeneity matrix in Figure~\ref{fig:matrix}.

\textbf{Estimating interaction heterogeneity via crossmodal information transfer.} We are also interested in interaction heterogeneity: specifically, how differently should I fuse modalities $\{X_1,X_2\}$ versus $\{X_3,X_4\}$? We therefore extend to crossmodal transfer by comparing the difference in performance between a multimodal model pretrained on $(X_1,X_2;Y)$ before transfer to $(X_3,X_4;Y)$, versus those trained directly on the target task $(X_3,X_4;Y)$. In other words, we measure the difference in loss between
\begin{align}
    \theta_{12} &= \argmin_\theta \mathbb{E}_{p(x_1,x_2,y)} \ell(f(y|x_1,x_2;\theta), y), \\
    \mathcal{L}_{12 \rightarrow 34}^* &= \min_\theta \mathbb{E}_{p(x_3,x_4,y)} \ell(f(y|x_3,x_4;\theta \leftarrow \theta_{12}), y),
\end{align}
and direct training
\begin{align}
    \mathcal{L}_{34}^* &= \min_\theta \mathbb{E}_{p(x_3,x_4,y)} \ell(f(y|x_3,x_4;\theta), y),
\end{align}
to obtain $T(X_1,X_2 \rightarrow X_3,X_4; Y) = \mathcal{L}_{12 \rightarrow 34}^* - \mathcal{L}_{34}^*$. The distance $d(X_1, X_2; X_3, X_4)$ after aggregation over tasks and transfer directions estimates the interaction heterogeneity between $\{X_1,X_2\}$ and $\{X_3,X_4\}$.

\textbf{Modality and interaction heterogeneity matrix.} Finally, we construct a modality heterogeneity matrix $M_U(i,j) = d(X_i; X_j)$ and an interaction heterogeneity matrix (technically $4$D-tensor) $M_C(i,j,k,\ell) = d(X_i, X_j; X_k, X_\ell)$. Determining parameter groupings to balance both total performance and parameter efficiency can be solved via agglomerative hierarchical clustering where modalities are nodes and heterogeneity measurements are edges. The number of clusters $k$ is treated as a hyperparameter dependent on the parameter budget (see clustering examples in \S\ref{sec:groups}). Clustering on the modality heterogeneity matrix $M_U$ results in a grouping of modalities based on similarity (e.g., $\mathcal{U}_1 = \{X_1,X_2,X_4\}, \mathcal{U}_2 = \{X_3\}, \mathcal{U}_3 = \{X_5\}$), and likewise for the crossmodal matrix $M_C$ (e.g., $\mathcal{C}_1 = \{ \{X_1,X_2\}, \{X_1,X_3\}, \{X_4,X_5\} \}, \mathcal{C}_2 = \{ \{X_2,X_3\}, \mathcal{C}_3 = \{ \{X_4,X_6\}, \{X_5,X_6\} \}$, and so on.

\textbf{Computational complexity.} In a high-modality setting, suppose we are given a suite of modalities and tasks of the form $\{(X_1,X_2,Y_1), (X_1,X_3,X_4,Y_2), ...\}$ and so on, where there are a total of $M$ unique modality and task pairs $\{(X_1,Y_1), (X_2, Y_1), (X_1,Y_2), (X_3,Y_2), (X_4,Y_2), ...\}$. In practice, the number of unique (pairwise) interaction and task pairs $\{(X_1,X_2,Y_1), (X_1,X_3,Y_2), ...\}$ is also $O(M)$, since the maximum number of modalities jointly observed for a task is never above a constant (at most $4$ in all real-world datasets, and often $2$ or $3$). As an example in Figure~\ref{fig:matrix}, our experiments involve $M=10$ modality and task pairs (across $4$ tasks defined on $2,2,3$ and $3$ modalities respectively), and $8 = {2 \choose 2} + {2 \choose 2} + {3 \choose 2} + {3 \choose 2}$ interaction and task pairs.

The modality heterogeneity matrix for $M$ unique modality and task pairs has $M(M-1)/2$ unique entries after removing the upper triangular portion due to symmetry and diagonal entries since $d(X_i,X_i) = 0$. Computing these $M(M-1)/2$ entries exactly requires one to first train $M$ unimodal models (to estimate the $M$ $\mathcal{L}_{m}^*$ terms) before fine-tuning $M(M-1)$ transfer models (to estimate the $M(M-1)$ $\mathcal{L}_{m \rightarrow n}^*$ terms), for a total of $M^2$ pre-trained and fine-tuned models. The interaction heterogeneity matrix also requires $O(M^2)$ models for exact computation. However, we find that a key approximation can be made in practice: the heterogeneity matrices are highly structured due to distances approximately satisfying the triangle inequality, which implies that we do not need to compute all entries and instead rely on low-rank reconstruction from partial entries in practice. In our experiments, even using a low-rank approximation of $r=3$ is sufficient to approximate the entire matrix. This suggests that we do not need to exhaustively measure unimodal and interaction transfer between all modality pairs to enjoy the benefits of our proposed approach. Instead, running a random sample of $O(M)$ pairs of heterogeneity values, and imputing the rest of the heterogeneity matrix, is sufficient in practice. Please see an example heterogeneity quantification for real-world datasets in \S\ref{sec:groups}.

\subsection{Capturing heterogeneity and homogeneity in \names}
\label{sec:design}

Using these insights, we now describe our architecture for a general model \names\ suitable for high-modality representation across many modalities and tasks (see Figure~\ref{fig:gen}). Training the \names\ model consists of $2$ main steps (see Figure~\ref{fig:tune}): (1) \textit{homogeneous pre-training} of a fully shared model across all modalities, before (2) \textit{heterogeneity-aware fine-tuning} to respect modality and interaction heterogeneity.

\begin{figure}[t]
\centering
\includegraphics[width=0.95\linewidth]{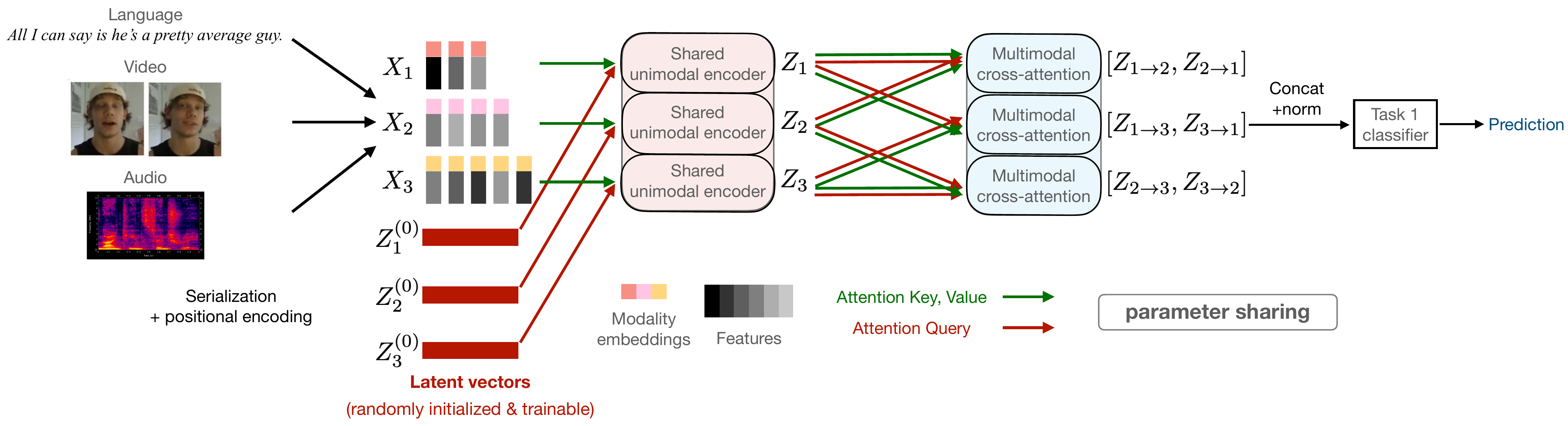}
\caption{\names\ architecture: Given arbitrary modalities, (1) the inputs are standardized into a sequence and padded, (2) modality embeddings and positional encodings are added to the input sequence, (3) a single shared unimodal Perceiver encoder is applied to all modalities to learn modality-agnostic representations, (4) each pair of unimodal representations is fed through a shared multimodal cross-attention layer to learn multimodal representations, and finally (5) all outputs are concatenated, batch-normalized, and fed into task-specific classification heads.}
\label{fig:gen}
\vspace{-4mm}
\end{figure}

\textbf{Homogeneous pre-training}. We first design a homogeneous multimodal model fully shared across all modalities and tasks with the following key components (see Figure~\ref{fig:gen})

\textit{1. Standardized input sequence}: We first standardize modalities as a sequence of embeddings, as is already done for sequential data such as text, audio, and time series, and recently adapted for image patches too~\citep{dosovitskiy2020image}. For tables, sets, and graphs we treat each element in the table/set/graph as an element in the sequence. The end result is a standardized input data $X_m$ of dimension $t_m \times d_m$, where $t_m$ is a modality and task-specific input sequence length, and $d_m$ is a modality and task-specific input dimension.

\textit{2. Modality-specific embedding and positional encoding.} For each distinct modality $m \in M$ (which may appear across multiple tasks), we define a one-hot modality embedding $\mathbf{e}_m \in \mathbb{R}^{|M|}$, where $|M|$ is the total number of distinct modalities, to identify common modalities across different tasks for information sharing. We also introduce Fourier feature positional encodings $\mathbf{p}_m \in \mathbb{R}^{t_m \times d_{pm}}$, where $d_{pm}$ is the positional encoding dimension, to capture positional information across each modality. For multimodal tasks where a common dimension is shared across time (e.g., videos/time series), we apply a common positional encoding to capture the common time dimension (i.e., the first image frame occurs at the same time as the first word and first audio segment). Finally, the processed modality $m$ is given by concatenating $X_m = X_m \oplus \mathbf{e}_m \oplus \mathbf{p}_m \oplus \mathbf{0}_m$ (i.e., the input sequence, modality embedding, positional encodings and zero-padding) into a standard dimension $t_m \times d_{all}$. $d_{all} = \max_{m\in M}(d_m+|M|+d_{pm})$ where $d_m$ is the channel size of modality $m$, $d_{pm}$ is the positional encoding size of modality $m$, and $|M|$ is the modality encoding size (i.e., the total number of involved modalities).

\textit{3. Shared unimodal networks.} Now that we have standardized all modalities into a common format, we design a general unimodal encoder with parameters $\mathbb{U}$ via a Transformer-based Perceiver block~\citep{jaegle2021perceiver}.
Our model recursively trains a latent array $Z_m$ of shape $d_{LN} \times d_{LS}$ (where $d_{LN}$ is the sequence length/number of latent vectors and $d_{LS}$ is the latent dimension) that is random initialized as $Z_m^{(0)}$. For each layer $L$ starting with a previously-computed representation $Z_m^{(L-1)}$, we first perform cross-attention from the processed input ($X_m$ of shape $t_m \times d_{all}$) to $Z_m^{(L-1)}$ obtaining an intermediate representation $\tilde{Z}_m^{(L)}$, before self-attention and feed-forward layers on $\tilde{Z}_m^{(L)}$ resulting in a new representation $Z_m^{(L)}$ for input to the next layer:
\begin{align}
    \tilde{Z}_m^{(L)} &= \textsc{Cross Attention}(Z_m^{(L-1)}, X_m) = \textrm{softmax} \left( \frac{Q_c K_c^\top}{\sqrt{d_{LS}}} \right) V_c = \textrm{softmax} \left( \frac{Z_m^{(L-1)} W_{Q_c} W_{V_c}^\top X_m^\top}{\sqrt{d_{LS}}} \right) X_m W_{V_c}, \\
    Z_m^{(L)} &= \textsc{Self Attention}(\tilde{Z}_m^{(L)}) = \textrm{softmax} \left( \frac{Q_s K_s^\top}{\sqrt{d_{LS}}} \right) V_s = \textrm{softmax} \left( \frac{\tilde{Z}_m^{(L)} W_{Q_s} W_{V_s}^\top \tilde{Z}_m^{(L)\top}}{\sqrt{d_{LS}}} \right) \tilde{Z}_m^{(L)} W_{V_s},
\end{align}
with trainable cross-attention parameters $W_{Q_c} \in \mathbb{R}^{d_{LS} \times d_{LS}}, W_{K_c} \in \mathbb{R}^{d_{all} \times d_{LS}}, W_{V_c} \in \mathbb{R}^{d_{all} \times d_{LS}}$ and self-attention parameters $W_{Q_s} \in \mathbb{R}^{d_{LS} \times d_{LS}}, W_{K_s} \in \mathbb{R}^{d_{LS} \times d_{LS}}, W_{V_s} \in \mathbb{R}^{d_{LS} \times d_{LS}}$.
Repeating cross- and self-attention between the latent vector and the input modality summarizes the relationships between modality elements into the latent vector, resulting in a final unimodal representation $Z_m \in \mathbb{R}^{d_{LN} \times d_{LS}}$. Summarizing all information into a common $d_{LN} \times d_{LS}$ latent array regardless of the input shape $t_m \times d_{all}$ results in total runtime only linear with respect to the size of $t_m$ and $d_{all}$ which scales to high-modality scenarios.

\textit{4. Shared crossmodal networks.} To learn multimodal representations, we use a shared Crossmodal Transformer block with parameters $\mathbb{C}$~\citep{tsai2019multimodal,lu2019vilbert}. Given $2$ unimodal representations $Z_1$ and $Z_2$ of common shape $d_{LN} \times d_{LS}$ learned from unimodal Perceiver encoders, a Crossmodal Transformer (CT) block uses crossmodal self-attention by setting the input layer query $Q = Z_1$ and keys and values $K,V = Z_2$ to learn attention from $Z_2$ to $Z_1$, and a separate block to capture the attention in the opposite direction.
\begin{align}
    Z_{2 \rightarrow 1} = \textsc{Cross Attention}(Z_1, Z_2) = \textrm{softmax} \left( \frac{Q_1 K_2^\top}{\sqrt{d_k}} \right) V_2 = \textrm{softmax} \left( \frac{Z_1 W_{Q_1} W_{V_2}^\top Z_2^\top}{\sqrt{d_k}} \right) Z_2 W_{V_2},
\end{align}
and vice-versa for $Z_{1 \rightarrow 2}$, with parameters $W_{Q_1},W_{Q_2} \in \mathbb{R}^{d_{LS} \times d_k}, W_{K_1},W_{K_2} \in \mathbb{R}^{d_{LS} \times d_k}, W_{V_1},W_{V_2} \in \mathbb{R}^{d_{LS} \times d_k}$. This step enables one modality's elements to discover bidirectional interactions with another, resulting in a final multimodal representation $Z_\textrm{mm} = \left[ Z_{1 \rightarrow 2}, Z_{2 \rightarrow 1} \right]$ of shape $d_{LS} \times 2 d_k$. For each layer, we first perform cross-attention followed by self-attention and feed-forward functions. For tasks with more than $2$ modalities, a Crossmodal Transformer block is applied for each pair of modalities before concatenating all representations.

\textit{5. Task-specific classifier and multitask pre-training.} Finally, on top of $Z_\textrm{mm}$, we use a separate linear classification layer per task. To enable information sharing across modalities and tasks, homogeneous pre-training is performed across a diverse set of datasets in a multitask manner by optimizing a weighted sum of losses over tasks. The result is a single set of shared unimodal parameters $\mathbb{U}^*$ that encodes all modalities, and a single set of shared crossmodal parameters $\mathbb{C}^*$ that captures all pairwise interactions between modality pairs, along with all modality-specific embeddings $\mathbb{E}^*$ and task-specific classifiers $\mathbb{T}^*$.

\begin{figure*}[tbp]
\centering
\vspace{-0mm}
\includegraphics[width=0.9\linewidth]{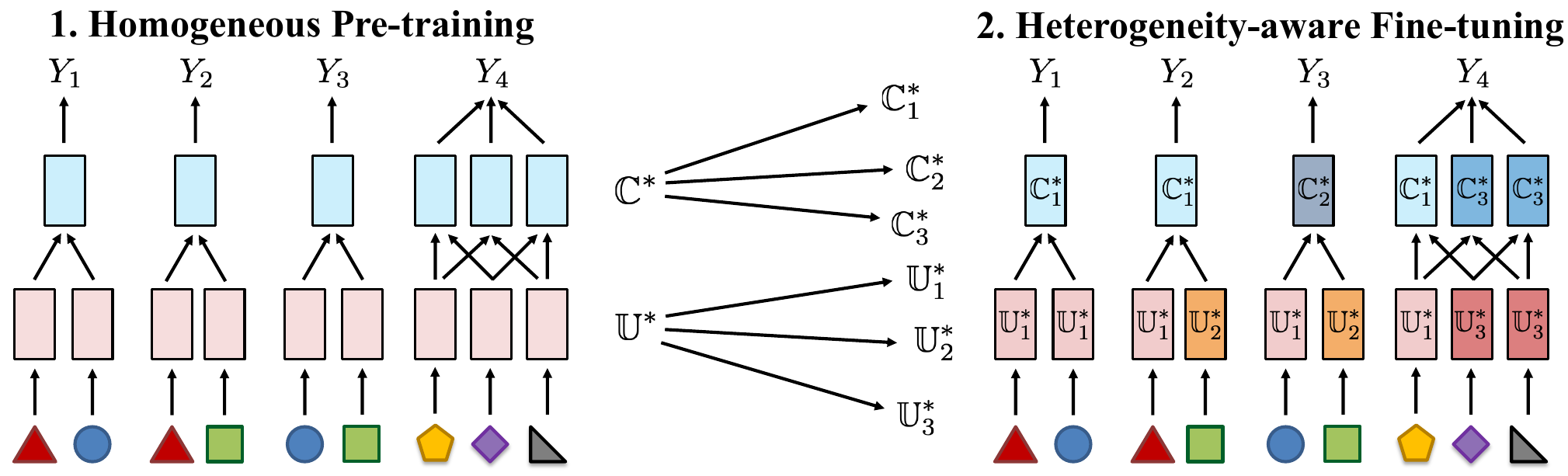}
\caption{\textbf{\names\ training} involves $2$ steps: (1) \textit{homogeneous pre-training} of a fully shared model across all modalities, before (2) \textit{heterogeneity-aware fine-tuning} of modality and interaction parameters in different groups to respect modality and interaction heterogeneity respectively.}
\label{fig:tune}
\end{figure*}

\textbf{Heterogeneity-aware fine-tuning.} Finally, we account for heterogeneity by grouping unimodal parameters based on modalities that we know to be similar from \S\ref{sec:hetero} (e.g., setting $\mathbb{U}_1 = \{U_1,U_2\}, \mathbb{U}_2 = \{U_3\}, \mathbb{U}_3 = \{U_4,U_5,U_6\}$), and likewise for the crossmodal parameters (e.g., $\mathbb{C}_1 = \{C_{12},C_{13},C_{14}\}, \mathbb{C}_2 = \{C_{23},C_{15}\}, \mathbb{C}_3 = \{C_{24},...\}$). From Figure~\ref{fig:tune}, these parameter groups are first initialized with the homogeneous model $\mathbb{U}^*$ and $\mathbb{C}^*$ before separate fine-tuning, which results in final parameters $\mathbb{U}^* \rightarrow \{\mathbb{U}_1^*, \mathbb{U}_2^*, ...\}$ and $\mathbb{C}^* \rightarrow \{\mathbb{C}_1^*, \mathbb{C}_2^*, ...\}$. The modality embeddings $\mathbb{E}^*$ and task classifiers $\mathbb{T}^*$ are jointly fine-tuned as well. Fine-tuning is also performed in a multitask manner by optimizing a weighted sum of supervised losses across all modalities and tasks.

\vspace{-2mm}
\section{Experiments}
\label{sec:experiments}
\vspace{-2mm}

\textbf{Setup}: In this section, we design experiments to analyze the multitask, transfer, and generalization capabilities of \names.
We use a large collection of multimodal datasets provided in MultiBench~\citep{liang2021multibench} spanning $10$ modalities, $15$ prediction tasks, and $5$ research areas. We trained $3$ multitask models across combinations of these datasets (see Table~\ref{tab:setup} for details).
Overall, the total size of datasets involved in our experiments exceeds $370,000$ and covers diverse modalities such as images, video, audio, text, time-series, robotics sensors, sets, and tables, prediction tasks spanning the image-caption matching, robot pose, object pose, robot contact, design interfaces, digits, humor, sentiment, emotions, mortality rate, and ICD-$9$ codes from the research areas of affective computing, healthcare, multimedia, robotics, and HCI.

\begin{table*}[]
\fontsize{9}{11}\selectfont
\setlength\tabcolsep{2.0pt}
\vspace{-0mm}
\caption{We investigate a multitask setup to evaluate the performance of \names\ across different modality inputs and prediction objectives. The total size of datasets involved in our experiments exceeds $370,000$ and covers diverse modalities, tasks, and research areas.}
\centering
\footnotesize
\vspace{1mm}
\begin{tabular}{l|ccccc}
\hline \hline
Datasets & Modalities & Size & Prediction task & Research Area \\
\Xhline{0.5\arrayrulewidth}
\textsc{ENRICO} & $\{ \textrm{image}, \textrm{set} \}$ & $1,460$ & design interface & HCI \\
\textsc{UR-FUNNY} & $\{ \textrm{text}, \textrm{video}, \textrm{audio} \}$ & $16,514$ & humor & Affective Computing \\
\textsc{MOSEI} & $\{ \textrm{text}, \textrm{video}, \textrm{audio} \}$ & $22,777$ & sentiment, emotions & Affective Computing \\
\textsc{MIMIC} & $\{ \textrm{time-series}, \textrm{table} \}$ & $36,212$ & mortality, ICD-$9$ codes & Healthcare \\
\textsc{Push} & $\{ \textrm{image}, \textrm{force}, \textrm{proprioception}, \textrm{control} \}$ & $37,990$ & object pose & Robotics \\
\textsc{AV-MNIST} & $\{ \textrm{image}, \textrm{audio} \}$ & $70,000$ & digit & Multimedia \\
\textsc{V\&T} & $\{ \textrm{image}, \textrm{force}, \textrm{proprioception}, \textrm{depth} \}$ & $147,000$ & contact, robot pose & Robotics \\
\hline \hline
\end{tabular}

\vspace{-4mm}
\label{tab:setup}
\end{table*}

\begin{figure*}[t]
\centering
\vspace{-0mm}
\includegraphics[width=\linewidth]{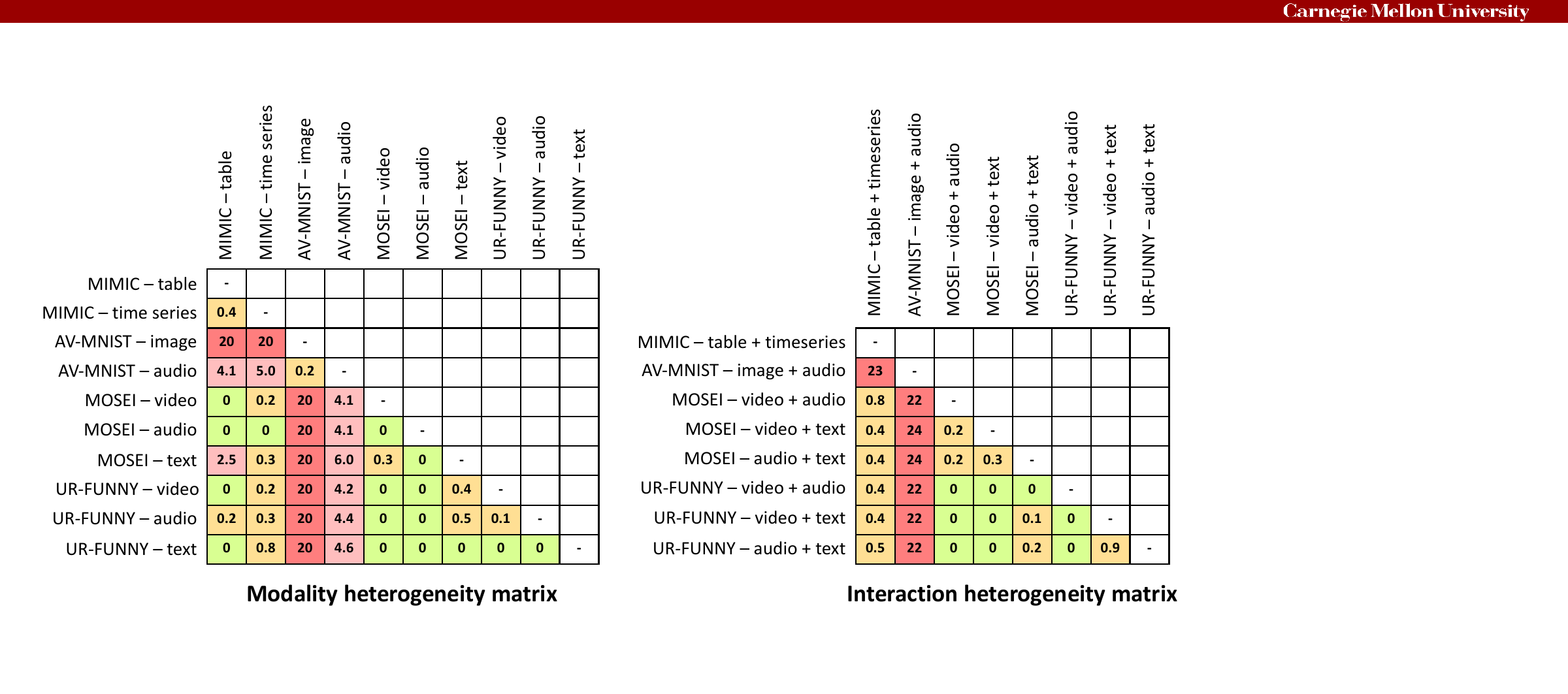}
\caption{Modality and interaction heterogeneity matrices color coded by distances, with green showing smaller distances and dark red larger distances. We find clear task outliers (\textsc{AV-MNIST} has high difficulty transferring to others), and that there is generally more interaction heterogeneity than unimodal heterogeneity. Otherwise, the same modality and modality pairs across different tasks are generally similar to each other.}
\label{fig:matrix}
\vspace{-4mm}
\end{figure*}

\subsection{Heterogeneity measurements and parameter groups}
\label{sec:groups}

We begin with a study of the heterogeneity matrices in Figure~\ref{fig:matrix} and the resulting parameter groups.

\textbf{Modality heterogeneity:} We first notice that the modalities from \textsc{AV-MNIST} only transfer well to each other and has high difficulty transferring to other modalities from the other datasets. The same modality across different tasks is generally similar to each other (e.g., text between \textsc{UR-FUNNY} and \textsc{MOSEI}, audio between \textsc{UR-FUNNY} and \textsc{MOSEI}). The text modality in \textsc{UR-FUNNY} seems to be close to most other modalities, and likewise for the tabular modality in \textsc{MIMIC}. It is also worth noting that the video and audio modalities are not the most informative in \textsc{MOSEI}, and predictions are dominated by language~\citep{zadeh2017tensor}, which may explain their general homogeneity with respect to other modalities.

\textbf{Interaction heterogeneity:} There is generally more interaction heterogeneity than unimodal, implying that the interactions between modality pairs tend to be more unique.
Again, we notice the general poor transfer from the modality pair (image+audio) in \textsc{AV-MNIST} to other pairs, and the general strong transfer from (audio+text) in \textsc{UR-FUNNY} to the rest, which shows a relationship between modality and interaction heterogeneity.
We also find that the same modality pairs (video+text) and (video+audio) shows crossmodal similarity across both datasets they appear in: \textsc{MOSEI} and \textsc{UR-FUNNY}. Finally, while the triplet of crossmodal pairs in \textsc{MOSEI} are quite different from each other, those in \textsc{UR-FUNNY} are more similar.

Using these measurements, we show the final groups of parameters obtained after clustering the matrices for different values of $k$. As an example, for $|\mathbb{U}|=3, |\mathbb{C}|=3, k=6$, the groups are $\mathcal{U}_1 = \{$\textsc{AV-MNIST} image, \textsc{AV-MNIST} audio$\}$, $\mathcal{U}_2 = \{$\textsc{MIMIC} table, \textsc{MOSEI} video, \textsc{MOSEI} audio$\}$, $\mathcal{U}_3 = \{$\textsc{MIMIC} timeseries, \textsc{MOSEI} text, \textsc{UR-FUNNY} text, \textsc{UR-FUNNY} video, \textsc{UR-FUNNY} audio$\}$, and $\mathcal{C}_1 = \{$\textsc{AV-MNIST} image+audio$\}$, $\mathcal{C}_2 = \{$\textsc{MOSEI} video+audio$\}$, and $\mathcal{C}_3 = \{$\textsc{MIMIC} table+timeseries, \textsc{MOSEI} video+text, \textsc{MOSEI} audio+text, \textsc{UR-FUNNY} video+text, \textsc{UR-FUNNY} video+audio, \textsc{UR-FUNNY} audio+text$\}$.

Finally, we observe the low-rank nature of the heterogeneity matrices due to symmetry and approximate triangle inequality, such that even using a low-rank approximation of $r=3$ is sufficient to approximate the entire matrix. This suggests that we do not need to exhaustively measure unimodal and interaction transfer between all modality pairs to enjoy the benefits of our proposed approach.

\subsection{Qualitative results}

We now present our results on the multitask, transfer, and generalization capabilities of \names\ using performance and efficiency metrics. Henceforth, we will refer to the following models:

(1) \textbf{\names\ share none} refers to individual copies of \names\ models, one for each task.

(2) \textbf{\names\ share all} refers to one single \names\ model fully shared across all modalities and tasks.

(3) \textbf{\names} refers to the full heterogeneity-aware \names\ model across all modalities and tasks with learned parameter groupings based on heterogeneity measurements.

\begin{wrapfigure}{r}{7cm}
\centering
\includegraphics[width=\linewidth]{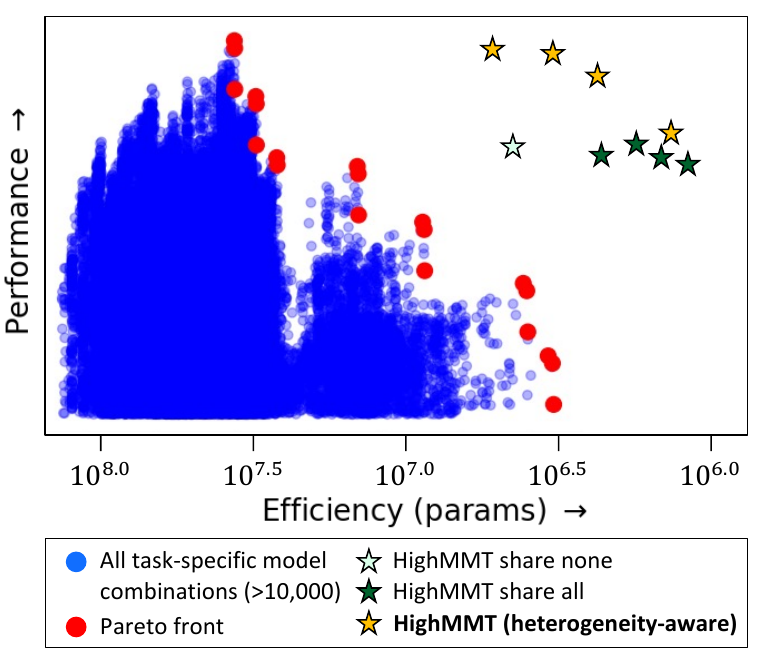}
\caption{\textbf{Overall tradeoff.} \names\ pushes forward the Pareto front of performance and efficiency as compared to all possible ($>10^{5}$) combinations of task-specific models across multiple datasets~\citep{liang2021multibench}. The $x$-axis denotes (inverted) total parameters and $y$-axis denotes performance scaled to a $0-1$ range before averaging across datasets.}
\label{fig:overall}
\vspace{-2mm}
\end{wrapfigure}

\textbf{Multitask performance and efficiency.} In Figure~\ref{fig:overall}, we summarize the overall tradeoff between performance and efficiency using existing task-specific models and variants of \names. The blue dots represent all possible combinations of task-specific models across multiple datasets (summarized in MultiBench~\citep{liang2021multibench}, $>10^{5}$ total combinations) with their overall performance (scaled to a $0-1$ range before averaging across datasets) and overall efficiency (inverted total number of parameters). The red dots represent the state-of-the-art Pareto front: points that are not strictly dominated in both performance and efficiency. In light green, separate single-task \names\ models (share none) already improve parameter efficiency as compared to standard Multimodal Transformers~\citep{lu2019vilbert,tsai2019multimodal}. In dark green is \names\ (share all) trained in a homogeneous multitask manner (i.e., with full parameter sharing across unimodal and multimodal layers within and across tasks), which further pushes forward the Pareto front by improving both performance and efficiency. Finally, in orange, \names\ with heterogeneity-aware fine-tuning achieves significantly better tradeoffs between performance and efficiency, with efficiency and consistently high performance across multiple modalities and tasks.

\begin{table}
\fontsize{9}{11}\selectfont
\setlength\tabcolsep{2.0pt}
\caption{Tuning the number of parameter groups results in controlled tradeoffs between parameters and performance.}
\centering
\footnotesize
\begin{tabular}{l|cccc}
\hline \hline
Clusters & Performance $\uparrow$ & Params (M) $\downarrow$ \\
\hline
2 (share all) & $68.4\pm 0.4$ & $1.07$ \\
4 & $68.8 \pm 0.5$ & $1.24$ \\ 
6 & $70.1 \pm 0.2$& $2.47$ \\ 
7 & $71.0 \pm 0.1$ & $3.11$ \\
9 & $71.2 \pm 0.2$ & $4.23$ \\
\hline \hline
\end{tabular}

\vspace{-2mm}
\label{tab:clusters}
\end{table}

The suite of \names\ models is obtained by tuning $k$, the total number of unimodal and crossmodal parameter groups (i.e., the number of clusters when clustering heterogeneity matrices). $k$ can be seen as a hyper-parameter depending on the computational budget, with smaller $k$ implying more parameter sharing on lower budgets and vice-versa. In Table~\ref{tab:clusters}, we show the effect of $k$ on average performance and total parameters. We test $k$ in the range $\{2,4,6,7,9\}$, with $|\mathbb{U}|=1, |\mathbb{C}|=1$, $|\mathbb{U}|=3, |\mathbb{C}|=1$, $|\mathbb{U}|=3, |\mathbb{C}|=3$, $|\mathbb{U}|=3, |\mathbb{C}|=4$, and $|\mathbb{U}|=4, |\mathbb{C}|=5$ respectively where $|\mathbb{U}|, |\mathbb{C}|$ denote the number of unimodal and crossmodal parameter groups.
We see a controllable tradeoff: starting with a fully shared model and increasing the number of parameter groups, we also see steadily improving performance approaching task-specific state-of-the-art models. Overall, optimizing for performance results in a model as strong as current state-of-the-art models while using $8\times$ fewer total parameters. Optimizing for efficiency results in a model that reaches within $96\%$ of current state-of-the-art performance but using $30\times$ fewer total parameters (mean and deviation over $10$ runs).

\begin{table}
\fontsize{9}{11}\selectfont
\setlength\tabcolsep{1.0pt}
\caption{\textbf{Cross-modal few-shot transfer to new modalities and tasks.} We train multitask \names\ on $1/2/3$ datasets and find that it generalizes few-shot to new modalities and tasks on the $4$th dataset, with improved performance over single-task training on the $4$th dataset. Cross-modal transfer improves with more pretraining tasks and works best on the smallest target tasks (\textsc{UR-FUNNY}).}
\centering
\footnotesize
\begin{tabular}{l|cccc}
\hline \hline
\multirow{2}{*}{\# Source tasks} & \multicolumn{4}{c}{Target task} \\
& \textsc{UR-FUNNY} & \textsc{MOSEI} & \textsc{MIMIC} & \textsc{AV-MNIST} \\
\hline
$0$ (no transfer) & $63.1 \pm 0.5$ & $79.0 \pm 0.5$ & $67.7 \pm 0.6$ & $70.3 \pm 0.4$ \\
$1$ & $63.5 \pm 0.5$ & $79.2 \pm 0.3$ & $67.9 \pm 0.5$ & $70.5 \pm 0.4$ \\
$2$ & $64.0 \pm 0.7$ & $79.3 \pm 0.5$ & $68.0 \pm 0.8$ & $70.5 \pm 0.4$ \\
$3$ & $\mathbf{64.7 \pm 0.4}$ & $\mathbf{79.6 \pm 0.6}$ & $\mathbf{68.4 \pm 0.6}$ & $70.6 \pm 0.4$ \\
\hline \hline
\end{tabular}

\vspace{-2mm}
\label{tab:transfer}
\end{table}

\textbf{Positive transfer to new modalities and tasks.}
\names\ also offers opportunities to study whether we can \textit{transfer} knowledge between completely different modalities and tasks. Starting with the collection of $4$ datasets in the order \textsc{MOSEI}, \textsc{AV-MNIST}, \textsc{MIMIC}, and \textsc{UR-FUNNY} ranked by largest dataset size (total of datapoints and memory storage per datapoint), we pre-train a fully-shared \names\ model on $1/2/3$ of the $4$ tasks before fine-tuning on the fourth task only (e.g., train on \textsc{MOSEI} and transfer to \textsc{UR-FUNNY}, on \textsc{MOSEI}+\textsc{AV-MNIST} then transfer to \textsc{UR-FUNNY}, and on \textsc{MOSEI}+\textsc{AV-MNIST}+\textsc{MIMIC} then transfer to \textsc{UR-FUNNY}, and likewise for transfer to the other $3$ datasets.

From Table~\ref{tab:transfer}, we found that on all four combinations of multitask pretraining and fine-tuning, weights learned from other multimodal tasks generalize well to new modalities and tasks, improving performance over single target-task training (mean and standard deviation over $10$ runs).
When we increase the number of pretraining datasets, we observe a consistent improvement in fine-tuned target task performance. There is an inverse correlation between target task size and performance improvement: the smallest dataset, \textsc{UR-FUNNY}, benefited the most ($+2.4\%$) from transfer learning from $0$ to $3$ multitask datasets. This implies that our multimodal pretraining-fine-tuning paradigm is useful for low-resource target modalities and tasks.

\color{black}

Finally, we compare transfer learning performance across different levels of partial observability. While one would expect the transfer to \textsc{MIMIC} to be the hardest due to its modality set $\{ \textrm{time-series}, \textrm{table} \}$ being completely disjoint from the remaining $3$ datasets, we still observe a $+0.8\%$ gain as compared to single-task training. Therefore, \names\ can generalize to new modalities and tasks. Unsurprisingly, for datasets with more overlap (e.g., \textsc{UR-FUNNY} with complete overlap in $\{ \textrm{text}, \textrm{video}, \textrm{audio} \}$ with respect to pretraining), we find larger improvements using transfer learning over single-task models ($+2.4\%$).

\begin{table*}[]
\centering
\fontsize{9}{11}\selectfont
\setlength\tabcolsep{1.0pt}
\caption{\names\ achieves strong performance on overall performance and efficiency (mean and deviation over $10$ runs), sometimes even beating (shown in \textbf{bold}) the task-specific state-of-the-art, especially on the relatively understudied modalities (time-series, robotics sensors, and sets) from the robotics (\textsc{Push}, \textsc{V\&T}) HCI (\textsc{ENRICO}), and healthcare (\textsc{MIMIC}) research areas, while using \textbf{$\mathbf{10\times}$ fewer parameters} due to parameter sharing and multitask learning. SOTA captures the max performance and parameters of more than $20$ task-specific multimodal models: [1] \textsc{GradBlend}~\citep{wang2020makes}, [2] \textsc{LF-LSTM}~\citep{ding2022multimodal}, [3] \textsc{LF}~\citep{gadzicki2020early}, [4] \textsc{MulT}~\citep{tsai2019multimodal}, [5] \textsc{MFAS}~\citep{perez2019mfas}, [6] \textsc{MFM}~\citep{yang2022disentangled}, and [7] \textsc{LRTF}~\citep{zeng2022multimodal}.}
\centering
\footnotesize

\begin{tabular}{l|ccccccc}
\hline \hline
Model & \textsc{ENRICO} $\uparrow$ & \textsc{Push} $\downarrow$ & \textsc{V\&T} $\uparrow$ & \textsc{UR-FUNNY} $\uparrow$ & \textsc{MOSEI} $\uparrow$ & \textsc{MIMIC} $\uparrow$ & \textsc{AV-MNIST} $\uparrow$ \\
\hline
SOTA & $51.0 \pm 1.4$[1] & $0.290 \pm 0.1$[2] & $93.6 \pm 0.1$[3] & $66.7 \pm 0.3$[4] & $\mathbf{82.1 \pm 0.5}$[4] & $68.9 \pm 0.5$[6,7] & $\mathbf{72.8 \pm 0.2}$[5] \\
\names & $\mathbf{52.7 \pm 0.6}$ & $\mathbf{0.277 \pm 0.1}$ & $\mathbf{96.3 \pm 0.2}$ & $66.2 \pm 0.4$ & $80.2 \pm 0.2$ & $68.2 \pm 0.3$ & $71.1 \pm 0.2$ \\
\hline \hline
\end{tabular}

\vspace{2mm}

\begin{tabular}{l|ccccccc}
\hline \hline
Model & Params (M) $\downarrow$ \\
\hline
SOTA & $32.3$ \\
\names & $\mathbf{3.01}$ \\
\hline \hline
\end{tabular}

\vspace{-2mm}
\label{tab:sota}
\end{table*}

\textbf{Comparison with task-specific state-of-the-art.} In Table~\ref{tab:sota}, we compare multitask performance and efficiency with task-specific state-of-the-art models. We achieve performance within the range of published models (and usually close to the individual task-specific state-of-the-art) in MultiBench, which tallies more than $20$ recent multimodal models in each task's literature~\citep{liang2021multibench}. In fact, \names\ even sets new state-of-the-art results on several datasets, especially on the relatively understudied modalities (time-series, force and proprioception sensors, and sets) from the robotics (\textsc{Push}, \textsc{V\&T}) and HCI (\textsc{ENRICO}) research areas. On top of strong performance, the main benefit lies in using fewer total parameters as compared to separate task-specific models - more than $10\times$ reduction. Since this reduction grows with the number of tasks, our approach is scalable to high-modality scenarios.

\textbf{Partial-observability.} Observe \names\ performance on partially-observable modality subsets (i.e., target task involving modalities not present in the other tasks): from Table~\ref{tab:sota}, we find that the model performs well on the \textsc{MIMIC} dataset despite its modality set $\{ \textrm{time-series}, \textrm{table} \}$ being completely disjoint from the remaining $3$ datasets - we obtain similar performance across both multitask and single-task models ($68.2 \pm 0.3\%$ vs $68.9 \pm 0.5\%$). We find that \names\ multitask also works on \textsc{ENRICO} dataset in HCI ($52.7 \pm 0.6\%$ multitask vs $51.0 \pm 1.4\%$ single-task) despite it having completely disjoint modality inputs.

\textbf{Multitask fusion and retrieval.} We perform multitask training over multimodal fusion in \textsc{AV-MNIST} and retrieval in \textsc{CIFAR-ESC}. While fusion emphasizes information integration, retrieval focuses on aligning corresponding elements expressed through different views of the data~\citep{liang2022foundations}. Even across these vastly different prediction tasks, we find that multitask training ($60.5\%$ retrieval accuracy) improves upon single-task training ($58.8\%$). Not only have the unimodal networks simultaneously processed different modalities, but the crossmodal network has captured correspondences useful for both fusion and retrieval.

\begin{table*}[]
\fontsize{9}{11}\selectfont
\setlength\tabcolsep{1.0pt}
\caption{We conduct in-depth \textbf{ablation studies} and find strong evidence for (1) having separate unimodal and interaction layers, (2) determining parameter sharing via feature transfer, and (3) homogeneous pre-training before heterogeneity-aware fine-tuning into parameter groups (mean and standard deviation over $10$ runs).}
\centering
\footnotesize
\begin{tabular}{l|l|cccc|c}
\hline \hline
& Model & \textsc{UR-FUNNY} $\uparrow$ & \textsc{MOSEI} $\uparrow$ & \textsc{MIMIC} $\uparrow$ & \textsc{AV-MNIST} $\uparrow$ & Ave $\uparrow$ \\
\hline
Full model & \names & $\mathbf{66.2 \pm 0.4}$ & $\mathbf{80.2 \pm 0.2}$ & $\mathbf{68.2 \pm 0.3}$ & $\mathbf{71.1 \pm 0.2}$ & $\mathbf{71.4 \pm 0.3}$ \\
\hline
\multirow{3}{*}{\makecell[l]{Architecture\\ablations}} & - w/o embeddings & $63.0 \pm 1.2$ & $79.0 \pm 0.7$ & $67.1 \pm 1.2$ & $70.3 \pm 0.7$ & $69.8 \pm 0.3$ \\
& - w/o unimodal & $57.9 \pm 0.3$ & $61.9 \pm 2.1$ & $63.0 \pm 0.9$ & $59.5 \pm 1.4$ & $60.6 \pm 0.7$ \\
& - w/o crossmodal~\citep{reed2022generalist} & $63.8 \pm 1.0$ & $79.5 \pm 0.5$ & $\mathbf{67.9 \pm 0.4}$ & $70.4 \pm 0.5$ & $70.4 \pm 0.5$ \\
\hline
\multirow{6}{*}{\makecell[l]{Param sharing\\ablations}} & - share none~\citep{liang2021multibench} & $63.7 \pm 0.7$ & $79.4 \pm 0.4$ & $67.7 \pm 0.7$ & $70.4 \pm 0.1$ & $70.2\pm 0.3$ \\
& - share unimodal~\citep{reed2022generalist} & $62.5\pm 1.3$ & $79.0\pm 1.1$ & $63.4\pm 1.4$ & $70.1\pm 0.7$ & $68.8\pm 0.8$ \\
& - share crossmodal~\citep{akbari2021vatt} & $63.0\pm 1.1$ & $79.5\pm 0.3$ & $64.3\pm0.3$ & $70.1\pm 0.9$ & $69.2\pm0.3$ \\
& - share all~\citep{singh2022flava} & $63.1\pm 0.7$ & $79.2\pm 0.3$ & $63.7\pm 1.6$ & $68.6\pm 0.6$ & $68.7\pm 0.5$ \\
& - random difference & $62.9\pm 0.9$ & $79.5\pm 0.6$ & $67.6\pm 0.3$ & $70.4\pm 0.2$ & $70.1\pm 0.3$ \\
& - feature difference~\citep{sun2016return} & $64.0\pm 1.0$ & $79.4\pm 0.3$ & $\mathbf{67.9\pm 0.3}$ & $70.1\pm 0.4$ & $70.4\pm 0.2$ \\
\hline
\multirow{1}{*}{Training ablations} & - w/o homogeneous pretraining & $61.2 \pm 0.1$ & $78.5 \pm 0.1$ & $64.8 \pm 0.1$ & $\mathbf{71.1 \pm 0.2}$ & $69.9 \pm 0.1$ \\
\hline \hline
\end{tabular}

\vspace{-4mm}
\label{tab:ablation}
\end{table*}

\subsection{Ablation studies}

In this subsection, we carefully ablate the model architectures, parameter sharing, and training decisions.

\textbf{Architectural ablations.} We first analyze each architectural component of \names: (1) \textit{w/o embeddings} removes the only modality-specific component in the model - the modality embeddings. We set embeddings for all modalities to be the same to test whether a modality-specific component is necessary to capture heterogeneity across input data sources, (2) \textit{w/o unimodal} removes the unimodal encoder and directly applies the cross-attention layer, and \textit{w/o crossmodal} replaces the crossmodal layer with a concatenation of unimodal features and a linear classification layer. The latter resembles the most direct multimodal extension of existing work in shared unimodal encoders like Perceiver~\citep{jaegle2021perceiver}, MultiModel~\citep{kaiser2017one}, ViT-BERT~\citep{li2021towards} or PolyViT~\citep{likhosherstov2022polyvit}. From Table~\ref{tab:ablation}, removing any of the $3$ components in \names\ results in worse performance. The unimodal encoder is particularly important.

\textbf{Param sharing ablations.} We further ablate with respect to possible parameter sharing settings in \names: (1) \textit{share none} uses separate unimodal and multimodal layers reminiscent of typical single-task multimodal transformers~\citep{tsai2019multimodal,lu2019vilbert,hendricks2021decoupling}, (2-3) \textit{share unimodal (crossmodal)} only shares the unimodal (crossmodal) layer during multitask training, (4) \textit{share all} shares all parameters without accounting for possible heterogeneity~\citep{reed2022generalist}, (5) \textit{random difference} determines $k$ parameter groups randomly rather than via heterogeneity measurements, (6) \textit{feature difference} uses feature-level divergences on jointly trained unimodal encoders (i.e., $\| U(X_1)-U(X_2)\|_2^2$) rather than transfer performance to measure heterogeneity as is commonly done in transfer learning and domain adaptation~\citep{daume2007frustratingly,sun2016return}. From Table~\ref{tab:ablation}, our proposed heterogeneity-aware parameter grouping results in the best overall performance as compared to fully shared, fully separate, or parameter grouping informed by other heterogeneity measures such as random or feature distance.

\textbf{Training ablations.} Finally, we explore \textit{w/o homogeneous pretraining}: directly learning a model with parameter groups as selected by our approach as opposed to performing homogeneous pre-training before fine-tuning them into parameter groups. From Table~\ref{tab:ablation}, we find that this ablation underperforms - training parameter groups from scratch overfits to smaller datasets which hurts overall performance.

\subsection{Understanding homogeneity and heterogeneity in \names}

We now take a deeper empirical analysis to better understand \names, through parameter overlap and interference experiments.

\textbf{Parameter overlap.} Starting with a trained multitask \names, we use a gradient-based method~\citep{han2020explaining} to determine how much each parameter is involved in a specific task. For each task $T$ and parameter $\theta \in \Theta$ in multitask model $M_\Theta$, we compute the involvement $I_T (\theta) = \mathbb{E}_{(x,y) \in T} | \nabla_\theta M_\Theta(y|x) |$ where $M_\Theta(y|x)$ is the predicted probability of correct target $y$ by $M_\Theta$ given $x$ as input. In other words, this measures the absolute gradient with respect to $\theta$ when predicting $y$ given $x$ in task $T$. A higher absolute gradient implies ``activated'' neurons and vice-versa for gradients closer to $0$. This enables us to compute the extent a parameter $\theta$ is involved for each task. The \textit{number of tasks} a given parameter $\theta$ is involved in can then be approximated by thresholding and summing up $n(\theta) = \sum_T \left( \mathbbm{1} \{ I_T (\theta) > \epsilon \max(I_1 (\theta),I_2 (\theta),I_3 (\theta),I_4 (\theta) \} \right)$ which returns an integer from $1$ to $4$. We chose a threshold $\epsilon$ such that parameters are classified as active about half the time on average, which occurs at $\epsilon = 0.2$.

\begin{wraptable}{r}{8cm}
\fontsize{9}{11}\selectfont
\setlength\tabcolsep{2.0pt}
\caption{We find evidence of significant \textbf{parameter overlap} across unimodal encoders: $>92\%$ of neurons are involved in at least $3$ of the $4$ tasks, while the multimodal layers are more task-specific: only $10\%$ of neurons are involved in $3$ or $4$ tasks.}
\centering
\footnotesize
\begin{tabular}{l|cccc}
\hline \hline
\multirow{2}{*}{Component} & \multicolumn{4}{c}{Number of involved tasks} \\
& $1$ & $2$ & $3$ & $4$ \\
\hline
Unimodal layers & $2.8\%$ & $5.1\%$ & $\mathbf{61.1}\%$ & $\mathbf{31.1}\%$ \\
Crossmodal layers & $\mathbf{48.8}\%$ & $\mathbf{39.7}\%$ & $9.9\%$ & $1.6\%$ \\
\hline \hline
\end{tabular}
\vspace{-0mm}
\label{tab:sharing}
\end{wraptable}

Since we are interested in the level of parameter overlap in the shared unimodal encoder and multimodal layer, we set $\theta$ as these $2$ modules and report results in Table~\ref{tab:sharing}. There is evidence of significant parameter overlap across unimodal encoders: more than $92\%$ of neurons are involved in at least $3$ of the $4$ tasks. On the other hand, there is not nearly as much parameter overlap in the multimodal layer: only $10\%$ of neurons are involved in $3$ or $4$ tasks. Hence, it seems like the unimodal encoders learn task-agnostic representations, but the subsequent multimodal layers (closer to task-specific classifiers) capture more task-specific information. This also reinforces our observation in \S\ref{sec:groups} that there is generally more interaction heterogeneity than modality heterogeneity, which suggests using fewer unimodal parameter groups and more crossmodal parameter groups.

\definecolor{gg}{RGB}{15,250,15}
\definecolor{rr}{RGB}{250,15,15}
\definecolor{yy}{RGB}{230,185,25}

\begin{table}
\fontsize{9}{11}\selectfont
\setlength\tabcolsep{2.0pt}
\caption{ \textbf{Parameter interference}: we observe different performance drops on each task (columns) after training on one task with flipped labels (rows). Training the shared unimodal encoders causes the most harm, which implies that unimodal encoders contain more shared neurons sensitive to task changes. \colorbox{rr}{Red} for drops greater than $20\%$, \colorbox{yy}{yellow} for drops between $10$ and $20\%$, and \colorbox{gg}{green} for drops below $10\%$.}
\centering
\begin{tabular}{l|cccc}
\hline \hline
\multicolumn{5}{c}{(a) Training entire model} \\
\hline
Flipped task & \multicolumn{1}{c}{\textsc{UR-FUNNY}} & \multicolumn{1}{c}{\textsc{MOSEI}} & \multicolumn{1}{c}{\textsc{MIMIC}} & \multicolumn{1}{c}{\textsc{AV-MNIST}}  \\
\hline
\textsc{UR-FUNNY} & \cellcolor{rr} $-24.6$ & \cellcolor{gg} $-8.83$ & \cellcolor{yy} $-10.6$ & \cellcolor{rr} $-57.7$ \\
\textsc{MOSEI} & \cellcolor{gg} $-4.07$ & \cellcolor{rr} $-59.7$ & \cellcolor{rr} $-20.3$ & \cellcolor{rr} $-53.2$ \\
\textsc{MIMIC} & \cellcolor{gg} $-3.59$ & \cellcolor{gg} $-5.83$ & \cellcolor{rr} $-33.1$ & \cellcolor{rr} $-37.5$ \\
\textsc{AV-MNIST} & \cellcolor{gg} $-3.50$ & \cellcolor{gg} $-1.23$ & \cellcolor{gg} $-4.87$ & \cellcolor{rr} $-68.9$ \\
\hline \hline
\end{tabular}

\vspace{2mm}

\begin{tabular}{l|cccc}
\hline \hline
\multicolumn{5}{c}{(b) Only training unimodal encoder} \\
\hline
Flipped task & \multicolumn{1}{c}{\textsc{UR-FUNNY}} & \multicolumn{1}{c}{\textsc{MOSEI}} & \multicolumn{1}{c}{\textsc{MIMIC}} & \multicolumn{1}{c}{\textsc{AV-MNIST}}  \\
\hline
\textsc{UR-FUNNY} & \cellcolor{rr} $-23.8$ & \cellcolor{yy} $-10.1$ & \cellcolor{yy} $-12.8$ & \cellcolor{rr} $-58.4$  \\
\textsc{MOSEI} & \cellcolor{gg} $-5.77$ & \cellcolor{rr} $-57.6$ & \cellcolor{rr} $-21.1$ & \cellcolor{rr} $-52.7$ \\
\textsc{MIMIC} & \cellcolor{gg} $-3.03$ & \cellcolor{gg} $-3.54$ & \cellcolor{rr} $-35.0$ & \cellcolor{rr} $-56.3$ \\
\textsc{AV-MNIST} & \cellcolor{gg} $-2.94$ & \cellcolor{gg} $-7.82$ & \cellcolor{rr} $-53.6$ & \cellcolor{rr} $-69.3$ \\
\hline \hline
\end{tabular}

\vspace{2mm}

\begin{tabular}{l|cccc}
\hline \hline
\multicolumn{5}{c}{(c) Only training multimodal layer} \\
\hline
Flipped task & \multicolumn{1}{c}{\textsc{UR-FUNNY}} & \multicolumn{1}{c}{\textsc{MOSEI}} & \multicolumn{1}{c}{\textsc{MIMIC}} & \multicolumn{1}{c}{\textsc{AV-MNIST}}  \\
\hline
\textsc{UR-FUNNY} & \cellcolor{rr} $-25.2$ & \cellcolor{gg} $-8.34$ & \cellcolor{gg} $-2.67$ & \cellcolor{gg} $-8.16$ \\
\textsc{MOSEI} & \cellcolor{gg} $0.47$ & \cellcolor{rr} $-59.6$ & \cellcolor{yy} $-19.8$ & \cellcolor{gg} $-8.19$  \\
\textsc{MIMIC} & \cellcolor{gg} $0.19$ & \cellcolor{gg} $-0.76$ & \cellcolor{rr} $-35.2$ & \cellcolor{gg} $-4.87$ \\
\textsc{AV-MNIST} & \cellcolor{gg} $-1.61$ & \cellcolor{gg} $-1.48$ & \cellcolor{gg} $-2.23$ & \cellcolor{rr} $-69.1$ \\
\hline \hline
\end{tabular}
\vspace{-2mm}
\label{tab:destructive}
\end{table}

\textbf{Parameter interference.} Another empirical proof for parameter sharing in multitask models is the phenomenon of \textit{parameter interference}: to what extent do parameters interfere with each other across tasks? We perform an experiment to investigate parameter interference: we pick one task and flip the labels in its training set, train the multitask model on the modified training set, and see how the incorrectly labeled task affects performance on other tasks. This experiment provides evidence of information sharing: if the multitask model does not share information (i.e., the model learns independent subspaces for each task), then one would not observe negative interference from one noisy dataset. We study negative interference under $3$ configurations of training (a) the whole model; (b) only the unimodal encoder, and (c) only the multimodal layer on the flipped training set.

From Table~\ref{tab:destructive}, certain tasks are more affected by negative interference (e.g., \textsc{AV-MNIST}), while some tasks are not influenced as much (e.g., \textsc{UR-FUNNY}). Again, this reflects our heterogeneity measurements in \S\ref{sec:groups}, where \textsc{AV-MNIST} displays high heterogeneity.
Furthermore, performance drops due to training the unimodal encoders are the most significant, which corroborates with our parameter overlap and heterogeneity analysis that unimodal encoders contain more entangled parameters which are more sensitive to task changes. On the other hand, multimodal layers contain more disentangled parameters, which results in higher heterogeneity measurements and needs more separate parameter groups.

\vspace{-2mm}
\section{Related Work}
\vspace{-2mm}

\textbf{Multimodal Transformers} have emerged as strong models for representation learning. Building upon the Transformer~\citep{vaswani2017attention}, multimodal extensions use either full self-attention over modalities concatenated across the sequence dimension~\citep{li2019visualbert,sun2019videobert,Su2020VLBERT,chen2020uniter} or a cross-modal attention layer~\citep{lu2019vilbert,tsai2019multimodal,tan2019lxmert}, and are useful for sequential data by automatically aligning and capturing complementary features at different time-steps~\citep{tsai2019multimodal,yao2020multimodal,lee2020parameter}. Self-supervised multimodal pretraining has emerged as an effective way to train these architectures, with the aim of learning representations from large-scale unlabeled multimodal data before transferring to downstream tasks via fine-tuning~\citep{lu2019vilbert,li2019visualbert,Su2020VLBERT}. These pretraining objectives typically consist of unimodal masked prediction, crossmodal masked prediction, and multimodal alignment prediction~\citep{hendricks2021decoupling}.

\textbf{Unified encoder for unimodal learning.} Several works such as Perceiver~\citep{jaegle2021perceiverio,jaegle2021perceiver}, MultiModel~\citep{kaiser2017one}, ViT-BERT~\citep{li2021towards}, and PolyViT~\citep{likhosherstov2022polyvit} have explored the possibility of using the same architecture for different inputs on unimodal tasks (i.e., language, image, video, or audio-only). The Transformer architecture has emerged as a popular choice due to its suitability for serialized inputs such as text~\citep{devlin2019bert}, images~\citep{dosovitskiy2020image}, video~\citep{sun2019videobert}, and time-series data~\citep{lim2021temporal}, a phenomenon further observed by~\citet{lu2021pretrained} where a single Transformer pretrained on text transfers to sequence modeling and image classification. While these serve as building blocks in our model, our focus is on a general-purpose multimodal model for multitask and transfer learning across different subsets of modalities rather than unimodal tasks.

\textbf{Multimodal multitask and transfer learning.} There have also been several attempts to build a single model that works well on a suite of multimodal tasks~\citep{li2019visualbert,lu2019vilbert,Su2020VLBERT,cho2021vlt5,reed2022generalist}. For example, UniT~\citep{hu2021transformer}, VLBERT~\citep{Su2020VLBERT}, ViLBERT~\citep{lu2019vilbert}, and VL-T5~\citep{cho2021vlt5} are all unifying models for vision-and-language tasks. VATT~\citep{akbari2021vatt} jointly trains a shared model on video, audio, and text data to perform audio-only, video-only, and image-text retrieval tasks. FLAVA~\citep{singh2022flava} found that pretraining a shared model with unpaired images, unpaired text, and image-text pairs results in strong performance on image-only, text-only, and image-text multimodal tasks, while~\citet{reed2022generalist} scales up a single Transformer model for image, text, and decision-making tasks.
However, all of these train a single model for all tasks, without investigating how heterogeneity can necessitate partial parameter sharing.
On the transfer side, while more research has focused on transfer within the same modality with external information~\citep{socher2013zero,DBLP:journals/corr/abs-1903-11101,NIPS2019_8731,zadeh2020foundations},~\citet{liang2021cross} is the only work that studies transfer to completely new modalities. However, they require paired data collection and modality-specific modeling. Our work goes beyond the commonly studied language, vision, and audio modalities to relatively understudied ones (e.g., tabular data, time-series, sensors, graphs, and set data). Furthermore, we show the possibility of generalizing to new modality subsets. Finally, our work also complements studies of transfer learning in a single modality~\citep{standley2020tasks,wu2022information,zamir2018taskonomy}, where insights from task heterogeneity have informed multitask approaches, as well as multisensor fusion in various domains such as healthcare~\citep{muhammad2021comprehensive} and robotics~\citep{suzuki2022survey,taniguchi2023world}.

\vspace{-2mm}
\section{Conclusion}
\vspace{-2mm}

We propose an information transfer approach for estimating modality and interaction heterogeneity, a key component towards automatically determining which modalities should be processed and fused jointly for efficient representation learning in high-modality scenarios. Our resulting model, \names\ dynamically determines the optimal parameter groupings balancing total performance and parameter efficiency, simultaneously achieves strong results on modalities (text, image, video, audio, time-series, sensors, tables, and sets) and tasks from different research areas, and transfers to new modalities and tasks during fine-tuning. We release our code and benchmarks which we hope will present a unified platform for subsequent analysis.

\chapter{Conclusion}
\label{chap:conclusion}
In this thesis, we advanced the foundations of multimodal machine learning by highlighting its key principles and core challenges. In the bulk of the thesis, we outlined our progress towards understanding the foundations of multimodal interactions and new modeling methods for generalizable representation learning across many input modalities and tasks. This concluding chapter provides a summary of the main contributions, discusses potential limitations, and outlines future research directions in multimodal artificial intelligence.

\section{Summary of Thesis Contributions}

Multimodal artificial intelligence is one of the most exciting subareas of artificial intelligence research today, and has the potential to make major impacts in autonomous agents with digital, physical, and social capabilities. This thesis aims to pave a foundation for multimodal artificial intelligence so that future students and researchers are able to better understand the breadth and depth of multimodal research today, are equipped with the scientific fundamentals required to perform cutting-edge research in this field, and are up-to-date with practical methods for machine learning from real-world multimodal datasets.

To summarize the contributions of this thesis, we began (in Section~\ref{chap:background}) by outlining the theoretical and computational foundations of multimodal machine learning by synthesizing a broad range of theoretical frameworks and application domains from both historical and recent perspectives. This foundation involves three key principles of modality \textit{heterogeneity}, \textit{connections}, and \textit{interactions} often present in multimodal problems which brings unique challenges to machine learning, which we outline through a taxonomy of six core challenges: \textit{representation}, \textit{alignment}, \textit{reasoning}, \textit{generation}, \textit{transference}, and \textit{quantification}. This taxonomy enables researchers to navigate the breadth of recent technical achievements and enables us to identify key open problems for future research.

In this first major part of this thesis, we build a foundation for multimodal interactions: the basic principle of how modalities combine to give rise to new information for a task. Section~\ref{chap:foundations1} presented an information-theoretic framework formalizing how \textit{modalities interact} with each other to give rise to new information for a task, which can be decomposed into redundancy, uniqueness, and synergy~\cite{liang2023quantifying}.
Using this theoretical framework, we proposed two practical estimators to quantify the interactions in real-world datasets. Quantifying the types of interactions a multimodal task requires enables researchers to understand their data and choose the right model to learn interactions in a principled way. Using this foundation of multimodal interactions, we design new self-supervised approaches to learn these interactions~\cite{liang2023factorized} (Section~\ref{chap:foundations2}), visualization tools for practitioners to analyze whether their model has succeeded in learning~\cite{liang2023multiviz} (Section~\ref{chap:foundations3}), and new guidelines for practitioners to decide which modality to collect for maximum increase in performance~\cite{liang2023multimodal} (Section~\ref{chap:foundations4}).

In the second major part of this thesis, we design practical multimodal foundation models that generalize over many modalities and tasks, which presents a step toward grounding large language models to real-world sensory modalities such as videos, physical sensors, and medical data. Section~\ref{chap:models1} introduced \multibench, a unified large-scale benchmark across a wide range of modalities, tasks, and research areas enabling research towards multimodal foundation models~\cite{liang2021multibench}. Section~\ref{chap:models2} presented the \textit{cross-modal attention}~\cite{liang2018multimodal,chen2017multimodal} and \textit{multimodal transformers}~\cite{tsai2019multimodal} architectures that are suitable for learning the interactions across many elements in modality sequences such as text, videos, time-series, and sensors. Finally, Section~\ref{chap:models3} showed how we can scale these architectures on \multibench\ to create general-purpose multimodal multitask models across a variety of tasks, including collaborating with practitioners to apply these models for real-world impact on affective computing, mental health, and cancer prognosis.

Together, our contributions deliver fundamental methodological and practical insights in multimodal learning, presenting approaches that are principled and explainable to practitioners while also capturing the benefits of scale across many modalities and tasks. Some of the work done during the PhD but not included in this thesis also paves a way towards improving the robustness, safety, and efficiency of multimodal models for real-world deployment.

\section{Limitations and Future Directions}
\label{sec:future}

Finally, we conclude this thesis by identifying the following future research challenges in multimodal artificial intelligence:

\textbf{Representation}: Learning multimodal representations is the cornerstone of multimodal machine learning. There has been substantial progress towards increasingly expressive and performant multimodal representations. However, there remain key challenges in their theoretical understanding and generalization beyond image and text. 

\textit{Theoretical and empirical frameworks}: How can we formally define the three core principles of heterogeneity, connections, and interactions? Can we quantify their presence in multimodal datasets and models, and understand whether current multimodal representation learning methods are suitable for learning different interactions? Answering these fundamental questions will lead to a better understanding of the capabilities and limitations of current multimodal representations, and inspire the development of new methods in a principled manner.

\textit{Beyond additive and multiplicative cross-modal interactions}: While recent work has been successful at modeling multiplicative interactions of increasing order, how can we capture causal, logical, and temporal connections and interactions? What is the right type of data and domain knowledge necessary to model these relationships? Modeling these interactions in a principled manner could lead to systems that are more robust, compositional, and explainable than those based fully on neural networks.

\textit{Tabular, sensors, and time-series}: Existing work has shown success in learning image, text, and audio-visual representations. However, tabular and time-series data are prevalent in many real-world applications such as healthcare and autonomous vehicles. How can we learn multimodal interactions between the best encoders for tabular and sensor data, which may not be based on deep learning (e.g., decision trees, time-series analysis), and neural network representations that are state-of-the-art for the text and image modalities?

\textit{Brain and multimodal perception}. There are many core insights regarding multimodal processing to be gained from human cognition, including the brain's multimodal properties~\cite{kosslyn2010multimodal} and mental imagery~\cite{nanay2018multimodal}. How does the human brain represent different modalities, how is multisensory integration performed, and how can these insights inform multimodal learning? In the other direction, what are opportunities in processing high-resolution brain signals such as fMRI and MEG/EEG, and how can multimodal learning help in the future analysis of data collected in neuroscience?

\textbf{Alignment}: There remain important challenges in aligning modality elements when these elements are extremely fine-grained in nature and exhibit long-range patterns across time.

\textit{Memory and long-term interactions}. Many current multimodal benchmarks only have a short temporal dimension, which has limited the demand for models that can accurately process long-range sequences and learn long-range interactions. Capturing long-term interactions presents challenges since it is difficult to semantically relate information when they occur very far apart in time or space and raises complexity issues. How can we design models (perhaps with memory mechanisms) to ensure that these long-term cross-modal interactions are captured?

\textbf{Reasoning}: Today's multimodal systems, especially those based on deep learning or large language models, are still not capable of robust and complex reasoning. We outline two challenges in compositional and interactive reasoning.

\textit{Multimodal compositionality}. How can we understand the reasoning process of trained models, especially regarding how they combine information from modality elements? This challenge of compositional generalization is difficult since many compositions of elements are typically not present during training, and the possible number of compositions increases exponentially with the number of elements~\cite{thrush2022winoground}. How can we best test for compositionality, and what reasoning approaches can enable compositional generalization?

\textit{Multimodal embodiment and interaction}. Most of today's multimodal systems are trained to make predictions without the capability to take actions in the world. The next generation of these systems will be those that can plan actions, imagine the effect these actions will have on the world, and choose the right sequence of actions over a long period of time to solve complex tasks. We have begun to build these interactive multimodal agents for the virtual world, such as processing multimedia web data to help humans with web tasks like online shopping, travel bookings, and content management. Building multisensory robotic systems that can actions in the real world, while respecting safety and robustness, is another long-term future direction.

\textbf{Generation}: The incredible advances of generative AI have inspired many future directions in generating multimedia content. 

\textit{Multimodal creation}. Synchronized creation of realistic video, text, and audio remains a challenge. These systems can be applied for entertainment, such as generating music videos, virtual avatar characters, virtual humans, and more. It is also likely that better multimodal generative models of the world can serve as world models to train planning and sequential decision making agents.

\textit{Real-world ethical concerns}. However, the recent success in generation has brought ethical concerns regarding their use. For example, large-scale pretrained language models can generate text denigrating to particular social groups~\cite{sheng2019woman}, toxic speech~\cite{gehman2020realtoxicityprompts}, and sensitive pretraining data~\cite{carlini2021extracting}. Future work should study how these risks are potentially amplified or reduced when the dataset is multimodal, and whether there are ethical issues specific to multimodal generation.

\textbf{Transference}: Advances in foundation models have also enabled increasingly general-purpose models that can transfer information and knowledge across a wide range of modalities and tasks. This opens up new directions in high-modality learning.

\textit{High-modality learning} aims to learn representations from an especially large number of heterogeneous data sources, which is a common feature of many real-world multimodal systems such as self-driving cars and IoT~\cite{huang2019multimodal}. More modalities introduce more dimensions of heterogeneity, incur complexity challenges in unimodal and multimodal processing, and require dealing with non-parallel data (i.e., not all modalities are present at the same time).

\textbf{Quantification}: Finally, we highlight several important lines of future work in quantifying and understanding key design decisions in the multimodal learning process.

\textit{Modality utility, tradeoffs, and selection}. How can we formalize why modalities can be useful or potentially harmful for a task? There are also challenges in quantifying \textit{modality and social biases} and \textit{robustness} to imperfect, noisy, and out-of-distribution modalities. Future work should come up with formal guidelines to compare these tradeoffs and select the optimal set of modalities balancing performance with these other potential concerns, which can help practitioners decide the right modalities to work with.

\textit{Explainability and interpretability}. Before models can be safely used by real-world stakeholders in domains such as medicine, autonomous systems, and user interfaces, we need to understand how to interpret their inner workings. How can we evaluate whether these phenomena are accurately interpreted? These challenges are exacerbated for relatively understudied modalities beyond language and vision, where the modalities themselves are not easy to visualize. Finally, how can we tailor these explanations, possibly in a \textit{human-in-the-loop} manner, to inform real-world decision-making?

In conclusion, we believe that this thesis can lay the theoretical and practical foundations for multimodal machine learning and inspire future work towards these open problems.

\backmatter



\footnotesize
\bibliographystyle{plainnat}
\bibliography{references} 

\end{document}